\documentclass[onefignum,onetabnum]{siamonline190516}

\usepackage{lipsum}
\usepackage{amsfonts}
\usepackage{graphicx}
\usepackage{epstopdf}
\usepackage{algorithmic}
\ifpdf
  \DeclareGraphicsExtensions{.eps,.pdf,.png,.jpg}
\else
  \DeclareGraphicsExtensions{.eps}
\fi

\usepackage{enumitem}
\setlist[enumerate]{leftmargin=.5in}
\setlist[itemize]{leftmargin=.5in}

\newsiamremark{remark}{Remark}
\newsiamremark{hypothesis}{Hypothesis}
\crefname{hypothesis}{Hypothesis}{Hypotheses}
\newsiamthm{claim}{Claim}

\usepackage{amssymb}
\usepackage{amsmath}
\usepackage{mathtools}
\usepackage{subfig}

\renewcommand{\vec}[1]{\mathbf{#1}}
\newcommand{\vecgreek}[1]{\boldsymbol{#1}}
\newcommand{\mtx}[1]{\mathbf{#1}}

\DeclareMathOperator*{\argmin}{arg\,min}

\DeclareMathOperator{\Tr}{Tr}
\newcommand{\mtxtrace}[1]{\Tr\left\lbrace{#1}\right\rbrace}
\newcommand{\expectation}[1]{\mathbb{E}\left[{{#1}}\right]} 
\newcommand{\expectationwrt}[2]{\mathbb{E}_{#2}\left[{{#1}}\right]}

\newcommand{\Ltwonorm}[1]{\left\Vert{{#1}}\right\Vert_2^2}

\headers{Double Double Descent}{Yehuda Dar and Richard G.~Baraniuk}

\title{Double Double Descent: On Generalization Errors in Transfer Learning between Linear Regression Tasks}

\author{Yehuda Dar\thanks{Electrical and Computer Engineering Department, Rice University
  (\email{ydar@rice.edu}).}
\and Richard G.~Baraniuk\thanks{Electrical and Computer Engineering Department, Rice University 
  (\email{richb@rice.edu}).}}

\usepackage{amsopn}

\ifpdf
\hypersetup{
  pdftitle={Double Double Descent: On Generalization Errors in Transfer Learning between Linear Regression Tasks},
  pdfauthor={Y. Dar and R. G. Baraniuk}
}
\fi

\begin{document}

\maketitle

\begin{abstract}
We study the \textit{transfer learning} process between two linear regression problems. 
An important and timely special case is when the regressors are \textit{overparameterized} and perfectly \textit{interpolate} their training data. 
We examine a parameter transfer mechanism whereby a subset of the parameters of the target task solution are constrained to the values learned for a related source task. 
We analytically characterize the generalization error of the target task in terms of the salient factors in the transfer learning architecture, i.e., the number of examples available, the number of (free) parameters in each of the tasks, the number of parameters transferred from the source to target task, and the relation between the two tasks. 
Our non-asymptotic analysis shows that the generalization error of the target task follows a \textit{two-dimensional double descent} trend (with respect to the number of free parameters in each of the tasks) that is controlled by the transfer learning factors. 
Our analysis points to specific cases where the transfer of parameters is beneficial as a substitute for extra overparameterization (i.e., additional free parameters in the target task). 
Specifically, we show that the usefulness of a transfer learning setting is fragile and depends on a delicate interplay among the set of transferred parameters, the relation between the tasks, and the true solution. 
We also demonstrate that overparameterized transfer learning is not necessarily more beneficial when the source task is closer or identical to the target task. 
\end{abstract}



	\section{Introduction}
\label{sec:introduction}

Transfer learning \cite{pan2009survey} is a prominent strategy to address a machine learning task of interest using information and parameters already learned and/or available for a related task. Such 
designs significantly aid training of overparameterized models like deep neural networks (e.g., \cite{bengio2012deep,shin2016deep,long2017deep}), which are inherently challenging due to the vast number of parameters compared to the number of training data examples. There are various ways to integrate the previously-learned information from the source task in the learning process of the target task; often this is done by taking subsets of parameters (e.g., layers in neural networks) learned for the source task and plugging them in the target task model as parameter subsets that can be set fixed, finely tuned, or serve as  non-random initialization for a thorough learning process. Obviously, transfer learning is useful only if the source and target tasks are sufficiently related with respect to the transfer mechanism utilized, e.g., \cite{rosenstein2005transfer,zamir2018taskonomy,kornblith2019better}.
Moreover, finding a successful transfer learning setting for deep neural networks was shown in \cite{raghu2019transfusion} to be a delicate engineering task.
The importance of transfer learning in contemporary practice should motivate fundamental understanding of its main aspects via \textit{analytical} frameworks that may consider linear structures (see, e.g., \cite{lampinen2018analytic}).

In general, the impressive success of overparameterized architectures for supervised learning have raised fundamental questions on the classical role of the bias-variance tradeoff that guided the traditional designs towards seemingly-optimal underparameterized models \cite{breiman1983many}. Empirical studies from recent years  \cite{spigler2018jamming,geiger2019scaling,belkin2019reconciling} have demonstrated the phenomenon that overparameterized supervised learning corresponds to a generalization error curve with a \textit{double descent} trend (with respect to the number of parameters in the learned model). This double descent shape means that the generalization error peaks when the learned model starts to interpolate the training data (i.e., to achieve zero training error), but then the error continuously decreases as the overparmeterization increases, often arriving to a global minimum that outperforms the best underparameterized solution. This phenomenon has been studied theoretically from the linear regression perspective in an extensive series of papers, e.g.,  in \cite{belkin2020two,hastie2019surprises,xu2019number,mei2019generalization,bartlett2020benign,muthukumar2020harmless}. The next stage is to provide corresponding fundamental understanding to learning problems beyond a single fully-supervised regression problem (see, for example, the study in \cite{dar2020subspace} on overparameterized linear subspace fitting in unsupervised and semi-supervised settings).

In this paper we study the fundamentals of the natural meeting point between {\em overparameterized models} and the \textit{transfer learning} concept. 
Our analytical framework is based on the least squares solutions to two related linear regression problems: the first is a source task whose solution has been found independently, and the second is a target task that is addressed using the solution already available for the source task. 
Specifically, the target task is carried out while keeping a subset of its parameters fixed to values transferred from the source task solution. Accordingly, the target task includes three types of parameters: free to-be-optimized parameters, transferred parameters set fixed to values from the source task, and parameters fixed to zeros (which in our case correspond to the elimination of input features). The mixture of the parameter types defines the parameterization level (i.e., the relation between the number of free parameters and the number of examples given) and the transfer-learning level (i.e., the portion of transferred parameters among the solution layout).

We conduct a non-asymptotic statistical analysis of the generalization errors in this transfer learning structure where the minimum $\ell_2$-norm solutions are used when the source and/or target tasks are overparameterized. Clearly, since the source task is solved independently, its generalization error follows a regular (one-dimensional) double descent shape with respect to the number of examples and free parameters available in the source task. Hence, our main contribution and interest are in the characterization of the generalization error of the target task that is carried out using the transfer learning approach described above. We show that the generalization error of the target task follows a double descent trend that depends on the double descent shape of the source task and on the transfer learning factors such as the number of parameters transferred and the relation between the source and target tasks. 
We also examine the generalization error of the target task as a function of two quantities: the number of free parameters in the source task and the number of free parameters in the target task. This interpretation presents the generalization error of the target task as having a \textit{two-dimensional double descent} trend. 

We also show how the generalization error of the target task is affected by the \textit{specific set} of transferred parameters and its delicate interplay with the forms of the true solution and the source-target task relation. 
By that, we provide an analytical theory to the fragile nature of successful transfer learning designs. 
We demonstrate that the practical approach of arbitrary selection of transferred parameters signifies the fragile nature of successful parameter transfer settings, especially when the true (unknown) parameters have a sparse form in the feature space. 

We characterize the settings where transferring a set of parameters is more beneficial (in the sense of improved generalization in the target task) than defining them as additional free parameters or zeroing them. 
We also prove that the transfer of parameters from an overparameterized solution of a source task is not necessarily optimal when the source task is closer or identical to the target task. 

The majority of this paper is focused on the analytical and empirical study of the minimum $\ell_2$-norm solution in our transfer learning setting. Yet, we also empirically examine the utilization of the minimum $\ell_1$-norm solution and the ridge regression approach in our transfer learning framework. 
The minimum $\ell_1$-norm (interpolating) solution also induces generalization errors that follow a double descent shape, and can outperform the minimum $\ell_2$-norm solution especially if the true parameters are sparse. 
The ridge regression method includes a regularization term that prevents interpolation and, when properly tuned, eliminates the double descent peak. Ridge regression can outperform the interpolating solutions for a wide range of overparameterization levels. However, at the proximity of maximal overparameterization, the minimum $\ell_2$-norm solution performs comparably to the ridge regression. 
This demonstrates that interpolation at extreme overparameterization levels can substitute properly tuned regularization in our transfer learning setting.

\subsection{Related Work}
Despite the prevalence of transfer learning in contemporary practice, there are only few analytical theories for this topic. First, Lampinen and Ganguli \cite{lampinen2018analytic} studied the optimization dynamics in transfer learning of multi-layer linear models, which is a different research objective than our focus on double descent phenomena. 
Interestingly, when we posted the first version of our work on arXiv in June 2020, there was no literature on double descent phenomena and interpolating solutions in transfer learning. 
Later on, Dhifallah and Lu~\cite{dhifallah2021phase} studied single-layer nonlinear models that are suitable for both regression and classification problems, where the source task is ridge regularized and therefore prevents interpolation and double descent phenomena.
Gerace at al.~\cite{gerace2021probing} studied a binary classification problem that is addressed by transfer learning of the first layer in a two-layer model that includes nonlinearities. The learning settings in \cite{gerace2021probing} include regularization on both the source and target task and therefore attenuate some of the double descent behavior. The analysis in \cite{gerace2021probing} requires numerical optimizations and empirical estimation of covariance matrices as inputs for their asymptotic formulations. 

Remotely from overparamterized learning but related to transfer learning theory, Obst et al.~\cite{obst2021transfer} studied fine-tuning (based on gradient descent) between linear regression problems in underparameterized settings. 

To summarize, each of the existing transfer learning theories employs different analytical tools, assumptions, and accordingly presents a unique analytical perspective on the topic. 
We focus on transfer learning between linear regression problems for Gaussian data and without regularization. This allows us to consider the minimum $\ell_2$-norm solution and explicitly characterize the generalization error and its double descent behavior in a detailed, completely closed-form formulation that is unavailable elsewhere. This also contributes to our discussion on beneficial transfer learning and lets us to analytically characterize the optimal source task to transfer from. 	


\subsection{Paper Organization}
This paper is organized as follows. 
In Section \ref{sec:Transfer Learning in the Linear Regression Case: Problem Definition} we define the transfer learning architecture examined in this paper. In Section \ref{sec:The Double Descent Phenomenon in Transfer Learning} we analytically characterize the double descent phenomenon in our transfer learning setting. 
In Section \ref{sec:When is Transfer Learning Beneficial} we analyze the conditions for beneficial transfer of parameters compared to the alternatives of free and zeroed parameters. 
In Section \ref{sec:Additional Insights The Optimal H in a Componentwise Task Relation} we characterize the optimal source task to transfer from. 
Sections \ref{sec:The Double Descent Phenomenon in Transfer Learning}-\ref{sec:Additional Insights The Optimal H in a Componentwise Task Relation} are focused on the minimum $\ell_2$-norm solution in the overparameterized regime of our transfer learning setting; in Section \ref{sec:Additional Linear Regression Methods} we examine transfer learning in conjunction with the minimum $\ell_1$-norm solution and the ridge regression method.
Section \ref{sec:Conclusion} concludes the paper. The Supplementary Materials include all of the proofs and mathematical developments as well as additional details and results for the empirical part of the paper. 


\section{Transfer Learning between Linear Regression Tasks: Problem Definition}
\label{sec:Transfer Learning in the Linear Regression Case: Problem Definition}

\subsection{Source Task: Data Model and Solution Form}
We start with the \textit{source} data model, where a $d$-dimensional Gaussian input vector $\vec{z}\sim \mathcal{N}\left(\vec{0},\mtx{I}_d\right)$ is connected to a response value $v\in\mathbb{R}$ via the noisy linear model 
\begin{equation}
	\label{eq:source data model}
	v = \vec{z}^T \vecgreek{\theta} + \xi, 
\end{equation}
where $\xi\sim\mathcal{N}\left(0,\sigma_{\xi}^2\right)$ is a Gaussian noise component independent of $\vec{z}$, $\sigma_{\xi}>0$, and $\vecgreek{\theta}\in\mathbb{R}^d$ is an unknown vector.  
The data user is unfamiliar with the distribution of $\left(\vec{z},v\right)$, however gets a dataset of $\widetilde{n}$ independent and identically distributed (i.i.d.) draws of $\left(\vec{z},v\right)$ pairs denoted as ${\widetilde{\mathcal{D}}\triangleq\Big\{ { \left({\vec{z}^{(i)},v^{(i)}}\right) }\Big\}_{i=1}^{\widetilde{n}}}$. The $\widetilde{n}$ data samples can be rearranged as ${\mtx{Z}\triangleq \lbrack {\vec{z}^{(1)}, \dots, \vec{z}^{(\widetilde{n})}} \rbrack^{T}}$ and $\vec{v}\triangleq \lbrack{ v^{(1)}, \dots, v^{(\widetilde{n})} } \rbrack^{T}$ that satisfy the relation  ${\vec{v} = \mtx{Z} \vecgreek{\theta} + \vecgreek{\xi}}$ where ${\vecgreek{\xi}\triangleq \lbrack{ {\xi}^{(1)}, \dots, {\xi}^{(\widetilde{n})} } \rbrack^{T}}$ is an unknown noise vector that its $i^{\rm th}$ component ${\xi}^{(i)}$ participates in the relation ${v^{(i)} = \vec{z}^{{(i)},T} \vecgreek{\theta} + \xi^{(i)}}$ underlying the $i^{\rm th}$ data sample. 

The \textit{source} task is defined for a new (out of sample) data pair $\left( { \vec{z}^{(\rm test)}, v^{(\rm test)} } \right)$ drawn from the distribution induced by (\ref{eq:source data model}) independently of the $\widetilde{n}$ examples in $\widetilde{\mathcal{D}}$. For a given $\vec{z}^{(\rm test)}$, the source task is to estimate the response value $v^{(\rm test)}$ by the value $\widehat{v}$ that minimizes the corresponding out-of-sample squared error (i.e., the generalization error of the \textit{source} task) 
\begin{equation}
	\label{eq:out of sample error - source data class}
	\widetilde{\mathcal{E}}_{\rm out} \triangleq \expectation{ \left( \widehat{v} - v^{(\rm test)} \right)^2  } =   \sigma_{\xi}^2 + \expectation{ \left \Vert { \widehat{\vecgreek{\theta}} - \vecgreek{\theta} } \right \Vert _2^2 } 
\end{equation}
where the second equality stems from the data model in (\ref{eq:source data model}) and the corresponding linear form of ${\widehat{v} = \vec{z}^{({\rm test}),T}\widehat{\vecgreek{\theta}}}$ where $\widehat{\vecgreek{\theta}}$ estimates $\vecgreek{\theta}$ based on $\widetilde{\mathcal{D}}$. 

To address the source task based on the $\widetilde{n}$ examples, one should choose the number of free parameters in the estimate $\widehat{\vecgreek{\theta}}\in \mathbb{R}^{d} $. Consider a predetermined layout where $\widetilde{p}$ out of the $d$ components of $\widehat{\vecgreek{\theta}}$ are free to be optimized, whereas the remaining $d-\widetilde{p}$ components are constrained to zero values. The coordinates of the free parameters are specified in the set ${\mathcal{S}\triangleq\lbrace { s_1, \dots, s_{\widetilde{p}} }\rbrace}$ where ${1 \le s_1 < \dots < s_{\widetilde{p}} \le d}$ and the complementary set ${\mathcal{S}^{\rm c} \triangleq \lbrace{1,\dots,d}\rbrace \setminus \mathcal{S}}$ contains the coordinates constrained to be zero valued. 
We define the $\rvert{\mathcal{S}}\lvert\times d$ matrix $\mtx{Q}_{\mathcal{S}}$ as the linear operator that extracts from a $d$-dimensional vector its $\rvert{\mathcal{S}}\lvert$-dimensional subvector of components residing at the coordinates specified in $\mathcal{S}$. Specifically, the values of the $\left (k, s_k \right) $ components ($k=1,\dots,\rvert{\mathcal{S}}\lvert$) of $\mtx{Q}_{\mathcal{S}}$ are ones and the other components of $\mtx{Q}_{\mathcal{S}}$ are zeros. The definition given here for $\mtx{Q}_{\mathcal{S}}$ can be adapted also to other sets of coordinates (e.g., $\mtx{Q}_{\mathcal{S}^{\rm c}}$ for $\mathcal{S}^{\rm c}$) as denoted by the subscript of $\mtx{Q}$. 
We now turn to formulate the \textit{source} task using the linear regression form of 
\begin{align} 
\label{eq:constrained linear regression - source data class}
\widehat{\vecgreek{\theta}} = \argmin_{\vec{r}\in\mathbb{R}^{d}} \left \Vert  \vec{v} - \mtx{Z}\vec{r} \right \Vert _2^2
~~~\text{subject to}~~\mtx{Q}_{\mathcal{S}^{\rm c}} \vec{r} = \vec{0}
\end{align}
that its minimum $\ell_2$-norm solution (see details in Appendix \ref{appendix:subsec:Mathematical Development of Estimate of Theta}) is
\begin{equation}
\label{eq:constrained linear regression - solution - source data class}
{\widehat{\vecgreek{\theta}}=\mtx{Q}_{\mathcal{S}}^T \mtx{Z}_{\mathcal{S}}^{+} \vec{v}}
\end{equation}
where $\mtx{Z}_{\mathcal{S}}^{+}$ is the pseudoinverse of $\mtx{Z}_{\mathcal{S}}\triangleq \mtx{Z} \mtx{Q}_{\mathcal{S}}^{T}$. Note that $\mtx{Z}_{\mathcal{S}}$ is a $\widetilde{n} \times \widetilde{p}$ matrix that its $i^{\rm th}$ row is formed by the $\widetilde{p}$ components of $\vec{z}^{(i)}$ specified by the coordinates in $\mathcal{S}$, namely, only $\widetilde{p}$ out of the $d$ features of the input data vectors are utilized. 
Hence, in the underparameterized case of ${\widetilde{p}\le\widetilde{n}}$ the solution (\ref{eq:constrained linear regression - solution - source data class}) almost surely reduces to the unique least squares form of 
\begin{equation}
\label{eq:constrained linear regression - underparameterized unique solution - source data class}
{\widehat{\vecgreek{\theta}}=\mtx{Q}_{\mathcal{S}}^T \left(\mtx{Z}_{\mathcal{S}}^{T} \mtx{Z}_{\mathcal{S}}\right)^{-1} \mtx{Z}_{\mathcal{S}}^{T} \vec{v}}.
\end{equation}
Moreover, in both (\ref{eq:constrained linear regression - solution - source data class})-(\ref{eq:constrained linear regression - underparameterized unique solution - source data class}), $\widehat{\vecgreek{\theta}}$ is a $d$-dimensional vector that may have nonzero values only in the $\widetilde{p}$ coordinates specified in $\mathcal{S}$ (this can be easily observed by noting that for an arbitrary ${\vec{w}\in\mathbb{R}^{\rvert{\mathcal{S}}\lvert}}$, the vector ${\vec{u}=\mtx{Q}_{\mathcal{S}}^T\vec{w}}$ is a $d$-dimensional vector that its components satisfy ${u_{s_k}=w_k}$ for ${k=1,...,\rvert{\mathcal{S}}\lvert}$ and ${u_j=0}$ for $j\notin \mathcal{S}$). 
While the specific optimization form in (\ref{eq:constrained linear regression - source data class}) was not explicit in previous studies of non-asymptotic settings, e.g., \cite{breiman1983many,belkin2020two}, the solution in (\ref{eq:constrained linear regression - solution - source data class}) coincides with theirs and, thus, the formulation of the generalization error of our \textit{source} task (which is a linear regression problem that, by itself, does not have any transfer learning aspect) is available from \cite{breiman1983many,belkin2020two} and provided in Appendix \ref{appendix:subsec:Double Descent Formulation for Source Task} in our notations for completeness of presentation.

\subsection{Target Task: Data Model and Solution using Transfer Learning}
\label{subsec: Target Task}

A second data class, which is our main interest, is modeled by  $\left(\vec{x},y\right)\in\mathbb{R}^{d}\times\mathbb{R}$ that satisfy 
\begin{equation}
	\label{eq:target data model}
	y = \vec{x}^T \vecgreek{\beta} + \epsilon
\end{equation}
where ${\vec{x}\sim \mathcal{N}\left(\vec{0},\mtx{I}_d\right)}$ is a Gaussian input vector including $d$ features, ${\epsilon\sim\mathcal{N}\left(0,\sigma_{\epsilon}^2\right)}$ is a Gaussian noise component independent of $\vec{x}$, $\sigma_{\epsilon}>0$, and ${\vecgreek{\beta}\in\mathbb{R}^d}$ is an unknown vector related to the $\vecgreek{\theta}$ from (\ref{eq:source data model}) via 
\begin{equation}
	\label{eq:theta-beta relation}
	\vecgreek{\theta} = \mtx{H}\vecgreek{\beta} + \vecgreek{\eta}
\end{equation}
where ${\mtx{H}\in\mathbb{R}^{d\times d}}$ is a deterministic matrix and ${\vecgreek{\eta}\sim\mathcal{N}\left(\vec{0},\sigma_{\eta}^2\mtx{I}_d\right)}$ is a Gaussian noise vector with ${\sigma_{\eta}\ge 0}$. Here $\vecgreek{\eta}$, $\vec{x}$, $\epsilon$, $\vec{z}$ and $\xi$ are independent. 
The data user does not know the distribution of $\left(\vec{x},y\right)$ but receives a small dataset of $n$ i.i.d.\ draws of ${\left(\vec{x},y\right)}$ pairs denoted as ${\mathcal{D}\triangleq\Big\lbrace { \left(\vec{x}^{(i)},y^{(i)}\right) }\Big\rbrace_{i=1}^{n}}$. The $n$ data samples can be organized in a $n\times d$ matrix of input variables ${\mtx{X}\triangleq \lbrack {\vec{x}^{(1)}, \dots, \vec{x}^{(n)}} \rbrack^{T}}$ and a $n\times 1$ vector of responses ${\vec{y}\triangleq \lbrack{ y^{(1)}, \dots, y^{(n)} } \rbrack^{T}}$ that together satisfy the relation  ${\vec{y} = \mtx{X} \vecgreek{\beta} + \vecgreek{\epsilon}}$ where ${\vecgreek{\epsilon}\triangleq \lbrack{ {\epsilon}^{(1)}, \dots, {\epsilon}^{(n)} } \rbrack^{T}}$ is an unknown noise vector that its $i^{\rm th}$ component ${\epsilon}^{(i)}$ is involved in the connection ${y^{(i)} = \vec{x}^{{(i)},T} \vecgreek{\beta} + \epsilon^{(i)}}$ underlying the $i^{\rm th}$ example pair.

The \textit{target} task considers a new (out of sample) data pair ${\left( { \vec{x}^{(\rm test)}, y^{(\rm test)} } \right)}$ drawn from the model in (\ref{eq:target data model}) independently of the training examples in $\mathcal{D}$. Given $\vec{x}^{(\rm test)}$, the goal is to establish an estimate $\widehat{y}$ of the response value ${y^{(\rm test)}}$ such that the out-of-sample squared error, i.e., the generalization error of the \textit{target} task,  
\begin{equation}
	\label{eq:out of sample error - target data class - beta form}
	\mathcal{E}_{\rm out} \triangleq \expectation{ \left( \widehat{y} - y^{(\rm test)} \right)^2 } = \sigma_{\epsilon}^2 + \expectation{ \left \Vert { \widehat{\vecgreek{\beta}} - \vecgreek{\beta} } \right \Vert _2^2 }
\end{equation}
is minimized, where ${\widehat{y} = \vec{x}^{({\rm test}),T}\widehat{\vecgreek{\beta}}}$, and the second equality stems from the data model in (\ref{eq:target data model}). 

The target task is addressed via linear regression that seeks for an estimate ${\widehat{\vecgreek{\beta}}\in\mathbb{R}^{d}}$ with a layout including three disjoint sets of coordinates ${\mathcal{F}, \mathcal{T}, \mathcal{Z}}$ that satisfy ${\mathcal{F}\cup \mathcal{T} \cup \mathcal{Z} = \{{1,\dots,d}\}}$ and correspond to three types of parameters:
\begin{itemize}
	\item $p$ parameters are \textit{free} to be optimized and their coordinates are specified in $\mathcal{F}$.
	\item $t$ parameters are \textit{transferred} from the co-located coordinates of the estimate $\widehat{\vecgreek{\theta}}$ already formed for the source task. 
	Only the free parameters of the \textit{source} task are relevant to be transferred to the target task and, therefore, $\mathcal{T}\subseteq\mathcal{S}$ and $t\in\{0,\dots,\widetilde{p}\}$. The transferred parameters are taken as is from $\widehat{\vecgreek{\theta}}$ and set fixed in the corresponding coordinates of $\widehat{\vecgreek{\beta}}$, i.e., for $k\in\mathcal{T}$, ${ \widehat{{\beta}}_{k} = \widehat{{\theta}}_{k} }$ where $\widehat{{\beta}}_{k}$ and $\widehat{{\theta}}_{k}$ are the $k^{\rm th}$ components of $\widehat{\vecgreek{\beta}}$ and $\widehat{\vecgreek{\theta}}$, respectively. 
	\item $\ell$ parameters are set to \textit{zeros}. Their coordinates are included in ${\mathcal{Z}}$ and effectively correspond to ignoring features in the same coordinates of the input vectors.
\end{itemize}
Clearly, the layout should satisfy $p+t+\ell = d$. 
Then, the constrained linear regression problem for the target task is formulated as 
\begin{align} 
	\label{eq:constrained linear regression - target task}
	&\widehat{\vecgreek{\beta}} = \argmin_{\vec{b}\in\mathbb{R}^{d}} \left \Vert  \vec{y} - \mtx{X}\vec{b} \right \Vert _2^2
	\\ \nonumber
	&\mathmakebox[5em][l]{\text{subject to}}\mtx{Q}_{\mathcal{T}} \vec{b} = \mtx{Q}_{\mathcal{T}}\widehat{\vecgreek{\theta}}
	\\ \nonumber
	&\qquad\qquad\quad\mtx{Q}_{\mathcal{Z}} \vec{b} = \vec{0}
\end{align}
where $\mtx{Q}_{\mathcal{T}}$ and $\mtx{Q}_{\mathcal{Z}}$ are the linear operators extracting the subvectors corresponding to the coordinates in ${\mathcal{T}}$ and ${\mathcal{Z}}$, respectively, from $d$-dimensional vectors.
Here $\widehat{\vecgreek{\theta}}\in \mathbb{R}^{d}$ is the \textit{precomputed} estimate for the source task and considered a constant vector for the purpose of the target task. The examined transfer learning structure includes a single computation of the source task (\ref{eq:constrained linear regression - source data class}), followed by a single computation of the target task (\ref{eq:constrained linear regression - target task}) that produces the eventual estimate of interest $\widehat{\vecgreek{\beta}}$ using the given $\widehat{\vecgreek{\theta}}$. 
The minimum $\ell_2$-norm solution of the target task in (\ref{eq:constrained linear regression - target task}) is (see details in Appendix \ref{appendix:subsec:Mathematical Development of Estimate of Beta}) 
\begin{equation}
	\label{eq:constrained linear regression - solution - target task}
	{\widehat{\vecgreek{\beta}}=\mtx{Q}_{\mathcal{F}}^T \mtx{X}_{\mathcal{F}}^{+} \left( \vec{y} - \mtx{X}_{\mathcal{T}} \widehat{\vecgreek{\theta}}_{\mathcal{T}} \right)} + \mtx{Q}_{\mathcal{T}}^T\widehat{\vecgreek{\theta}}_{\mathcal{T}}
\end{equation}
where $\widehat{\vecgreek{\theta}}_{\mathcal{T}} \triangleq \mtx{Q}_{\mathcal{T}}\widehat{\vecgreek{\theta}}$, $\mtx{X}_{\mathcal{T}}\triangleq \mtx{X} \mtx{Q}_{\mathcal{T}}^T$, and $\mtx{X}_{\mathcal{F}}^{+}$ is the pseudoinverse of $\mtx{X}_{\mathcal{F}}\triangleq \mtx{X} \mtx{Q}_{\mathcal{F}}^T$. 
In the underparameterized case of ${p\le n}$ the solution (\ref{eq:constrained linear regression - solution - target task}) almost surely reduces to the unique least squares form of 
\begin{equation}
\label{eq:constrained linear regression - underparameterized unique solution - target task}
{\widehat{\vecgreek{\beta}}=\mtx{Q}_{\mathcal{F}}^T \left(\mtx{X}_{\mathcal{F}}^{T}\mtx{X}_{\mathcal{F}}\right)^{-1}\mtx{X}_{\mathcal{F}}^{T} \left( \vec{y} - \mtx{X}_{\mathcal{T}} \widehat{\vecgreek{\theta}}_{\mathcal{T}} \right) + \mtx{Q}_{\mathcal{T}}^T\widehat{\vecgreek{\theta}}_{\mathcal{T}}}.
\end{equation}

Note that the desired layout is indeed implemented by the $\widehat{\vecgreek{\beta}}$ forms in (\ref{eq:constrained linear regression - solution - target task}), (\ref{eq:constrained linear regression - underparameterized unique solution - target task}): the components corresponding to $\mathcal{Z}$ are zeros, the components corresponding to $\mathcal{T}$ are taken as is from $\widehat{\vecgreek{\theta}}$, and only the $p$ coordinates specified in $\mathcal{F}$ are adjusted for the purpose of minimizing the in-sample error in the optimization cost of (\ref{eq:constrained linear regression - target task}) while considering the transferred parameters. 
In this paper we study the generalization ability of overparameterized solutions (i.e., when $p>n$) to the \textit{target} task formulated in (\ref{eq:constrained linear regression - target task}): In Sections \ref{sec:The Double Descent Phenomenon in Transfer Learning}-\ref{sec:Additional Insights The Optimal H in a Componentwise Task Relation} we analyze the minimum $\ell_2$-norm solution (\ref{eq:constrained linear regression - solution - target task}) and in Section \ref{sec:Additional Linear Regression Methods} we examine other solutions.

\section{The Double Descent Phenomenon in Transfer Learning}
\label{sec:The Double Descent Phenomenon in Transfer Learning}

\subsection{Analytical Characterization of the Double Descent Phenomenon}
\label{subsec:Analytical Characterization of the Double Descent Phenomenon}

Consider an \textit{overall layout of coordinate subsets} ${\mathcal{L}\triangleq \lbrace{ \mathcal{S}, \mathcal{F}, \mathcal{T}, \mathcal{Z} }\rbrace}$.
The following theorem formulates the generalization error of the target task that is solved using a set of $t$ parameters that are transferred as is from their co-located coordinates (indicated by $\mathcal{T}$) at the source task solution.
\begin{theorem}
	\label{theorem:out of sample error - target task - specific layout}
	Let ${\mathcal{L} = \lbrace{ \mathcal{S}, \mathcal{F}, \mathcal{T}, \mathcal{Z} }\rbrace}$ be a deterministic (i.e., non-random) coordinate layout. 
	Then, the out-of-sample error of the target task has the form of 
	\begin{align}
		\label{eq:out of sample error - target task - theorem - general decomposition form - specific layout}
		&\mathcal{E}_{\rm out}^{(\mathcal{L})}  = \begin{cases}
			\mathmakebox[25em][l]{\frac{n-1}{n-p-1}\left( \Ltwonorm{\vecgreek{\beta}_{\mathcal{Z}}} +  \sigma_{\epsilon}^{2} + \mathcal{E}_{\rm transfer}^{(\mathcal{T},\mathcal{S})}\right) }  \text{for } p \le n-2,   
			\\
			\mathmakebox[25em][l]{\infty} \text{for } n-1 \le p \le n+1,
			\\
			\mathmakebox[25em][l]{\frac{p-1}{p-n-1} \left( \Ltwonorm{\vecgreek{\beta}_{\mathcal{Z}}} +  \sigma_{\epsilon}^{2} + \mathcal{E}_{\rm transfer}^{(\mathcal{T},\mathcal{S})}\right)+ \left({1-\frac{n}{p}}\right)\Ltwonorm{\vecgreek{\beta}_{\mathcal{F}}}} \text{for } p \ge n+2,
		\end{cases}
	\end{align}		
	where 
	\begin{align}
		\label{eq:out of sample error - target task - theorem - transfer term - general form - specific layout}
		&\mathcal{E}_{\rm transfer}^{(\mathcal{T},\mathcal{S})} \triangleq    
		\underbrace{\Ltwonorm{\expectation{\widehat{\vecgreek{\theta}}_{\mathcal{T}}} - \vecgreek{\beta}_{\mathcal{T}}}}_{\text{transfer bias}} + 
		\underbrace{\expectation{\Ltwonorm{\widehat{\vecgreek{\theta}}_{\mathcal{T}} - \expectation{\widehat{\vecgreek{\theta}}_{\mathcal{T}}} }}}_{\text{transfer variance}}
	\end{align}	
	is the error in the transferred coordinates in the target task solution. 
\end{theorem}
The last theorem is proved using non-asymptotic properties of Wishart matrices (see Appendix \ref{appendix:sec:Transfer of Specific Sets of Parameters: Proofs}). 

The error formulation in (\ref{eq:out of sample error - target task - theorem - general decomposition form - specific layout}) expresses a double descent form in terms of $p$ and $n$, similarly to the formulation given in \cite{belkin2020two} for a linear regression task without transfer learning. However, here (\ref{eq:out of sample error - target task - theorem - general decomposition form - specific layout}) introduces a new term, $\mathcal{E}_{\rm transfer}^{(\mathcal{T},\mathcal{S})}$, that encapsulates the transfer learning aspect of our setting. 

The formulation of the transfer error term in (\ref{eq:out of sample error - target task - theorem - transfer term - general form - specific layout}) demonstrates a bias-variance decomposition where the bias in the transferred parameters is measured with respect to the true parameters of the \textit{target} task, and the variance depends only on the source task. 
This bias-variance decomposition is affected by the coordinates of the transferred parameters (i.e., $\mathcal{T}$) and also indirectly (through $\widehat{\vecgreek{\theta}}_{\mathcal{T}}$) by the coordinates of the free parameters in the source task (i.e., $\mathcal{S}$), as we will see next.

The following corollary explicitly formulates the transfer bias and variance terms and their detailed dependency on the \textit{parameterization level} of the \textit{source} task, i.e., the $\widetilde{p},\widetilde{n}$ pair, and the set $\mathcal{S}$ of free parameters in the source task  (see proof in Appendix \ref{appendix:subsec:Proof of Corollary 6}).  
\begin{corollary}
	\label{corollary:out of sample error - target task - specific layout - detailed - specific layout}
	The transfer bias term from (\ref{eq:out of sample error - target task - theorem - transfer term - general form - specific layout}) can be written as 
	\begin{equation}
	{\rm{Bias}}_{\mathcal{T}}^{2} \triangleq  \Ltwonorm{\expectation{\widehat{\vecgreek{\theta}}_{\mathcal{T}}} - \vecgreek{\beta}_{\mathcal{T}}} =  \Ltwonorm{\mtx{Q}_{\mathcal{T}}\left({r\mtx{H}-\mtx{I}_{d}}\right)\vecgreek{\beta} }
	~~~\text{where}~~ r \triangleq  \begin{cases}	
	\mathmakebox[2em][l]{ 1 }    \text{for }  \widetilde{p} \le \widetilde{n},  
	\\
	\mathmakebox[2em][l]{ \frac{\widetilde{n}}{\widetilde{p}} }    \text{for }  \widetilde{p} > \widetilde{n}. 
	\end{cases} 
	\label{eq:out of sample error - target task - corollary - transfer bias - detailed - specific layout}
	\end{equation}	
	The transfer variance term from (\ref{eq:out of sample error - target task - theorem - transfer term - general form - specific layout}) can be formulated as 
	
	\begin{align}
	&{\rm{Var}}_{\mathcal{T},\mathcal{S}} \triangleq \expectation{\Ltwonorm{\widehat{\vecgreek{\theta}}_{\mathcal{T}} - \expectation{\widehat{\vecgreek{\theta}}_{\mathcal{T}}} } } 
	\nonumber\\	\nonumber 
	&=  \begin{cases}	
	\mathmakebox[26em][l]{ t\left( \sigma_{\eta}^2 + \frac{ \zeta_{\mathcal{S}^{c}} + \left(d-\widetilde{p}\right)\sigma_{\eta}^2 + \sigma_{\xi}^2 }{\widetilde{n} - \widetilde{p} - 1 } \right)}    \text{for }  1\le\widetilde{p} \le \widetilde{n}-2,  
	\\
	\mathmakebox[26em][l]{\infty} \text{for } \widetilde{n}-1 \le \widetilde{p} \le \widetilde{n}+1,
	\\
	\mathmakebox[26em][l]{\frac{\widetilde{n}}{\widetilde{p}}  \left( 
		\frac{ \left({\widetilde{p}-\widetilde{n}}\right)t \zeta_{\mathcal{S}\setminus\mathcal{T}} +  	 \left( \left({\widetilde{p}-\widetilde{n}}\right)t - 1 + \frac{\widetilde{n}}{\widetilde{p}}\right) \zeta_{\mathcal{T}} }{\widetilde{p}^2 - 1} + 	t\left( \sigma_{\eta}^2 + \frac{ \zeta_{\mathcal{S}^{c}}  + \left(d-\widetilde{p}\right)\sigma_{\eta}^2 + \sigma_{\xi}^2 }{\widetilde{p} - \widetilde{n} - 1 } \right) \right) }  \text{for } \widetilde{p} \ge \widetilde{n}+2, 
	\end{cases}
	\\
	\label{eq:out of sample error - target task - corollary - transfer variance - detailed - specific layout}
	\end{align}	
	where ${\zeta_{\mathcal{T}}\triangleq\Ltwonorm{ \mtx{Q}_{\mathcal{T}}\mtx{H}\vecgreek{\beta}}}$, ${\zeta_{\mathcal{S}\setminus\mathcal{T}}\triangleq\Ltwonorm{ \mtx{Q}_{\mathcal{S}\setminus\mathcal{T}}\mtx{H}\vecgreek{\beta}}}$, and  ${\zeta_{\mathcal{S}^{c}}\triangleq\Ltwonorm{ \mtx{Q}_{\mathcal{S}^{c}}\mtx{H}\vecgreek{\beta}}}$. 
\end{corollary}
Note that the out-of-sample error formulation in (\ref{eq:out of sample error - target task - theorem - general decomposition form - specific layout}) depends on the parameterization level of the \textit{target} task (i.e., the $p,n$ pair) and also on the parameterization level of the \textit{source} task (i.e., the $\widetilde{p},\widetilde{n}$ pair) via the transfer bias and variance terms that are formulated in (\ref{eq:out of sample error - target task - corollary - transfer bias - detailed - specific layout})-(\ref{eq:out of sample error - target task - corollary - transfer variance - detailed - specific layout}).  
The formulations in Corollary \ref{corollary:out of sample error - target task - specific layout - detailed - specific layout} imply that the target error peaks not only around $p=n$, but also around $\widetilde{p}=\widetilde{n}$ when transfer learning is applied ($t>0$). This induces the \textit{double double descent} behavior that will be demonstrated in the next subsection. Later, in Sections \ref{sec:When is Transfer Learning Beneficial}-\ref{sec:Additional Insights The Optimal H in a Componentwise Task Relation} we will analyze the formulations in Theorem \ref{theorem:out of sample error - target task - specific layout} and Corollary \ref{corollary:out of sample error - target task - specific layout - detailed - specific layout} in more detail and characterize the conditions for beneficial transfer learning.

\subsection{On-Average Analysis of Arbitrarily Selected Parameters}
\label{subsec:On-Average Analysis of Arbitrarily Selected Parameters}

In this subsection, we consider the \textit{overall layout of coordinate subsets} ${\mathcal{L}\triangleq \lbrace{ \mathcal{S}, \mathcal{F}, \mathcal{T}, \mathcal{Z} }\rbrace}$ as a random structure. This will let us to formulate the expected value (with respect to $\mathcal{L}$) of the generalization error of interest. 
The simplified setting in this subsection provides useful insights towards Section \ref{subsec:The Fragile Nature of Transfer Learning: Analysis of a Single Layout of Arbitrarily Selected Parameters} where we return to analyze the transfer of a set of parameters which is induced by a single layout ${\mathcal{L}}$ (i.e., in Section \ref{subsec:The Fragile Nature of Transfer Learning: Analysis of a Single Layout of Arbitrarily Selected Parameters} we will return to use Theorem \ref{theorem:out of sample error - target task - specific layout} and Corollary \ref{corollary:out of sample error - target task - specific layout - detailed - specific layout} where there is no expectation over a random $\mathcal{L}$). 

For given $d$, $\widetilde{p}$, $p$, and $t$, we consider a uniform distribution of the coordinate layout $\mathcal{L}$ which is defined as follows. 
\begin{definition}
	\label{definition:coordinate subset layout - uniformly distributed}
	A coordinate subset layout ${\mathcal{L} = \lbrace{ \mathcal{S}, \mathcal{F}, \mathcal{T}, \mathcal{Z} }\rbrace}$ that is ${\lbrace{ \widetilde{p}, p, t }\rbrace}$-uniformly distributed, for  ${\widetilde{p}\in \lbrace{1,\dots,d}\rbrace}$ and ${\left( {p,t}\right)\in \left\{{0,\dots,d}\right\}\times \left\{{0,\dots,\widetilde{p}}\right\}}$ such that ${p+t\le d}$, satisfies: 
	$\mathcal{S}$ is uniformly chosen at random from all the subsets of $\widetilde{p}$ unique coordinates of $\{{1,\dots,d}\}$. 
	Given $\mathcal{S}$, the target-task coordinate layout ${\lbrace{ \mathcal{F}, \mathcal{T}, \mathcal{Z} }\rbrace}$ is uniformly chosen at random from all the layouts where $\mathcal{F}$, $\mathcal{T}$, $\mathcal{Z}$ are three disjoint sets of coordinates that satisfy ${\mathcal{F}\cup \mathcal{T} \cup \mathcal{Z} = \{{1,\dots,d}\}}$ such that $\rvert{\mathcal{F}}\lvert = p$, $\rvert{\mathcal{T}}\lvert = t$ and $\mathcal{T}\subseteq\mathcal{S}$, and $\rvert{\mathcal{Z}}\lvert = d-p-t$.
\end{definition}
Then, the following formulates the out-of-sample error of the target task under expectation with respect to a uniformly distributed $\mathcal{L}$ (more details are provided in Appendix \ref{appendix:subsec:proof of Corollary 4}).
\begin{corollary}
	\label{corollary:out of sample error - target task}
	Let ${\mathcal{L} = \lbrace{ \mathcal{S}, \mathcal{F}, \mathcal{T}, \mathcal{Z} }\rbrace}$ be a coordinate subset layout that is ${\lbrace{ \widetilde{p}, p, t }\rbrace}$-uniformly distributed. Then, the expected out-of-sample error of the target task has the form of 
	\begin{align}
		\label{eq:out of sample error - target task - theorem - general decomposition form}
		&\expectationwrt{ \mathcal{E}_{\rm out} }{\mathcal{L}} = 
		\begin{cases}
			\mathmakebox[23em][l]{\frac{n-1}{n-p-1}\left( (1 - \frac{p+t}{d})\Ltwonorm{\vecgreek{\beta}} + \sigma_{\epsilon}^2 + \mathcal{E}_{\rm transfer}\right)}  \text{for } p \le n-2,  
			\\
			\mathmakebox[23em][l]{\infty} \text{for } n-1 \le p \le n+1,
			\\
			\mathmakebox[23em][l]{\frac{p-1}{p-n-1} \left({ (1 - \frac{p+t}{d}) \Ltwonorm{\vecgreek{\beta}}+ \sigma_{\epsilon}^2  + \mathcal{E}_{\rm transfer} } \right) + \frac{ p - n}{d} \Ltwonorm{\vecgreek{\beta}} }  \text{for } p \ge n+2, 
		\end{cases}
	\end{align}		
	where 
	\begin{align}
		\label{eq:out of sample error - target task - theorem - transfer term - src error form}
		&\mathcal{E}_{\rm transfer} \triangleq \expectationwrt{ \mathcal{E}_{\rm transfer}^{(\mathcal{T},\mathcal{S})} }{\mathcal{L}} = \expectationwrt{\rm{Bias}_{\mathcal{T}}^{2}}{\mathcal{L}} + \expectationwrt{\rm{Var}_{\mathcal{T},\mathcal{S}}}{\mathcal{L}}
	\end{align}
		\begin{align}
	\label{eq:transfer bias - on average}
	&\expectationwrt{\rm{Bias}_{\mathcal{T}}^{2}}{\mathcal{L}} = \frac{t}{d}  \Ltwonorm{\left({r\mtx{H}-\mtx{I}_{d}}\right)\vecgreek{\beta} }
	~~~\text{where}~~ r \triangleq  \begin{cases}	
	\mathmakebox[2em][l]{ 1 }    \text{for }  \widetilde{p} \le \widetilde{n},  
	\\
	\mathmakebox[2em][l]{ \frac{\widetilde{n}}{\widetilde{p}} }    \text{for }  \widetilde{p} > \widetilde{n}. 
	\end{cases}
	\end{align}				
	\begin{align}
	&\expectationwrt{\rm{Var}_{\mathcal{T},\mathcal{S}}}{\mathcal{L}}  
=  \begin{cases}	
	\mathmakebox[22em][l]{ t\left( \sigma_{\eta}^2 + \frac{ \left(1-\frac{\widetilde{p}}{d}\right)\zeta + \left(d-\widetilde{p}\right)\sigma_{\eta}^2 + \sigma_{\xi}^2 }{\widetilde{n} - \widetilde{p} - 1 } \right)}    \text{for }  1\le\widetilde{p} \le \widetilde{n}-2,  
	\\
	\mathmakebox[22em][l]{\infty} \text{for } \widetilde{n}-1 \le \widetilde{p} \le \widetilde{n}+1,
	\\
	\mathmakebox[22em][l]{\frac{\widetilde{n}}{\widetilde{p}}t  \left( 
		\frac{ \left({\widetilde{p}-\widetilde{n}}\right) \widetilde{p}  + \frac{\widetilde{n}}{\widetilde{p}} - 1 }{d\left(\widetilde{p}^2 - 1\right)}\zeta +  \sigma_{\eta}^2 + \frac{ \left(1-\frac{\widetilde{p}}{d}\right)\zeta  + \left(d-\widetilde{p}\right)\sigma_{\eta}^2 + \sigma_{\xi}^2 }{\widetilde{p} - \widetilde{n} - 1 }  \right) }  \text{for } \widetilde{p} \ge \widetilde{n}+2, 
	\end{cases}
	\label{eq:transfer variance - on average}
	\end{align}		
	and ${\zeta\triangleq\Ltwonorm{\mtx{H}\vecgreek{\beta}}}$. 
\end{corollary}

\begin{figure}[t]
		\subfloat[{\small$\mtx{H}$:\hspace{.4em}local\hspace{.4em}averaging neighborhood size  5}]{\label{fig:H_average5_target_generalization_errors_vs_p_eta_0_p_tilde_is_d}\includegraphics[width=0.24\textwidth]{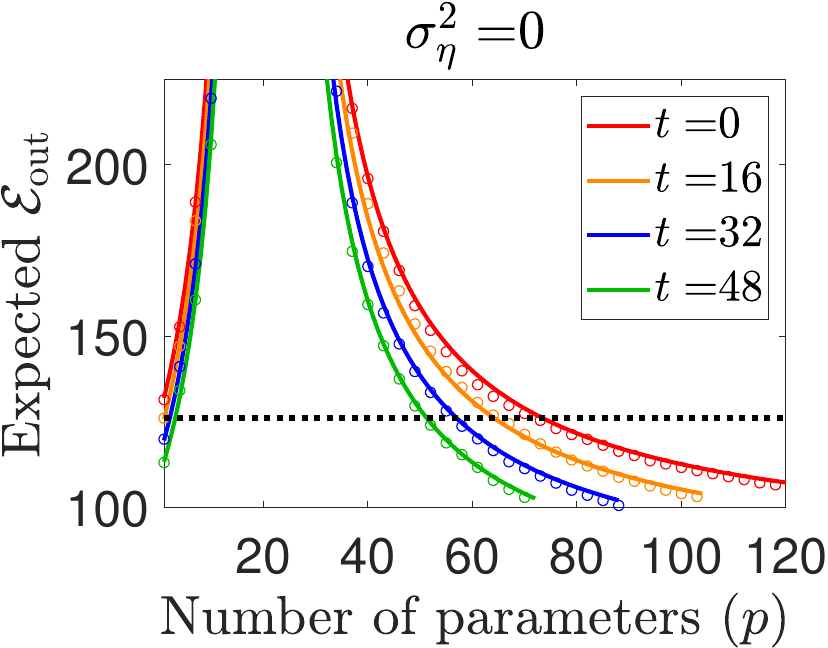}}
		~
		\subfloat[{\small$\mtx{H}$:\hspace{.4em}local\hspace{.4em}averaging neighborhood size  5}]{\label{fig:H_average5_target_generalization_errors_vs_p_eta_0.2_p_tilde_is_d}%
			\includegraphics[width=0.24\textwidth]{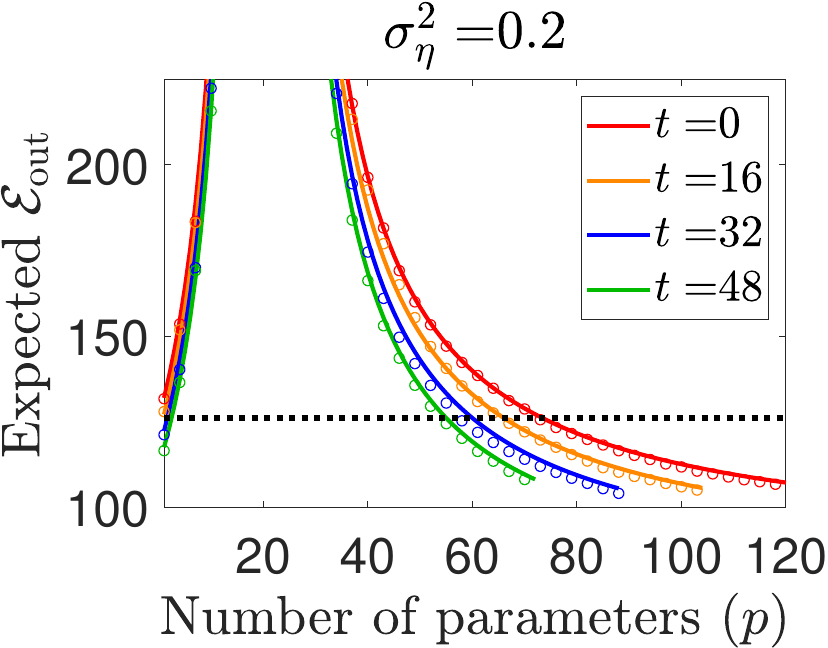}}
		~
		\subfloat[{\small$\mtx{H}$:\hspace{.4em}local\hspace{.4em}averaging neighborhood size  5}]{\label{fig:H_average5_target_generalization_errors_vs_p_eta_0.9_p_tilde_is_d}%
			\includegraphics[width=0.24\textwidth]{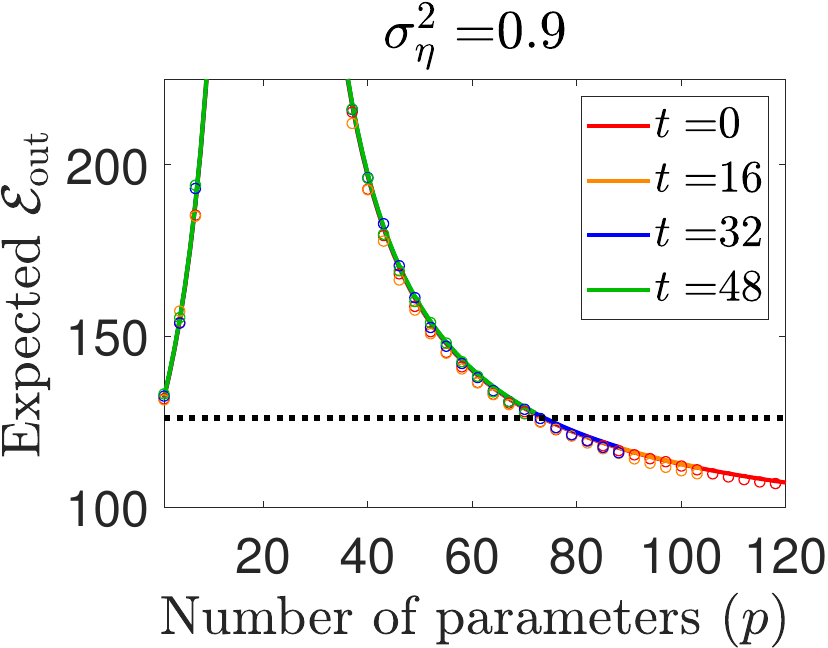}} 
		~
		\subfloat[{\small$\mtx{H}$:\hspace{.4em}local\hspace{.4em}averaging neighborhood size  5}]{\label{fig:H_average5_target_generalization_errors_vs_p_eta_2_p_tilde_120}%
			\includegraphics[width=0.24\textwidth]{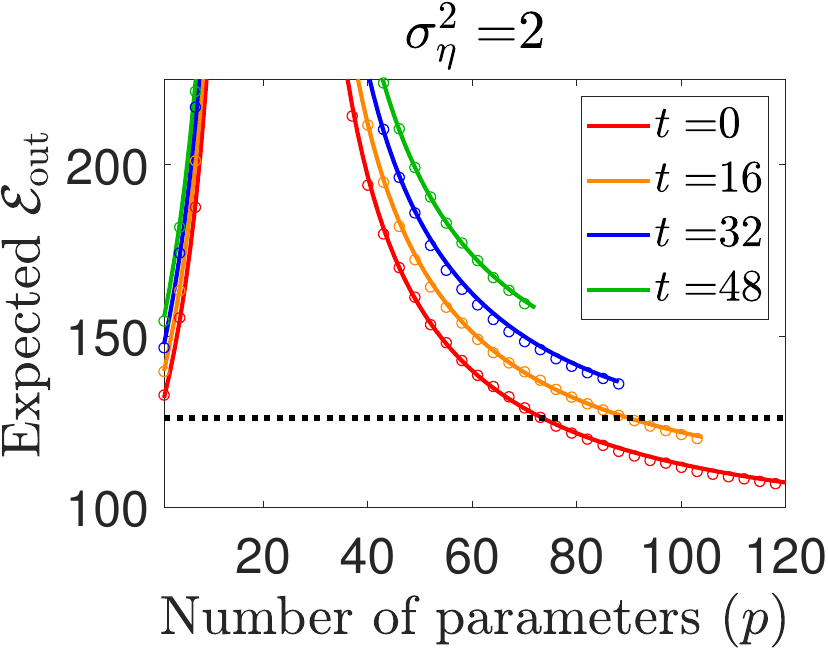}}
		\\~\\~\\
		\subfloat[{\small$\mtx{H}=\mtx{I}_d$}]{\label{fig:H_average1_target_generalization_errors_vs_p_eta_0_2_p_tilde_80}\includegraphics[width=0.24\textwidth]{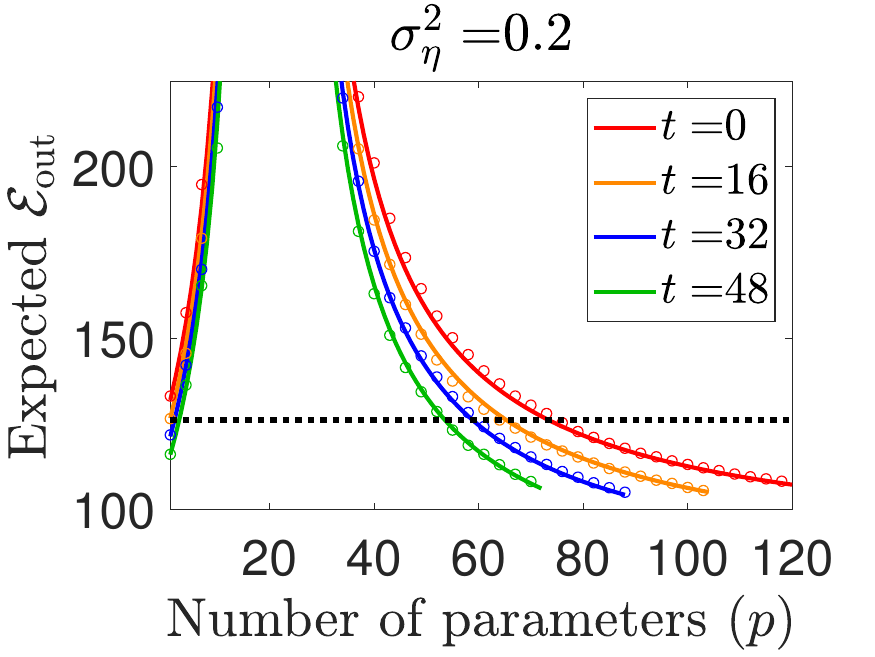}}
		~
		\subfloat[{\small$\mtx{H}$:\hspace{.4em}local\hspace{.4em}averaging neighborhood size  19}]{\label{fig:H_average19_target_generalization_errors_vs_p_eta_0_2_p_tilde_80}%
			\includegraphics[width=0.24\textwidth]{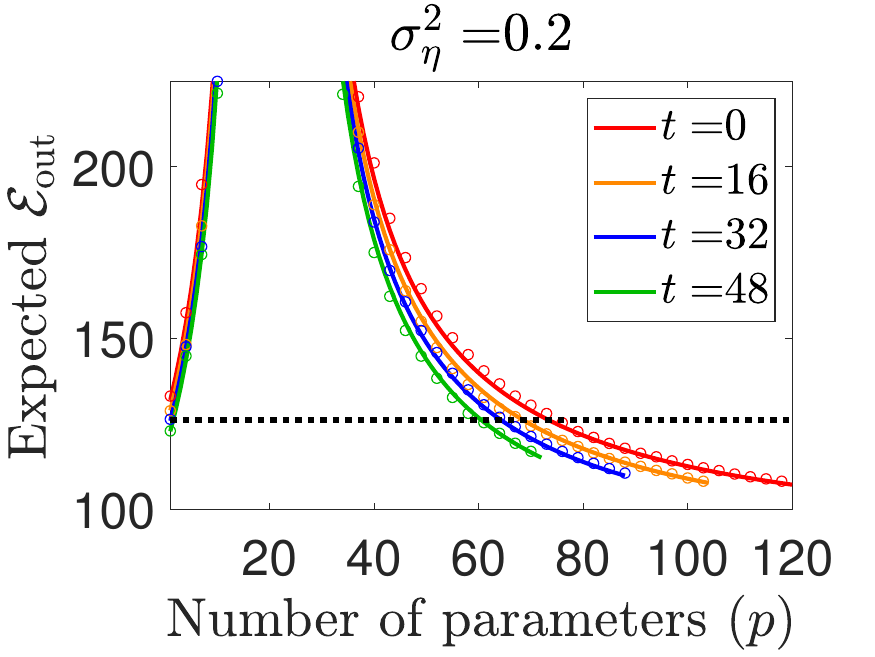}}
		~
		\subfloat[{\small$\mtx{H}$:\hspace{.4em}local\hspace{.4em}averaging neighborhood size  45}]{\label{fig:H_average45_target_generalization_errors_vs_p_eta_0_2_p_tilde_80}%
			\includegraphics[width=0.24\textwidth]{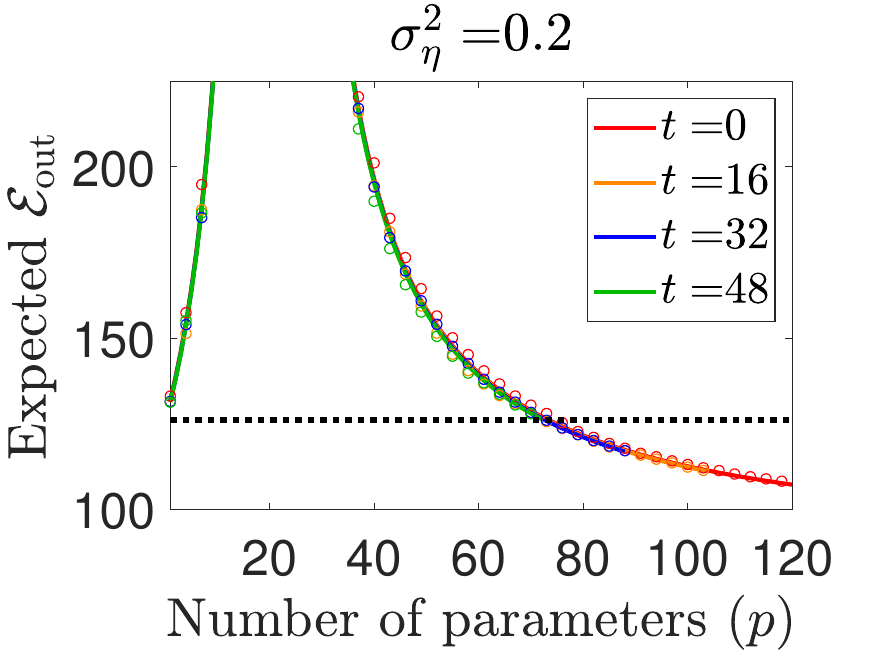}} 
		~
		\subfloat[{\small$\mtx{H}$:\hspace{.4em}overall\hspace{.4em}averaging neighborhood size  120}]{\label{fig:H_average80_target_generalization_errors_vs_p_eta_0_2_p_tilde_80}%
			\includegraphics[width=0.24\textwidth]{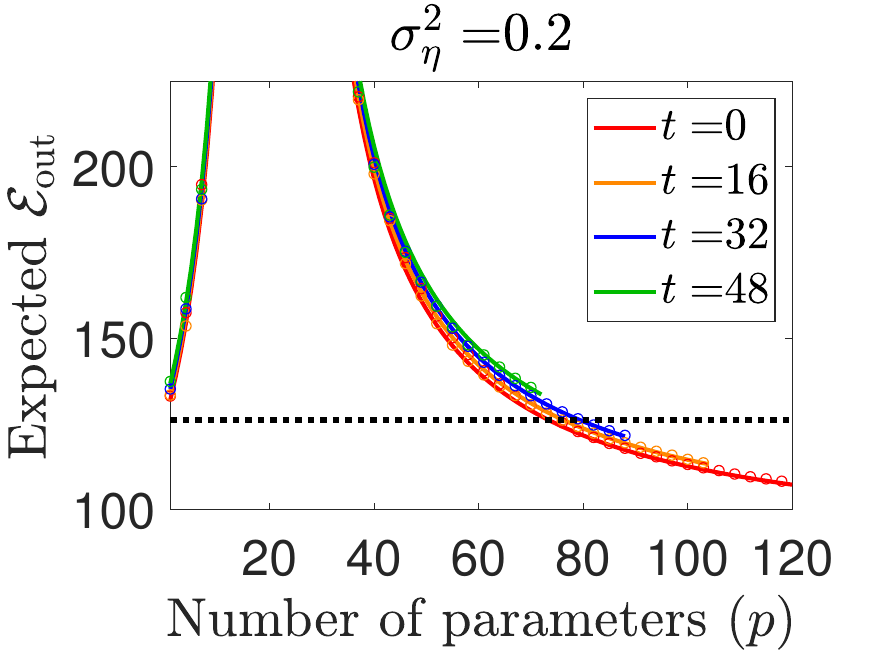}}
\caption{The expected generalization error of the target task, $\expectationwrt{ \mathcal{E}_{\rm out} }{\mathcal{L}}$, with respect to the number of free parameters (in the target task). The analytical values, computed using Corollary \ref{corollary:out of sample error - target task}, are presented using solid-line curves, and the respective empirical results obtained from averaging over 250 experiments are denoted by circle markers. 
		The horizontal dotted lines denote the error level of the null estimate. 
		Each subfigure considers a different case of the source-target task relation (\ref{eq:theta-beta relation}) with a different pair of $\sigma_{\eta}^2$ and $\mtx{H}$. The second row of subfigures corresponds to $\mtx{H}$ operators that perform local averaging, each subfigure (e)-(h) is w.r.t.~a different size of local averaging neighborhood.   Each curve color refers to a different number of transferred parameters. }
		\label{fig:target_generalization_errors_vs_p__on_average}
\end{figure}

Figure \ref{fig:target_generalization_errors_vs_p__on_average} presents the curves of $\expectationwrt{ \mathcal{E}_{\rm out} }{\mathcal{L}}$ with respect to the number of free parameters $p$ in the target task, whereas the source task has $\widetilde{p}=d$ free parameters. In Fig.~\ref{fig:target_generalization_errors_vs_p__on_average}, the solid-line curves correspond to analytical values induced by Corollary \ref{corollary:out of sample error - target task}, and the respective empirically computed values are denoted by circles (all the presented results are for $d=120$, ${n=20}$, $\widetilde{n}=50$, ${\| \vecgreek{\beta} \|_2^2 = d}$, $\sigma_{\epsilon}^2 = 0.05\cdot d$, $\sigma_{\xi}^2 = 0.025\cdot d$. See additional details in Appendix \ref{appendix:sec:Empirical Results for Section 3 Additional Details and Demonstrations}. The number of free parameters $p$ is upper bounded by $d-t$ that gets smaller for a larger number of transferred parameters $t$ (see, in Fig.~\ref{fig:target_generalization_errors_vs_p__on_average}, the earlier stopping of the curves when $t$ is larger).  Observe that the generalization error peaks at $p=n$ and, then, decreases as $p$ grows in the overparameterized range of $p>n+1$. We identify this behavior as a \textit{double descent} phenomenon, but without the first descent in the underparameterized range (double descent curves without the first descent are common when the parameters are selected arbitrarily, for example, see the results in \cite{belkin2020two,dar2020subspace}).

The error of a trivial solution in the form of the null estimate, i.e., the estimate $\widehat{\vecgreek{\beta}}$ is all zeros, is presented as black dotted horizontal lines in the figures. In various settings where the source and target tasks are sufficiently related and the number of parameters is sufficiently far from the peak of the double descent error curve, the examined transfer learning method (with $t>0$ arbitrarily selected parameters) outperforms the null estimate.

Each subfigure in Fig.~\ref{fig:target_generalization_errors_vs_p__on_average} considers a different task relation with a different pair of noise level $\sigma_{\eta}^2$ and operator $\mtx{H}$. 
The first row of subfigures in Fig.~\ref{fig:target_generalization_errors_vs_p__on_average} emphasizes the effect of the noise variance $\sigma_{\eta}^2$ in the task relation model on the generalization errors in the target task.
The second row of subfigures in Fig.~\ref{fig:target_generalization_errors_vs_p__on_average} emphasizes the effect of the linear operator $\mtx{H}$ in the task relation model on the generalization errors in the target task.

We can interpret the results in Figure \ref{fig:target_generalization_errors_vs_p__on_average} as examples for important cases of transfer learning settings. 
Figs.~\ref{fig:H_average5_target_generalization_errors_vs_p_eta_0_p_tilde_is_d},\ref{fig:H_average1_target_generalization_errors_vs_p_eta_0_2_p_tilde_80} correspond to transfer learning between two \textit{highly related tasks}, therefore, transfer learning is \textit{beneficial} in the sense that for a given $p\notin\{n-1,n,n+1\}$ the error decreases as $t$ increases (i.e., as more parameters are transferred instead of being omitted). 
Figs.~\ref{fig:H_average5_target_generalization_errors_vs_p_eta_0.2_p_tilde_is_d},\ref{fig:H_average19_target_generalization_errors_vs_p_eta_0_2_p_tilde_80} correspond to transfer learning between two \textit{moderately related tasks}, hence, transfer learning is still \textit{beneficial, but less} than in the former case of highly related tasks.
Figs.~\ref{fig:H_average5_target_generalization_errors_vs_p_eta_0.9_p_tilde_is_d},\ref{fig:H_average45_target_generalization_errors_vs_p_eta_0_2_p_tilde_80} correspond to transfer learning between two \textit{unrelated tasks} (although not extremely different), hence, transfer learning is \textit{useless}, but not harmful (i.e., for a given $p$, the number of transferred parameters $t$ does not affect the out-of-sample error).
Figs.~\ref{fig:H_average5_target_generalization_errors_vs_p_eta_2_p_tilde_120},\ref{fig:H_average80_target_generalization_errors_vs_p_eta_0_2_p_tilde_80} correspond to transfer learning between two \textit{very different tasks} and, accordingly, transfer learning \textit{degrades} the generalization performance (namely, for a given $p$, transferring more parameters increases the out-of-sample error).

\begin{figure}[t]
	\subfloat[Dimensions as in Fig.~ \ref{fig:H_average5_target_generalization_errors_vs_p_eta_0.2_p_tilde_is_d} ]{\label{fig:H_average5_target_generalization_errors_vs_p_eta_0.2_p_tilde_is_d_w_std_1x120}%
		\includegraphics[width=0.31\textwidth]{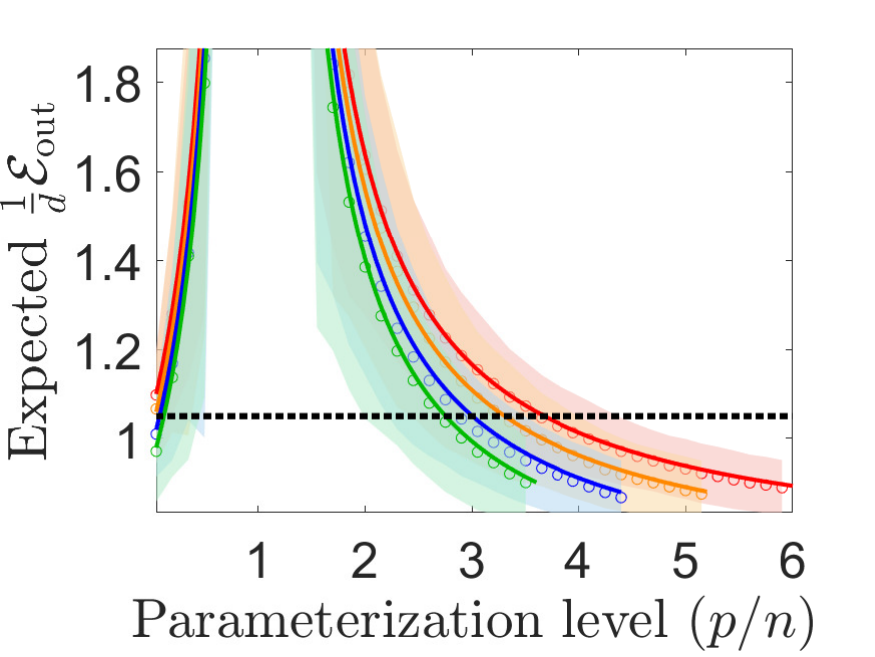}}
	~
	\subfloat[Dimensions 3 times larger]{\label{fig:H_average5_target_generalization_errors_vs_p_eta_0.2_p_tilde_is_d_w_std_3x120}%
		\includegraphics[width=0.31\textwidth]{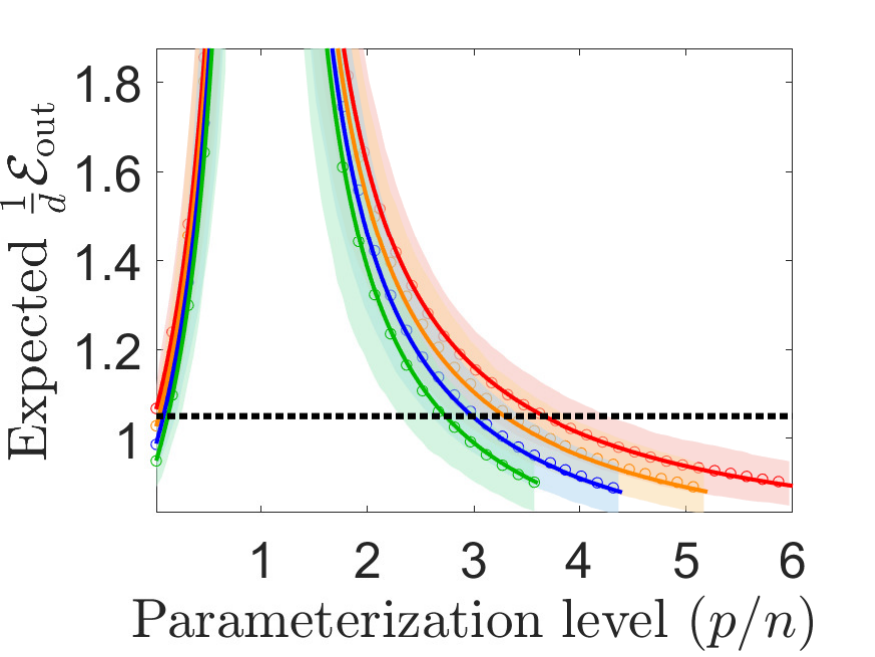}}
	\subfloat[Dimensions 5 times larger]{\label{fig:H_average5_target_generalization_errors_vs_p_eta_0.2_p_tilde_is_d_w_std_5x120}%
		\includegraphics[width=0.31\textwidth]{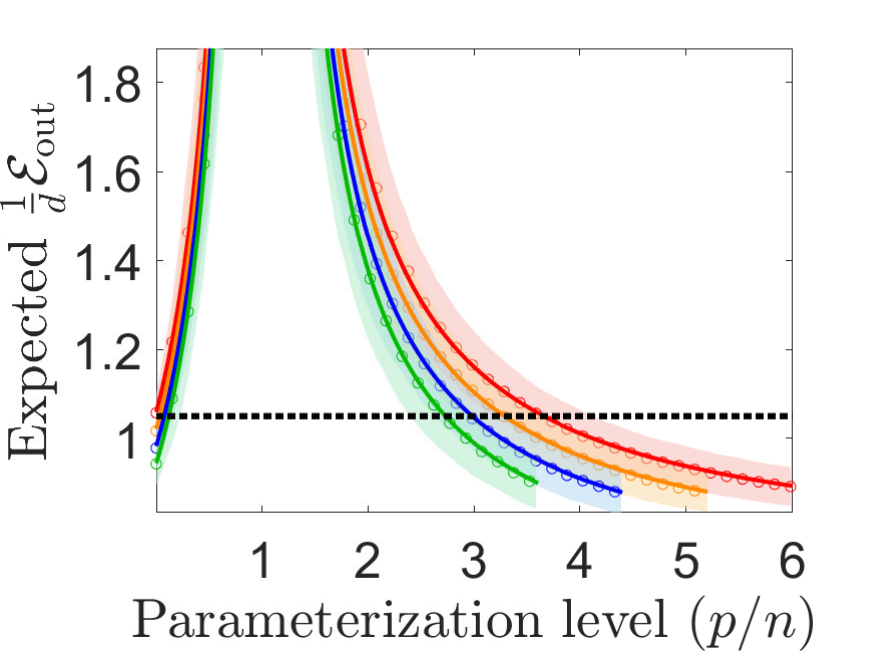}}
	\caption{The concentration of the generalization error at three proportional scales of the same problem. The empirical standard deviations are denoted as shaded areas in colors corresponding to the on-average error curves (solid lines and markers denote the analytical and empirical evaluations of the expected error, respectively). 
	Subfigure (a) corresponds to the setting of Fig.~\ref{fig:H_average5_target_generalization_errors_vs_p_eta_0.2_p_tilde_is_d}. 
	Subfigure (b) corresponds to a setting where all the dimensions and dimension-dependent quantities are 3 times their values in Fig.~\ref{fig:H_average5_target_generalization_errors_vs_p_eta_0.2_p_tilde_is_d} (see a more detailed explanation in the text).
	Subfigure (c) corresponds to a setting where all dimensions and dimension-dependent quantities are 5 times their values in Fig.~\ref{fig:H_average5_target_generalization_errors_vs_p_eta_0.2_p_tilde_is_d}.
	Lines, markers and areas in red correspond to $t=0$ (no parameters are transferred); orange corresponds to transferring $t=16\times\frac{d}{120}$ parameters; blue corresponds to $t=32\times\frac{d}{120}$; green corresponds  to $t=48\times\frac{d}{120}$. Note that $\frac{d}{120}$ equals to 1, 3, 5, in (a), (b), (c), respectively. The axes in this figure are normalized to be dimension-independent.}
	\label{fig:target_generalization_errors_vs_p__on_average_with_standard_deviations}
\end{figure}

Figure \ref{fig:target_generalization_errors_vs_p__on_average_with_standard_deviations} demonstrates the empirical standard deviations of the generalization errors (denoted as shaded areas around the curves and markers that denote the average errors). 
The results show that the empirical errors are more concentrated around their (theoretical and empirical) expectations as the dimensions of the problem (e.g., input dimension, number of data examples and parameters) are proportionally increased. 
For this example, we consider the setting of Fig.~\ref{fig:H_average5_target_generalization_errors_vs_p_eta_0.2_p_tilde_is_d} in three proportional scales of the dimensions and dimension-dependent quantities:
\begin{itemize}
	\item Fig.~\ref{fig:H_average5_target_generalization_errors_vs_p_eta_0.2_p_tilde_is_d_w_std_1x120} corresponds to the same dimensions as in Fig.~\ref{fig:H_average5_target_generalization_errors_vs_p_eta_0.2_p_tilde_is_d}: $d=120$, $n=20$, $\widetilde{n}=50$, $t\in\{0,16,32,48\}$, and local averaging with neighborhood size 5.
	
	\item Fig.~\ref{fig:H_average5_target_generalization_errors_vs_p_eta_0.2_p_tilde_is_d_w_std_3x120} corresponds to dimensions that are proportionally increased 3 times: $d=360$, $n=60$, $\widetilde{n}=150$, $t\in\{0,48,96,144\}$, and local averaging with neighborhood size 15.
	
	\item Fig.~\ref{fig:H_average5_target_generalization_errors_vs_p_eta_0.2_p_tilde_is_d_w_std_5x120} corresponds to dimensions that are proportionally increased 5 times: $d=600$, $n=100$, $\widetilde{n}=250$, $t\in\{0,80,160,240\}$, and local averaging with neighborhood size 25.	
\end{itemize} 
In all the settings in Figure \ref{fig:target_generalization_errors_vs_p__on_average_with_standard_deviations}, $\Ltwonorm{\vecgreek{\beta}}=d$ and the noise levels are  $\sigma_{\epsilon}^2 = 0.05\times d$, $\sigma_{\xi}^2 = 0.025\times d$, $\sigma_{\eta}^2 = 0.2$.
More examples are provided in Fig.~\ref{appendix:fig:target_generalization_errors_vs_p__on_average_with_standard_deviations_full_comparison}.


\begin{figure}
	\begin{center}
		\subfloat[]{\includegraphics[width=0.235\textwidth]{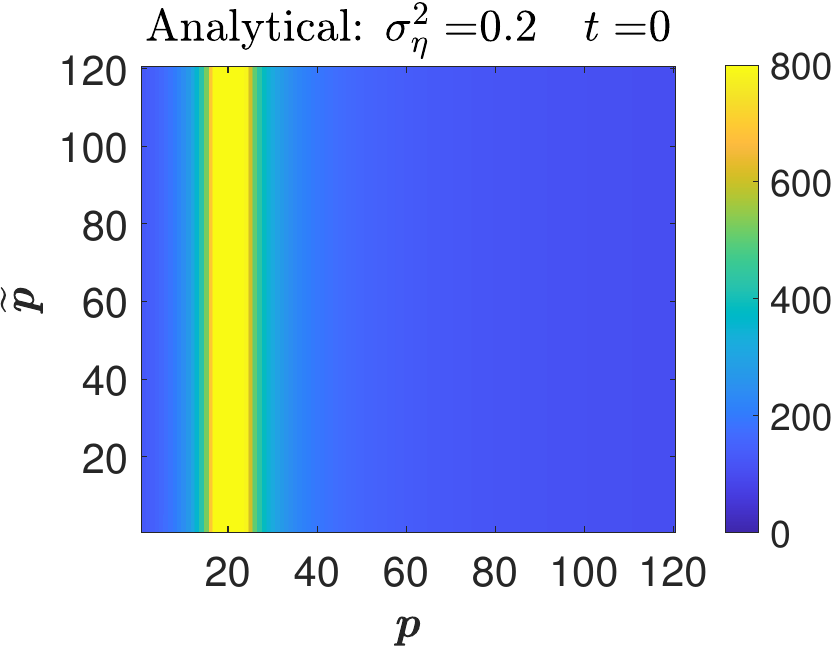}
			\label{fig:analytical_target_generalization_errors_eta_0.2_t0}} 		\subfloat[]{\includegraphics[width=0.235\textwidth]{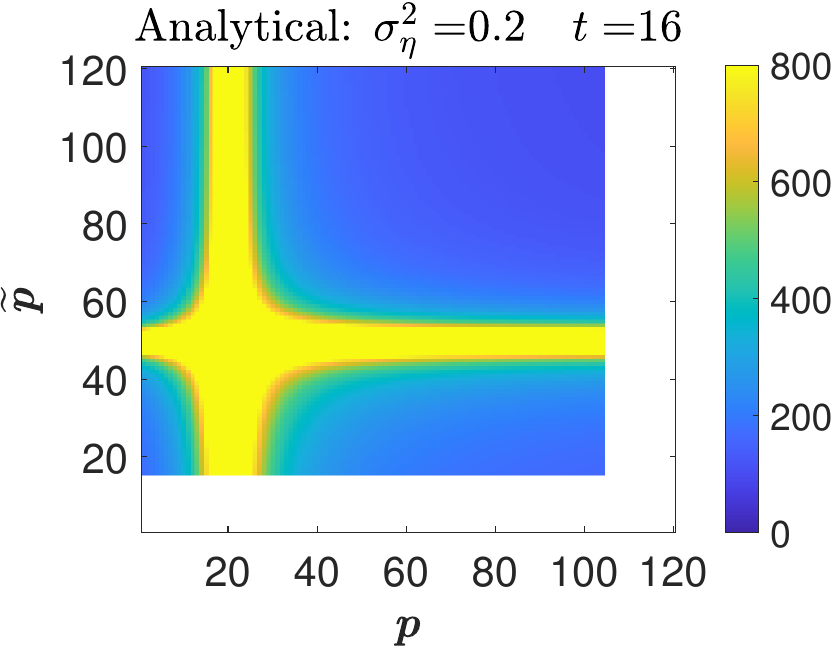}
			\label{fig:analytical_target_generalization_errors_eta_0.2_t16}}
		\subfloat[]{\includegraphics[width=0.235\textwidth]{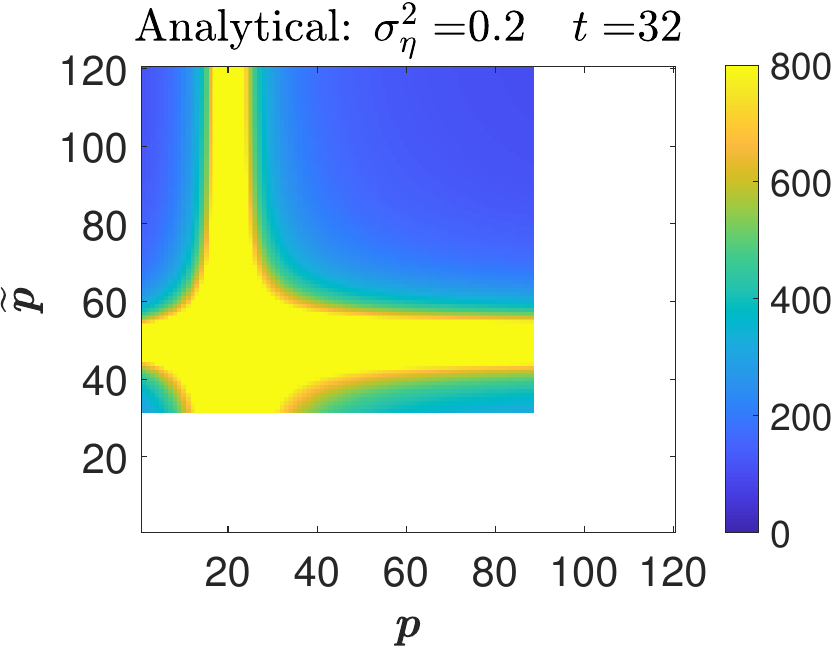}
			\label{fig:analytical_target_generalization_errors_eta_0.2_t32}} 		\subfloat[]{\includegraphics[width=0.235\textwidth]{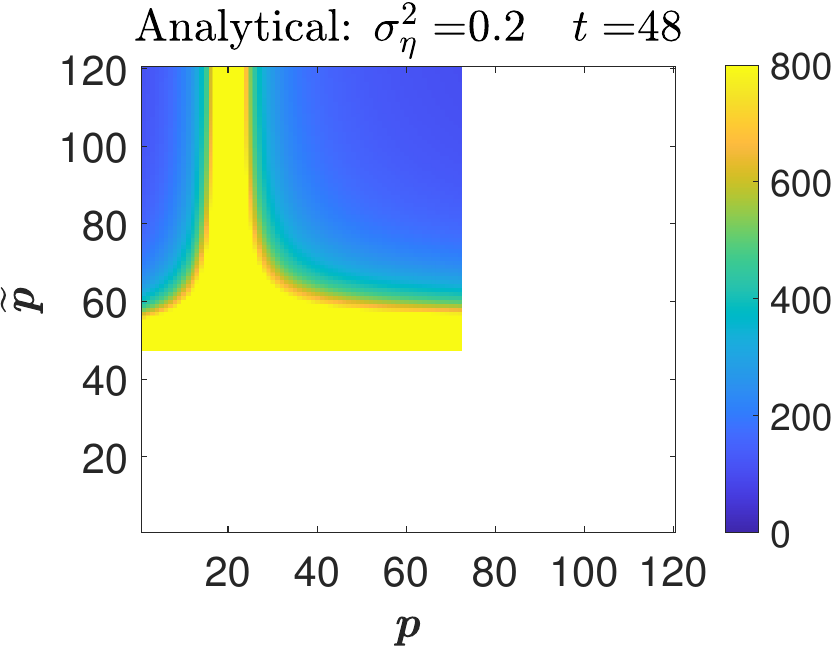}
			\label{fig:analytical_target_generalization_errors_eta_0.2_t48}}
		\caption{Analytical evaluation of the expected generalization error of the target task, $\expectationwrt{ \mathcal{E}_{\rm out} }{\mathcal{L}}$, with respect to the number of free parameters $\widetilde{p}$ and $p$ (in the source and target tasks, respectively). Each subfigure considers a different number of transferred parameters $t$. 
			The white regions correspond to $\left(\widetilde{p},p\right)$ settings eliminated by the value of $t$ in the specific subfigure. The yellow-colored areas correspond to values greater or equal to 800. 
			All of the subfigures are for $\sigma_{\eta}^2 = 0.2$, $\mtx{H}$ a local averaging operator with neighborhood of 5 samples, and $\vecgreek{\beta}$ that has a piecewise-constant form (Fig.~\ref{appendix:fig:piecewise_constant_beta_structure}). 
			See Fig.~\ref{appendix:fig:target_generalization_errors_p_vs_p_tilde_2D_planes} for settings with additional values of $\sigma_{\eta}^2$.
			See Fig.~\ref{appendix:fig:empirical_target_generalization_errors_p_vs_p_tilde_2D_planes} for the corresponding empirical evaluation. }
		\label{fig:target_generalization_errors_p_vs_p_tilde_2D_planes}
	\end{center}
	\vspace*{-5mm}
\end{figure}

By considering the generalization error formula from Theorem \ref{theorem:out of sample error - target task - specific layout} as a function of $\widetilde{p}$ and $p$ (i.e., the number of free parameters in the source and target tasks, respectively) we receive a \textit{two-dimensional double descent} behavior as presented in Fig.~\ref{fig:target_generalization_errors_p_vs_p_tilde_2D_planes} and its extended version Fig.~\ref{appendix:fig:target_generalization_errors_p_vs_p_tilde_2D_planes} in Appendix \ref{subsec:double double descent additional Results for the On-Average Analysis} that presents results for additional pairs of $t$ and $\sigma_{\eta}^2$. 
The results show a double descent trend along the $p$ axis (with a peak at $p=n$) and also, when parameter transfer is applied (i.e., $t>0$), a double descent trend along the $\widetilde{p}$ axis (with a peak at $\widetilde{p}=\widetilde{n}$).
Our solution structure implies that $\widetilde{p}\in\{t,\dots,d\}$ and $p\in\{0,\dots,d-t\}$, hence, a larger number of transferred parameters $t$ eliminates a larger portion of the underparameterized range of the source task and also eliminates a larger portion of the overparameterized range of the target task (see in Fig.~\ref{fig:target_generalization_errors_p_vs_p_tilde_2D_planes} the white eliminated regions that grow with $t$). When $t$ is high, the wide elimination of portions from the $(\widetilde{p},p)$-plane hinders the complete form of the two-dimensional double descent phenomenon (see, e.g.,  Fig.~\ref{fig:analytical_target_generalization_errors_eta_0.2_t48}). 

Conceptually, we can observe a \textit{tradeoff between overparamterized learning and transfer learning} where parameters are transferred as is from their co-located coordinates of the source task solution: an increased transfer of parameters limits the level of overparameterization applicable in the target task and, in turn, this may limit the overall potential gains from the transfer learning. Yet, when the source task is \textit{sufficiently related} to the target task (see, e.g., Figs.~\ref{fig:H_average5_target_generalization_errors_vs_p_eta_0_p_tilde_is_d},\ref{fig:H_average5_target_generalization_errors_vs_p_eta_0.2_p_tilde_is_d}), the parameter transfer can compensate (sometimes only partially) for an insufficient number of free parameters (in the target task). 
The last claim is also evident in Figs.~\ref{fig:H_average5_target_generalization_errors_vs_p_eta_0_p_tilde_is_d},\ref{fig:H_average5_target_generalization_errors_vs_p_eta_0.2_p_tilde_is_d},\ref{fig:H_average1_target_generalization_errors_vs_p_eta_0_2_p_tilde_80},\ref{fig:H_average19_target_generalization_errors_vs_p_eta_0_2_p_tilde_80} where, for $p>n+1$, there is a range of generalization error values that is achievable by several settings of $(p,t)$ pairs (i.e., specific error levels can be attained by curves of different colors in the same subfigure). 
E.g., in Fig.~\ref{fig:H_average5_target_generalization_errors_vs_p_eta_0.2_p_tilde_is_d} the error achieved by $p=112$ free parameters and no parameter transfer can be also achieved using $p=70$ free parameters and $t=48$ parameters transferred from the source task. 

\subsection{The Fragile Nature of Transfer Learning: Analysis of a Single Layout of Arbitrarily Selected Parameters}
\label{subsec:The Fragile Nature of Transfer Learning: Analysis of a Single Layout of Arbitrarily Selected Parameters}

Section \ref{subsec:On-Average Analysis of Arbitrarily Selected Parameters} considers an \text{on-average} error for a random coordinate layout. We now turn to discuss the generalization behavior of a \textit{single} coordinate layout that was formed arbitrarily (i.e., without using any knowledge on the problem setting). Hence, we return to consider the generalization error $\mathcal{E}_{\rm out}^{(\mathcal{L})}$ for a given layout $\mathcal{L}$ using Theorem \ref{theorem:out of sample error - target task - specific layout} and Corollary \ref{corollary:out of sample error - target task - specific layout - detailed - specific layout} from Section \ref{subsec:Analytical Characterization of the Double Descent Phenomenon}.

\begin{figure}[t]
	\subfloat[]{\includegraphics[width=0.24\textwidth]{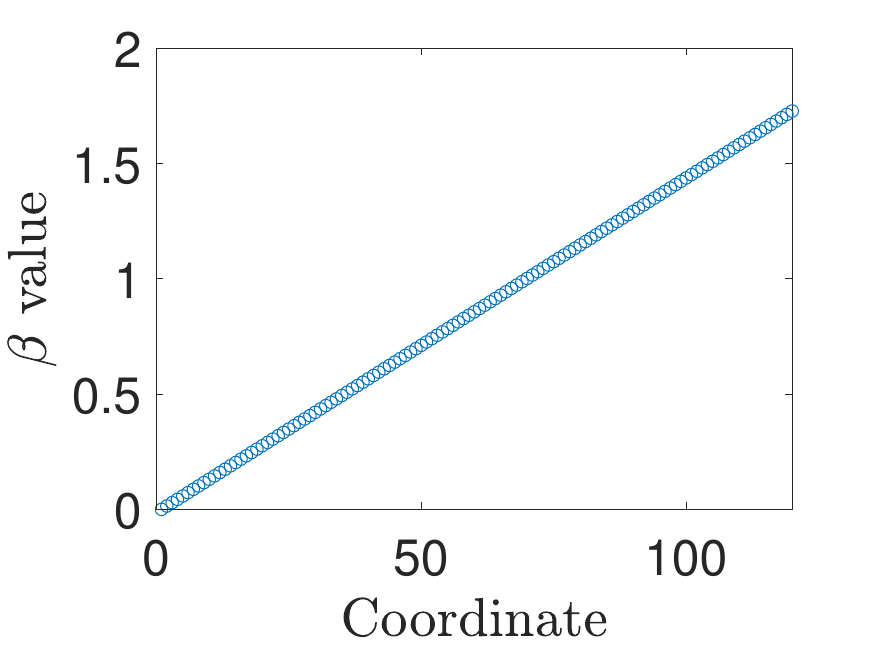} \label{appendix:fig:linear_beta_graph}}
	\subfloat[{\small$\mtx{H}=\mtx{I}_{d}$}]{\includegraphics[width=0.24\textwidth]{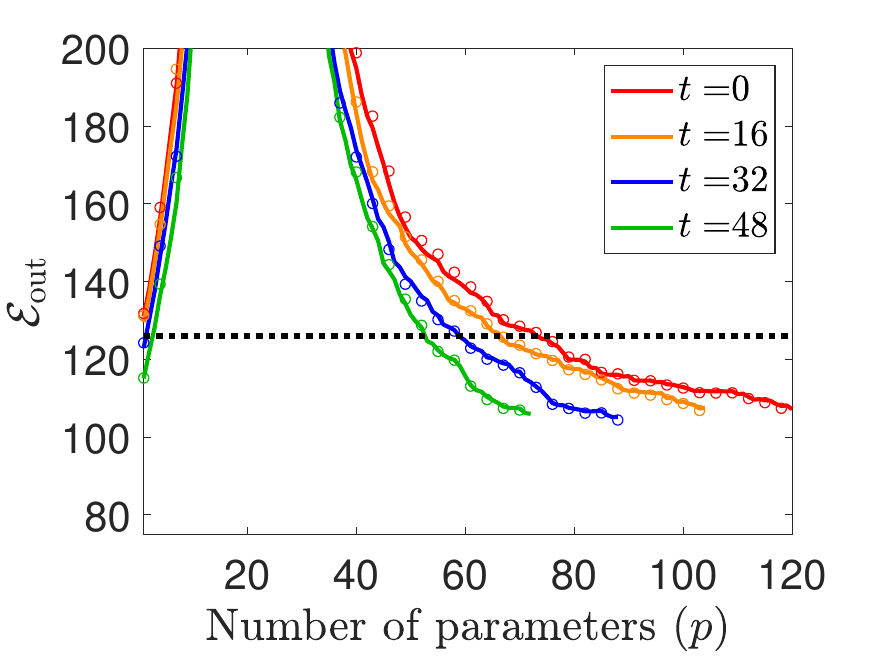} \label{fig:specific_linear_beta__H_is_I__target_generalization_errors_vs_p_eta_0_2_p_tilde_is_d}}
	\subfloat[{\small$\mtx{H}$:~Discrete derivative}]{\includegraphics[width=0.24\textwidth]{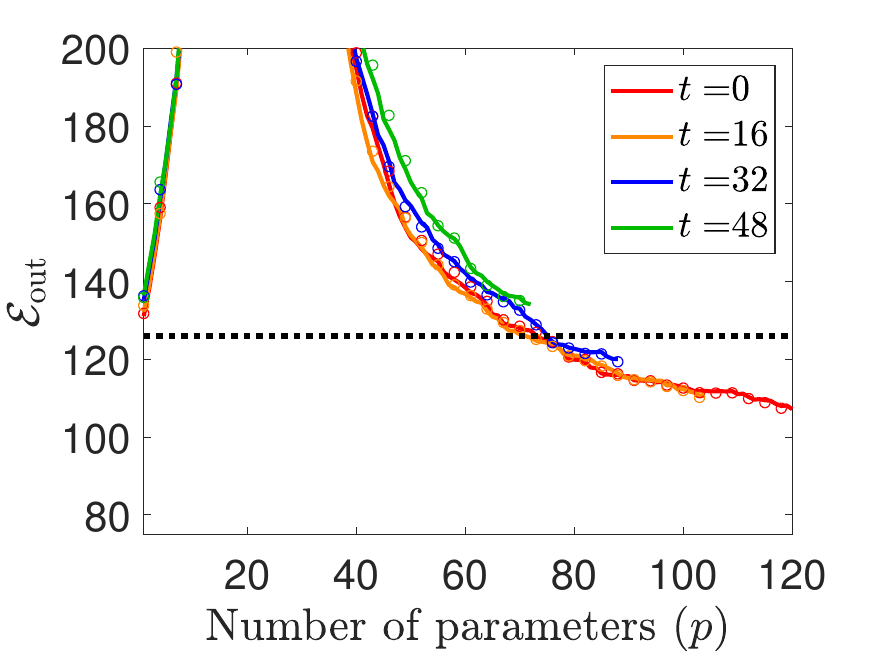} \label{fig:specific_linear_beta__H_is_derivative__target_generalization_errors_vs_p_eta_0_2_p_tilde_is_d}}
	\subfloat[{\small$\mtx{H}$:~Local averaging}]{\includegraphics[width=0.24\textwidth]{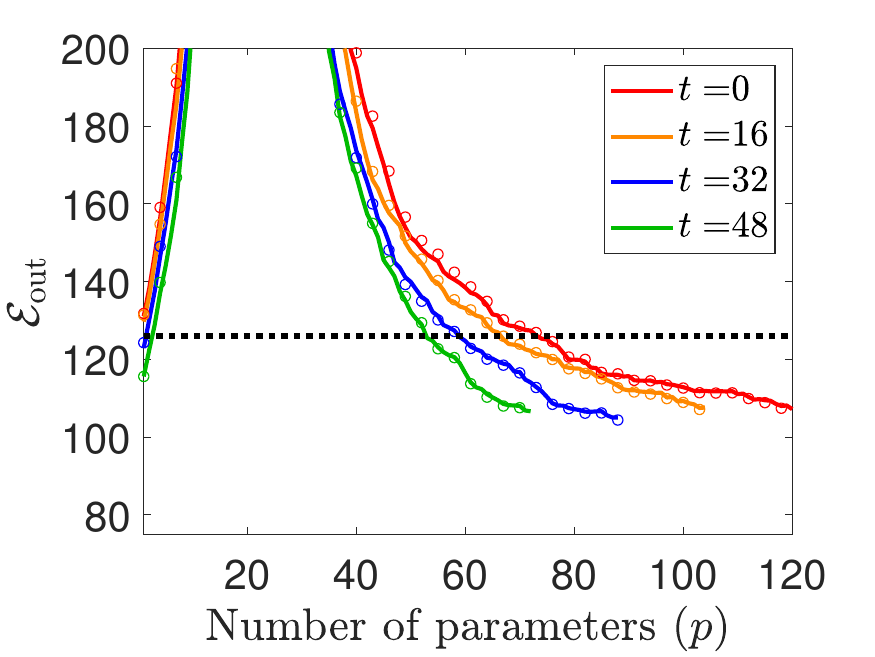} \label{fig:specific_linear_beta__H_is_averaging11__target_generalization_errors_vs_p_eta_0_2_p_tilde_is_d}}
	\\
	\subfloat[]{\includegraphics[width=0.24\textwidth]{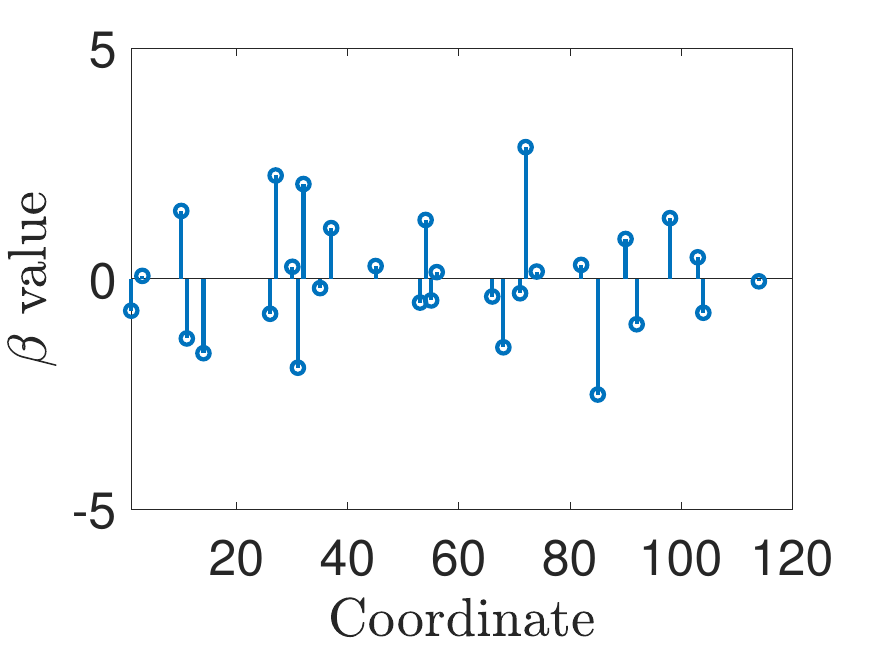} \label{appendix:fig:sparse_beta_graph}}
	\subfloat[{\small$\mtx{H}=\mtx{I}_{d}$}]{\includegraphics[width=0.24\textwidth]{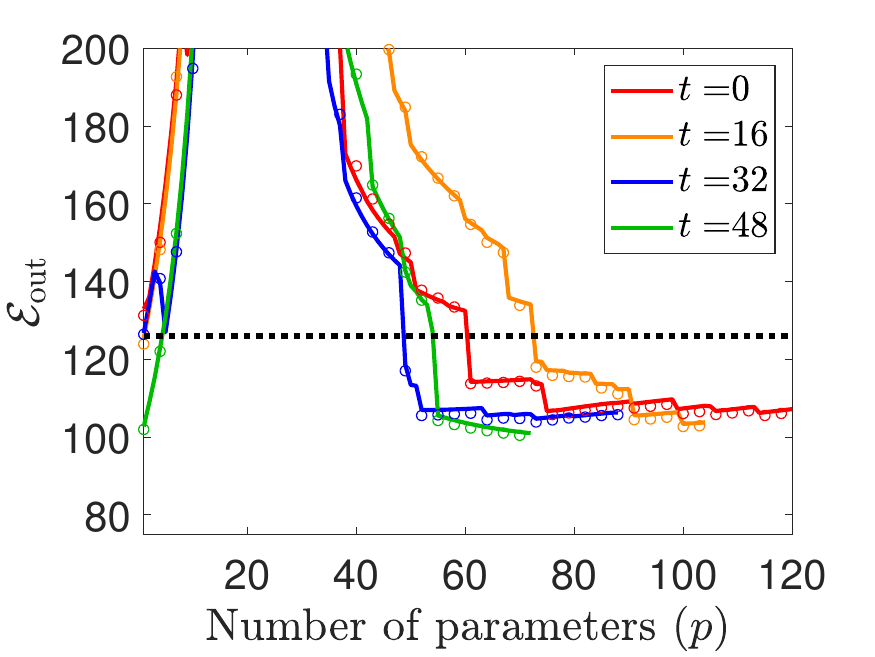} \label{fig:specific_sparse_beta__H_is_I__target_generalization_errors_vs_p_eta_0_2_p_tilde_is_d}}
	\subfloat[{\small$\mtx{H}$:~Discrete derivative}]{\includegraphics[width=0.24\textwidth]{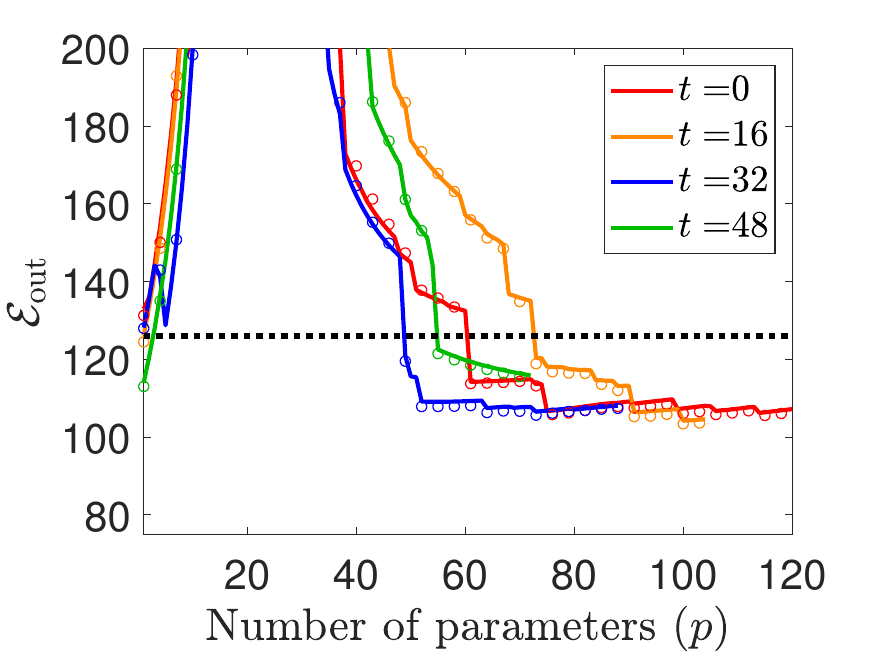} \label{fig:specific_sparse_beta__H_is_derivative__target_generalization_errors_vs_p_eta_0_2_p_tilde_is_d}}
	\subfloat[{\small$\mtx{H}$:~Local averaging}]{\includegraphics[width=0.24\textwidth]{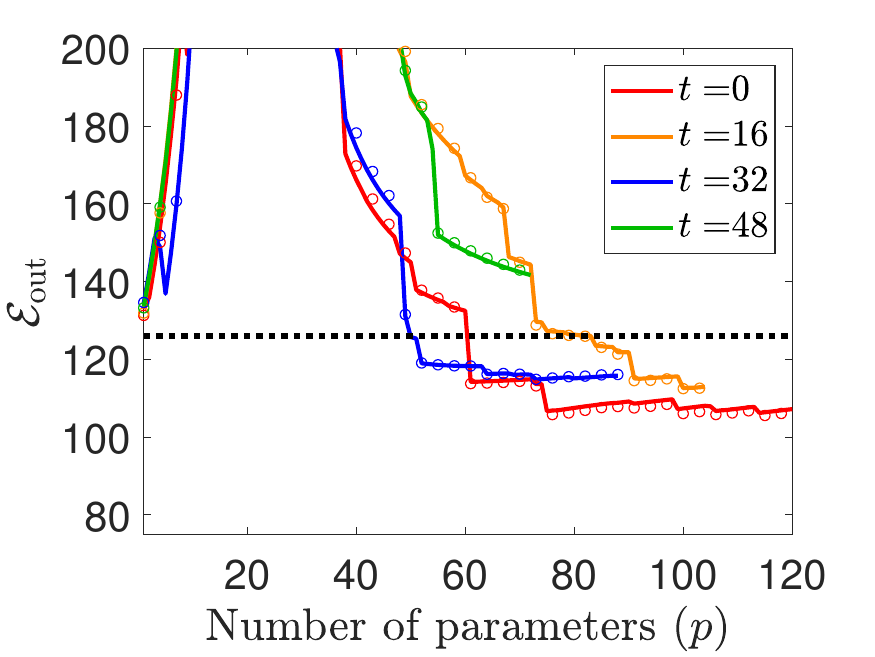} \label{fig:specific_sparse_beta__H_is_averaging11__target_generalization_errors_vs_p_eta_0_2_p_tilde_is_d}}
	\caption{Analytical (solid lines) and empirical (circle markers) values of $\mathcal{E}_{\rm out}^{(\mathcal{L})}$ for an arbitrary coordinate layout $\mathcal{L}$.  
		In the first row of subfigures the true $\vecgreek{\beta}$ has feature-domain values that follow a linear form , see (a).
		In the second row of subfigures the true $\vecgreek{\beta}$ has a sparse form in the feature domain with non-zero values at $30$ coordinates selected randomly out of the $d=120$, see (e). In each row there are three error plots for different circulant forms of the operator $\mtx{H}$ (here the local averaging is defined for a neighborhood of 11 samples). 
		Here $\sigma_{\eta}^2 = 0.2$, $d=120$, ${n=20}$, $\widetilde{n}=50$, $\| \vecgreek{\beta} \|_2^2 = d$, $\sigma_{\epsilon}^2 = 0.05\cdot d$, $\sigma_{\xi}^2 = 0.025\cdot d$, and $\widetilde{p}=d$  for all settings.
	}
	\label{fig:error_curves_for_specific_layouts__main_text_sample}
\end{figure}

Figure \ref{fig:error_curves_for_specific_layouts__main_text_sample} shows the curves of $\mathcal{E}_{\rm out}^{(\mathcal{L})} $ for specific coordinate layouts $\mathcal{L}$ that evolve with respect to the number of free parameters $p$ in the target task (for more examples see Figures \ref{appendix:fig:target_generalization_errors_vs_p__specific_extended_for_linear_beta}-\ref{appendix:fig:target_generalization_errors_vs_p__specific_extended_for_sparse_beta} in the Appendices).  The excellent fit of the analytical results (that were computed using Theorem \ref{theorem:out of sample error - target task - specific layout} and Corollary \ref{corollary:out of sample error - target task - specific layout - detailed - specific layout}) to the empirical values (computed by averaging over 250 experiments with the same evolution of $\mathcal{L}$) is evident. 
The effect of the specific coordinate layout is clearly visible by the less-smooth curves (compared to the on-average results in Fig.~\ref{fig:target_generalization_errors_vs_p__on_average}).  
We examine two different cases for the true $\vecgreek{\beta}$: a linearly-increasing (Fig.~\ref{appendix:fig:linear_beta_graph}) and a sparse (Fig.~\ref{appendix:fig:sparse_beta_graph}) layout of values, both have the same $\ell_2$ norm. The difference in the true $\vecgreek{\beta}$ forms yields error curves that significantly differ despite the use of the same sequential construction of the coordinate layouts with respect to $p$ (e.g., compare Figs.~\ref{fig:specific_sparse_beta__H_is_I__target_generalization_errors_vs_p_eta_0_2_p_tilde_is_d} and \ref{fig:specific_linear_beta__H_is_I__target_generalization_errors_vs_p_eta_0_2_p_tilde_is_d}). 
The operator $\mtx{H}$ in the task relation model greatly affects the generalization error curves as evident from comparing our results for different types of $\mtx{H}$: an identity, local averaging (with neighborhood size 11), and discrete derivative operators (e.g., compare subfigures within the first row of  Fig.~\ref{fig:error_curves_for_specific_layouts__main_text_sample}). 
The results clearly show that the interplay among the structures of $\mtx{H}$, $\vecgreek{\beta}$, and the coordinate layout significantly affects the generalization performance.

Our results also exhibit that an arbitrarily selected set $\mathcal{T}$ of $t$ transferred parameters can be the best setting for a given set $\mathcal{F}$ of $p$ free parameters but not necessarily for an extended set $\mathcal{F}'\supset\mathcal{F}$ of $p'>p$ free parameters. This is especially evident when the true parameters have a sparse form over an unknown support in the feature domain (i.e., true parameters with non-zero values are scarce and their coordinates are unknown). 
For example, see  Fig.~\ref{fig:specific_sparse_beta__H_is_derivative__target_generalization_errors_vs_p_eta_0_2_p_tilde_is_d} where the green colored curve does not consistently maintain its relative vertical order with the red color curve at the overparameterized range of solutions; as a second example see Fig.~\ref{fig:specific_sparse_beta__H_is_averaging11__target_generalization_errors_vs_p_eta_0_2_p_tilde_is_d} and observe that the blue and red curves do not maintain their vertical order in the overparameterized range. This exemplifies that, when arbitrary selection of parameters is employed due to unknown task relation, transfer learning settings can be fragile and hence finding a successful setting may require delicate, trial and error engineering. Therefore, our theory qualitatively explains similar practical aspects in deep neural networks (see, for example, \cite{raghu2019transfusion}).

\section{When is Transfer Learning Beneficial?}
\label{sec:When is Transfer Learning Beneficial}

The formulation of the generalization error in the target task (Theorem \ref{theorem:out of sample error - target task - specific layout} and Corollary \ref{corollary:out of sample error - target task - specific layout - detailed - specific layout}) shows that the benefits from parameter transfer depend on various aspects of the learning setting. 
In this section we characterize the conditions for beneficial transfer from the viewpoint of the following question: Given a setting where $\mathcal{T}$ is the intended set of coordinates for parameter transfer, can avoiding this parameter transfer improve generalization?

\subsection{Benefits in Transferred versus Zeroed Parameters}
\label{subsec:When is Transferring Parameters More Beneficial Than Zeroing Them}

To accurately evaluate the difference in the out-of-sample error of the target task, consider the following definitions. First, recall the coordinate layout $\mathcal{L}=\{\mathcal{S},\mathcal{F},\mathcal{T},\mathcal{Z}\}$ where  $\lvert\mathcal{S}\rvert = \widetilde{p}$, $\lvert\mathcal{F}\rvert = p$, $\lvert\mathcal{T}\rvert = t$.  
Second, we define a coordinate layout $\mathcal{L}'=\{\mathcal{S},\mathcal{F},\mathcal{T}',\mathcal{Z}'\}$ which is a modified version of $\mathcal{L}$ without transferred parameters, specifically, $\mathcal{T}'=\emptyset$ and $\mathcal{Z}'=\mathcal{Z}\cup\mathcal{T}$. 
Namely, $\mathcal{L}'$ is obtained by zeroing all the parameters that are transferred in $\mathcal{L}$. 
Then, we define the following error difference due to transferring the parameters in $\mathcal{T}$ \textit{instead of zeroing} them: 
\begin{equation}
	\label{eq:error difference term - transfer versus zero}
	\Delta\mathcal{E}_{\rm TvsZ}^{(\mathcal{L})} \triangleq  \mathcal{E}_{\rm out}^{(\mathcal{L})} - \mathcal{E}_{\rm out}^{(\mathcal{L}')}
\end{equation}
where $\mathcal{E}_{\rm out}^{(\mathcal{L})}$ and $\mathcal{E}_{\rm out}^{(\mathcal{L}')}$ are the out-of-sample errors in the target task for the coordinate layouts $\mathcal{L}$ and $\mathcal{L}'$, respectively. Using  Theorem \ref{theorem:out of sample error - target task - specific layout} we can write (\ref{eq:error difference term - transfer versus zero}) as 
\begin{equation}
	\label{eq:error difference term - transfer versus zero - detailed}
	\Delta\mathcal{E}_{\rm TvsZ}^{(\mathcal{L})} = \left( \mathcal{E}_{\rm transfer}^{(\mathcal{T},\mathcal{S})} - \Ltwonorm{\vecgreek{\beta}_{\mathcal{T}}} \right) \times \begin{cases}
		\mathmakebox[5em][l]{\frac{n-1}{n-p-1}}  \text{for } p \le n-2,   
		\\
		\mathmakebox[5em][l]{\frac{p-1}{p-n-1}} \text{for } p \ge n+2,
	\end{cases}
\end{equation}
where $\mathcal{E}_{\rm transfer}^{(\mathcal{T},\mathcal{S})}$ was defined and formulated in Theorem \ref{theorem:out of sample error - target task - specific layout} and Corollary \ref{corollary:out of sample error - target task - specific layout - detailed - specific layout}. 
Note that $\Delta\mathcal{E}_{\rm TvsZ}^{(\mathcal{L})}$ is undefined for $p\in\{n-1,n,n+1\}$. 
The definition in (\ref{eq:error difference term - transfer versus zero}) implies that transferring the parameters in $\mathcal{T}$ is beneficial (over zeroing these coordinates in addition to the coordinates in $\mathcal{Z}$) if ${\Delta\mathcal{E}_{\rm TvsZ}^{(\mathcal{L})}<0}$. 
We define 
\begin{equation}
\label{eq:error difference term - transfer versus zero - auxiliary term}
\Delta\mathcal{E}_{\rm transfer}^{(\mathcal{T},\mathcal{S})}\triangleq \mathcal{E}_{\rm transfer}^{(\mathcal{T},\mathcal{S})} - \Ltwonorm{\vecgreek{\beta}_{\mathcal{T}}}. 
\end{equation}
Then, according to (\ref{eq:error difference term - transfer versus zero - detailed}), ${\mathcal{E}_{\rm TvsZ}^{(\mathcal{L})}<0}$ occurs when $p\notin\{n-1,n,n+1\}$ and $\Delta\mathcal{E}_{\rm transfer}^{(\mathcal{T},\mathcal{S})} <0$.

\begin{figure}[t]
	\begin{center}		
		\subfloat[{\small$\mtx{H}=\frac{1}{2}\mtx{I}_{d}$}]{\includegraphics[width=0.31\textwidth]{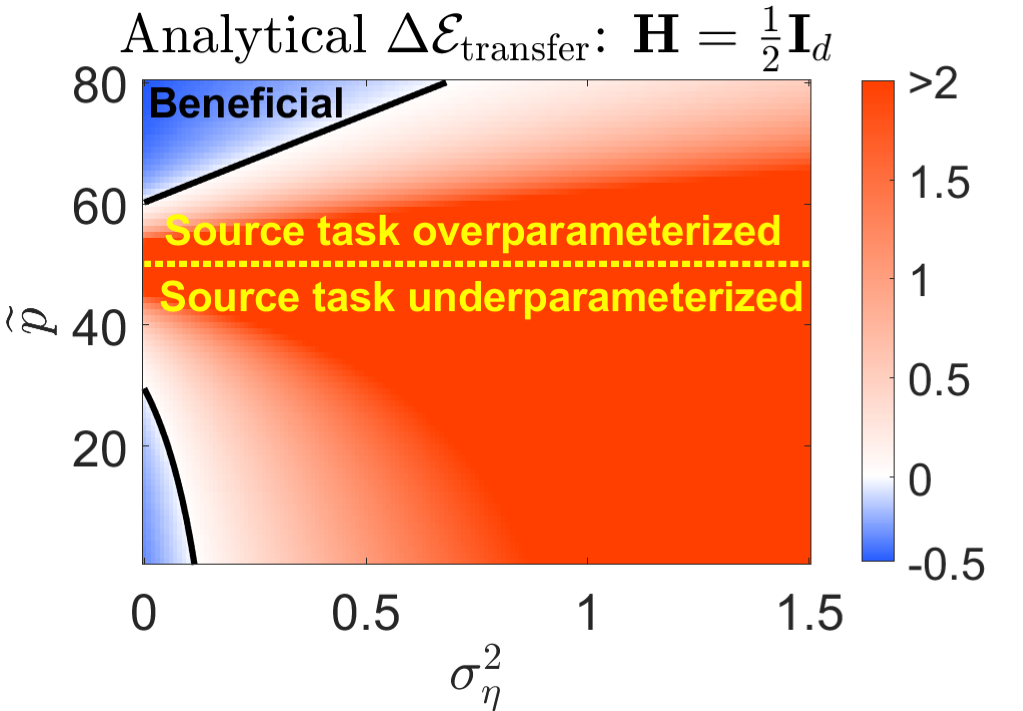}\label{fig:transfer_learning_usefulness_plane__H_is_0_5_I_analytic}}
		~~
		\subfloat[{\small$\mtx{H}=\mtx{I}_{d}$}]{\includegraphics[width=0.31\textwidth]{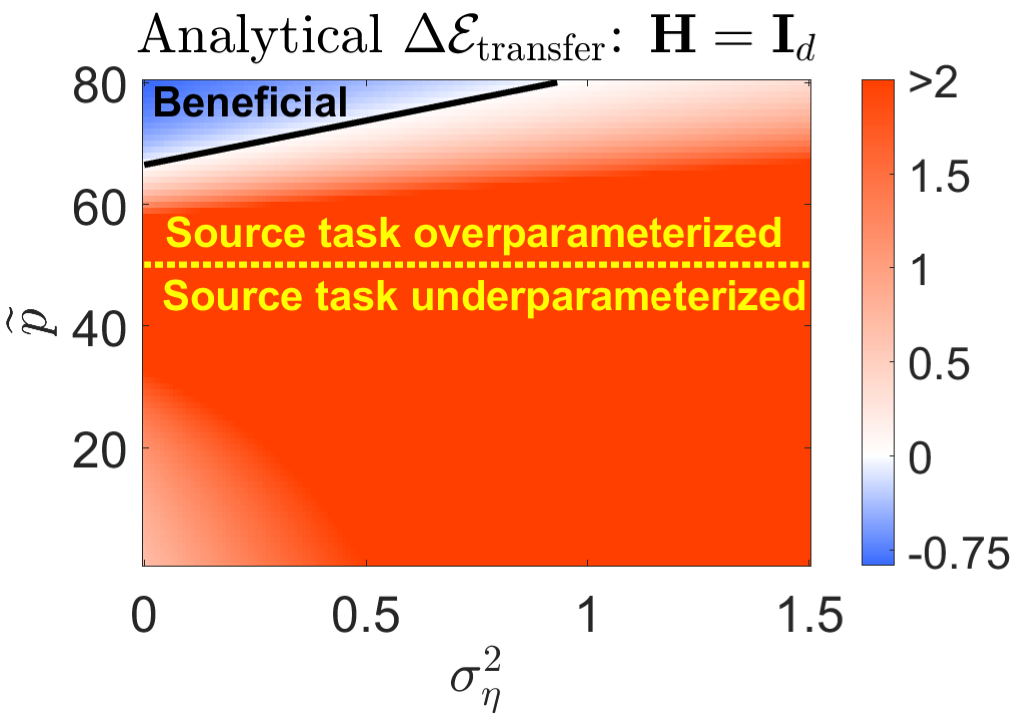}\label{fig:transfer_learning_usefulness_plane__H_is_I_analytic}} 	
		~~	\subfloat[{\small$\mtx{H}=\frac{3}{2}\mtx{I}_{d}$}]{\includegraphics[width=0.31\textwidth]{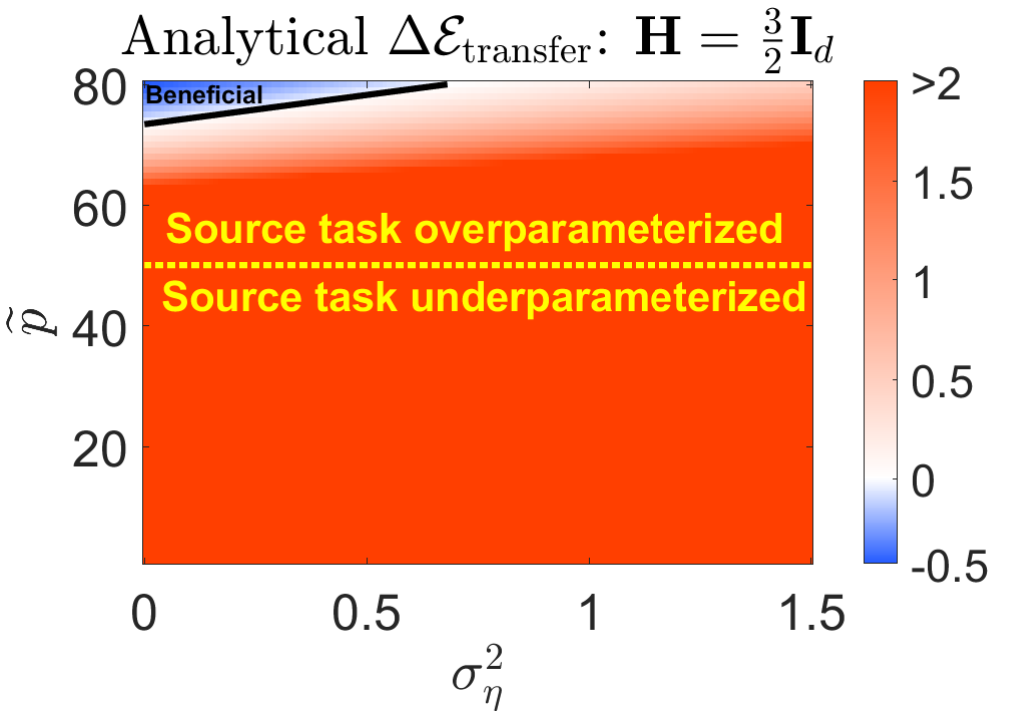}\label{fig:transfer_learning_usefulness_plane__H_is_1_5_I_analytic}}
		\\~\\~\\
		\subfloat[{\small$\mtx{H}$: local averaging \hspace{2em}neighborhood size  3}]{\includegraphics[width=0.31\textwidth]{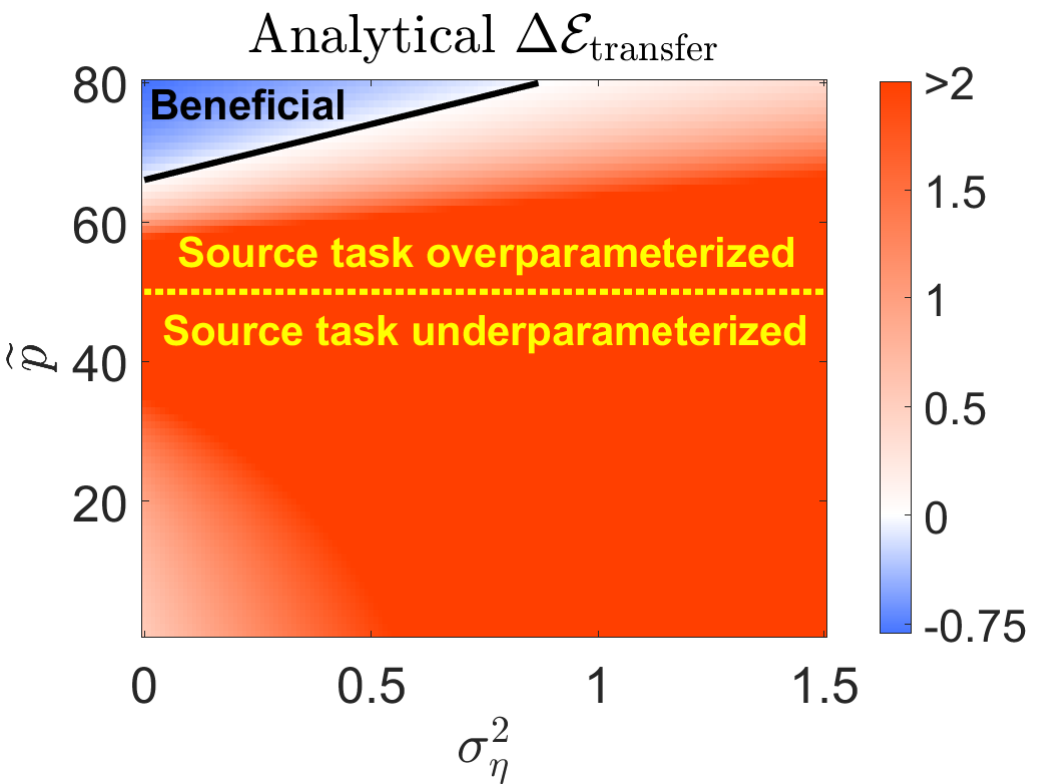}\label{fig:transfer_learning_usefulness_plane__H_average3_analytic}}
		~~
		\subfloat[{\small$\mtx{H}$: local averaging \hspace{2em}neighborhood size  15}]{\includegraphics[width=0.31\textwidth]{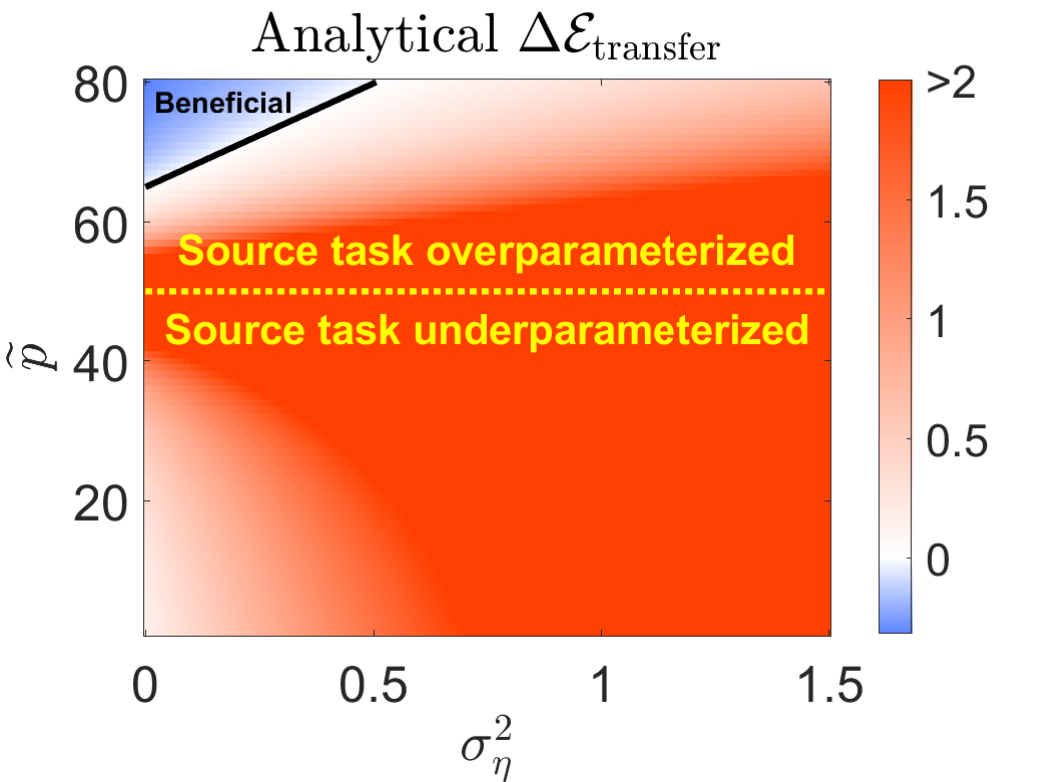} \label{fig:transfer_learning_usefulness_plane__H_average15_analytic}}
		~~
		\subfloat[{\small$\mtx{H}$: local averaging \hspace{4em}neighborhood size  59}]{\includegraphics[width=0.31\textwidth]{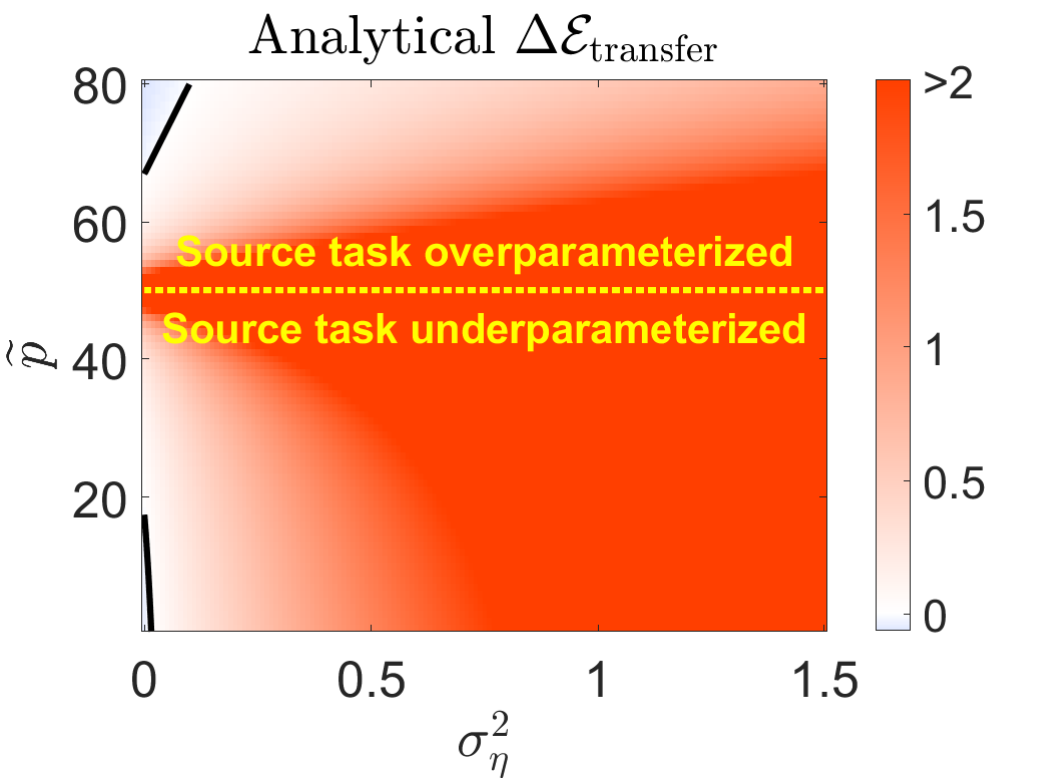} \label{fig:transfer_learning_usefulness_plane__H_average59_analytic}}
		\caption{The analytical values of $\Delta\mathcal{E}_{\rm transfer}$  (here normalized by $t$) as a function of $\widetilde{p}$ and $\sigma_{\eta}^2$. The positive and negative values of $\Delta\mathcal{E}_{\rm transfer}$ appear in color scales of red and blue, respectively. 
			The regions of negative values (appear in shades of blue) correspond to beneficial transfer of parameters (compared to zeroing them). 
			The positive values were truncated at the value of 2 for the clarity of visualization. 
			Each subfigure corresponds to a different task relation model induced by the definitions of $\mtx{H}$ as: \textit{(a)-(c)} different scalings of the identity matrix, \textit{(d)-(f)} circulant matrices that correspond to local averaging operators with different neighborhood sizes.			
			For all the subfigures, $d=80$, $\widetilde{n}=50$, $\sigma_{\xi}^2 = 0.025\cdot d$, $\| \vecgreek{\beta} \|_2^2 = d$. The components of $\vecgreek{\beta}$ have a linear form (see Fig.~\ref{appendix:fig:linear_beta_graph}) in \textit{(a)-(c)} and a piecewise-constant form (see Fig.~\ref{appendix:fig:piecewise_constant_beta_structure}) in \textit{(d)-(f)}.  See corresponding empirical results in Figs.~\ref{appendix:fig:transfer_learning_usefulness_plane - H is averaging}-\ref{fig:transfer_learning_usefulness_plane - H is I}.}
		\label{fig:transfer_learning_usefulness_plane - H averaging}
	\end{center}
	\vspace*{-5mm}
\end{figure}

The examples in Fig.~\ref{fig:transfer_learning_usefulness_plane - H averaging} show the values of $\Delta\mathcal{E}_{\rm transfer}\triangleq\expectationwrt{\Delta\mathcal{E}_{\rm transfer}^{(\mathcal{T},\mathcal{S})}}{\mathcal{L}}$ due to transferring an arbitrarily selected parameter. The error difference is presented with respect to the number $\widetilde{p}$ of free parameters in the source task and the variance $\sigma_{\eta}^2$ of the noise in the task relation model. 
Each subfigure in Fig.~\ref{fig:transfer_learning_usefulness_plane - H averaging} shows results for a different definition of $\mtx{H}$. 
All the subfigures demonstrate that transferring an arbitrarily selected parameter is more beneficial when the solution to the source task is highly overparameterized; indeed, then the generalization error in the source task itself is lower due to its own double descent phenomena (e.g., recall the behavior of the error along the vertical axis in Fig.~\ref{fig:analytical_target_generalization_errors_eta_0.2_t16}-\ref{fig:analytical_target_generalization_errors_eta_0.2_t48}). 
As expected, a low level of noise in the task relation is also important for beneficial transfer. 


The examples in Fig.~\ref{fig:transfer_learning_usefulness_plane - H averaging} also show a somewhat unexpected behavior: parameter transfer from an \textit{underparameterized} source task can be more beneficial as the tasks are less related, e.g., compare the lower left corners in Figures \ref{fig:transfer_learning_usefulness_plane__H_is_0_5_I_analytic} and \ref{fig:transfer_learning_usefulness_plane__H_is_I_analytic}. 
To understand this observation mathematically one may recall the transfer bias and variance formulations in Corollary \ref{corollary:out of sample error - target task - specific layout - detailed - specific layout} and notice the following effects of $\mtx{H}$, which in Figs.~\ref{fig:transfer_learning_usefulness_plane__H_is_0_5_I_analytic}-\ref{fig:transfer_learning_usefulness_plane__H_is_I_analytic} has the form of $\mtx{H}=c\mtx{I}_d$ where $c\in(0,1)$. First, the transfer bias increases as $c\in(0,1)$ gets smaller; also note that the transfer bias sums only over the $\lvert\mathcal{T}\rvert = t$ transferred coordinates. Second, the (underparameterized) transfer variance decreases as $c\in(0,1)$ gets smaller; here note that the transfer variance is affected by $\lvert\mathcal{S}^{c}\rvert = d-\widetilde{p}$ coordinates of $\mtx{H}=c\mtx{I}_d$ in addition to the number $t$ of transferred parameters. In the experiment setting of  Fig.~\ref{fig:transfer_learning_usefulness_plane__H_is_0_5_I_analytic}, the source task is underparameterized with a sufficiently large ${d-\widetilde{p}}$ value such that the reduction in the transfer variance dominates the increase in the transfer bias, resulting in beneficial transfer. 
 

More generally, the lesson here is that beneficial transfer learning can be achieved also in non-intuitive settings where the tasks are not necessarily highly related. 
In Section \ref{sec:Additional Insights The Optimal H in a Componentwise Task Relation} we will provide additional insights into counter-intuitive beneficial cases. 


\subsection{Benefits in Transferred versus Free Parameters}
\label{subsec:Benefits in Transferred versus Free Parameters}


Now we turn to discuss the question of when the transfer of the parameters in $\mathcal{T}$ is more beneficial than defining them as free for optimization.

Consider the coordinate layout $\mathcal{L}=\{\mathcal{S},\mathcal{F},\mathcal{T},\mathcal{Z}\}$ where, as usual,  $\lvert\mathcal{S}\rvert=\widetilde{p}$, $\lvert\mathcal{F}\rvert = p$, $\lvert\mathcal{T}\rvert = t$.  
Here we also define a second coordinate layout $\mathcal{L}''=\{\mathcal{S},\mathcal{F}'',\mathcal{T}'',\mathcal{Z}\}$ which is a modified version of $\mathcal{L}$ without transferred parameters, specifically, $\mathcal{T}''=\emptyset$ and $\mathcal{F}''=\mathcal{F}\cup\mathcal{T}$. 
Hence, the consideration of $\mathcal{L}''$ reflects the case where all the transferred parameters in $\mathcal{L}$ are replaced by free parameters. 
Then, we define the following error difference term due to transferring the parameters in $\mathcal{T}$ \textit{instead of allocating them as free parameters}: 
\begin{equation}
	\label{eq:error difference term - transfer versus free}
	\Delta\mathcal{E}_{\rm TvsF}^{(\mathcal{L})} \triangleq  \mathcal{E}_{\rm out}^{(\mathcal{L})} - \mathcal{E}_{\rm out}^{(\mathcal{L}'')}
\end{equation}
where $\mathcal{E}_{\rm out}^{(\mathcal{L})}$ and $\mathcal{E}_{\rm out}^{(\mathcal{L}'')}$ are the out-of-sample errors in the target task (recall Theorem \ref{theorem:out of sample error - target task - specific layout}) for the coordinate layouts $\mathcal{L}$ and $\mathcal{L}''$, respectively. 
The definition in (\ref{eq:error difference term - transfer versus free}) implies that transferring the parameters in $\mathcal{T}$ is beneficial (over allocating these coordinates as $t$ free parameters in addition to the $p$ free parameters in $\mathcal{F}$) if ${\Delta\mathcal{E}_{\rm TvsF}^{(\mathcal{L})}<0}$.

We now turn to examine the effects of $p$ and $t$ on beneficial transfer in the sense of ${\Delta\mathcal{E}_{\rm TvsF}^{(\mathcal{L})}<0}$. For this, note that $\mathcal{L}''$ includes $p''=p+t$ free parameters and hence is not necessarily in the same parameterization regime as $\mathcal{L}$ that includes $p$ free parameters. 
The parameterization regimes of $\mathcal{L}$ and $\mathcal{L}''$ affect the behavior of the benefits from parameter transfer, as described by the following corollaries of Theorem \ref{theorem:out of sample error - target task - specific layout} (these corollaries are simply proved by setting Eq.~(\ref{eq:out of sample error - target task - theorem - general decomposition form - specific layout}) in  (\ref{eq:error difference term - transfer versus free}) and reorganizing the formulations). 
\begin{corollary}
	\label{corollary:benefits in transfer versus free - both underparameterized}
	For $p+t\le n-2$, namely, when both $\mathcal{L}$ and $\mathcal{L}''$ correspond to underparameterized settings, the error difference due to transferred versus free parameters is 
	\begin{equation}
		\label{eq:benefits in transfer versus free - both underparameterized - error difference}
		\Delta\mathcal{E}_{\rm TvsF}^{(\mathcal{L})} = \frac{n-1}{n-p-1}\left({ \mathcal{E}_{\rm transfer}^{(\mathcal{T},\mathcal{S})}  - \frac{t}{n-p-t-1}\left({\Ltwonorm{\vecgreek{\beta}_{\mathcal{Z}}}+\sigma_{\epsilon}^2}\right)} \right).
	\end{equation}	
	Accordingly, the transfer of the $t>0$ parameters in $\mathcal{T}\subseteq\mathcal{S}$ is more beneficial than allocating them as free parameters (i.e., $\Delta\mathcal{E}_{\rm TvsF}^{(\mathcal{T},\mathcal{S})}<0$) if 	
	\begin{equation}
		\label{eq:benefits in transfer versus free - both underparameterized - condition for benefits}
		\mathcal{E}_{\rm transfer}^{(\mathcal{T},\mathcal{S})} <    \frac{t}{n-p-t-1}\left({\Ltwonorm{\vecgreek{\beta}_{\mathcal{Z}}}+\sigma_{\epsilon}^2}\right).
	\end{equation}
\end{corollary}

\begin{corollary}
	\label{corollary:benefits in transfer versus free - Ltag is overparameterized}
	For $p+t\ge n+2$, namely, when $\mathcal{L}''$ corresponds to an overparameterized setting, the error difference due to transferred versus free parameters is 
	\begin{equation}
		\label{eq:benefits in transfer versus free - Ltag is overparameterized - error difference - general}
		\Delta\mathcal{E}_{\rm TvsF}^{(\mathcal{L})} = \gamma_0 \mathcal{E}_{\rm transfer}^{(\mathcal{T},\mathcal{S})} 
		+\gamma_1 \Ltwonorm{\vecgreek{\beta}_{\mathcal{F}}} + \gamma_2 \Ltwonorm{\vecgreek{\beta}_{\mathcal{T}}}	
		+\gamma_3 \left({\Ltwonorm{\vecgreek{\beta}_{\mathcal{Z}}}+\sigma_{\epsilon}^2}\right)
	\end{equation}	
	where the coefficients $\gamma_0$,$\gamma_1$,$\gamma_2$,$\gamma_3$ depend on the parameterization level of $\mathcal{L}$: 
	\begin{itemize}
		\item For an underparameterized $\mathcal{L}$ with $p\le n-2$:
		
		$\gamma_0=\frac{n-1}{n-p-1}$, ${\gamma_1=-\left( 1-\frac{n}{p+t}\right)}$, ${\gamma_2=-\left( 1-\frac{n}{p+t}\right)}$, ${\gamma_3=-\frac{n(n-1)-p(p+t-1)}{(p+t-n-1)(n-p-1)}}$.
		\item For an overparameterized $\mathcal{L}$ with $p\ge n+2$:
		
		$\gamma_0=\frac{p-1}{p-n-1}$, $\gamma_1=- \frac{nt}{p(p+t)}$, ${\gamma_2=-\left( 1-\frac{n}{p+t}\right)}$, ${\gamma_3=\frac{nt}{(p+t-n-1)(p-n-1)}}$.
	\end{itemize}
	Accordingly, the transfer of the $t>0$ parameters in $\mathcal{T}\subseteq\mathcal{S}$ is more beneficial than allocating them as free parameters (i.e., $\Delta\mathcal{E}_{\rm TvsF}^{(\mathcal{L})}<0$) if 	
	\begin{equation}
		\label{eq:benefits in transfer versus free - Ltag is overparameterized - condition for benefits - general}
		\mathcal{E}_{\rm transfer}^{(\mathcal{T},\mathcal{S})} <   -\frac{\gamma_1}{\gamma_0} \Ltwonorm{\vecgreek{\beta}_{\mathcal{F}}}  -\frac{\gamma_2}{\gamma_0} \Ltwonorm{\vecgreek{\beta}_{\mathcal{T}}}	
		-\frac{\gamma_3}{\gamma_0} \left({\Ltwonorm{\vecgreek{\beta}_{\mathcal{Z}}}+\sigma_{\epsilon}^2}\right).
	\end{equation}
\end{corollary}

Clearly, the formulation of $\Delta\mathcal{E}_{\rm TvsF}^{(\mathcal{L})}$ is more intricate in the case of overparameterized $\mathcal{L}''$ in Corollary \ref{corollary:benefits in transfer versus free - Ltag is overparameterized} than in the case of underparameterized $\mathcal{L}$ and $\mathcal{L}''$ in Corollary \ref{corollary:benefits in transfer versus free - both underparameterized}.  Specifically, the set of free parameters $\mathcal{F}$ appears in (\ref{eq:benefits in transfer versus free - Ltag is overparameterized - error difference - general})-(\ref{eq:benefits in transfer versus free - Ltag is overparameterized - condition for benefits - general}) in addition to the sets $\mathcal{T}$ and $\mathcal{Z}$. 


\begin{figure}[t]
	\begin{center}		
		\subfloat[{\small$\mtx{H}$: local averaging \hspace{4em}neighborhood size  3}]{\includegraphics[height=0.3\textheight]{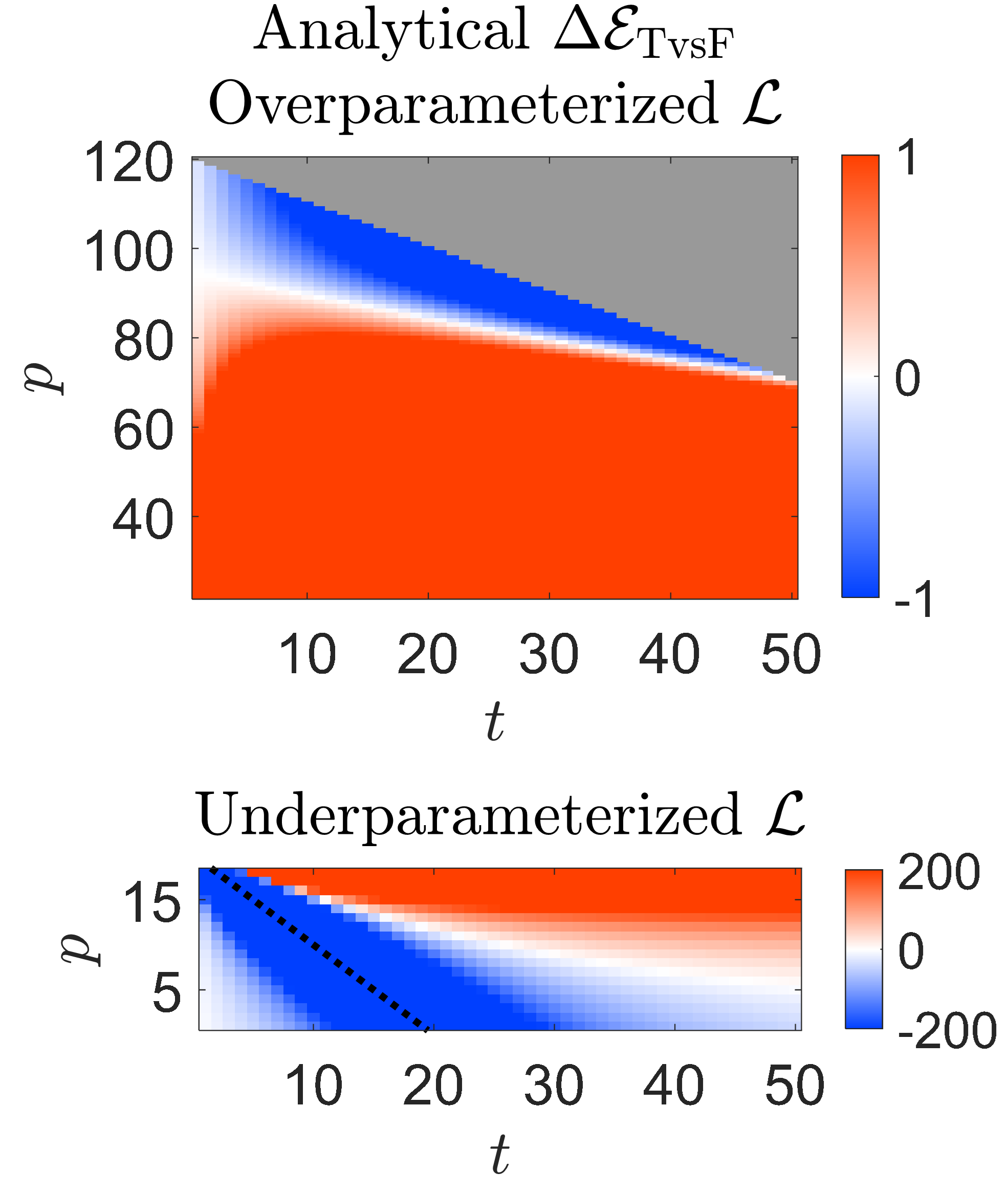}\label{fig:usefulness_transfer_vs_free_betaPWC_averaging3}}
		~~
		\subfloat[{\small$\mtx{H}$: local averaging \hspace{2em}neighborhood size  15}]{\includegraphics[height=0.3\textheight]{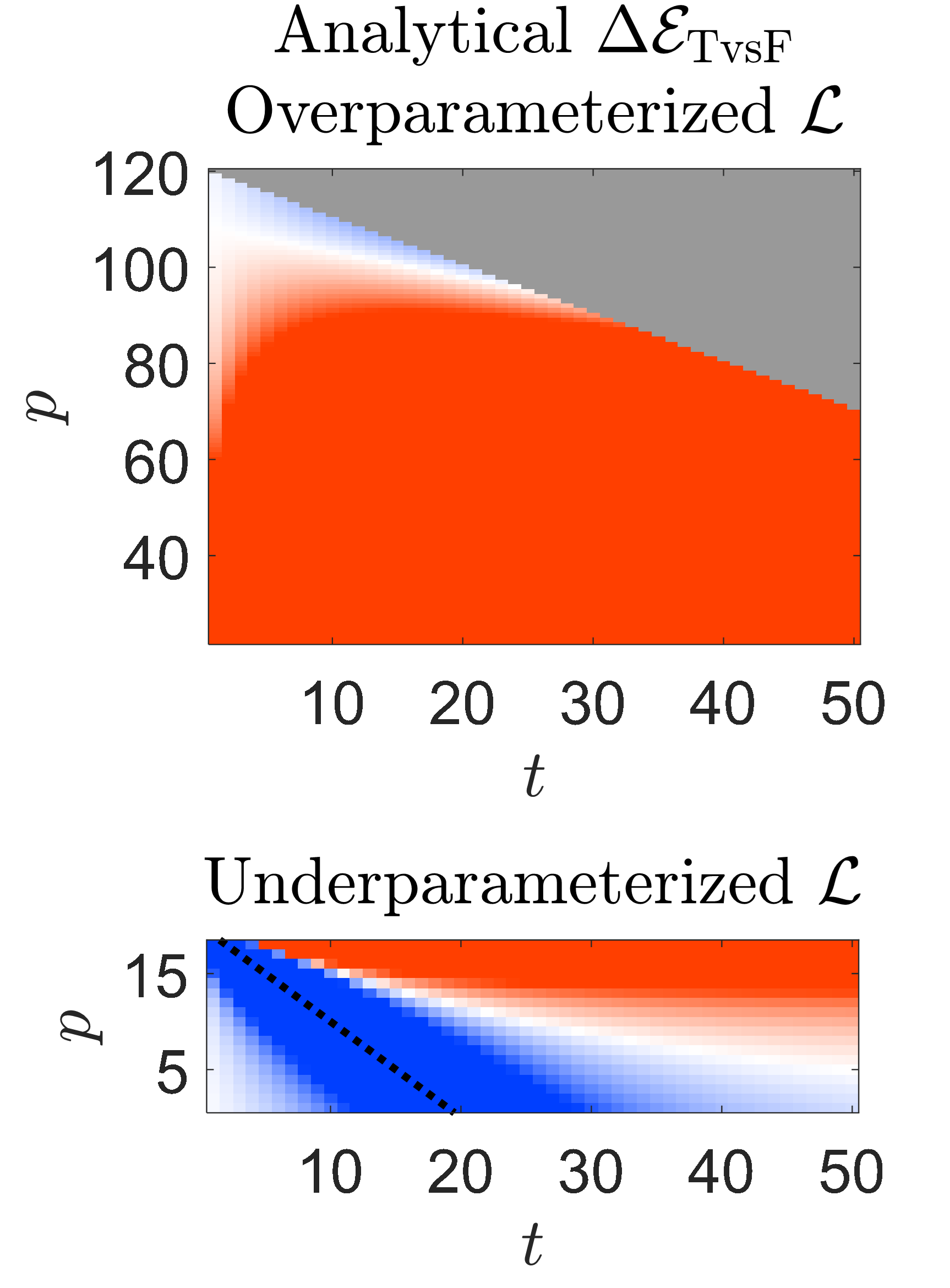}\label{fig:usefulness_transfer_vs_free_betaPWC_averaging15}}
		~~
		\subfloat[{\small$\mtx{H}=5\mtx{I}_d$}]{\includegraphics[height=0.3\textheight]{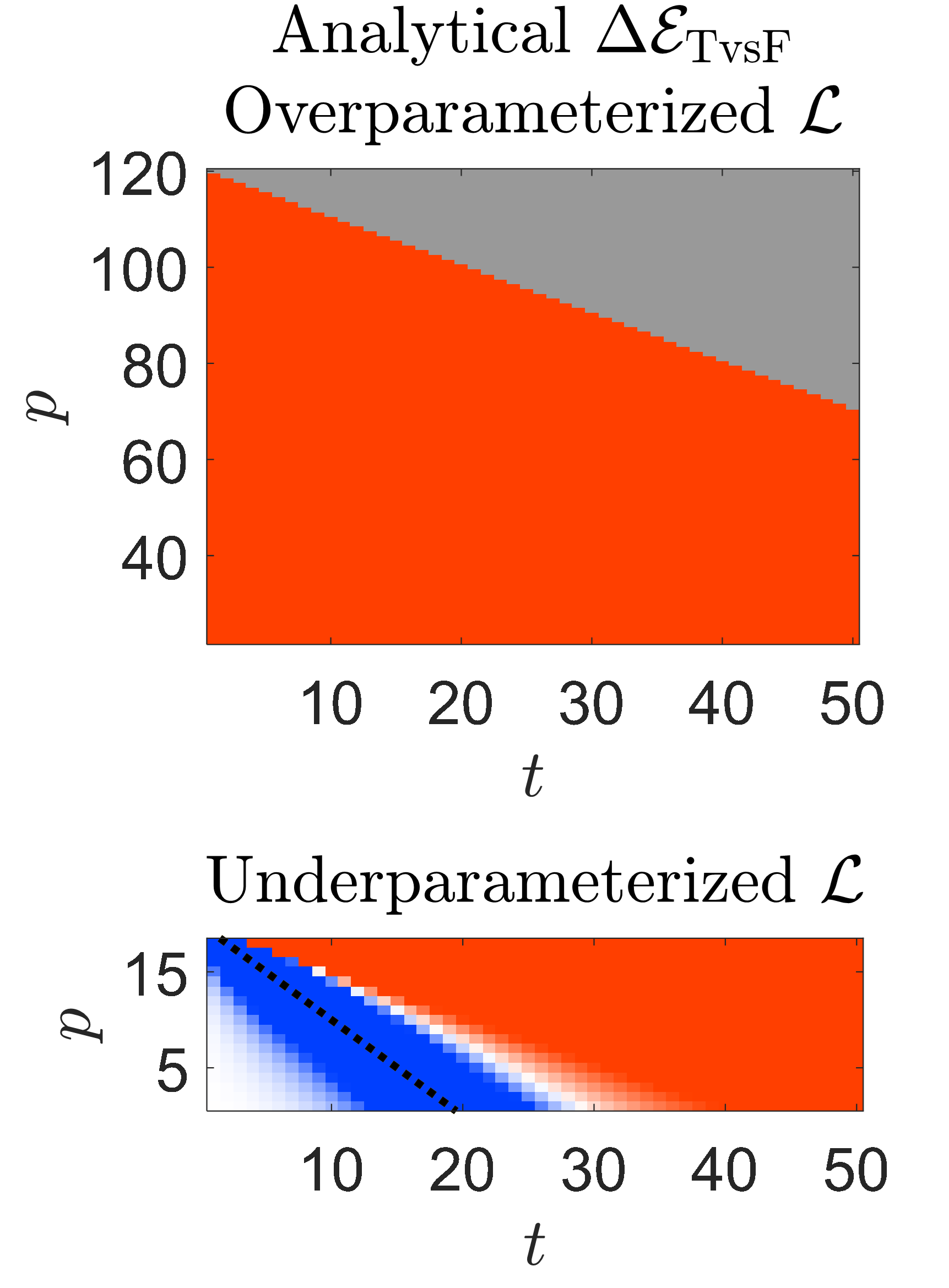}\label{fig:usefulness_transfer_vs_free_betaPWC_H_is_5I}}
		\caption{The analytical values of $\Delta\mathcal{E}_{\rm TvsF}$ (namely, the expected error difference due to transfer of an arbitrarily-selected set of $t$ parameters versus setting them as free parameters) as a function of $t$ and $p$. The positive and negative values of $\Delta\mathcal{E}_{\rm TvsF}$ appear in color scales of red and blue, respectively. 
			The regions of negative values (appear in shades of blue) correspond to beneficial transfer of parameters (compared to defining them as free parameters). 
			The gray regions correspond to $p+t>d$ where parameter transfer cannot be performed.
			For better visual clarity, the underparameterized and overparameterized cases of $\mathcal{L}$ are shown in different subfigures with significantly different range of values. 
			Also, the positive values were truncated at the values of 1 and 200 in the overparameterized and underparameterized cases, respectively. Corresponding truncations are applied on the negative values at the values of -1 and -200. 
			The dotted black line corresponds to $p+t=n$, which is the interpolation threshold of the auxiliary layout $\mathcal{L}''$. 
			Each column of subfigures corresponds to a different task relation model induced by the definitions of $\mtx{H}$.
			For all the subfigures, $d=120$, $\widetilde{p}=d$, $\widetilde{n}=50$, $\sigma_{\xi}^2 = 0.025\cdot d$, $n=20$, $\sigma_{\epsilon}^2 = 0.05\cdot d$, $\| \vecgreek{\beta} \|_2^2 = d$ where $\vecgreek{\beta}$ components have a piecewise-constant form (see Fig.~\ref{appendix:fig:piecewise_constant_beta_structure}). See corresponding empirical results in Fig.~\ref{fig:transfer_learning_usefulness_plane - transfer versus free - empirical} in Appendix \ref{appendix:subsec:Empirical Results for Benefits in Transferred versus Free Parameters}.}
		\label{fig:transfer_learning_usefulness_plane - transfer versus free}
	\end{center}
	\vspace*{-5mm}
\end{figure}

To intuitively understand the behavior of $\Delta\mathcal{E}_{\rm TvsF}^{(\mathcal{L})}$, consider expectation over coordinates that are selected uniformly at random (recall Definition \ref{definition:coordinate subset layout - uniformly distributed}). Specifically, we define $\Delta\mathcal{E}_{\rm TvsF}\triangleq \expectationwrt{\Delta\mathcal{E}_{\rm TvsF}^{(\mathcal{L})}}{\mathcal{L}}$ while noting that $\mathcal{L}''$ is deterministically related to $\mathcal{L}$ and hence it is sufficient to consider the expectation over $\mathcal{L}$. 
Figure \ref{fig:transfer_learning_usefulness_plane - transfer versus free} shows the value of $\Delta\mathcal{E}_{\rm TvsF}$ as a function of $p$ and $t$ for several operators $\mtx{H}$ in the task relation.
Figure \ref{fig:transfer_learning_usefulness_plane - transfer versus free} demonstrates the following typical behaviors: 
\begin{itemize}
	\item For \textit{underparameterized} $\mathcal{L}$ and $\mathcal{L}''$ (i.e., $p+t\le n-2$, which corresponds to the region to the left of the dotted black lines in the bottom subfigures in Fig.~\ref{fig:transfer_learning_usefulness_plane - transfer versus free}): Parameter transfer is more likely to be beneficial, and with larger gains, as $p+t$ increases towards $n-2$. Moreover, when (\ref{eq:benefits in transfer versus free - both underparameterized - condition for benefits}) is satisfied, it is beneficial to have more transferred than free parameters (i.e., having $t>p$ for a fixed sum of $p+t$). 
	The intuition here is that, when we are constrained to the underparameterized regime in both options (i.e., $\mathcal{L}$ and $\mathcal{L}''$), transferring parameters instead of having more free parameters can keep the error farther away from the peak of the double descent curve, which yields a beneficial transfer. Interestingly, due to the significant peak of the generalization error around the interpolation threshold, this behavior also occurs when the source and target tasks are quite different (see Fig.~\ref{fig:usefulness_transfer_vs_free_betaPWC_H_is_5I}). 
	
	\item For \textit{underparameterized} $\mathcal{L}$ and \textit{overparameterized} $\mathcal{L}''$ (i.e., $p\le n-2$ and $p+t\ge n+2$, which correspond to the region to the right of the dotted black lines in the bottom subfigures in Fig.~\ref{fig:transfer_learning_usefulness_plane - transfer versus free}): 
	Parameter transfer is more likely to degrade as $p$ increases towards $n-2$ while $p+t\ge n+2$. Moreover, increasing the number $t$ of transferred parameters degrades the benefits from transfer learning. 
	The intuition here is that, when the transfer option $\mathcal{L}$ is underparameterized and the no-transfer option $\mathcal{L}''$ is overparameterized, any increase in $p+t$ makes $\mathcal{L}''$ more overparameterized and hence farther away from the double descent peak (hence having an improved generalization ability). 
	
	\item For \textit{overparameterized} $\mathcal{L}$ and $\mathcal{L}''$ (i.e., $p\ge n+2$, which corresponds to the upper subfigures in Fig.~\ref{fig:transfer_learning_usefulness_plane - transfer versus free}): 
	Here, the important behavior is that, for a given $\mathcal{T}$, parameter transfer improves as the number of free parameters $p$ increases. 
	Of course that this improvement becomes beneficial (at high overparameterization levels) only when the source task is sufficiently related to the target task (see, e.g., Figures \ref{fig:usefulness_transfer_vs_free_betaPWC_averaging3} and \ref{fig:usefulness_transfer_vs_free_betaPWC_averaging15}).
	The intuition here is that, when both options (i.e., $\mathcal{L}$ and $\mathcal{L}''$) are in the overparameterized regime then the increase in $p$ moves both $\mathcal{L}$ and $\mathcal{L}''$ away from the peak of the double descent. The transfer learning option $\mathcal{L}$ is closer to the double descent peak and hence typically improves more (due to the steeper slope at the corresponding part of the generalization error curve, e.g., see in Fig.~\ref{fig:target_generalization_errors_vs_p__on_average}) than the no-transfer case $\mathcal{L}''$. 	
	
\end{itemize}

\section{The Optimal $\mtx{H}$ in a Componentwise Task Relation}
\label{sec:Additional Insights The Optimal H in a Componentwise Task Relation}

Theorem \ref{theorem:out of sample error - target task - specific layout} shows that the task relation aspect is encapsulated in the term $\mathcal{E}_{\rm transfer}^{(\mathcal{S},\mathcal{T})}$. Consequently, we demonstrated in Section \ref{sec:When is Transfer Learning Beneficial} that $\mathcal{E}_{\rm transfer}^{(\mathcal{S},\mathcal{T})}$ greatly affects the potential benefits from parameter transfer compared to both zero and free parameters. Specifically, $\Delta\mathcal{E}_{\rm TvsZ}^{(\mathcal{L})}$ and $\Delta\mathcal{E}_{\rm TvsF}^{(\mathcal{L})}$ both decrease as $\mathcal{E}_{\rm transfer}^{(\mathcal{S},\mathcal{T})}$ decreases. 
This motivates us to characterize the optimal task relation in the sense of minimum $\mathcal{E}_{\rm transfer}^{(\mathcal{S},\mathcal{T})}$ for a given coordinate layout $\mathcal{L}$. Namely, we characterize the best source task to transfer from when the other aspects of the parameter transfer are fixed.

Consider an operator $\mtx{H}$ that has the diagonal form of 
\begin{equation}
	\label{eq:diagonal form of H}
	\mtx{H}={\rm diag}\{\lambda_{\mtx{H}}^{(1)},\dots,\lambda_{\mtx{H}}^{(d)}\}
\end{equation}
where $\{\lambda_{\mtx{H}}^{(j)}\}_{j=1}^{d}\in\mathbb{R}$. 
We refer to $\{\lambda_{\mtx{H}}^{(j)}\}_{j=1}^{d}$ as the eigenvalues of $\mtx{H}$ due to the fact that applying the same orthonormal rotation on the feature spaces of the source and target tasks can induce a non-diagonal $\mtx{H}$ that its eigenvalues are the same $\{\lambda_{\mtx{H}}^{(j)}\}_{j=1}^{d}$ as in (\ref{eq:diagonal form of H}). 
The diagonal form in (\ref{eq:diagonal form of H}) implies that the task relation in Eq.~(\ref{eq:theta-beta relation}) reduces to the componentwise form of 
\begin{equation}
\label{eq:theta-beta relation - componentwise}
\theta^{(j)} = \lambda_{\mtx{H}}^{(j)} \beta^{(j)} + \eta^{(j)},~~~j=1,\dots,d 
\end{equation}
where $\theta^{(j)}$ and $\beta^{(j)}$ are the $j^{\rm th}$ components of the true parameter vectors of the source and target tasks, respectively, and $\eta^{(j)}$ is the $j^{\rm th}$ noise component in $\vecgreek{\eta}$. 
Then, we can further develop the expressions from Corollary \ref{corollary:out of sample error - target task - specific layout - detailed - specific layout}. First, the transfer bias term from (\ref{eq:out of sample error - target task - corollary - transfer bias - detailed - specific layout}) can be written as 
\begin{equation}
{\rm{Bias}}_{\mathcal{T}}^{2} =  {\sum_{j\in\mathcal{T}}{\left(r\lambda_{\mtx{H}}^{(j)} - 1\right)^2 \left( \beta^{(j)} \right)^2} }
~~~\text{where}~~ r \triangleq  \begin{cases}	
\mathmakebox[2em][l]{ 1 }    \text{for }  \widetilde{p} \le \widetilde{n},  
\\
\mathmakebox[2em][l]{ \frac{\widetilde{n}}{\widetilde{p}} }    \text{for }  \widetilde{p} > \widetilde{n}. 
\end{cases}  
\label{eq:componentwise transfer bias}
\end{equation}	
Second, the transfer variance term $\rm{Var}_{\mathcal{T},\mathcal{S}}$ from (\ref{eq:out of sample error - target task - corollary - transfer variance - detailed - specific layout}) can be also rewritten using 
\begin{align}
	\label{eq: zeta_S in terms of H eigenvalues}
	&\zeta_{\mathcal{T}}=\sum_{j\in\mathcal{T}} {\left({\lambda_{\mtx{H}}^{(j)} \beta^{(j)} }\right)^2},~~~\zeta_{\mathcal{S}\setminus\mathcal{T}}=\sum_{j\in\mathcal{S}\setminus\mathcal{T}} {\left({\lambda_{\mtx{H}}^{(j)} \beta^{(j)} }\right)^2},~~~\zeta_{\mathcal{S}^{c}}=\sum_{j\in\mathcal{S}^{c}} {\left({\lambda_{\mtx{H}}^{(j)} \beta^{(j)} }\right)^2}.
\end{align} 
Consequently, we get the following result (proof is provided in Appendix \ref{subsec:Proof of Theorem optimal eigenvalues of H}). 
\begin{theorem}
	\label{theorem:optimal eigenvalues of H}
	Consider a task relation operator of the form $\mtx{H}={\rm diag}\{\lambda_{\mtx{H}}^{(1)},\dots,\lambda_{\mtx{H}}^{(d)}\}$. Also,  $\widetilde{p}\notin\{\widetilde{n}-1,\widetilde{n},\widetilde{n}+1\}$. Then, for given coordinate sets $\mathcal{S}$ (of size $\widetilde{p}$) and $\mathcal{T}\subseteq\mathcal{S}$ (of size $t\le\widetilde{p}$), the transfer error term $\mathcal{E}_{\rm transfer}^{(\mathcal{T},\mathcal{S})}$ attains its minimal value with respect to the eigenvalues of $\mtx{H}$ at 
	\begin{align}
		\label{eq: optimal H eigenvalue in T}
		&\text{for}~j\in\mathcal{T},~\beta^{(j)}\neq 0: &&\lambda_{\mtx{H}}^{(j)} = \begin{cases}	
			1        & \text{for }  \widetilde{p} \le \widetilde{n}-2,  \\
			\frac{\widetilde{p}^2 - 1}{\widetilde{n}\widetilde{p}-1+ t \left(\widetilde{p}-\widetilde{n}\right)}          & \text{for } \widetilde{p} \ge \widetilde{n}+2,
		\end{cases}
		\\
		\label{eq: optimal H eigenvalue in SminusT}
		&\text{for}~j\in\mathcal{S}\setminus\mathcal{T},~\beta^{(j)}\neq 0: &&\lambda_{\mtx{H}}^{(j)} = \begin{cases}	
			\text{any value}        & \text{for }  \widetilde{p} \le \widetilde{n}-2,  \\
			0          & \text{for } \widetilde{p} \ge \widetilde{n}+2,
		\end{cases}
		\\
		\label{eq: optimal H eigenvalue in Sc}
		&\text{for}~j\in\mathcal{S}^{c},~\beta^{(j)}\neq 0: &&\lambda_{\mtx{H}}^{(j)} = 0.
	\end{align}
	For $j\in\{1,\dots,d\}$ where ${\beta^{(j)}=0}$, $\lambda_{\mtx{H}}^{(j)}$ can have any value.
\end{theorem}

Let us interpret the meaning of Theorem \ref{theorem:optimal eigenvalues of H} for the case of $\beta^{(j)}\neq 0$ for $j\in\{1,\dots,d\}$. 
The theorem shows that the linear operator in the optimal task relation has two different characterizations depending on whether the \textit{source} task is under or over parameterized: 
\begin{itemize}
	\item For an \textit{underparameterized source} task it is best that the true parameters in the transferred coordinates of the source task are the \textit{same} (up to the additive noise terms from $\vecgreek{\eta}$) as the corresponding true parameters of the target task (see (\ref{eq: optimal H eigenvalue in T}) and recall (\ref{eq:theta-beta relation - componentwise})).
	
	\item For an \textit{overparameterized source} task it is best that the true parameters in the transferred coordinates of the source task are \textit{amplified} versions by a factor of $\frac{\widetilde{p}^2 - 1}{\widetilde{n}\widetilde{p}-1+ t \left(\widetilde{p}-\widetilde{n}\right)}$ (see (\ref{eq: optimal H eigenvalue in T})) of the corresponding true parameters of the target task.
	This amplification intends to partially compensate for the increased transfer bias in the case of overparameterized source task (see the $\widetilde{p}>\widetilde{n}$ case in (\ref{eq:componentwise transfer bias})). Indeed, the optimal amplification increases as the source task is more overparameterized (i.e., as $p$ increases towards $d$). 
	While this amplification reduces the transfer bias, it increases the transfer variance (see (\ref{eq:out of sample error - target task - corollary - transfer variance - detailed - specific layout}) and (\ref{eq: zeta_S in terms of H eigenvalues})) and hence the amplification is somewhat restrained. 
	 Also, the optimal amplification $\frac{\widetilde{p}^2 - 1}{\widetilde{n}\widetilde{p}-1+ t \left(\widetilde{p}-\widetilde{n}\right)}$  decreases as the number $t$ of transferred parameters increases. Specifically, when $t=\widetilde{p}$ (i.e., all the free parameters of the source task are transferred to the target task) the optimal $\lambda_{\mtx{H}}^{(j)}$ values for $j\in\mathcal{T}$ become 1 and there is no amplification. 
\end{itemize}

Eq.~(\ref{eq: optimal H eigenvalue in SminusT}) considers the coordinates of the free parameters of the source task that are \textit{not} transferred (i.e., $\mathcal{S}\setminus\mathcal{T}$) and shows that the optimal true parameters of the source task in these coordinates are zeros in the overparameterized regime due to the dependency of the transfer variance on the true parameters of the source task at $\mathcal{S}\setminus\mathcal{T}$ (see (\ref{eq:out of sample error - target task - corollary - transfer variance - detailed - specific layout}) and (\ref{eq: zeta_S in terms of H eigenvalues})).  
In the underparameterized regime of the source task, there is no such dependency on  $\vecgreek{\theta}_{\mathcal{S}\setminus\mathcal{T}}$ and hence these true parameters may have any value without affecting the minimization of $\mathcal{E}_{\rm transfer}^{(\mathcal{T},\mathcal{S})}$. 

According to Eq.~(\ref{eq: optimal H eigenvalue in Sc}), in both the under and over parameterized cases, it is best that the true parameters of the source task are all zeros in the coordinates that do not participate in the source task solution (i.e., $\mathcal{S}^{c}$).  
This optimal form reduces the misspecification in the solution of the source task to originate only in the task relation noise $\vecgreek{\eta}_{\mathcal{S}^{c}}$. Avoiding these misspecifications makes the estimate of each free source parameter closer to its true value (i.e., instead of also trying to compensate for omitted parameters that are informative) and, hence, this improves the relevance of the parameters $\widehat{\vecgreek{\theta}}_{\mathcal{T}}$ that are transferred ``as is" to the target task from the co-located coordinates in the source task. 

\begin{figure}[t]
	\begin{center}		
		\subfloat[{\small$\sigma_{\eta}^2 = 0.2$;  $\vecgreek{\beta}$ has the form in~Fig.~\ref{fig:beta_all_ones_usefulness}}]{\includegraphics[width=0.31\textwidth]{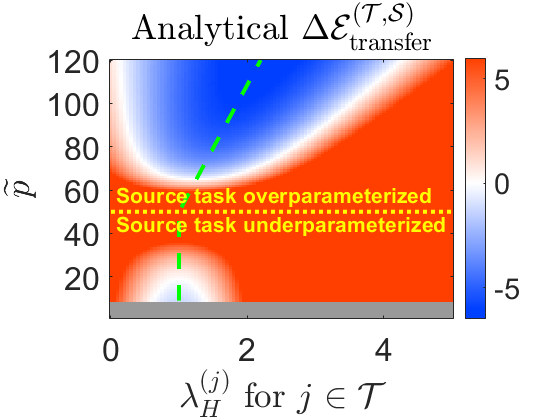}\label{fig:eigenvals_usefulness_beta_all__ones_t8__eta0_2}}
		\subfloat[{\small$\sigma_{\eta}^2 = 1$; $\vecgreek{\beta}$ has the form in~Fig.~\ref{fig:beta_all_ones_usefulness} }]{\includegraphics[width=0.28\textwidth]{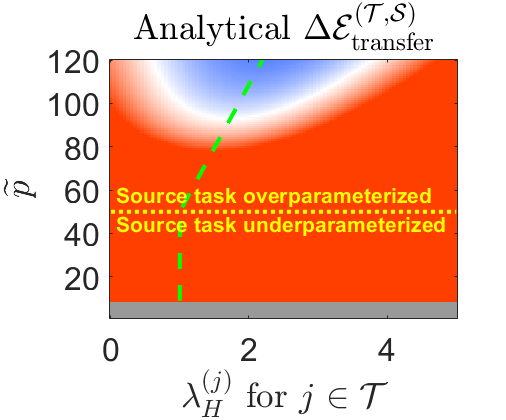}\label{fig:eigenvals_usefulness_beta_all__ones_t8__eta1}}			
		\subfloat[{\small$\sigma_{\eta}^2 = 0.2$;  $\vecgreek{\beta}$ has the form in~Fig.~\ref{fig:beta_high_in_non_transferred__usefulness} }]{\includegraphics[width=0.28\textwidth]{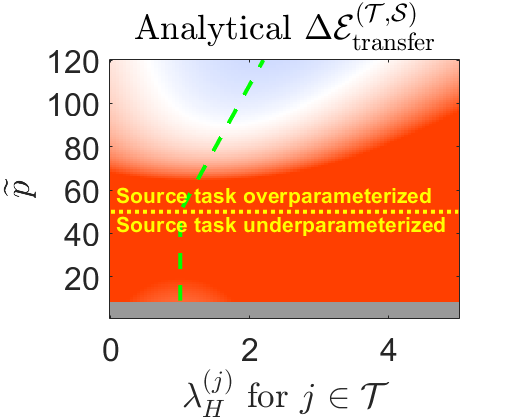}\label{fig:eigenvals_usefulness_beta_high_in_non_transferred_t8__eta0_2}}
		\\~\\~\\
		\subfloat[$\mtx{H}$ eigenvalues]{\includegraphics[width=0.28\textwidth]{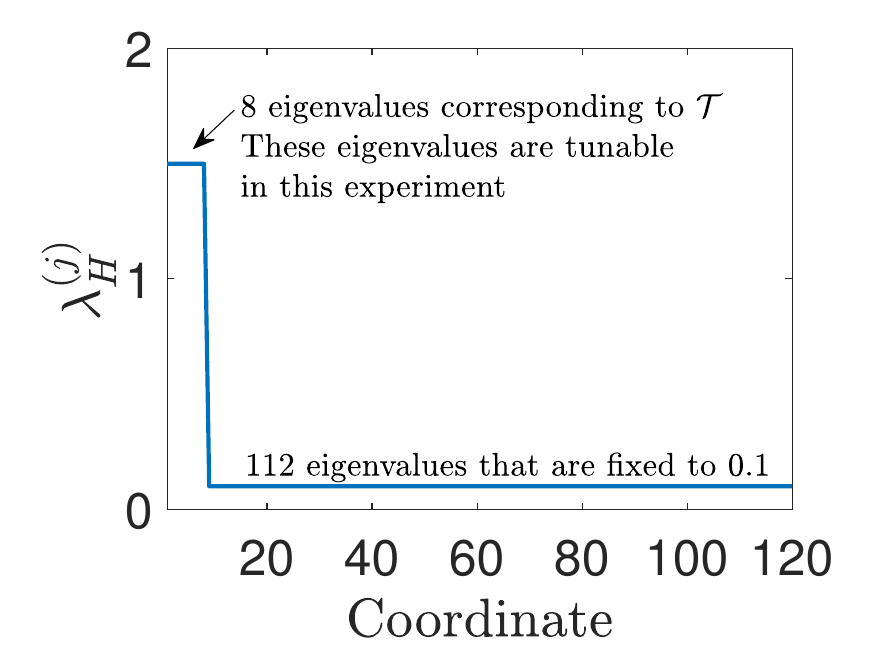}\label{fig:usefulness_H_eigenvalues_structure}}
		\subfloat[$\vecgreek{\beta}$ with all components ones]{\includegraphics[width=0.28\textwidth]{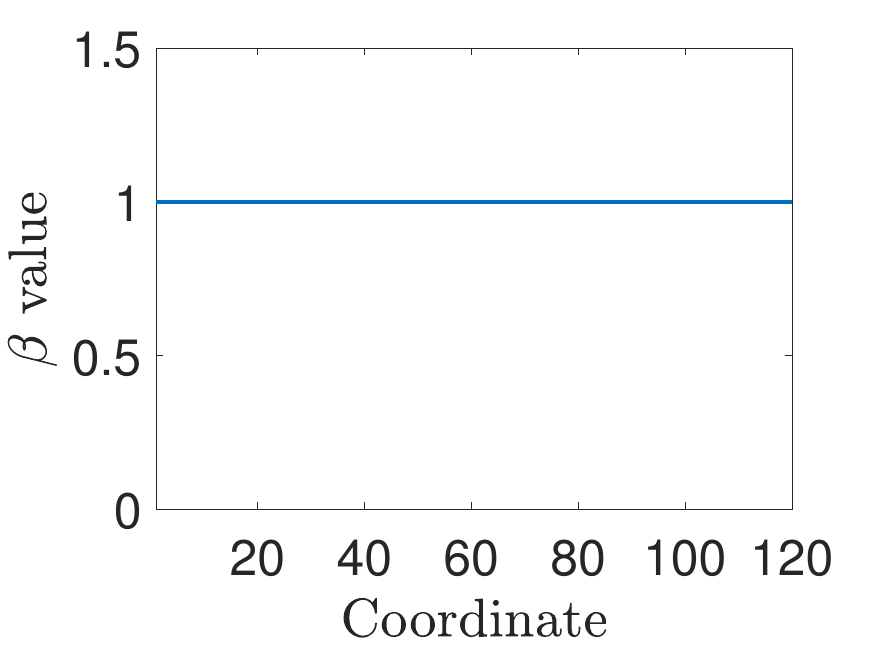}\label{fig:beta_all_ones_usefulness}}			
		\subfloat[$\vecgreek{\beta}$ with components in $\mathcal{T}^{c}$ are x2 larger than in $\mathcal{T}$]{\includegraphics[width=0.28\textwidth]{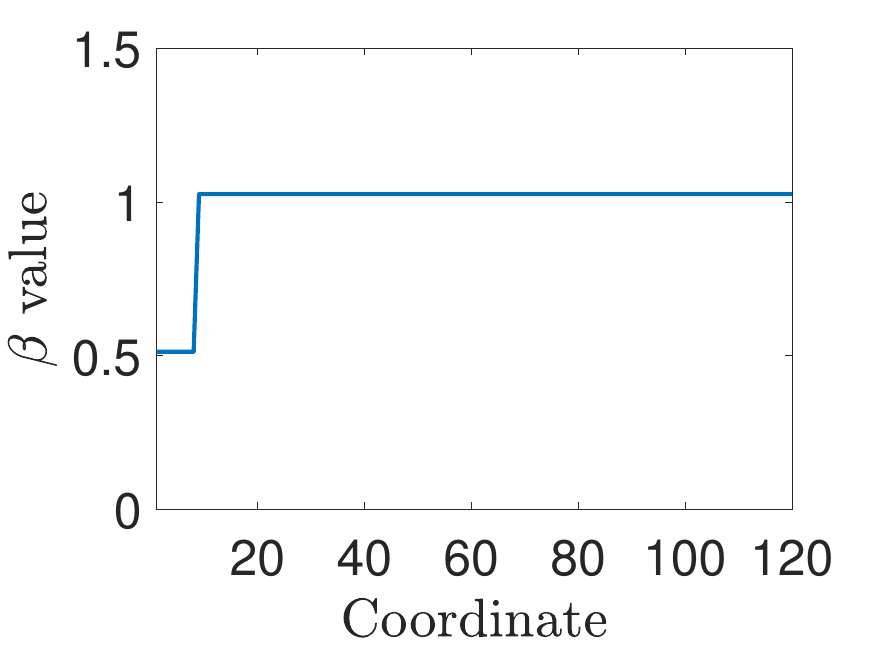}\label{fig:beta_high_in_non_transferred__usefulness}}
		\caption{The analytical values of $\Delta\mathcal{E}_{\rm transfer}^{(\mathcal{T},\mathcal{S})}$ as a function of $\widetilde{p}$ and a value that determines the eigenvalues of $\mtx{H}$ in $\mathcal{T}$.
			Here $\mathcal{T}=\{1,\dots,8\}$. The operator $\mtx{H}$ is diagonal with a main diagonal in the form described in subfigure \textit{(d)}.
			In subfigures \textit{(a)-(c)}, the positive and negative values of $\Delta\mathcal{E}_{\rm transfer}^{(\mathcal{T},\mathcal{S})}$ appear in color scales of red and blue, respectively. The color scales are the same for all the subfigures in the first row.
			The regions of negative values (appear in shades of blue) correspond to beneficial transfer of parameters. 
			The positive values were truncated at the value of 6 for the clarity of visualization. The gray regions correspond to $\widetilde{p}<t$ where parameter transfer cannot be performed. The dashed green lines (in all subfigures) denote the optimal eigenvalues (corresponding to $\mathcal{T}$) as formulated in Theorem \ref{theorem:optimal eigenvalues of H}. 			
			Here, $d=120$, $\widetilde{n}=50$, $\sigma_{\xi}^2 = 0.025\cdot d$. 
			Subfigure \textit{(e)} shows the form of $\vecgreek{\beta}$ in the experiments of subfigures \textit{(a),(b)}. Subfigure \textit{(f)} shows the form of $\vecgreek{\beta}$ in the experiments of subfigure \textit{(c)}. 
			The corresponding empirical evaluations are provided in Fig.~\ref{appendix:fig:transfer_learning_usefulness_plane - in terms of H eigenvalues - empirical} in Appendix \ref{appendix:subsec:Empirical Results for The Optimal H in a Componentwise Task Relation}.
		}
		\label{fig:transfer_learning_usefulness_plane - in terms of H eigenvalues}
	\end{center}
	\vspace*{-5mm}
\end{figure}

Theorem \ref{theorem:optimal eigenvalues of H} describes the eigenvalues of $\mtx{H}$ that minimize $\mathcal{E}_{\rm transfer}^{(\mathcal{T},\mathcal{S})}$. However, $\mathcal{E}_{\rm transfer}^{(\mathcal{T},\mathcal{S})}$ also depends on additional factors such as the noise level in the task relation and the true parameters $\vecgreek{\beta}$ of the target task, which determine whether transferring the parameters in $\mathcal{T}$ is preferred, e.g., over zeroing them. Thus, evaluation of the sign of $\Delta\mathcal{E}_{\rm TvsZ}^{(\mathcal{L})}$ is required for understanding whether the optimal eigenvalues of $\mtx{H}$ induce beneficial transfer and, if so, how much can the eigenvalues deviate from their optimal values while still having a beneficial transfer. 

In Fig.~\ref{fig:transfer_learning_usefulness_plane - in terms of H eigenvalues} we show the analytical values of $\Delta\mathcal{E}_{\rm transfer}^{(\mathcal{T},\mathcal{S})}$ (recall the definition in (\ref{eq:error difference term - transfer versus zero - auxiliary term})) as a function of $\widetilde{p}$ and the eigenvalues of $\mtx{H}$ that correspond to the transferred parameters (see Fig.~\ref{appendix:fig:transfer_learning_usefulness_plane - in terms of H eigenvalues - empirical} in Appendix \ref{appendix:subsec:Empirical Results for The Optimal H in a Componentwise Task Relation} for the corresponding empirical evaluations). Specifically, we consider a family of $\mtx{H}$ operators that are diagonal in the feature domain and have the following step-shaped structure for their eigenvalues (see Fig.~\ref{fig:usefulness_H_eigenvalues_structure}): the first 8 eigenvalues are all equal to a value that varies among the $\mtx{H}$ operators in this family; the next 112 eigenvalues are $0.1$ for any $\mtx{H}$ operator in this family. Here we consider transfer of the first eight parameters, i.e., $\mathcal{T}=\{1,\dots,8\}$ that correspond to the eigenvalues with a tunable value in our evaluations (this is reflected by the horizontal axes in Figs.~\ref{fig:eigenvals_usefulness_beta_all__ones_t8__eta0_2}-\ref{fig:eigenvals_usefulness_beta_high_in_non_transferred_t8__eta0_2}).
The results in Fig.~\ref{fig:transfer_learning_usefulness_plane - in terms of H eigenvalues} demonstrate that the range of eigenvalues that induce beneficial transfer increases together with overparameterization (see the wider blue regions around the green dashed line that denotes the optimal eigenvalue from Theorem \ref{theorem:optimal eigenvalues of H}). Fig.~\ref{fig:eigenvals_usefulness_beta_all__ones_t8__eta0_2} shows that benefits can be also obtained in underparameterized settings where the eigenvalues (in $\mathcal{T}$) are around 1, however, these eigenvalue regions are smaller and produce lower benefits than their overparameterized alternatives. 
The effect of the noise in the task model is also apparent by comparing Fig.~\ref{fig:eigenvals_usefulness_beta_all__ones_t8__eta0_2} to Fig.~\ref{fig:eigenvals_usefulness_beta_all__ones_t8__eta1} that shows results for the same settings but with a higher $\sigma_{\eta}^2$. Specifically, the higher noise level in the task relation reduces the size of the beneficial regions and their gains in the overparameterized range, and leaves the underparameterized range without any beneficial regions. 
Next, compare Fig.~\ref{fig:eigenvals_usefulness_beta_all__ones_t8__eta0_2} to Fig.~\ref{fig:eigenvals_usefulness_beta_high_in_non_transferred_t8__eta0_2} that shows results for the same settings but with $\vecgreek{\beta}$ of the less favorable structure in Fig.~\ref{fig:beta_high_in_non_transferred__usefulness} instead of the structure in Fig.~\ref{fig:beta_all_ones_usefulness}.
This shows how the structure of the true parameters in $\vecgreek{\beta}$ can also affect the parameter transfer performance.

\section{Additional Linear Regression Methods}
\label{sec:Additional Linear Regression Methods}

The previous sections presented analytical and empirical results on transfer learning between two least squares solutions that take their minimum $\ell_2$-norm forms (\ref{eq:constrained linear regression - solution - source data class}), (\ref{eq:constrained linear regression - solution - target task}) in the overparameterized regime. 
In this section we examine the same transfer learning mechanism (i.e., transferring co-located parameters from the already-learned source model to the to-be-learned target model) for two more kinds of solutions to linear regression: least squares with the minimum $\ell_1$-norm form in the overparameterized regime, and ridge regression. 

\subsection{The Minimum $\ell_1$-Norm Interpolator}
\label{subsec:The Minimum l1-Norm Interpolator}

This setting also emerges from the optimization problems formulated in (\ref{eq:constrained linear regression - source data class}) and (\ref{eq:constrained linear regression - target task}) for the source and target tasks, respectively. In the underparameterized regime, the source and target tasks almost surely have the unique least squares solutions as provided in (\ref{eq:constrained linear regression - underparameterized unique solution - source data class}) and (\ref{eq:constrained linear regression - underparameterized unique solution - target task}). 

For an overparameterized source task, i.e., $\widetilde{p}>\widetilde{n}$, the problem (\ref{eq:constrained linear regression - source data class}) has infinite interpolating solutions among which we choose here the minimum $\ell_1$-norm solution: 
\begin{align} 
\label{eq:constrained linear regression - source task - overparameterized - min l1 norm}
&\widehat{\vecgreek{\theta}} = \argmin_{\vec{r}\in\mathbb{R}^{d}} \left \Vert r  \right \Vert _1
\\ \nonumber
&\mathmakebox[5em][l]{\text{subject to}}\mtx{Q}_{\mathcal{S}^{\rm c}} \vec{r} = \vec{0}
\\ \nonumber
&\qquad\qquad\quad\mtx{Z}\vec{r} = \vec{v}.
\end{align}
This solution equals to $\widehat{\vecgreek{\theta}}$ where $\widehat{\vecgreek{\theta}}_{\mathcal{S}^c}=\vec{0}$ and ${\widehat{\vecgreek{\theta}}_{\mathcal{S}}=\argmin_{\vec{k}\in\mathbb{R}^{\widetilde{p}}} \left \Vert  \vec{k} \right \Vert _1 {~\text{subject to}~} \mtx{Z}_{\mathcal{S}}\vec{k} = \vec{v}}$,
which is a basis pursuit problem \cite{chen2001atomic} that can be addressed via linear programming methods. 

For an overparameterized target task, i.e., $p>n$, the problem (\ref{eq:constrained linear regression - target task}) has infinite interpolating solutions, from which we choose here the minimum $\ell_1$-norm solution 
\begin{align} 
\label{eq:constrained linear regression - target task - overparameterized - min l1 norm}
&\widehat{\vecgreek{\beta}} = \argmin_{\vec{b}\in\mathbb{R}^{d}} \left \Vert  \vec{b} \right \Vert _1
\\ \nonumber
&\mathmakebox[5em][l]{\text{subject to}}\mtx{Q}_{\mathcal{T}} \vec{b} = \mtx{Q}_{\mathcal{T}}\widehat{\vecgreek{\theta}}
\\ \nonumber
&\qquad\qquad\quad\mtx{Q}_{\mathcal{Z}} \vec{b} = \vec{0}
\\ \nonumber
&\qquad\qquad\quad\mtx{X}\vec{b} = \vec{y}
\end{align}
that can be formulated also as $\widehat{\vecgreek{\beta}}$ where $\widehat{\vecgreek{\beta}}_{\mathcal{Z}}=\vec{0}$, $\widehat{\vecgreek{\beta}}_{\mathcal{T}}=\widehat{\vecgreek{\theta}}_{\mathcal{T}}$ and 
\begin{align} 
\label{eq:constrained linear regression - target task - overparameterized - min l1 norm - F subvector}
&\widehat{\vecgreek{\beta}}_{\mathcal{F}} = \argmin_{\vec{f}\in\mathbb{R}^{p}} \left \Vert  \vec{f} \right \Vert _1
\\ \nonumber
&\mathmakebox[5em][l]{\text{subject to}} \mtx{X}_{\mathcal{F}}\vec{f} = \vec{y} - \mtx{X}_{\mathcal{T}}\widehat{\vecgreek{\theta}}_{\mathcal{T}},
\end{align}
which is a basis pursuit problem that can be solved via linear programming. 

\begin{figure}[!ht]
	\parbox{0.95\textwidth}{
		\parbox{.18\textwidth}{%
			\subfloat{\includegraphics[width=0.17\textwidth]{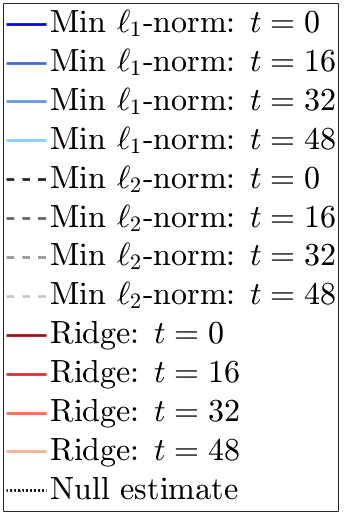}}
		}~~
		\parbox{.75\textwidth}{%
			\setcounter{subfigure}{0}
			\subfloat[{\small$\mtx{H}=\mtx{I}_{d}$}]{\includegraphics[width=0.24\textwidth]{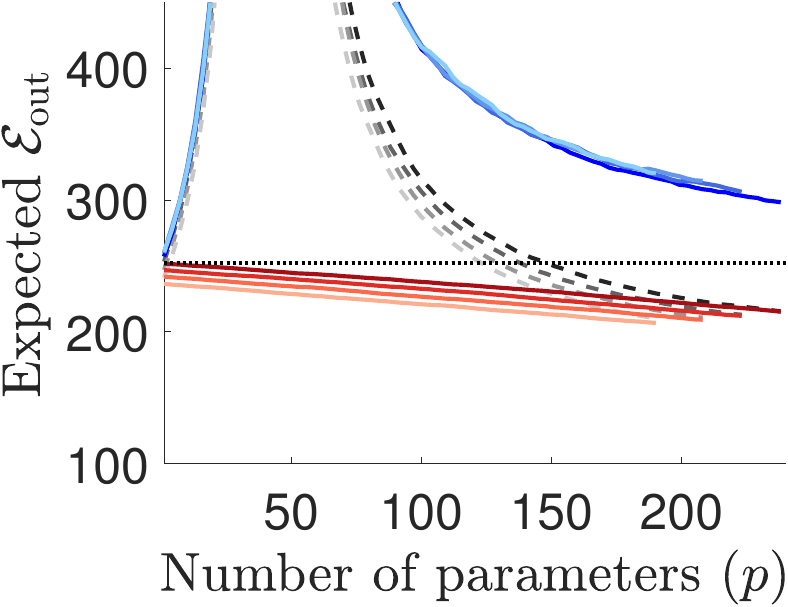} \label{fig:target_generalization_errors_vs_p_eta_0.2linearbeta_H_is_I_allL1L2Ridge}}
			\subfloat[{\small$\mtx{H}$:~Discrete derivative}]{\includegraphics[width=0.24\textwidth]{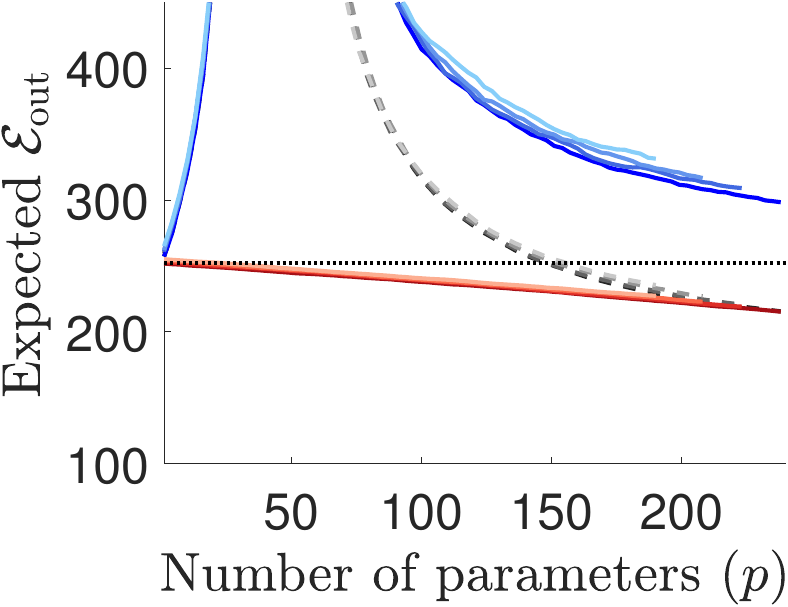} \label{fig:target_generalization_errors_vs_p_eta_0.2linearbeta_H_is_derivative_allL1L2Ridge}}
			\subfloat[{\small$\mtx{H}$:~Local averaging}]{\includegraphics[width=0.24\textwidth]{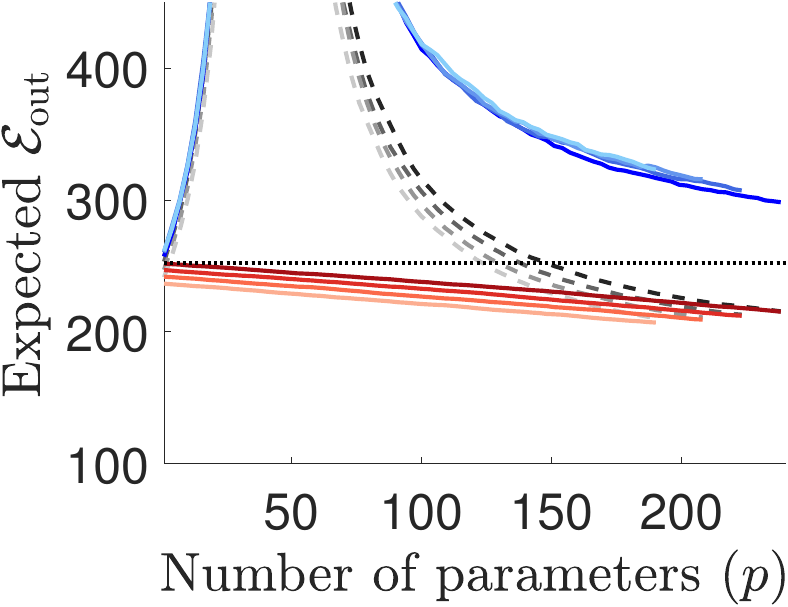} \label{fig:target_generalization_errors_vs_p_eta_0.2linearbeta_H_is_averaging11_allL1L2Ridge}}
			\\				
			\subfloat[{\small$\mtx{H}=\mtx{I}_{d}$}]{\includegraphics[width=0.24\textwidth]{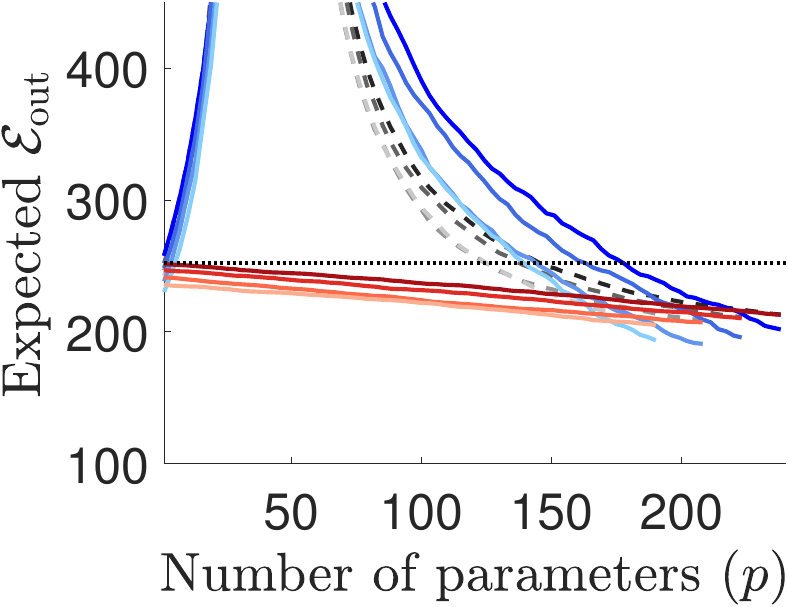} \label{fig:target_generalization_errors_vs_p_eta_0.2sparsebeta_H_is_I_allL1L2Ridge}}
			\subfloat[{\small$\mtx{H}$:~Discrete derivative}]{\includegraphics[width=0.24\textwidth]{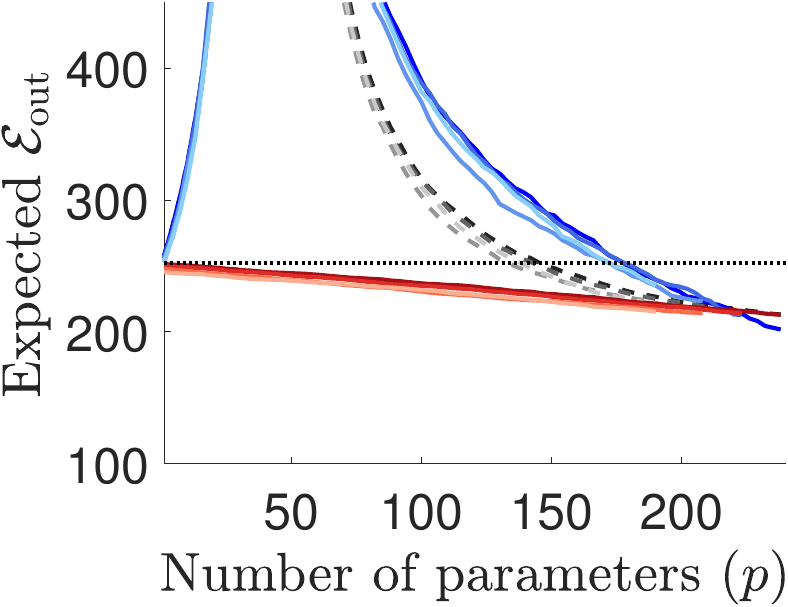} \label{fig:target_generalization_errors_vs_p_eta_0.2sparsebeta_H_is_derivative_allL1L2Ridge}}
			\subfloat[{\small$\mtx{H}$:~Local averaging}]{\includegraphics[width=0.24\textwidth]{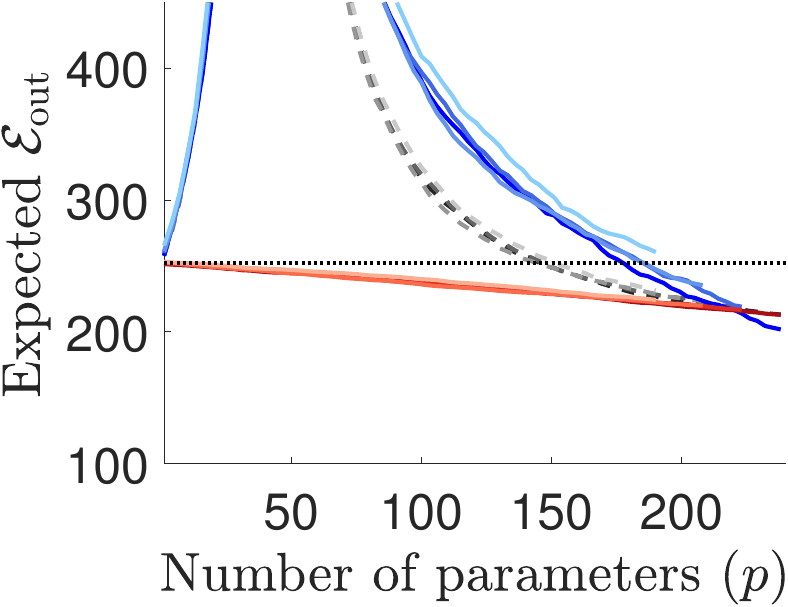} \label{fig:target_generalization_errors_vs_p_eta_0.2sparsebeta_H_is_averaging11_allL1L2Ridge}}				
			\\
			\\				
			\subfloat[{\small$\mtx{H}=\mtx{I}_{d}$}]{\includegraphics[width=0.24\textwidth]{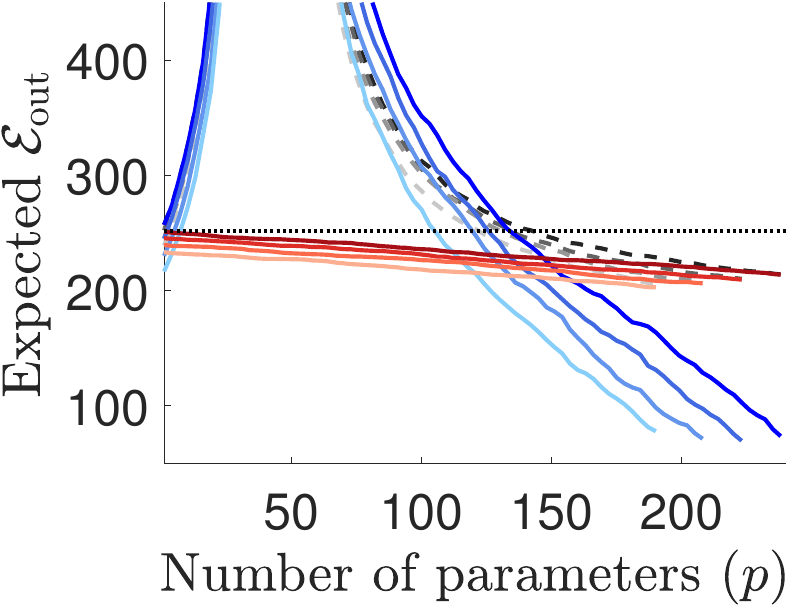} \label{fig:target_generalization_errors_vs_p_eta_0_2_VERYsparsebeta_H_is_I_allL1L2Ridge}}
			\subfloat[{\small$\mtx{H}$:~Discrete derivative}]{\includegraphics[width=0.24\textwidth]{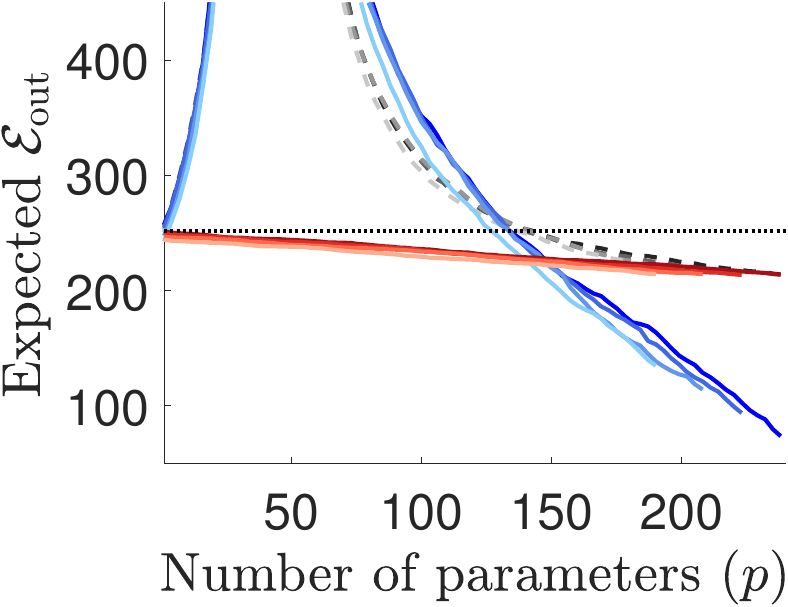} \label{fig:target_generalization_errors_vs_p_eta_0_2_VERYsparsebeta_H_is_derivative_allL1L2Ridge}}
			\subfloat[{\small$\mtx{H}$:~Local averaging}]{\includegraphics[width=0.24\textwidth]{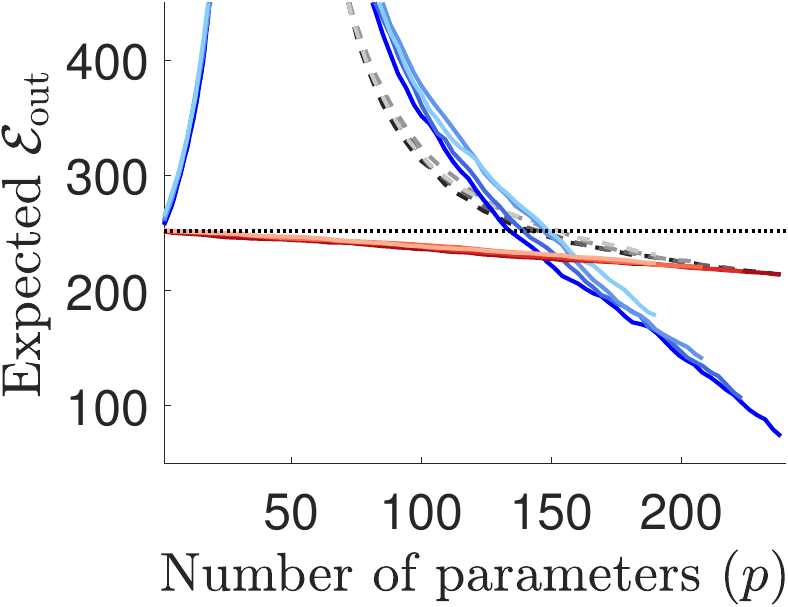} \label{fig:target_generalization_errors_vs_p_eta_0_2_VERYsparsebeta_H_is_averaging11_allL1L2Ridge}}
		}	 
	}
	\caption{The expected generalization error of the target task, $\expectationwrt{\mathcal{E}_{\rm out}}{\mathcal{L}}$, compared for minimum $\ell_2$-norm, minimum $\ell_1$-norm, and ridge solutions. In this figure all the errors are empirically evaluated. 
		The errors of the minimum $\ell_1$-norm solution are shown in curves of blue shades.
		The errors of the ridge regression are shown in curves of red shades.
		The errors of the minimum $\ell_2$-norm solution are shown in dashed curves of gray shades.
		For each of the solution types the darkest shade corresponds to no transfer ($t=0$), a lighter shade denotes more transferred parameters (larger $t$), and the lightest shade corresponds to $t=48$. 
		In the first row of subfigures the true $\vecgreek{\beta}$ has values that follow a linear form.
		In the second row of subfigures the true $\vecgreek{\beta}$ has a sparse form with non-zero values at $25\%$ of the coordinates (selected randomly out of the $d=240$). In the third row of subfigures the true $\vecgreek{\beta}$  is sparse with only $5\%$ non-zero values.  Examples for linear and sparse $\vecgreek{\beta}$ of a smaller dimension are provided in Figs.~\ref{appendix:fig:linear_beta_graph}, \ref{appendix:fig:sparse_beta_graph}. In each row there are three error plots for different circulant forms of the operator $\mtx{H}$ (here the local averaging is defined for a neighborhood of 11 samples). 
		Here $\sigma_{\eta}^2 = 0.2$, $d=240$, ${n=40}$, $\widetilde{n}=100$, $\| \vecgreek{\beta} \|_2^2 = d$, $\sigma_{\epsilon}^2 = 0.05\cdot d$, $\sigma_{\xi}^2 = 0.025\cdot d$, and $\widetilde{p}=d$  for all settings. }
	\label{fig:min l1 norm and ridge regression error curves - on average}
\end{figure}

The empirical out-of-sample errors of the minimum $\ell_1$-norm transfer learning are shown in Figure \ref{fig:min l1 norm and ridge regression error curves - on average} as blue curves where the shade of blue denotes the number of transferred parameters. 
The results demonstrate that the double descent behavior occurs also for the minimum $\ell_1$-norm solution. Double descent phenomena for minimum $\ell_1$-norm solutions to linear regression (without transfer learning) were studied in \cite{muthukumar2020harmless,mitra2019understanding}. 
The asymptotic analyses in \cite{li2021minimum,wang2022tight} show that a triple descent can be observed if the true parameters are sufficiently sparse. 
Here we do not clearly observe an additional error peak in the overparameterized regime, which could possibly be due to the misspecification strategy (i.e., zeroing parameters in predetermined coordinates) that we use for controlling the parameterization levels in our non-asymptotic setting. 
For $\vecgreek{\beta}$ of a linear shape the minimum $\ell_2$-norm solution is better than the minimum $\ell_1$-norm solution in the entire overparameterized range (see Figs.~\ref{fig:target_generalization_errors_vs_p_eta_0.2linearbeta_H_is_I_allL1L2Ridge}-\ref{fig:target_generalization_errors_vs_p_eta_0.2linearbeta_H_is_averaging11_allL1L2Ridge}). 
Nevertheless, for a sparse $\vecgreek{\beta}$ the minimum $\ell_1$-norm solution can provide the best performance in the high overparameterization levels due to its sparsity-promoting nature (see Figs.~\ref{fig:target_generalization_errors_vs_p_eta_0.2sparsebeta_H_is_I_allL1L2Ridge}, \ref{fig:target_generalization_errors_vs_p_eta_0_2_VERYsparsebeta_H_is_I_allL1L2Ridge}-\ref{fig:target_generalization_errors_vs_p_eta_0_2_VERYsparsebeta_H_is_averaging11_allL1L2Ridge}).

\subsection{Ridge Regression}
\label{subsec:Ridge Regression}

Now we turn to formulate our transfer learning approach for ridge regression. 
So far we considered the optimization problems (\ref{eq:constrained linear regression - source data class}) and (\ref{eq:constrained linear regression - target task}) that do not include explicit regularization terms in their cost functions. Hence, the ridge regression extension of (\ref{eq:constrained linear regression - source data class}) is 
\begin{align} 
\label{eq:constrained linear regression - source - ridge regression}
\widehat{\vecgreek{\theta}} = \argmin_{\vec{r}\in\mathbb{R}^{d}} \left \Vert  \vec{v} - \mtx{Z}\vec{r} \right \Vert _2^2 + \widetilde{\alpha}\left \Vert \vec{r} \right \Vert _2^2
~~~\text{subject to}~~\mtx{Q}_{\mathcal{S}^{\rm c}} \vec{r} = \vec{0},
\end{align}
where $\widetilde{\alpha}>0$ determines the level of ridge regularization. 
The formulation in (\ref{eq:constrained linear regression - source - ridge regression}) is equivalent to $\widehat{\vecgreek{\theta}}$ where $\widehat{\vecgreek{\theta}}_{\mathcal{S}^c}=0$ and 
\begin{align} 
\label{eq:constrained linear regression - source - ridge regression - free parameters}
\widehat{\vecgreek{\theta}}_{\mathcal{S}} = \argmin_{\vec{k}\in\mathbb{R}^{\widetilde{p}}} \left \Vert  \vec{v} - \mtx{Z}_{\mathcal{S}}\vec{k} \right \Vert _2^2 + \widetilde{\alpha}\left \Vert \vec{k} \right \Vert _2^2. 
\end{align}
The optimization in (\ref{eq:constrained linear regression - source - ridge regression - free parameters}) has a standard ridge regression form and, thus, its closed-form solution is 
\begin{align} 
\label{eq:constrained linear regression - source - ridge regression - free parameters - closed form}
\widehat{\vecgreek{\theta}}_{\mathcal{S}} = \left( \mtx{Z}_{\mathcal{S}}^T \mtx{Z}_{\mathcal{S}} + \widetilde{\alpha} \mtx{I}_{\widetilde{p}} \right)^{-1} \mtx{Z}_{\mathcal{S}}^T \vec{v}.
\end{align}
In appendix \ref{appendix:subsec:Ridge Regression Error Expression and Optimal Tuning for the Source Task} we formulate the out-of-sample error of the ridge regression solution to the source task, and show that optimal tuning is provided by $\widetilde{\alpha}=\frac{\widetilde{p}\left(\sigma_{\xi}^2 + \Ltwonorm{\vecgreek{\theta}_{\mathcal{S}^c}}\right)}{\Ltwonorm{\vecgreek{\theta}_{\mathcal{S}}}}$. In our experiments we assume that $\Ltwonorm{\vecgreek{\theta}_{\mathcal{S}}},\Ltwonorm{\vecgreek{\theta}_{\mathcal{S}^c}}$ are unknown and only $\Ltwonorm{\vecgreek{\theta}}$, $\sigma_{\xi}^2$ are known about the source task. Thus, we assume $\Ltwonorm{\vecgreek{\theta}_{\mathcal{S}}}\approx\frac{\widetilde{p}}{d}\Ltwonorm{\vecgreek{\theta}}, \Ltwonorm{\vecgreek{\theta}_{\mathcal{S}^c}}\approx\left(1-\frac{\widetilde{p}}{d}\right)\Ltwonorm{\vecgreek{\theta}}$ that let us to approximate the optimal tuning using  $\widetilde{\alpha}=\frac{d\sigma_{\xi}^2}{\Ltwonorm{\vecgreek{\theta}}}+d-\widetilde{p}$. 

Proceeding to the target task, the ridge regression extension of (\ref{eq:constrained linear regression - target task}) is 
\begin{align} 
\label{eq:constrained linear regression - target task - ridge regression}
&\widehat{\vecgreek{\beta}} = \argmin_{\vec{b}\in\mathbb{R}^{d}} \left \Vert  \vec{y} - \mtx{X}\vec{b} \right \Vert _2^2 + \alpha\Ltwonorm{b}
\\ \nonumber
&\mathmakebox[5em][l]{\text{subject to}}\mtx{Q}_{\mathcal{T}} \vec{b} = \mtx{Q}_{\mathcal{T}}\widehat{\vecgreek{\theta}}
\\ \nonumber
&\qquad\qquad\quad\mtx{Q}_{\mathcal{Z}} \vec{b} = \vec{0},
\end{align}
for $\alpha>0$. The solution of (\ref{eq:constrained linear regression - target task - ridge regression}) has the closed form of $\widehat{\vecgreek{\beta}}$ where $\widehat{\vecgreek{\beta}}_{\mathcal{Z}}=\vec{0}$, $\widehat{\vecgreek{\beta}}_{\mathcal{T}}=\widehat{\vecgreek{\theta}}_{\mathcal{T}}$, and 
\begin{align} 
\label{eq:constrained linear regression - target - ridge regression - free parameters - closed form}
\widehat{\vecgreek{\beta}}_{\mathcal{F}} = \left( \mtx{X}_{\mathcal{F}}^T \mtx{X}_{\mathcal{F}} + \alpha \mtx{I}_{p} \right)^{-1} \mtx{X}_{\mathcal{F}}^T \left( \vec{y} - \mtx{X}_{\mathcal{T}} \widehat{\vecgreek{\theta}}_{\mathcal{T}} \right).
\end{align}
In appendix \ref{appendix:subsec:Ridge Regression Error Expression and Optimal Tuning for the Target Task} we provide further details on (\ref{eq:constrained linear regression - target - ridge regression - free parameters - closed form}),  the corresponding out-of-sample error, and show that optimal tuning is given by $\alpha=\frac{p\left(\sigma_{\epsilon}^2 + \Ltwonorm{\vecgreek{\beta}_{\mathcal{Z}}} + \expectation{\Ltwonorm{\vecgreek{\beta}_{\mathcal{T}} - \widehat{\vecgreek{\theta}}_{\mathcal{T}}  }}\right)}{\Ltwonorm{\vecgreek{\beta}_{\mathcal{F}}}}$. 
In our experiments we approximate the optimal tuning by setting $\alpha = \frac{d\sigma_{\epsilon}^2}{\Ltwonorm{\vecgreek{\beta}}} + d - p - t + t\frac{\Ltwonorm{\left(\mtx{I}_d - \mtx{H}\right)\vecgreek{\beta}} + d\sigma_{\eta}^2}{\Ltwonorm{\vecgreek{\beta}}}$ that assumes the knowledge of $\Ltwonorm{\vecgreek{\beta}}$, $\Ltwonorm{\left(\mtx{I}_d - \mtx{H}\right)\vecgreek{\beta}}$, $\sigma_{\eta}$, $\sigma_{\epsilon}$. See Appendix \ref{appendix:subsec:Ridge Regression Error Expression and Optimal Tuning for the Target Task} for more details. 

\begin{figure}[!ht]
	\parbox{0.95\textwidth}{
		\parbox{.18\textwidth}{%
			\subfloat{\includegraphics[width=0.17\textwidth]{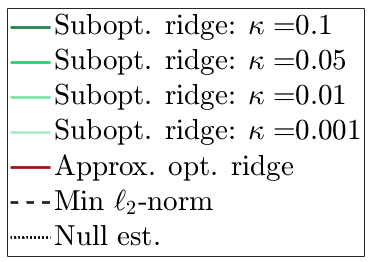}}
		}~~
		\parbox{.75\textwidth}{%
			\setcounter{subfigure}{0}
			\subfloat[{\small$t=0$}]{\includegraphics[width=0.24\textwidth]{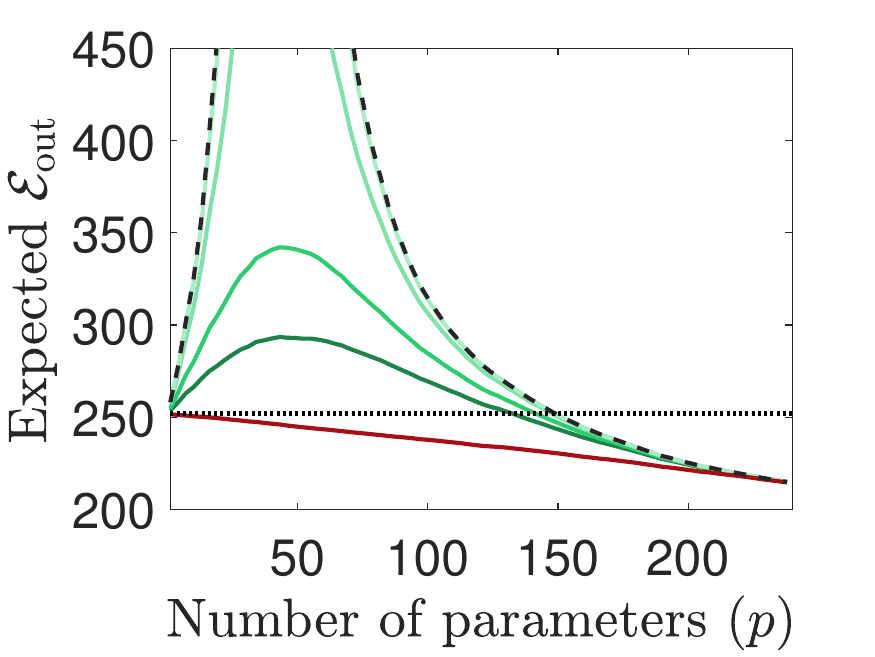} \label{fig:subopt_ridge_linearbeta_t0}}
			\subfloat[{\small$t=16$}]{\includegraphics[width=0.24\textwidth]{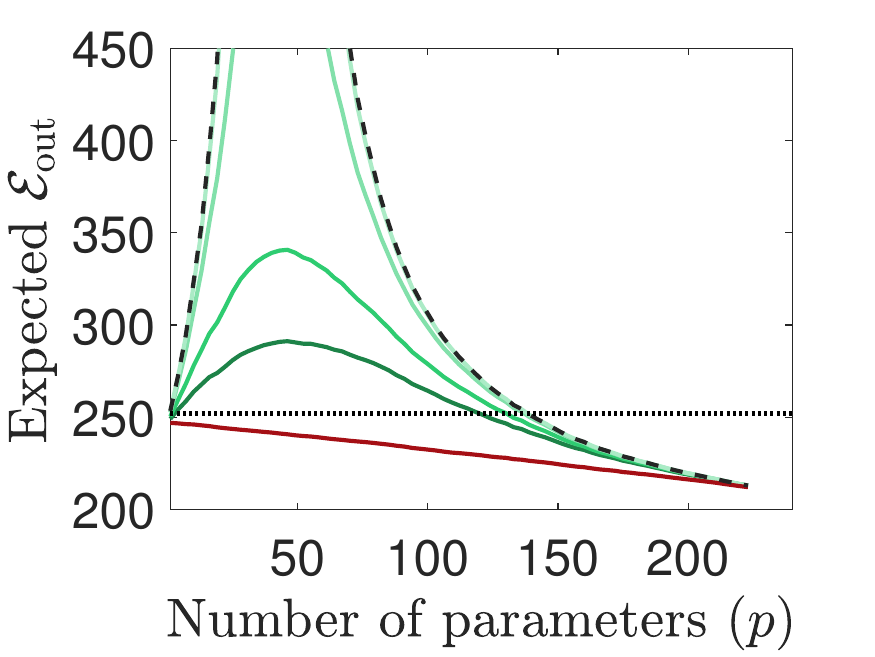} \label{fig:subopt_ridge_linearbeta_t16}}
			\subfloat[{\small$t=32$}]{\includegraphics[width=0.24\textwidth]{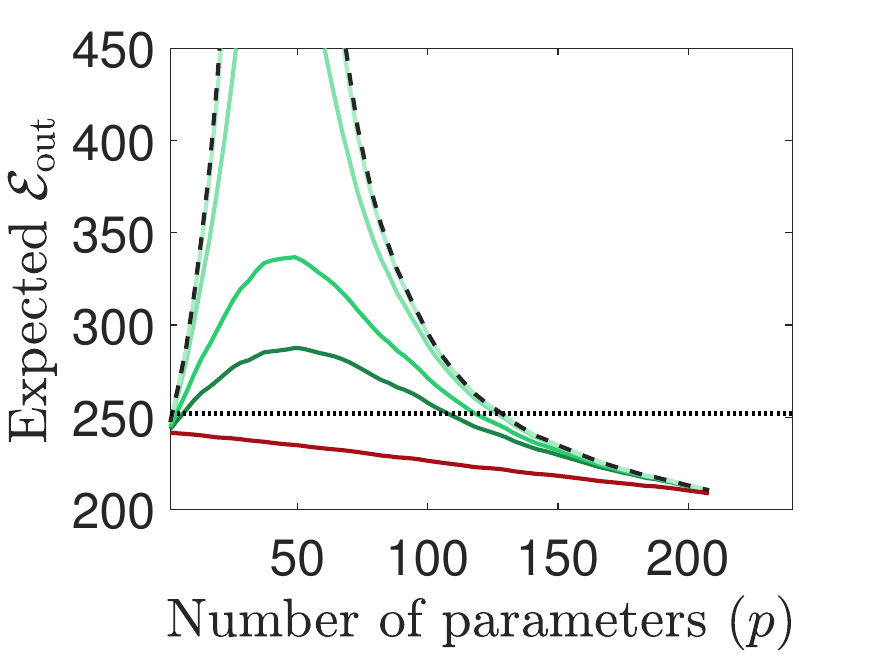} \label{fig:subopt_ridge_linearbeta_t32}}		
		}	 
	}
	\caption{The expected generalization error of the target task, $\expectationwrt{\mathcal{E}_{\rm out}}{\mathcal{L}}$, for transfer learning with suboptimally tuned ridge regularization. Each of the subfigures corresponds to a different number of transferred parameters $t$, and shows the error curves for four suboptimal ridge tunings where the parameters $\alpha$ and $\widetilde{\alpha}$ are $\kappa$ times their (approximately) optimal values. Here $\vecgreek{\beta}$ has linear shape, $\mtx{H}$ is a local averaging operator (over a neighborhood size 11), $\sigma_{\eta}^2 = 0.2$, $d=240$, ${n=40}$, $\widetilde{n}=100$, $\| \vecgreek{\beta} \|_2^2 = d$, $\sigma_{\epsilon}^2 = 0.05\cdot d$, $\sigma_{\xi}^2 = 0.025\cdot d$, and $\widetilde{p}=d$. }
	\label{fig:subopt ridge regression error curves - on average}
\end{figure}

The empirical out-of-sample errors for the ridge regression transfer learning appear in Figure \ref{fig:min l1 norm and ridge regression error curves - on average} as red curves where the shade of red denotes the number of transferred parameters. As expected, the ridge regularization (which is approximately optimally tuned) resolves the error peak and eliminates the double descent shape of the minimum norm interpolating solutions. In the case of suboptimally tuned ridge regularization the double descent peak may not be fully resolved (see Fig.~\ref{fig:subopt ridge regression error curves - on average}). 
For $\vecgreek{\beta}$ with a linear shape, the ridge regularization suits best among the examined methods at any parameterization level, except for the maximal overparameterization levels where the minimum $\ell_2$-norm solution performs comparably (see Figs.~\ref{fig:target_generalization_errors_vs_p_eta_0.2linearbeta_H_is_I_allL1L2Ridge}-\ref{fig:target_generalization_errors_vs_p_eta_0.2linearbeta_H_is_averaging11_allL1L2Ridge}). However, for a sparse $\vecgreek{\beta}$ the minimum $\ell_1$-norm outperforms the ridge solution for high overparameterization levels (see Figs.~\ref{fig:target_generalization_errors_vs_p_eta_0.2sparsebeta_H_is_I_allL1L2Ridge}-\ref{fig:target_generalization_errors_vs_p_eta_0.2sparsebeta_H_is_averaging11_allL1L2Ridge}).

\section{Conclusions}
\label{sec:Conclusion}

In this work we have established an analytical framework for the fundamental study of transfer learning in conjunction with overparameterized models. We used least squares solutions to linear regression problems for shedding clarifying light on 
the generalization performance induced for a target task addressed using parameters transferred from an already completed source task. 
We formulated the generalization error of the target task and presented its two-dimensional double descent shape as a function of the number of free parameters individually available in the source and target tasks. 
We characterized the conditions for a beneficial transfer of parameters and demonstrated its high sensitivity to the delicate interaction among crucial aspects such as the source-target task relation, the specific choice of transferred parameters, and the form of the true solution. 
We importantly showed that overparameterized transfer learning is not necessarily improved by using a source task which is closer or identical to the target task. 
Our focus was mainly on the analytical and empirical study of the minimum $\ell_2$-norm solution to overparameterized transfer learning. We also empirically examined the performance of the minimum $\ell_1$-norm solution and ridge regression in our transfer learning framework.
We believe that our work opens a new research direction for the fundamental understanding of the generalization ability of transfer learning designs. Future work may study the theory and practice of additional transfer learning layouts such as fine tuning of the transferred parameters, inclusion of various regularization methods, well-specified and other settings where the task relation model is known (to some extent) and utilized in the actual learning process.

\section*{Acknowledgments}
This work was supported by NSF grants CCF-1911094, IIS-1838177, and IIS-1730574; ONR grants N00014-18-12571, N00014-20-1-2534, and MURI N00014-20-1-2787; AFOSR grant FA9550-18-1-0478; and a Vannevar Bush Faculty Fellowship, ONR grant N00014-18-1-2047.

\appendix
\section*{Appendices}
The following appendices support the main paper as follows. 
Appendix \ref{appendix:sec:Mathematical Developments for Section 2} provides additional details on the mathematical developments leading to the formulations in Section \ref{sec:Transfer Learning in the Linear Regression Case: Problem Definition} of the main paper. 
In Appendix \ref{appendix:sec:Transfer of Specific Sets of Parameters: Proofs} we present the proofs of Theorem \ref{theorem:out of sample error - target task - specific layout} and Corollaries \ref{corollary:out of sample error - target task - specific layout - detailed - specific layout}, \ref{corollary:out of sample error - target task}  from Section \ref{sec:The Double Descent Phenomenon in Transfer Learning}, which formulate the generalization error of the target task. 
Appendix \ref{appendix:sec:Empirical Results for Section 3 Additional Details and Demonstrations} provides additional empirical results and details for Section \ref{sec:The Double Descent Phenomenon in Transfer Learning} of the main paper. 
In Appendices \ref{appendix:sec:Proofs and Additional Details on Parameter Transfer Usefulness in the Setting of Uniformly-Distributed Coordinate Layouts}, \ref{sec:The Optimal Componentwise Task Relation: Additional Analytical and Empirical Details} we provide analytical proofs and empirical results for Sections \ref{sec:When is Transfer Learning Beneficial}, \ref{sec:Additional Insights The Optimal H in a Componentwise Task Relation} of the main paper. In Appendix \ref{appendix:sec:Additional Details and Results for Section 6} we provide the mathematical developments for the ridge regression setting of our transfer learning problem from Section \ref{subsec:Ridge Regression} of the main paper. 


\section{Mathematical Developments for Section 2}
\label{appendix:sec:Mathematical Developments for Section 2}

\subsection{The Estimate \texorpdfstring{$\widehat{\vecgreek{\theta}}$}{~} in Eq.~(\ref{eq:constrained linear regression - solution - source data class})}
\label{appendix:subsec:Mathematical Development of Estimate of Theta}

Let us solve the optimization problem provided in (\ref{eq:constrained linear regression - source data class}). 
Using the relation ${\mtx{Q}_{\mathcal{S}}^{T} \mtx{Q}_{\mathcal{S}} + \mtx{Q}_{\mathcal{S}^{{\rm c}}}^{T} \mtx{Q}_{\mathcal{S}^{\rm c}} = \mtx{I}_{d}}$ we can rewrite (\ref{eq:constrained linear regression - source data class}) as 
\begin{align} 
\label{appendix:eq:constrained linear regression - source data class - decomposed optimization cost}
&\widehat{\vecgreek{\theta}} = \argmin_{\vec{r}\in\mathbb{R}^{d}} \left \Vert  \vec{v} - \mtx{Z}_{\mathcal{S}} \mtx{Q}_{\mathcal{S}}\vec{r} - \mtx{Z}_{\mathcal{S}^{{\rm c}}} \mtx{Q}_{\mathcal{S}^{\rm c}}\vec{r} \right \Vert _2^2
\\ \nonumber
&\text{subject to}~~\mtx{Q}_{\mathcal{S}^{\rm c}} \vec{r} = \vec{0}
\end{align}
where $\mtx{Z}_{\mathcal{S}} \triangleq \mtx{Z}\mtx{Q}_{\mathcal{S}}^{T}$ and $\mtx{Z}_{\mathcal{S}^{{\rm c}}} \triangleq \mtx{Z}\mtx{Q}_{\mathcal{S}^{{\rm c}}}^{T}$. By setting the equality constraint in the optimization cost, the problem in (\ref{appendix:eq:constrained linear regression - source data class - decomposed optimization cost}) becomes 
\begin{align} 
\label{appendix:eq:constrained linear regression - source data class - unconstrained form}
&\widehat{\vecgreek{\theta}} = \argmin_{\vec{r}\in\mathbb{R}^{d}} \left \Vert  \vec{v} - \mtx{Z}_{\mathcal{S}} \mtx{Q}_{\mathcal{S}}\vec{r}  \right \Vert _2^2
\\ \nonumber
&\text{subject to}~~\mtx{Q}_{\mathcal{S}^{\rm c}} \vec{r} = \vec{0}.
\end{align}
Without the equality constraint, (\ref{appendix:eq:constrained linear regression - source data class - unconstrained form}) is just an unconstrained least squares problem that its minimum $\ell_2$-norm solution is 
\begin{equation}
\label{appendix:eq:constrained linear regression - solution - source data class}
{\widehat{\vecgreek{\theta}}=\mtx{Q}_{\mathcal{S}}^T \mtx{Z}_{\mathcal{S}}^{+} \vec{v}}
\end{equation}
where $\mtx{Z}_{\mathcal{S}}^{+}$ is the Moore-Penrose pseudoinverse of $\mtx{Z}_{\mathcal{S}}$. Note that $\widehat{\vecgreek{\theta}}$ in (\ref{appendix:eq:constrained linear regression - solution - source data class}) satisfies the equality constraint in (\ref{appendix:eq:constrained linear regression - source data class - decomposed optimization cost}) and, therefore, (\ref{appendix:eq:constrained linear regression - solution - source data class}) is also the solution for the constrained optimization problems in (\ref{appendix:eq:constrained linear regression - source data class - decomposed optimization cost}), (\ref{appendix:eq:constrained linear regression - source data class - unconstrained form}), and (\ref{eq:constrained linear regression - source data class}).

\subsection{The Double Descent Formulation for the Generalization Error of the Source Task}
\label{appendix:subsec:Double Descent Formulation for Source Task}

The generalization error of a single linear regression problem (that includes noise) in non-asymptotic settings is provided in \cite{belkin2020two} for a given coordinate subset (i.e., deterministic $\mathcal{S}$ in our terms). The result from \cite{belkin2020two} can be written in our notations as 
\begin{equation}
\label{eq:out of sample error - source task - deterministic S - from reference}
\widetilde{\mathcal{E}}_{\rm out} =   \begin{cases}
\mathmakebox[18em][l]{{ \frac{\widetilde{n}-1}{\widetilde{n}-\widetilde{p}-1} }\left( \Ltwonorm{\vecgreek{\theta}_{\mathcal{S}^{\rm c}}} + \sigma_{\xi}^{2} \right)}  \text{for } \widetilde{p} \le \widetilde{n}-2,   
\\
\mathmakebox[18em][l]{\infty} \text{for } \widetilde{n}-1 \le \widetilde{p} \le \widetilde{n}+1,
\\
\mathmakebox[18em][l]{{ \frac{\widetilde{p}-1}{\widetilde{p}-\widetilde{n}-1} }\left( \Ltwonorm{\vecgreek{\theta}_{\mathcal{S}^{\rm c}}} + \sigma_{\xi}^{2} \right) +  { \frac{\widetilde{p}-\widetilde{n}}{\widetilde{p}} }\Ltwonorm{\vecgreek{\theta}_{\mathcal{S}}}} \text{for } \widetilde{p} \ge \widetilde{n}+2.
\end{cases}
\end{equation}

In case that the coordinate subset $\mathcal{S}$ is uniformly chosen at random from all the subsets of ${\widetilde{p}\in\{{1,\dots,d}\}}$ unique coordinates of ${\{{1,\dots,d}\}}$, then we get that ${\expectationwrt{\Ltwonorm{\vecgreek{\theta}_{\mathcal{S}}} }{\mathcal{S}}=\frac{\widetilde{p}}{d} \Ltwonorm{\vecgreek{\theta}}}$ and ${\expectationwrt{\Ltwonorm{\vecgreek{\theta}_{\mathcal{S}^{\rm c}}} }{\mathcal{S}}=\frac{d-\widetilde{p}}{d} \Ltwonorm{\vecgreek{\theta}}}$. Accordingly, the expectation over $\mathcal{S}$ of the generalization error of the source task leads to the following result 
\begin{align}
\label{eq:out of sample error - source task - expectation over S}
\expectationwrt{\widetilde{\mathcal{E}}_{\rm out} }{\mathcal{S}} =  \begin{cases}
\mathmakebox[18em][l]{\frac{\widetilde{n}-1}{\widetilde{n}-\widetilde{p}-1} \left( \left({1 - \frac{\widetilde{p}}{\widetilde{d}}}\right) \Ltwonorm{\vecgreek{\theta}} + \sigma_{\xi}^2\right)}  \text{for } \widetilde{p} \le \widetilde{n}-2, \\
\mathmakebox[18em][l]{\infty} \text{for } \widetilde{n}-1 \le \widetilde{p} \le \widetilde{n}+1,
\\
\mathmakebox[18em][l]{\frac{\widetilde{p}-1}{\widetilde{p}-\widetilde{n}-1} \left( \left({1 - \frac{\widetilde{p}}{\widetilde{d}}}\right) \Ltwonorm{\vecgreek{\theta}} + \sigma_{\xi}^2\right) + \frac{\widetilde{p} - \widetilde{n}}{d} \Ltwonorm{\vecgreek{\theta}} } \text{for } \widetilde{p} \ge \widetilde{n}+2. 
\end{cases}
\end{align}	

The formulation in (\ref{eq:out of sample error - source task - expectation over S}) considers $\vecgreek{\theta}$ as a deterministic vector. 
For the analysis of the target task, where the task relation model (\ref{eq:theta-beta relation}) is assumed to hold, it is also useful to formulate the expectation of the out-of-sample error of the source task with respect to both $\mathcal{S}$ and the noise vector $\vecgreek{\eta}$ from the task relation model. This leads us to to consider $\vecgreek{\theta}$ as a random vector and to formulate the following expectation. 
\begin{align}
\label{appendix:eq:out of sample error - source task - expectation over S and eta}
\expectationwrt{\widetilde{\mathcal{E}}_{\rm out} }{\mathcal{S},\vecgreek{\eta}} =  \begin{cases}
\mathmakebox[18em][l]{\frac{\widetilde{n}-1}{\widetilde{n}-\widetilde{p}-1} \left( \left({1 - \frac{\widetilde{p}}{\widetilde{d}}}\right)\kappa  + \sigma_{\xi}^2\right)}  \text{for } \widetilde{p} \le \widetilde{n}-2, \\
\mathmakebox[18em][l]{\infty} \text{for } \widetilde{n}-1 \le \widetilde{p} \le \widetilde{n}+1,
\\
\mathmakebox[18em][l]{\frac{\widetilde{p}-1}{\widetilde{p}-\widetilde{n}-1} \left( \left({1 - \frac{\widetilde{p}}{\widetilde{d}}}\right)\kappa + \sigma_{\xi}^2\right) + \frac{\widetilde{p} - \widetilde{n}}{d} \kappa } \text{for } \widetilde{p} \ge \widetilde{n}+2. 
\end{cases}
\end{align}	
where ${\kappa\triangleq \expectationwrt{\Ltwonorm{\vecgreek{\theta}}}{\vecgreek{\eta}} =  \Ltwonorm{ \mtx{H}\vecgreek{\beta}}  + d \sigma_{\eta}^2 }$.

\subsection{The Estimate \texorpdfstring{$\widehat{\vecgreek{\beta}}$}{~} in Eq.~(\ref{eq:constrained linear regression - solution - target task})}
\label{appendix:subsec:Mathematical Development of Estimate of Beta}

The optimization problem in (\ref{eq:constrained linear regression - target task}), given for the target task, can be addressed using the relation ${\mtx{Q}_{\mathcal{F}}^{T} \mtx{Q}_{\mathcal{F}} + \mtx{Q}_{\mathcal{T}}^{T} \mtx{Q}_{\mathcal{T}} + \mtx{Q}_{\mathcal{Z}}^{T} \mtx{Q}_{\mathcal{Z}} = \mtx{I}_{d}}$ and rewritten as 
\begin{align} 
\label{appendix:eq:constrained linear regression - target task - decomposed optimization cost}
&\widehat{\vecgreek{\beta}} = \argmin_{\vec{b}\in\mathbb{R}^{d}} \left \Vert  \vec{y} - \mtx{X}_{\mathcal{F}}\mtx{Q}_{\mathcal{F}} \vec{b} - \mtx{X}_{\mathcal{T}}\mtx{Q}_{\mathcal{T}} \vec{b} - \mtx{X}_{\mathcal{Z}}\mtx{Q}_{\mathcal{Z}} \vec{b} \right \Vert _2^2
\nonumber\\ \nonumber
&\mathmakebox[5em][l]{\text{subject to}}\mtx{Q}_{\mathcal{T}} \vec{b} = \mtx{Q}_{\mathcal{T}}\widehat{\vecgreek{\theta}}
\\ 
&\qquad\qquad\quad\mtx{Q}_{\mathcal{Z}} \vec{b} = \vec{0}
\end{align}
where $\mtx{X}_{\mathcal{F}} \triangleq \mtx{X}\mtx{Q}_{\mathcal{F}}^{T}$, $\mtx{X}_{\mathcal{T}} \triangleq \mtx{X}\mtx{Q}_{\mathcal{T}}^{T}$, and $\mtx{X}_{\mathcal{Z}} \triangleq \mtx{X}\mtx{Q}_{\mathcal{Z}}^{T}$. By setting the equality constraints of (\ref{appendix:eq:constrained linear regression - target task - decomposed optimization cost}) in its optimization cost, the problem (\ref{appendix:eq:constrained linear regression - target task - decomposed optimization cost}) can be translated into the form of 
\begin{align}
\label{appendix:eq:constrained linear regression - target task - decomposed optimization cost - constraints set in cost}
&\widehat{\vecgreek{\beta}} = \argmin_{\vec{b}\in\mathbb{R}^{d}} \left \Vert  \vec{y} - \mtx{X}_{\mathcal{T}}\mtx{Q}_{\mathcal{T}}\widehat{\vecgreek{\theta}} - \mtx{X}_{\mathcal{F}}\mtx{Q}_{\mathcal{F}} \vec{b}   \right \Vert _2^2 
\nonumber \\ \nonumber
&\mathmakebox[5em][l]{\text{subject to}}\mtx{Q}_{\mathcal{T}} \vec{b} = \mtx{Q}_{\mathcal{T}}\widehat{\vecgreek{\theta}}
\\ 
&\qquad\qquad\quad\mtx{Q}_{\mathcal{Z}} \vec{b} = \vec{0}.
\end{align}
The last optimization is a restricted least squares problem that can be solved using the method of Lagrange multipliers to show that 
\begin{equation}
\label{appendix:eq:constrained linear regression - solution - target task}
{\widehat{\vecgreek{\beta}}=\mtx{Q}_{\mathcal{F}}^T \mtx{X}_{\mathcal{F}}^{+} \left( \vec{y} - \mtx{X}_{\mathcal{T}} \widehat{\vecgreek{\theta}}_{\mathcal{T}} \right)} + \mtx{Q}_{\mathcal{T}}^T\widehat{\vecgreek{\theta}}_{\mathcal{T}}
\end{equation}
where $\widehat{\vecgreek{\theta}}_{\mathcal{T}} \triangleq \mtx{Q}_{\mathcal{T}}\widehat{\vecgreek{\theta}}$ and $\mtx{X}_{\mathcal{F}}^{+}$ is the Moore-Penrose pseudoinverse of $\mtx{X}_{\mathcal{F}}$.

\section{Proofs for Section \ref{sec:The Double Descent Phenomenon in Transfer Learning}}
\label{appendix:sec:Transfer of Specific Sets of Parameters: Proofs}

In this section we outline the proof of Theorem \ref{theorem:out of sample error - target task - specific layout} for the generalization error of the target task in the setting where a specific coordinate subset layout $\mathcal{L}$ determines the transferred set of parameters. 
We start in Section \ref{appendix:subsec:Auxiliary Results using Wishart Matrices} by providing auxiliary results that use non-asymptotic properties of Gaussian and Wishart matrices. 
Then, in Section \ref{appendix:subsec:Theorem 4.1 - Proof Outline} we prove Theorem \ref{theorem:out of sample error - target task - specific layout}, in Section \ref{appendix:subsec:Proof of Corollary 6} we prove Corollary \ref{corollary:out of sample error - target task - specific layout - detailed - specific layout}, and in Section \ref{appendix:subsec:proof of Corollary 4} we prove Corollary \ref{corollary:out of sample error - target task}.

\subsection{Auxiliary Results using Non-Asymptotic Properties of Gaussian and Wishart Matrices}
\label{appendix:subsec:Auxiliary Results using Wishart Matrices}

The random matrix ${\mtx{X}_{\mathcal{F}}\triangleq\mtx{X}\mtx{Q}_{\mathcal{F}}^{T}}$ is of size $n\times p$ and all its components are i.i.d.~standard Gaussian variables. Then, almost surely,  
\begin{equation}
\label{appendix:eq:lemma - mean of X_F projection matrix}
\expectation{  \mtx{X}_{\mathcal{F}}^{+} \mtx{X}_{\mathcal{F}} } = \mtx{I}_{p} \times \begin{cases}
1          & \text{for } p \le n, \\
\frac{n}{p}          & \text{for } p > n, 
\end{cases} 
\end{equation}
where $\mtx{X}_{\mathcal{F}}^{+} \mtx{X}_{\mathcal{F}}$ is the $p\times p$ projection operator onto the range of $\mtx{X}_{\mathcal{F}}$. 
Accordingly, let $\vec{a}\in\mathbb{R}^{p}$ be a random vector independent of the matrix ${\mtx{X}_{\mathcal{F}}}$ and, then, 
\begin{equation}
\label{appendix:eq:lemma - result of X_F projection and a random vector a}
\expectation{ \Ltwonorm{ \mtx{X}_{\mathcal{F}}^{+} \mtx{X}_{\mathcal{F}} \vec{a} } } = \expectation{ \Ltwonorm{ \vec{a} }} \times \begin{cases}
1          & \text{for } p \le n, \\
\frac{n}{p}          & \text{for } p > n.
\end{cases} 
\end{equation}

The components of ${\mtx{X}_{\mathcal{F}}}$ are i.i.d.~standard Gaussian variables, hence ${\mtx{X}_{\mathcal{F}}^{T} \mtx{X}_{\mathcal{F}}\sim \mathcal{W}_{p}\left({\mtx{I}_{p},n}\right)}$ is a $p\times p$ Wishart matrix with $n$ degrees of freedom, and ${\mtx{X}_{\mathcal{F}} \mtx{X}_{\mathcal{F}}^{T}\sim \mathcal{W}_{n}\left({\mtx{I}_{n},p}\right)}$ is a $n\times n$ Wishart matrix with $p$ degrees of freedom. 
The pseudoinverse of the $n\times n$ Wishart matrix (almost surely) satisfies 
\begin{align}
\label{appendix:eq:lemma - mean of squared pseudoinverse of X_F}
&\expectation{ \left( {\mtx{X}_{\mathcal{F}} \mtx{X}_{\mathcal{F}}^{T} }\right)^{+} } =  
\expectation{ \mtx{X}_{\mathcal{F}}^{+,T} \mtx{X}_{\mathcal{F}}^{+} } =  \mtx{I}_{n}\times \begin{cases}
\frac{1}{n-p-1} \cdot \frac{p}{n}          & \text{for } p \le n-2, \\
\infty& \text{for } n-1\le p \le n+1,  \\
\frac{1}{p-n-1}           & \text{for } p \ge n+2, 
\end{cases}  
\end{align}
where the result for ${p \ge n+2}$ corresponds to the common case of inverse Wishart matrix with more degrees of freedom than its dimension, and the result for ${p \le n-2}$ is based on constructions provided in the proof of Theorem 1.3 in \cite{breiman1983many}. 


Following (\ref{appendix:eq:lemma - mean of squared pseudoinverse of X_F}), let $\vec{u}\in\mathbb{R}^{n}$ be a random vector independent of ${\mtx{X}_{\mathcal{F}}}$. 
Then, 
\begin{align}
\label{appendix:eq:lemma - result of X_F and independent u}
&\expectation{ \Ltwonorm{ \mtx{X}_{\mathcal{F}}^{+} \vec{u} } } = \frac{1}{n}\expectation{ \Ltwonorm{ \vec{u} }} \times \begin{cases}
\frac{p}{n-p-1}          & \text{for } p \le n-2, \\
\infty& \text{for } n-1\le p \le n+1,  \\
\frac{n}{p-n-1}          & \text{for } p \ge n+2,
\end{cases} 
\end{align}
that specifically for $\vec{u}=\mtx{X}_{\mathcal{F}^{\rm c}}\vecgreek{\beta}_{\mathcal{F}^{\rm c}}$ becomes 
\begin{align}
\label{appendix:eq:lemma - result of X_F and X_Fc beta_Fc}
&\expectation{ \Ltwonorm{ \mtx{X}_{\mathcal{F}}^{+}\mtx{X}_{\mathcal{F}^{\rm c}} \vecgreek{\beta}_{\mathcal{F}^{\rm c}} } } = \expectation{ \Ltwonorm{ \vecgreek{\beta}_{\mathcal{F}^{\rm c}} }} \times \begin{cases}
\frac{p}{n-p-1}          & \text{for } p \le n-2, \\
\infty& \text{for } n-1\le p \le n+1,  \\
\frac{n}{p-n-1}          & \text{for } p \ge n+2. 
\end{cases} 
\end{align}

The results in (\ref{appendix:eq:lemma - mean of X_F projection matrix})-(\ref{appendix:eq:lemma - result of X_F and X_Fc beta_Fc}) are presented using notions of the \textit{target} task, specifically, using the data matrix $\mtx{X}$ and the coordinate subset $\mathcal{T}$. 
One can obtain the corresponding results for the \textit{source} task by updating (\ref{appendix:eq:lemma - mean of X_F projection matrix})-(\ref{appendix:eq:lemma - result of X_F and X_Fc beta_Fc}) by replacing $\mtx{X}$, $\mathcal{T}$, $n$ and $p$ with $\mtx{Z}$, $\mathcal{S}$, $\widetilde{n}$ and $\widetilde{p}$, respectively. For example, the result corresponding to (\ref{appendix:eq:lemma - mean of X_F projection matrix}) is 
\begin{equation}
\label{appendix:eq:lemma - mean of Z_S projection matrix}
\expectation{  \mtx{Z}_{\mathcal{S}}^{+} \mtx{Z}_{\mathcal{S}} } = \mtx{I}_{\widetilde{p}} \times \begin{cases}
1          & \text{for } \widetilde{p} \le \widetilde{n}, \\
\frac{\widetilde{n}}{\widetilde{p}}          & \text{for } \widetilde{p} > \widetilde{n}, 
\end{cases} 
\end{equation}
where $\mtx{Z}_{\mathcal{S}}^{+} \mtx{Z}_{\mathcal{S}}$ is the $\widetilde{p}\times \widetilde{p}$ projection operator onto the range of $\mtx{Z}_{\mathcal{S}}$.


The next auxiliary results consider a coordinate subset layout 
${\mathcal{L} = \lbrace{ \mathcal{S}, \mathcal{F}, \mathcal{T}, \mathcal{Z} }\rbrace}$ which is specific, i.e., non random, and therefore the induced operators such as  ${\mtx{Q}_{\mathcal{S}}}$, ${\mtx{Q}_{\mathcal{F}}}$, ${\mtx{Q}_{\mathcal{T}}}$, ${\mtx{Q}_{\mathcal{Z}}}$ are also fixed and do not have any random aspect. 
Recall that  $\mtx{Q}_{\mathcal{S}}^T{\mtx{Q}_{\mathcal{S}}}$ is a $d\times d$ diagonal matrix with its $j^{\rm th}$ diagonal component equals 1 if $j\in\mathcal{S}$ and 0 otherwise. Similarly holds for the other coordinate subsets. Accordingly, here the norms of vector forms such as  $\vecgreek{\beta}_{\mathcal{T}} \triangleq \mtx{Q}_{\mathcal{T}} \vecgreek{\beta}$, $\vecgreek{\beta}_{\mathcal{F}} \triangleq \mtx{Q}_{\mathcal{F}} \vecgreek{\beta}$, and $\vecgreek{\beta}_{\mathcal{Z}} \triangleq \mtx{Q}_{\mathcal{Z}} \vecgreek{\beta}$, are directly referred to as $\Ltwonorm{\vecgreek{\beta}_{\mathcal{T}}}$, $\Ltwonorm{\vecgreek{\beta}_{\mathcal{F}}}$, $\Ltwonorm{\vecgreek{\beta}_{\mathcal{Z}}}$, respectively. 

Recall that $\mathcal{T}\subseteq\mathcal{S}$. Then, for a deterministic vector $\vec{w}\in\mathbb{R}^{d}$, 
\begin{align}
&\expectation{ \Ltwonorm{ \mtx{Q}_{\mathcal{T}} \mtx{Q}_{\mathcal{S}}^{T} \mtx{Z}_{\mathcal{S}}^{+} \mtx{Z}_{\mathcal{S}} \mtx{Q}_{\mathcal{S}} \vec{w} } } =  \begin{cases}
\mathmakebox[21.5em][l] {{ \Ltwonorm{ \vec{w}_{\mathcal{T}} }} }  \text{for } \widetilde{p} \le \widetilde{n}, \\
\mathmakebox[21.5em][l] {\frac{\widetilde{n}}{\widetilde{p}\left({\widetilde{p}+1}\right)}\left({ \left({\widetilde{n}+\frac{\widetilde{n}-1}{\widetilde{p}-1}}\right){\Ltwonorm{ \vec{w}_{\mathcal{T}} }} + \left({1-\frac{\widetilde{n}-1}{\widetilde{p}-1}}\right) t {\Ltwonorm{ \vec{w}_{\mathcal{S}} }} }\right)} \text{for } \widetilde{p} > \widetilde{n}.
\end{cases} 
\label{appendix:eq:lemma - SPECIFIC - form 2}
\end{align}
\begin{align}
&\expectation{ \Ltwonorm{ \mtx{Q}_{\mathcal{T}} \mtx{Q}_{\mathcal{S}}^{T} \mtx{Z}_{\mathcal{S}}^{+} \mtx{Z}_{\mathcal{S}^{c}} \mtx{Q}_{\mathcal{S}^{c}} \vec{w} } } =  { \frac{t}{\widetilde{p}}\Ltwonorm{ \vec{w}_{\mathcal{S}^{c}} }} \times  { \begin{cases}
	\frac{\widetilde{p}}{\widetilde{n}-\widetilde{p}-1}          & \text{for } \widetilde{p} \le \widetilde{n}-2, \\
	\infty& \text{for } \widetilde{n}-1 \le \widetilde{p} \le \widetilde{n}+1,  \\
	\frac{\widetilde{n}}{\widetilde{p}-\widetilde{n}-1}          & \text{for } \widetilde{p} \ge \widetilde{n}+2.
	\end{cases} }
\label{appendix:eq:lemma - SPECIFIC - form 4}
\end{align}
For two deterministic vectors $\vec{w},\vec{a}\in\mathbb{R}^{d}$, 
\begin{align}
&\expectation{ \left\langle { \mtx{Q}_{\mathcal{T}}\vec{a},  \mtx{Q}_{\mathcal{T}} \mtx{Q}_{\mathcal{S}}^{T} \mtx{Z}_{\mathcal{S}}^{+} \mtx{Z}_{\mathcal{S}} \mtx{Q}_{\mathcal{S}} \vec{w}  }  \right\rangle   }   
=\left\langle { \vec{a}_{\mathcal{T}} ,  \vec{w}_{\mathcal{T}}  }  \right\rangle \times \begin{cases}
{ 1 }          & \text{for } \widetilde{p} \le \widetilde{n}, \\
\frac{\widetilde{n}}{\widetilde{p}}      & \text{for } \widetilde{p} > \widetilde{n}.
\end{cases} 
\label{appendix:eq:lemma - SPECIFIC - form 3}
\end{align}
For a deterministic vector $\vec{r}\in\mathbb{R}^{\widetilde{n}}$, 
\begin{align}
&\expectation{ \Ltwonorm{ \mtx{Q}_{\mathcal{T}} \mtx{Q}_{\mathcal{S}}^{T} \mtx{Z}_{\mathcal{S}}^{+} \vec{r} } } 
=  { \frac{t}{\widetilde{n}\widetilde{p}}\Ltwonorm{ \vec{r} }} \times  { \begin{cases}
	\frac{\widetilde{p}}{\widetilde{n}-\widetilde{p}-1}          & \text{for } \widetilde{p} \le \widetilde{n}-2, \\
	\infty& \text{for } \widetilde{n}-1 \le \widetilde{p} \le \widetilde{n}+1,  \\
	\frac{\widetilde{n}}{\widetilde{p}-\widetilde{n}-1}          & \text{for } \widetilde{p} \ge \widetilde{n}+2.
	\end{cases} } 
\label{appendix:eq:lemma - SPECIFIC - form 5}
\end{align}

In our case we have the $\widetilde{n}\times\widetilde{p}$ matrix $\mtx{Z}_{\mathcal{S}}$ that its components are i.i.d.~standard Gaussian variables, thus,  $\mtx{Z}_{\mathcal{S}}$ can be decomposed into a form that involves an independent Haar-distributed matrix, i.e., a random orthonormal matrix that is uniformly distributed over the set of orthonormal matrices of the relevant size.
This lets us to prove the results in (\ref{appendix:eq:lemma - SPECIFIC - form 2})-(\ref{appendix:eq:lemma - SPECIFIC - form 5}) using some algebra and the non-asymptotic properties of random Haar-distributed matrices, see examples for such properties in Lemma 2.5 in \cite{tulino2004random} and also in Proposition 1.2 in \cite{hiai2000asymptotic}.

\subsection{Proof Outline of Theorem \ref{theorem:out of sample error - target task - specific layout}}
\label{appendix:subsec:Theorem 4.1 - Proof Outline}

The generalization error $\mathcal{E}_{\rm out}$ of the target task was expressed in its basic form in Eq.~(\ref{eq:out of sample error - target data class - beta form}) for a specific coordinate subset layout ${\mathcal{L} = \lbrace{ \mathcal{S}, \mathcal{F}, \mathcal{T}, \mathcal{Z} }\rbrace}$. 
Please note that the expectations below do not include any expectation with respect to $\mathcal{L}$ or its components, which are non-random here. 

We start with the relevant decomposition of the error expression, namely, 
\begin{align}
\mathcal{E}_{\rm out} & =   \sigma_{\epsilon}^2 + \expectation{ \left \Vert { \widehat{\vecgreek{\beta}} - \vecgreek{\beta} } \right \Vert _2^2 }  
\nonumber \\ 
& =   \sigma_{\epsilon}^2 +  \Ltwonorm{\vecgreek{\beta}_{\mathcal{Z}}} 
+ 
\expectation{ \Ltwonorm{ \mtx{X}_{\mathcal{F}}^{+}\left({\vec{y} - \mtx{X}_{\mathcal{T}} \widehat{\vecgreek{\theta}}_{\mathcal{T}} }\right)  - \vecgreek{\beta}_{\mathcal{F}} } } + 
\expectation{ \Ltwonorm{ \widehat{\vecgreek{\theta}}_{\mathcal{T}}   - \vecgreek{\beta}_{\mathcal{T}} } }. 
\label{appendix:eq:out of sample error - target task - expectation for L - development - decomposition - SPECIFIC}
\end{align}
Then, we use the expression for the estimate $\widehat{\vecgreek{\theta}}$ given in (\ref{eq:constrained linear regression - solution - source data class}) and the relation ${\vec{y}=\mtx{X}\vecgreek{\beta}+\vecgreek{\epsilon}}$, to decompose the third term in (\ref{appendix:eq:out of sample error - target task - expectation for L - development - decomposition - SPECIFIC}) as follows 
\begin{align}
& \expectation{ \Ltwonorm{ \mtx{X}_{\mathcal{F}}^{+}\left({\vec{y} - \mtx{X}_{\mathcal{T}} \widehat{\vecgreek{\theta}}_{\mathcal{T}} }\right)  - \vecgreek{\beta}_{\mathcal{F}} } }
\nonumber\\\nonumber
&= \expectation{ \Ltwonorm{ \mtx{X}_{\mathcal{F}}^{+}\left({\vec{y} - \mtx{X}_{\mathcal{T}} \expectation{\widehat{\vecgreek{\theta}}_{\mathcal{T}}} }\right)  - \vecgreek{\beta}_{\mathcal{F}} } } + \expectation{ \Ltwonorm{ \mtx{X}_{\mathcal{F}}^{+}\mtx{X}_{\mathcal{T}}\left({ \widehat{\vecgreek{\theta}}_{\mathcal{T}} - \expectation{\widehat{\vecgreek{\theta}}_{\mathcal{T}}} }\right) } }
\nonumber\\\nonumber
&= \expectation{ \Ltwonorm{ \mtx{X}_{\mathcal{F}}^{+}\left({ \mtx{X}_{\mathcal{T}} \vecgreek{\beta}_{\mathcal{T}} + \mtx{X}_{\mathcal{T}^{\rm c}} \vecgreek{\beta}_{\mathcal{T}^{\rm c}} +\vecgreek{\epsilon}  - \mtx{X}_{\mathcal{T}} \expectation{\widehat{\vecgreek{\theta}}_{\mathcal{T}}} }\right)  - \vecgreek{\beta}_{\mathcal{F}} } } + \expectation{ \Ltwonorm{ \mtx{X}_{\mathcal{F}}^{+}\mtx{X}_{\mathcal{T}}\left({ \widehat{\vecgreek{\theta}}_{\mathcal{T}} - \expectation{\widehat{\vecgreek{\theta}}_{\mathcal{T}}} }\right) } }
\nonumber\\\nonumber
&= \expectation{ \Ltwonorm{ \mtx{X}_{\mathcal{F}}^{+}\left({ \mtx{X}_{\mathcal{T}^{\rm c}} \vecgreek{\beta}_{\mathcal{T}^{\rm c}} +\vecgreek{\epsilon} }\right)  - \vecgreek{\beta}_{\mathcal{F}} } } 
+ \expectation{ \Ltwonorm{ \mtx{X}_{\mathcal{F}}^{+}\mtx{X}_{\mathcal{T}}\left({ \vecgreek{\beta}_{\mathcal{T}} - \expectation{\widehat{\vecgreek{\theta}}_{\mathcal{T}}} }\right) } }
\nonumber\\\nonumber
&\quad+ \expectation{ \Ltwonorm{ \mtx{X}_{\mathcal{F}}^{+}\mtx{X}_{\mathcal{T}}\left({ \widehat{\vecgreek{\theta}}_{\mathcal{T}} - \expectation{\widehat{\vecgreek{\theta}}_{\mathcal{T}}} }\right) } }
\\
\label{appendix:eq:out of sample error - target task - expectation for L - development - third term development - SPECIFIC - new}
\end{align}
We further develop the last expression using the result in (\ref{appendix:eq:lemma - result of X_F and independent u}) and that ${\mtx{X}_{\mathcal{T}}^{T} \mtx{X}_{\mathcal{T}} \sim \mathcal{W}_{t}\left({\mtx{I}_{t},n}\right)}$ is a Wishart matrix with mean  ${\expectation{\mtx{X}_{\mathcal{T}}^{T} \mtx{X}_{\mathcal{T}}} = n \mtx{I}_{t} }$, and get 
\begin{align}
& \expectation{ \Ltwonorm{ \mtx{X}_{\mathcal{F}}^{+}\left({\vec{y} - \mtx{X}_{\mathcal{T}} \widehat{\vecgreek{\theta}}_{\mathcal{T}} }\right)  - \vecgreek{\beta}_{\mathcal{F}} } }
\nonumber\\\nonumber
& = \expectation{ \Ltwonorm{ \mtx{X}_{\mathcal{F}}^{+}\left({ \mtx{X}_{\mathcal{T}^{\rm c}} \vecgreek{\beta}_{\mathcal{T}^{\rm c}} +\vecgreek{\epsilon} }\right)  - \vecgreek{\beta}_{\mathcal{F}} } } 
\\ \nonumber
&+ \left({ { \Ltwonorm{  \vecgreek{\beta}_{\mathcal{T}} - \expectation{\widehat{\vecgreek{\theta}}_{\mathcal{T}}}  } }  + \expectation{ \Ltwonorm{  \widehat{\vecgreek{\theta}}_{\mathcal{T}} - \expectation{\widehat{\vecgreek{\theta}}_{\mathcal{T}}} } }
}\right)   \times \begin{cases}
\frac{p}{n-p-1}          & \text{for } p \le n-2, \\
\infty& \text{for } n-1 \le p \le n+1,  \\
\frac{n}{p-n-1}          & \text{for } p \ge n+2.\\
\end{cases}
\\
\label{appendix:eq:out of sample error - target task - expectation for L - development - third term development - SPECIFIC - new - 2}
\end{align}

We proceed to the fourth error term in (\ref{appendix:eq:out of sample error - target task - expectation for L - development - decomposition - SPECIFIC}), i.e., the error in the subvector induced by the specific $\mathcal{T}$ of interest, and develop its formulation as follows: 
\begin{align}
& \expectation{ \Ltwonorm{ \widehat{\vecgreek{\theta}}_{\mathcal{T}}   - \vecgreek{\beta}_{\mathcal{T}} } } = 
\nonumber\\ \nonumber
&= { \Ltwonorm{ \expectation{\widehat{\vecgreek{\theta}}_{\mathcal{T}}} - \vecgreek{\beta}_{\mathcal{T}} }} + \expectation{ \Ltwonorm{\widehat{\vecgreek{\theta}}_{\mathcal{T}} - \expectation{\widehat{\vecgreek{\theta}}_{\mathcal{T}}}}}
+2 \expectation{ \left({\widehat{\vecgreek{\theta}}_{\mathcal{T}} - \expectation{\widehat{\vecgreek{\theta}}_{\mathcal{T}}}}\right)^{T} \left({ \expectation{\widehat{\vecgreek{\theta}}_{\mathcal{T}}} - \vecgreek{\beta}_{\mathcal{T}} }\right) } 
\\ \nonumber
&= { \Ltwonorm{ \expectation{\widehat{\vecgreek{\theta}}_{\mathcal{T}}} - \vecgreek{\beta}_{\mathcal{T}} }} + \expectation{ \Ltwonorm{\widehat{\vecgreek{\theta}}_{\mathcal{T}} - \expectation{\widehat{\vecgreek{\theta}}_{\mathcal{T}}}}} 
\\ 
\label{appendix:eq:out of sample error - target task - expectation for L - development - fourth term development - SPECIFIC - new}
\end{align}

Then, setting (\ref{appendix:eq:out of sample error - target task - expectation for L - development - third term development - SPECIFIC - new - 2}) and (\ref{appendix:eq:out of sample error - target task - expectation for L - development - fourth term development - SPECIFIC - new}) in (\ref{appendix:eq:out of sample error - target task - expectation for L - development - decomposition - SPECIFIC}) gives 
\begin{align}
&\mathcal{E}_{\rm out} =   \sigma_{\epsilon}^2 +  \Ltwonorm{\vecgreek{\beta}_{\mathcal{Z}}} 
+  \expectation{ \Ltwonorm{ \mtx{X}_{\mathcal{F}}^{+}\left({ \mtx{X}_{\mathcal{T}^{\rm c}} \vecgreek{\beta}_{\mathcal{T}^{\rm c}} +\vecgreek{\epsilon} }\right)  - \vecgreek{\beta}_{\mathcal{F}} } }
\nonumber\\ \nonumber
&+ \left({{ \Ltwonorm{ \expectation{\widehat{\vecgreek{\theta}}_{\mathcal{T}}} - \vecgreek{\beta}_{\mathcal{T}} }} + \expectation{ \Ltwonorm{ \widehat{\vecgreek{\theta}}_{\mathcal{T}} - \expectation{\widehat{\vecgreek{\theta}}_{\mathcal{T}}} }} }\right) \left({1 + \begin{cases}
	\frac{p}{n-p-1}          & \text{for } p \le n-2, \\
	\infty& \text{for } n-1 \le p \le n+1,  \\
	\frac{n}{p-n-1}          & \text{for } p \ge n+2.\\
	\end{cases}}\right).
\\
\label{appendix:eq:out of sample error - target task - development - decomposition - SPECIFIC - new - bias-variance 1}
\end{align}
Also, 
\begin{align}
\expectation{ \Ltwonorm{ \mtx{X}_{\mathcal{F}}^{+}\left({ \mtx{X}_{\mathcal{T}^{\rm c}} \vecgreek{\beta}_{\mathcal{T}^{\rm c}} +\vecgreek{\epsilon} }\right)  - \vecgreek{\beta}_{\mathcal{F}} } } = \expectation{ \Ltwonorm{ \mtx{X}_{\mathcal{F}}^{+} \mtx{X}_{\mathcal{F}} \vecgreek{\beta}_{\mathcal{F}}   - \vecgreek{\beta}_{\mathcal{F}} } } + \expectation{ \Ltwonorm{ \mtx{X}_{\mathcal{F}}^{+}\left({ \mtx{X}_{\mathcal{Z}} \vecgreek{\beta}_{\mathcal{Z}} +\vecgreek{\epsilon} }\right) } }
\label{appendix:eq:out of sample error - target task - development - decomposition - SPECIFIC - new - bias-variance 2}
\end{align}
where the first term can be developed using  (\ref{appendix:eq:lemma - result of X_F projection and a random vector a}) into  
\begin{align}
\label{appendix:eq:out of sample error - target task - development - decomposition - SPECIFIC - new - bias-variance 2 - term 1}
&\expectation{ \Ltwonorm{ \mtx{X}_{\mathcal{F}}^{+} \mtx{X}_{\mathcal{F}} \vecgreek{\beta}_{\mathcal{F}}   - \vecgreek{\beta}_{\mathcal{F}} } } 
=  \Ltwonorm{\vecgreek{\beta}_{\mathcal{F}}} \times \begin{cases}
0          & \text{for } p \le n, \\
1-\frac{n}{p}          & \text{for } p > n, \\
\end{cases} 
\end{align}
and the second term in (\ref{appendix:eq:out of sample error - target task - development - decomposition - SPECIFIC - new - bias-variance 2}) can be rewritten using the result in (\ref{appendix:eq:lemma - result of X_F and independent u}) and that ${\mtx{X}_{\mathcal{Z}}^{T} \mtx{X}_{\mathcal{Z}} \sim \mathcal{W}_{d-p-t}\left({\mtx{I}_{d-p-t},n}\right)}$ is a Wishart matrix with mean  ${\expectation{\mtx{X}_{\mathcal{Z}}^{T} \mtx{X}_{\mathcal{Z}}} = n \mtx{I}_{d-p-t} }$: 
\begin{align}
&\expectation{ \Ltwonorm{ \mtx{X}_{\mathcal{F}}^{+} \left({\mtx{X}_{\mathcal{Z}} \vecgreek{\beta}_{\mathcal{Z}} +\vecgreek{\epsilon} }\right) } }  =  \left( { \Ltwonorm{\vecgreek{\beta}_{\mathcal{Z}}} + \sigma_{\epsilon}^{2}}\right) \times \begin{cases}
\frac{p}{n-p-1}          & \text{for } p \le n-2, \\
\infty& \text{for } n-1 \le p \le n+1,  \\
\frac{n}{p-n-1}          & \text{for } p \ge n+2. \\
\end{cases} 
\label{appendix:eq:out of sample error - target task - development - decomposition - SPECIFIC - new - bias-variance 2 - term 2}
\end{align}
Hence, (\ref{appendix:eq:out of sample error - target task - development - decomposition - SPECIFIC - new - bias-variance 2})-(\ref{appendix:eq:out of sample error - target task - development - decomposition - SPECIFIC - new - bias-variance 2 - term 2}) let us write (\ref{appendix:eq:out of sample error - target task - development - decomposition - SPECIFIC - new - bias-variance 1}) in the form which is provided in Theorem \ref{theorem:out of sample error - target task - specific layout}.

\subsection{Proof of Corollary \ref{corollary:out of sample error - target task - specific layout - detailed - specific layout}}
\label{appendix:subsec:Proof of Corollary 6}

To prove Corollary \ref{corollary:out of sample error - target task - specific layout - detailed - specific layout} we first formulate the expectation of the transferred parameters: 
\begin{align}
\label{eq:expectation of transferred parameters}
\expectation{ \widehat{\vecgreek{\theta}}_{\mathcal{T}}}= \expectation{ \mtx{Q}_{\mathcal{T}}\mtx{Q}_{\mathcal{S}}^T \mtx{Z}_{\mathcal{S}}^{+} \mtx{Z}\mtx{H}\vecgreek{\beta} }
&= 
\expectation{ \mtx{Q}_{\mathcal{T}}\mtx{Q}_{\mathcal{S}}^T \mtx{Z}_{\mathcal{S}}^{+} \mtx{Z}_{\mathcal{S}}\mtx{Q}_{\mathcal{S}}\mtx{H}\vecgreek{\beta} }
\nonumber\\ &= \mtx{Q}_{\mathcal{T}}\mtx{H}\vecgreek{\beta} \times  \begin{cases}
1         & \text{for } \widetilde{p} \le \widetilde{n}, \\
\frac{\widetilde{n}}{\widetilde{p}}      & \text{for } \widetilde{p} > \widetilde{n}, \\
\end{cases} 
\end{align}
where the last equality stems from (\ref{appendix:eq:lemma - mean of Z_S projection matrix}).
Consequently, we use (\ref{eq:expectation of transferred parameters}) to formulate the transfer bias term from Theorem \ref{theorem:out of sample error - target task - specific layout} as 
\begin{equation}
{\rm{Bias}}_{\mathcal{T}}^{2} =  \Ltwonorm{\expectation{\widehat{\vecgreek{\theta}}_{\mathcal{T}}} - \vecgreek{\beta}_{\mathcal{T}}} =  \Ltwonorm{\mtx{Q}_{\mathcal{T}}\left({r\mtx{H}-\mtx{I}_{d}}\right)\vecgreek{\beta} }
~~~\text{where}~~ r \triangleq  \begin{cases}	
\mathmakebox[2em][l]{ 1 }    \text{for }  \widetilde{p} \le \widetilde{n},  
\\
\mathmakebox[2em][l]{ \frac{\widetilde{n}}{\widetilde{p}} }    \text{for }  \widetilde{p} > \widetilde{n}, 
\end{cases} 
\label{eq:out of sample error - target task - corollary - transfer bias - detailed - specific layout - appendix}
\end{equation}
which corresponds to the formulation in Corollary \ref{corollary:out of sample error - target task - specific layout - detailed - specific layout}.

Now we turn to prove the transfer variance formulation from Corollary \ref{corollary:out of sample error - target task - specific layout - detailed - specific layout}. 
For a start, note that 
\begin{align}
{\rm{Var}}_{\mathcal{T},\mathcal{S}} = \expectation{\Ltwonorm{\widehat{\vecgreek{\theta}}_{\mathcal{T}} - \expectation{\widehat{\vecgreek{\theta}}_{\mathcal{T}}} } } & =  \expectation{\Ltwonorm{\widehat{\vecgreek{\theta}}_{\mathcal{T}} }}  - \Ltwonorm{\expectation{\widehat{\vecgreek{\theta}}_{\mathcal{T}}}} 
\nonumber\\ 
& = \expectation{\Ltwonorm{\widehat{\vecgreek{\theta}}_{\mathcal{T}} }} 
- \Ltwonorm{\mtx{Q}_{\mathcal{T}}\mtx{H}\vecgreek{\beta}} \times  \begin{cases}
1         & \text{for } \widetilde{p} \le \widetilde{n}, \\
\left({\frac{\widetilde{n}}{\widetilde{p}}}\right)^2      & \text{for } \widetilde{p} > \widetilde{n}. \\
\end{cases}
\label{eq:transfer variance - first general expression}
\end{align}	
To further develop the last expression, we will now explicitly formulate $\expectation{ \Ltwonorm{  \widehat{\vecgreek{\theta}}_{\mathcal{T}} } }$. 
Using the auxiliary notation $\vecgreek{\beta}^{(\mtx{H})}\triangleq\mtx{H}\vecgreek{\beta}$ we get 
\begin{align}
\label{appendix:eq:---------}
&\expectation{ \Ltwonorm{  \widehat{\vecgreek{\theta}}_{\mathcal{T}} } }=\expectation{ \Ltwonorm{ \mtx{Q}_{\mathcal{T}}\mtx{Q}_{\mathcal{S}}^T \mtx{Z}_{\mathcal{S}}^{+} \left({ \mtx{Z}\mtx{H}\vecgreek{\beta}+\mtx{Z}\vecgreek{\eta} + \vecgreek{\xi} }\right) } } 
\\ \nonumber
&=\expectation{ \Ltwonorm{  \mtx{Q}_{\mathcal{T}}\mtx{Q}_{\mathcal{S}}^T\mtx{Z}_{\mathcal{S}}^{+} \mtx{Z}_{\mathcal{S}} \left({ \vecgreek{\beta}^{(\mtx{H})}_{\mathcal{S}}+\vecgreek{\eta}_{\mathcal{S}} }\right) } } 
\\ \nonumber
&
\quad+ \expectation{ \Ltwonorm{  \mtx{Q}_{\mathcal{T}}\mtx{Q}_{\mathcal{S}}^T\mtx{Z}_{\mathcal{S}}^{+} \mtx{Z}_{\mathcal{S}^{\rm c}}\left({  \vecgreek{\beta}^{(\mtx{H})}_{\mathcal{S}^{\rm c}}
			+\vecgreek{\eta}_{\mathcal{S}^{\rm c}}}\right) } }
+ \expectation{ \Ltwonorm{  \mtx{Q}_{\mathcal{T}}\mtx{Q}_{\mathcal{S}}^T\mtx{Z}_{\mathcal{S}}^{+} \vecgreek{\xi} } }
\end{align}
that using (\ref{appendix:eq:lemma - SPECIFIC - form 2})-(\ref{appendix:eq:lemma - SPECIFIC - form 4}), (\ref{appendix:eq:lemma - SPECIFIC - form 5}) leads to 
\begin{align}
&  \expectation{ \Ltwonorm{  \widehat{\vecgreek{\theta}}_{\mathcal{T}} } } =
\nonumber\\ \nonumber
& =  \left({ \begin{cases}
	{\Ltwonorm{\vecgreek{\beta}^{(\mtx{H})}_{\mathcal{T}}} + t\sigma_{\eta}^{2}}          &\text{for } \widetilde{p} \le \widetilde{n}, \\
	\frac{\widetilde{n}}{\widetilde{p}\left(\widetilde{p}+1\right)}\left(  \left({ \widetilde{n}+\frac{\widetilde{n}-1}{\widetilde{p}-1} }\right)\left({\Ltwonorm{\vecgreek{\beta}^{(\mtx{H})}_{\mathcal{T}}} + t\sigma_{\eta}^{2}}\right) + \left({ 1-\frac{\widetilde{n}-1}{\widetilde{p}-1} }\right)t\left({\Ltwonorm{\vecgreek{\beta}^{(\mtx{H})}_{\mathcal{S}}} + \widetilde{p}\sigma_{\eta}^{2}}\right)     \right)  &\text{for } \widetilde{p} > \widetilde{n},
	\end{cases} }\right) 
\nonumber\\ \nonumber
& \quad + 
\frac{t}{\widetilde{p}}\left({{\Ltwonorm{\vecgreek{\beta}^{(\mtx{H})}_{\mathcal{S}^{c}}} + (d-\widetilde{p})\sigma_{\eta}^{2}} + \sigma_{\xi}^{2} }\right) 
\times \left({ \begin{cases}
	\frac{\widetilde{p}}{\widetilde{n}-\widetilde{p}-1}          & \text{for } \widetilde{p} \le \widetilde{n}-2, \\
	\infty& \text{for } \widetilde{n}-1 \le \widetilde{p} \le \widetilde{n}+1,  \\
	\frac{\widetilde{n}}{\widetilde{p}-\widetilde{n}-1}          & \text{for } \widetilde{p} \ge \widetilde{n}+2,
	\end{cases} }\right)
\nonumber \\ \nonumber
& =
\begin{cases}
{ \mathmakebox[27em][l] {{ \zeta_{\mathcal{T}}  + t \sigma_{\eta}^2 } + t\cdot\frac{ \zeta_{\mathcal{S}^{c}}  + (d-\widetilde{p}) \sigma_{\eta}^2 + \sigma_{\xi}^2 }{\widetilde{n} - \widetilde{p} - 1} } } \text{for } \widetilde{p} \le \widetilde{n}-2, \\
\mathmakebox[27em][l]{\infty} \text{for } \widetilde{n}-1 \le \widetilde{p} \le \widetilde{n}+1,  \\
\mathmakebox[27em][l] {\frac{\widetilde{n}}{\widetilde{p}} \left( 
	\frac{ \left({\widetilde{p}-\widetilde{n}}\right)t \left( \zeta_{\mathcal{S}} + \widetilde{p}\sigma_{\eta}^2 \right) + (\widetilde{n}\widetilde{p} - 1)\left( { \zeta_{\mathcal{T}}  + t \sigma_{\eta}^2 }\right)  }{\widetilde{p}^2 -  1} + t\cdot\frac{ \zeta_{\mathcal{S}^{c}}  + (d-\widetilde{p}) \sigma_{\eta}^2 + \sigma_{\xi}^2 }{\widetilde{p} - \widetilde{n} - 1}
	\right)} \text{for } \widetilde{p} \ge \widetilde{n}+2.
\end{cases} 
\\
\label{appendix:eq:out of sample error - target task - expectation for L - development - third term development - subterm 3 - expression for Z - SPECIFIC}
\end{align}
where  ${\zeta_{\mathcal{T}}\triangleq\Ltwonorm{ \mtx{Q}_{\mathcal{T}}\mtx{H}\vecgreek{\beta}} }$, ${\zeta_{\mathcal{S}^{c}}\triangleq\Ltwonorm{ \mtx{Q}_{\mathcal{S}^{c}}\mtx{H}\vecgreek{\beta}} }$ and  ${\zeta_{\mathcal{S}}\triangleq \Ltwonorm{\mtx{Q}_{\mathcal{S}}\mtx{H}\vecgreek{\beta} } }$. 
Then, we set (\ref{appendix:eq:out of sample error - target task - expectation for L - development - third term development - subterm 3 - expression for Z - SPECIFIC}) in (\ref{eq:transfer variance - first general expression}) and using some algebra obtain the transfer variance formulation in Corollary \ref{corollary:out of sample error - target task - specific layout - detailed - specific layout}.

\subsection{Proof of Corollary \ref{corollary:out of sample error - target task}}
\label{appendix:subsec:proof of Corollary 4}
Corollary \ref{corollary:out of sample error - target task} formulates the generalization error of the target task under the expectation over a coordinate layout $\mathcal{L}$ which is chosen uniformly at random. 
Recall Definition \ref{definition:coordinate subset layout - uniformly distributed} in the main text that characterizes a coordinate subset layout 
${\mathcal{L} = \lbrace{ \mathcal{S}, \mathcal{F}, \mathcal{T}, \mathcal{Z} }\rbrace}$ that is ${\lbrace{ \widetilde{p}, p, t }\rbrace}$-uniformly distributed, for  ${\widetilde{p}\in \lbrace{1,\dots,d}\rbrace}$ and $\left( {p,t}\right)\in \left\{{0,\dots,d}\right\}\times \left\{{0,\dots,\widetilde{p}}\right\}$ such that ${p+t\le d}$.
Here we provide several auxiliary results that are induced by this random structure and utilized in the proof of Corollary \ref{corollary:out of sample error - target task}.

For
$\mathcal{S}$ that is uniformly chosen at random from all the subsets of $\widetilde{p}$ unique coordinates of $\{{1,\dots,d}\}$, we get that the mean of the projection operator ${\mtx{Q}_{\mathcal{S}}^T{\mtx{Q}_{\mathcal{S}}}}$ is 
\begin{equation}
\label{appendix:eq:Proof of Theorem 3.1 - auxiliary results - expected projection of S}
\expectationwrt{\mtx{Q}_{\mathcal{S}}^T{\mtx{Q}_{\mathcal{S}}}}{\mathcal{L}} =\expectationwrt{\mtx{Q}_{\mathcal{S}}^T{\mtx{Q}_{\mathcal{S}}}}{\mathcal{S}} = \frac{\binom{d-1}{\widetilde{p}-1}}{\binom{d}{\widetilde{p}}} \mtx{I}_{d} = \frac{\widetilde{p}}{d} \mtx{I}_{d} 
\end{equation}
where we used the structure of  $\mtx{Q}_{\mathcal{S}}^T{\mtx{Q}_{\mathcal{S}}}$ that is a $d\times d$ diagonal matrix with its $j^{\rm th}$ diagonal component equals 1 if $j\in\mathcal{S}$ and 0 otherwise. 

Definition \ref{definition:coordinate subset layout - uniformly distributed} also specifies that, given $\mathcal{S}$, the target-task coordinate layout ${\lbrace{ \mathcal{F}, \mathcal{T}, \mathcal{Z} }\rbrace}$ is uniformly chosen at random from all the layouts where $\mathcal{F}$, $\mathcal{T}$, and $\mathcal{Z}$ are three disjoint sets of coordinates that satisfy ${\mathcal{F}\cup \mathcal{T} \cup \mathcal{Z} = \{{1,\dots,d}\}}$ such that $\rvert{\mathcal{F}}\lvert = p$, $\rvert{\mathcal{T}}\lvert = t$ and $\mathcal{T}\subseteq\mathcal{S}$, and $\rvert{\mathcal{Z}}\lvert = d-p-t$. 
Accordingly, 
\begin{align}
\label{appendix:eq:Proof of Theorem 3.1 - auxiliary results - expected projection of T}
\expectationwrt{\mtx{Q}_{\mathcal{T}}^T{\mtx{Q}_{\mathcal{T}}}}{\mathcal{L}} &=
\expectationwrt{\expectationwrt{\mtx{Q}_{\mathcal{T}}^T{\mtx{Q}_{\mathcal{T}}}}{\mathcal{L} | \mathcal{S}}}{S} 
\nonumber\\
&= 
\frac{\binom{\widetilde{p}-1}{t-1}}{\binom{\widetilde{p}}{t}} \expectationwrt{ \mtx{Q}_{\mathcal{S}}^T{\mtx{Q}_{\mathcal{S}}}}{S} 
\nonumber\\
&= \frac{t}{\widetilde{p}} \expectationwrt{\mtx{Q}_{\mathcal{S}}^T{\mtx{Q}_{\mathcal{S}}}}{S}
\nonumber\\
&= \frac{t}{d} \mtx{I}_{d}, 
\end{align}
and similarly 
\begin{align}
\label{appendix:eq:Proof of Theorem 3.1 - auxiliary results - expected projection of F}
\expectationwrt{\mtx{Q}_{\mathcal{F}}^T{\mtx{Q}_{\mathcal{F}}}}{\mathcal{L}}
& = \frac{p}{d} \mtx{I}_{d}, 
\\
\label{appendix:eq:Proof of Theorem 3.1 - auxiliary results - expected projection of Z}
\expectationwrt{\mtx{Q}_{\mathcal{Z}}^T{\mtx{Q}_{\mathcal{Z}}}}{\mathcal{L}}
& = \frac{d-p-t}{d} \mtx{I}_{d}. 
\end{align}
Another useful auxiliary result, based on the relation ${\mtx{Q}_{\mathcal{S}}}\mtx{Q}_{\mathcal{S}}^{T} = \mtx{I}_{\widetilde{p}}$ (carefully note the transpose appearance), is provided by 
\begin{align}
\label{appendix:eq:Proof of Theorem 3.1 - auxiliary results - expected projection of T mixed with S}
\expectationwrt{\mtx{Q}_{\mathcal{S}}\mtx{Q}_{\mathcal{T}}^T{\mtx{Q}_{\mathcal{T}}}\mtx{Q}_{\mathcal{S}}^{T} }{\mathcal{L}}  &= \expectationwrt{\mtx{Q}_{\mathcal{S}}\expectationwrt{\mtx{Q}_{\mathcal{T}}^T{\mtx{Q}_{\mathcal{T}}}}{\mathcal{L} | \mathcal{S}} \mtx{Q}_{\mathcal{S}}^{T} }{S} \nonumber\\ 
&= 
\frac{t}{\widetilde{p}} \expectationwrt{ \mtx{Q}_{\mathcal{S}} \mtx{Q}_{\mathcal{S}}^T{\mtx{Q}_{\mathcal{S}}}\mtx{Q}_{\mathcal{S}}^{T} }{S}  
\nonumber\\ 
&= \frac{t}{\widetilde{p}} \mtx{I}_{\widetilde{p}}. 
\end{align}

The results in (\ref{appendix:eq:Proof of Theorem 3.1 - auxiliary results - expected projection of T})--(\ref{appendix:eq:Proof of Theorem 3.1 - auxiliary results - expected projection of Z}) imply that 
\begin{align}
\label{appendix:eq:Proof of Theorem 3.1 - auxiliary results - expected norm of subvector of beta - T}
\expectationwrt{  \Ltwonorm{\vecgreek{\beta}_{\mathcal{T}}} }{\mathcal{L}} & = \vecgreek{\beta}^{T} \expectationwrt{\mtx{Q}_{\mathcal{T}}^T{\mtx{Q}_{\mathcal{T}}}}{\mathcal{L}} \vecgreek{\beta} =  
\frac{t}{d} \Ltwonorm{\vecgreek{\beta}},  
\\
\label{appendix:eq:Proof of Theorem 3.1 - auxiliary results - expected norm of subvector of beta - F}
\expectationwrt{  \Ltwonorm{\vecgreek{\beta}_{\mathcal{F}}} }{\mathcal{L}} & = \vecgreek{\beta}^{T} \expectationwrt{\mtx{Q}_{\mathcal{F}}^T{\mtx{Q}_{\mathcal{F}}}}{\mathcal{L}} \vecgreek{\beta} =  
\frac{p}{d} \Ltwonorm{\vecgreek{\beta}} ,  
\\
\label{appendix:eq:Proof of Theorem 3.1 - auxiliary results - expected norm of subvector of beta - Z}
\expectationwrt{  \Ltwonorm{ \vecgreek{\beta}_{\mathcal{Z}} } }{\mathcal{L}} & =  \vecgreek{\beta}^{T} \expectationwrt{\mtx{Q}_{\mathcal{Z}}^T{\mtx{Q}_{\mathcal{Z}}}}{\mathcal{L}} \vecgreek{\beta} = 
\frac{d-p-t}{d} \Ltwonorm{\vecgreek{\beta}}, 
\end{align}
where $\vecgreek{\beta}_{\mathcal{T}} \triangleq \mtx{Q}_{\mathcal{T}} \vecgreek{\beta}$, $\vecgreek{\beta}_{\mathcal{F}} \triangleq \mtx{Q}_{\mathcal{F}} \vecgreek{\beta}$, and $\vecgreek{\beta}_{\mathcal{Z}} \triangleq \mtx{Q}_{\mathcal{Z}} \vecgreek{\beta}$. 
Note that the expressions in (\ref{appendix:eq:Proof of Theorem 3.1 - auxiliary results - expected norm of subvector of beta - T})-(\ref{appendix:eq:Proof of Theorem 3.1 - auxiliary results - expected norm of subvector of beta - Z}) hold also for $d$-dimensional deterministic vectors other than $\vecgreek{\beta}$, e.g., (\ref{appendix:eq:Proof of Theorem 3.1 - auxiliary results - expected norm of subvector of beta - T})-(\ref{appendix:eq:Proof of Theorem 3.1 - auxiliary results - expected norm of subvector of beta - Z}) hold for $\vecgreek{\beta}^{(\mtx{H})}\triangleq \mtx{H}\vecgreek{\beta}$. 

Then, the auxiliary results in (\ref{appendix:eq:Proof of Theorem 3.1 - auxiliary results - expected projection of S})-(\ref{appendix:eq:Proof of Theorem 3.1 - auxiliary results - expected norm of subvector of beta - Z}) can be utilized to formulate the expectation over $\mathcal{L}$ of the analytical results in Theorem \ref{theorem:out of sample error - target task - specific layout} and, by that, proving Corollary \ref{corollary:out of sample error - target task}.

\section{Additional Results for Section \ref{sec:The Double Descent Phenomenon in Transfer Learning}}
\label{appendix:sec:Empirical Results for Section 3 Additional Details and Demonstrations}

\subsection{Additional Results for the On-Average Analysis in Section \ref{subsec:On-Average Analysis of Arbitrarily Selected Parameters}}
\label{subsec:double double descent additional Results for the On-Average Analysis}
In Fig.~\ref{appendix:fig:empirical_target_generalization_errors_p_vs_p_tilde_2D_planes} we present the empirically computed values of the out-of-sample squared error of the target task, $\expectationwrt{ \mathcal{E}_{\rm out} }{\mathcal{L}}$, with respect to the number of free parameters $\widetilde{p}$ and $p$ (in the source and target tasks, respectively). The empirical values in Fig.~\ref{appendix:fig:empirical_target_generalization_errors_p_vs_p_tilde_2D_planes} (and also the values denoted by circle markers in Fig.~\ref{fig:target_generalization_errors_vs_p__on_average} in Section \ref{subsec:On-Average Analysis of Arbitrarily Selected Parameters}) were obtained by averaging over 250 experiments where each experiment was carried out based on new realizations of the data matrices, noise components, and the sequential order of adding coordinates to subsets (such as $\mathcal{S}$) for the gradual increase of $\widetilde{p}$ and $p$ within each experiment. Note that the results in Fig.~\ref{fig:error_curves_for_specific_layouts__main_text_sample} do not include averaging over the sequential order of adding coordinates to subsets. 
Each single evaluation of the expectation of the squared error for an out-of-sample data pair ${\left( { \vec{x}^{(\rm test)}, y^{(\rm test)} } \right)}$ was empirically carried out by  averaging over a set of 1000 out-of-sample realizations of data pairs. 
Here $d=120$, $\widetilde{n}=50$, $n=20$, $\| \vecgreek{\beta} \|_2^2 = d$, $\sigma_{\epsilon}^2 = 0.05\cdot d$, $\sigma_{\xi}^2 = 0.025\cdot d$. 
The deterministic $\vecgreek{\beta}\in\mathbb{R}^{d}$ used in the experiments satisfies $\| \vecgreek{\beta} \|_2^2 = d$. 

\begin{figure}[t]
	\begin{center}
		\subfloat[]{\includegraphics[width=0.24\textwidth]{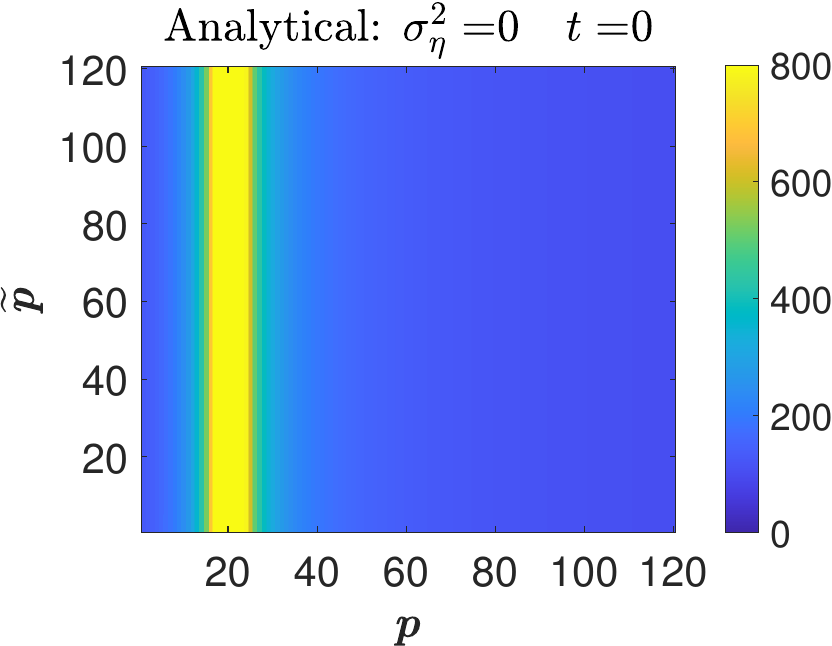}\label{appendix:fig:analytical_target_generalization_errors_eta_0_t0}}
		\subfloat[]{\includegraphics[width=0.24\textwidth]{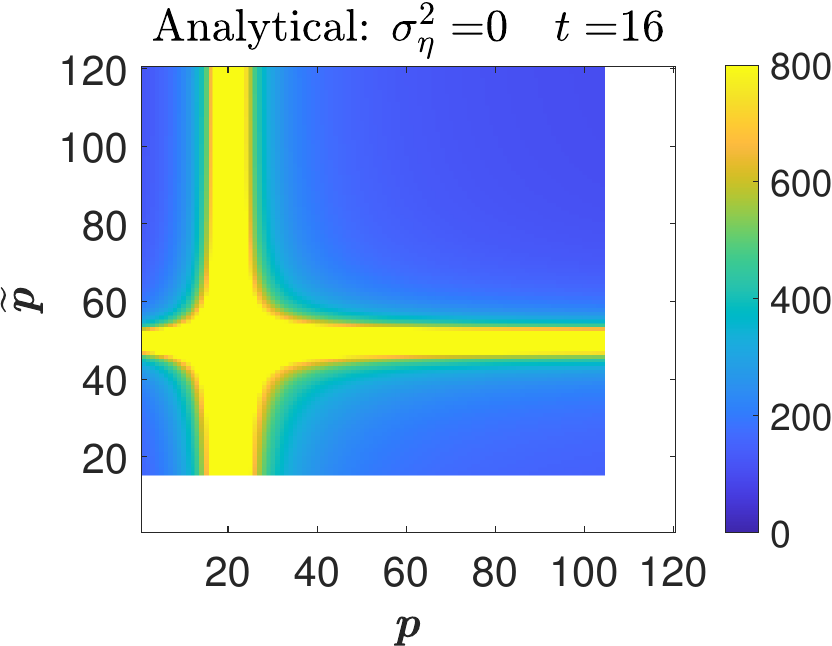} \label{appendix:fig:analytical_target_generalization_errors_eta_0_t16}}
		\subfloat[]{\includegraphics[width=0.24\textwidth]{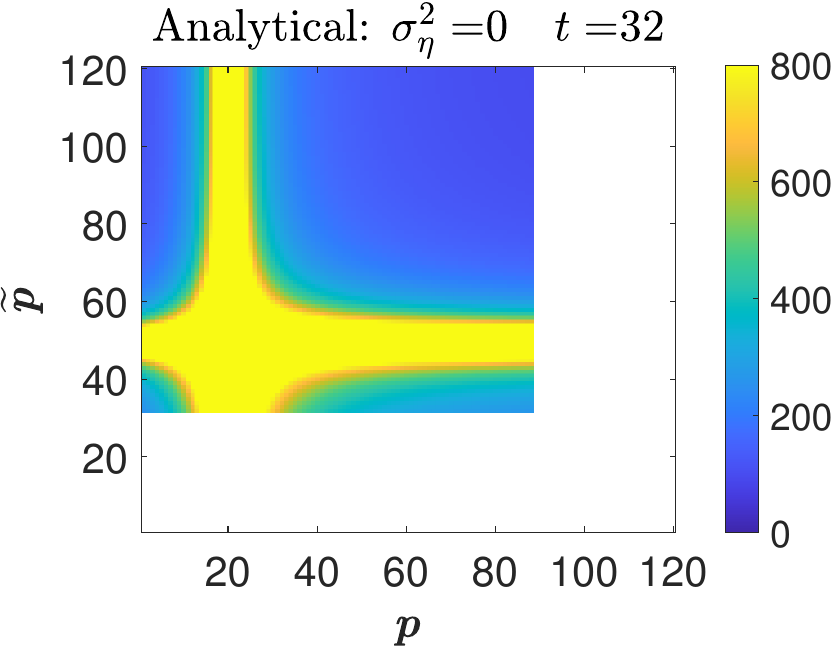} \label{appendix:fig:analytical_target_generalization_errors_eta_0_t32}}
		\subfloat[]{\includegraphics[width=0.24\textwidth]{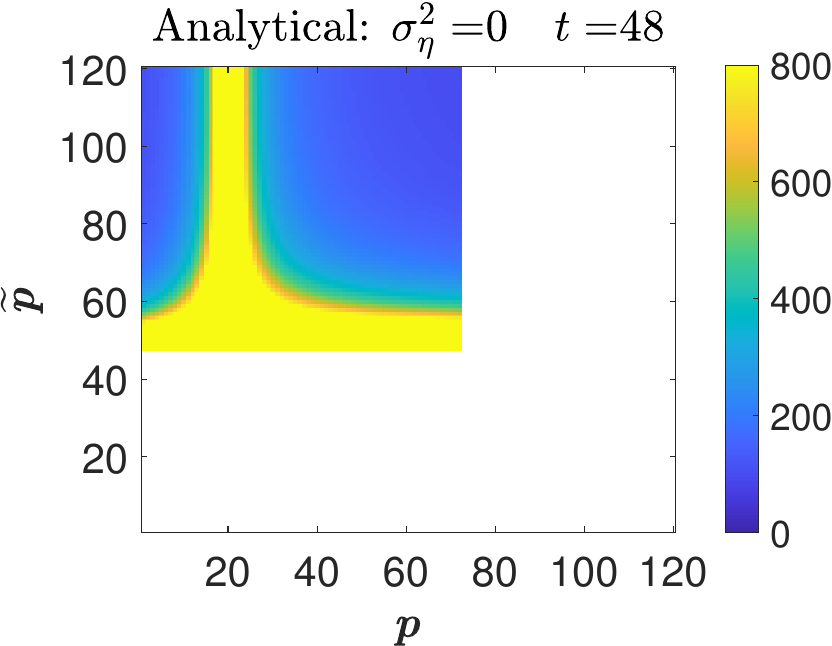} \label{appendix:fig:analytical_target_generalization_errors_eta_0_t48}}
		\\
		\subfloat[]{\includegraphics[width=0.24\textwidth]{figures/analytical_target_generalization_errors_eta_0_2_t0-eps-converted-to.pdf}\label{appendix:fig:analytical_target_generalization_errors_eta_0.2_t0}}
		\subfloat[]{\includegraphics[width=0.24\textwidth]{figures/analytical_target_generalization_errors_eta_0_2_t16-eps-converted-to.pdf} \label{appendix:fig:analytical_target_generalization_errors_eta_0.2_t16}}
		\subfloat[]{\includegraphics[width=0.24\textwidth]{figures/analytical_target_generalization_errors_eta_0_2_t32-eps-converted-to.pdf} \label{appendix:fig:analytical_target_generalization_errors_eta_0.2_t32}}
		\subfloat[]{\includegraphics[width=0.24\textwidth]{figures/analytical_target_generalization_errors_eta_0_2_t48-eps-converted-to.pdf} \label{appendix:fig:analytical_target_generalization_errors_eta_0.2_t48}}
		\\
		\subfloat[]{\includegraphics[width=0.24\textwidth]{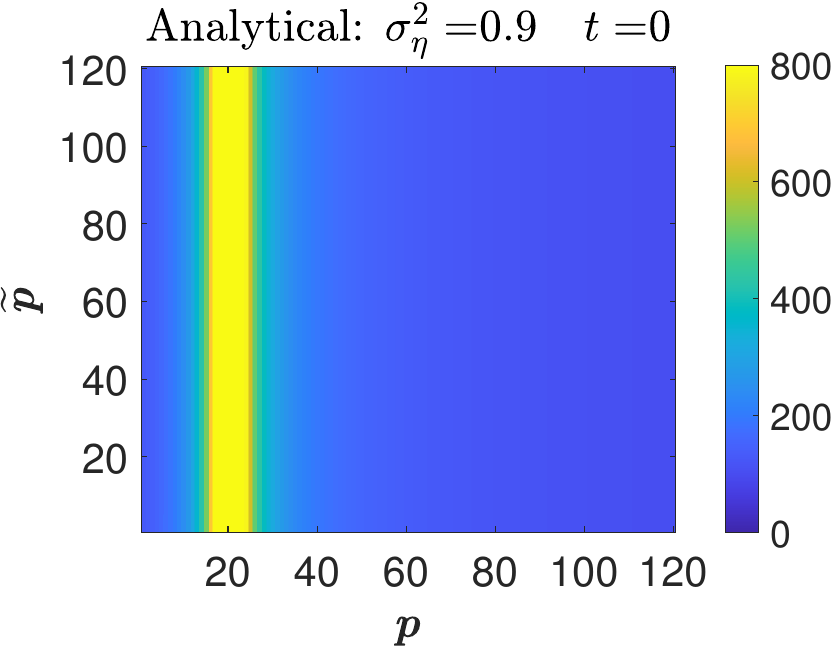}\label{appendix:fig:analytical_target_generalization_errors_eta_0.9_t0}}
		\subfloat[]{\includegraphics[width=0.24\textwidth]{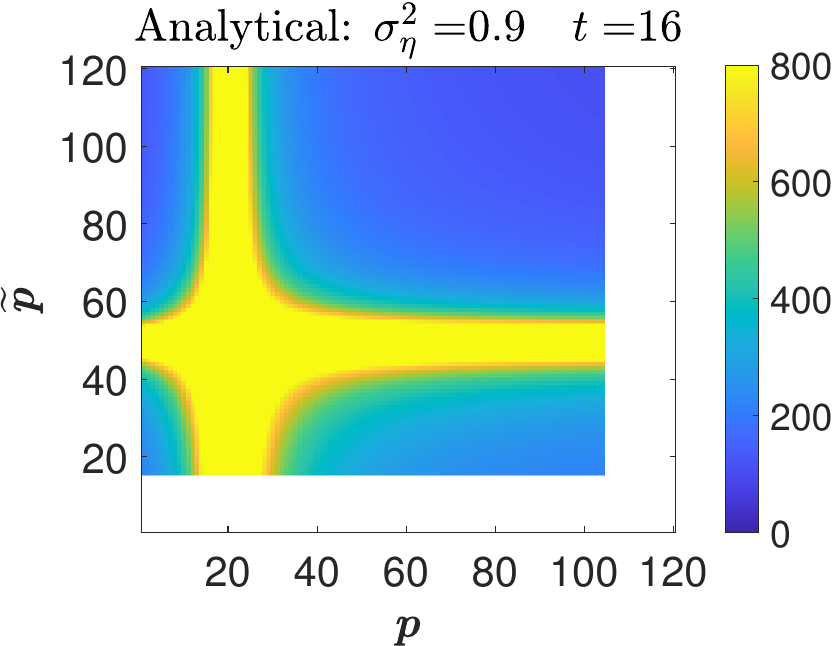} \label{appendix:fig:analytical_target_generalization_errors_eta_0.9_t16}}
		\subfloat[]{\includegraphics[width=0.24\textwidth]{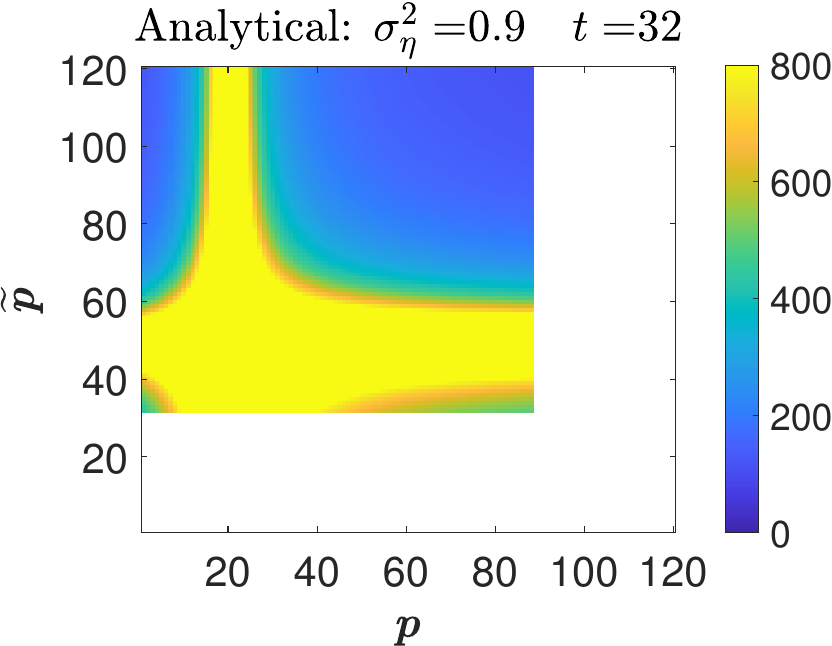} \label{appendix:fig:analytical_target_generalization_errors_eta_0.9_t32}}
		\subfloat[]{\includegraphics[width=0.24\textwidth]{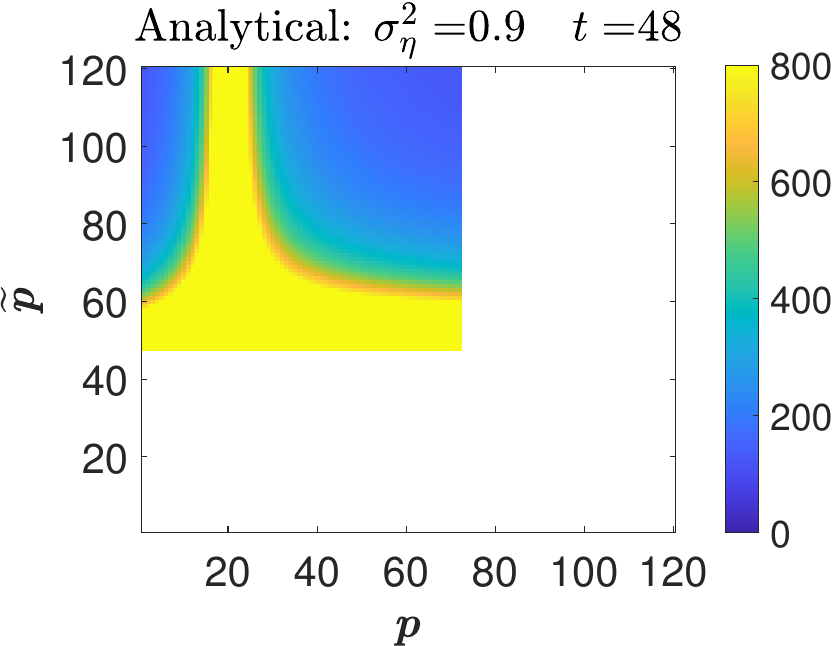} \label{appendix:fig:analytical_target_generalization_errors_eta_0.9_t48}}
		\\
		\subfloat[]{\includegraphics[width=0.24\textwidth]{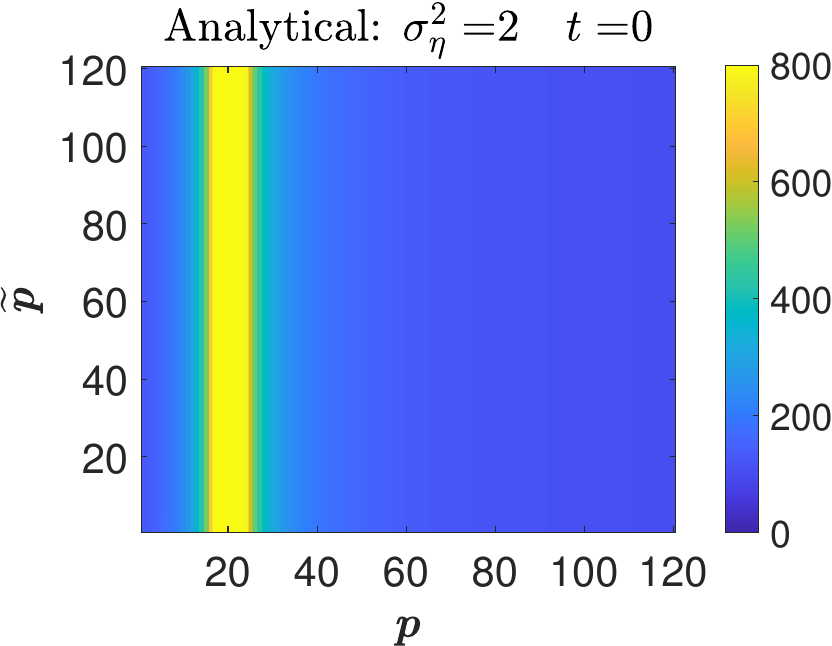}\label{appendix:fig:analytical_target_generalization_errors_eta_2_t0}}
		\subfloat[]{\includegraphics[width=0.24\textwidth]{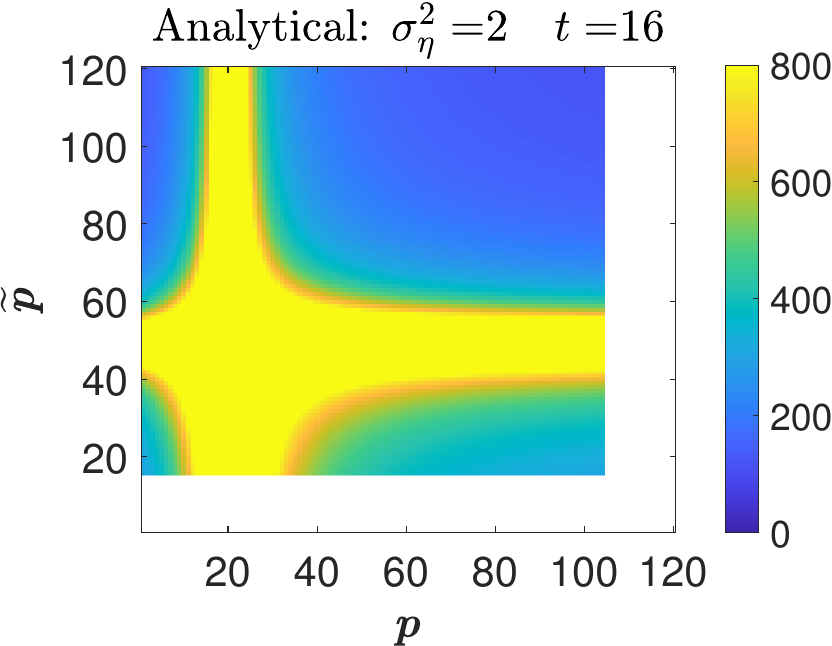} \label{appendix:fig:analytical_target_generalization_errors_eta_2_t16}}
		\subfloat[]{\includegraphics[width=0.24\textwidth]{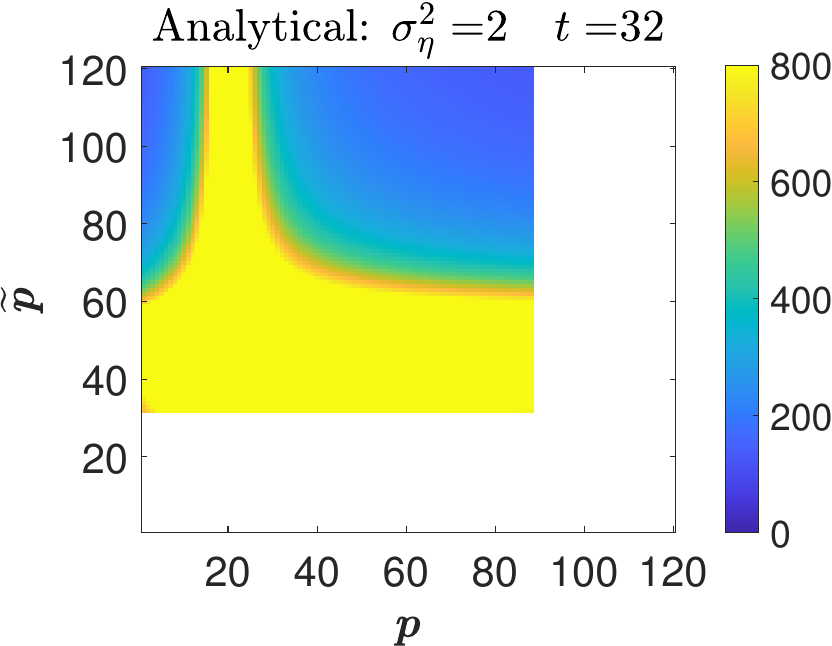} \label{appendix:fig:analytical_target_generalization_errors_eta_2_t32}}
		\subfloat[]{\includegraphics[width=0.24\textwidth]{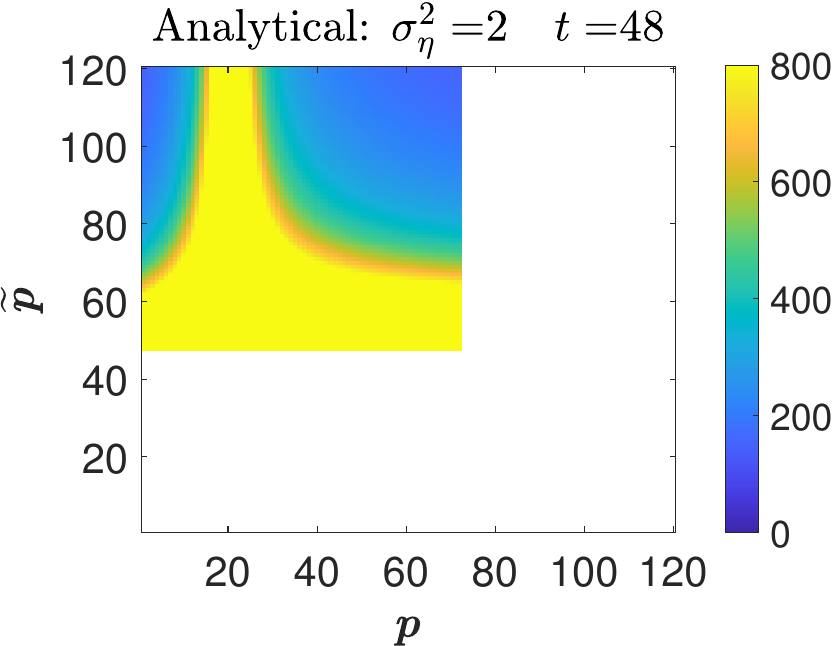} \label{appendix:fig:analytical_target_generalization_errors_eta_2_t48}}
		\caption{\textbf{Analytical} evaluation of the expected out-of-sample squared error of the target task, $\expectationwrt{ \mathcal{E}_{\rm out} }{\mathcal{L}}$, with respect to the number of free parameters $\widetilde{p}$ and $p$ (in the source and target tasks, respectively). Each row of subfigures considers a different case of the relation (\ref{eq:theta-beta relation}) between the source and target tasks in the form of a different noise variance $\sigma_{\eta}^2$ whereas $\mtx{H}$ is a local averaging operator with neighborhood size 5 for all. Each column of subfigures considers a different number of transferred parameters $t$. 
			Here $d=120$, $\widetilde{n}=50$, $n=20$, $\| \vecgreek{\beta} \|_2^2 = d$, $\sigma_{\epsilon}^2 = 0.05\cdot d$, $\sigma_{\xi}^2 = 0.025\cdot d$. The white regions correspond to $\left(\widetilde{p},p\right)$ settings eliminated by the value of $t$ in the specific subfigure. The yellow-colored areas correspond to values greater or equal to 800. See Fig.~\ref{appendix:fig:empirical_target_generalization_errors_p_vs_p_tilde_2D_planes} for the corresponding empirical evaluation.}
		\label{appendix:fig:target_generalization_errors_p_vs_p_tilde_2D_planes}
	\end{center}
	\vspace*{-5mm}
\end{figure}

\begin{figure}[t]
	\begin{center}
		\subfloat[]{\includegraphics[width=0.24\textwidth]{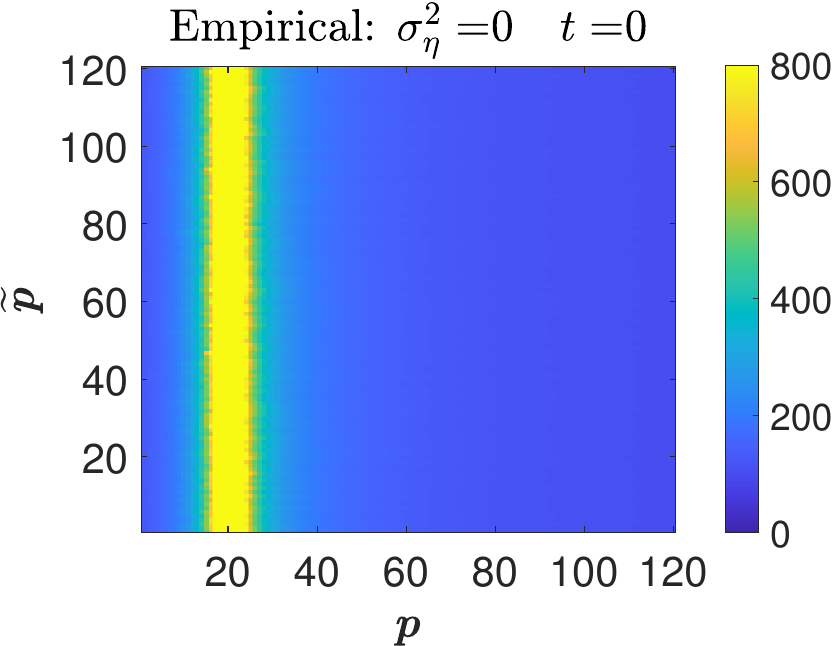}\label{fig:empirical_target_generalization_errors_eta_0_t0}}
		\subfloat[]{\includegraphics[width=0.24\textwidth]{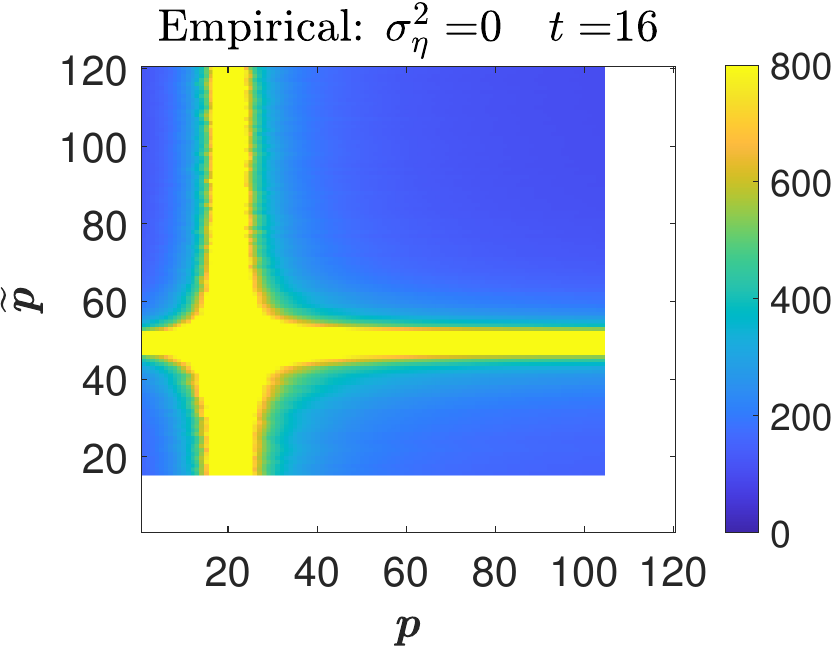} \label{fig:empirical_target_generalization_errors_eta_0_t16}}
		\subfloat[]{\includegraphics[width=0.24\textwidth]{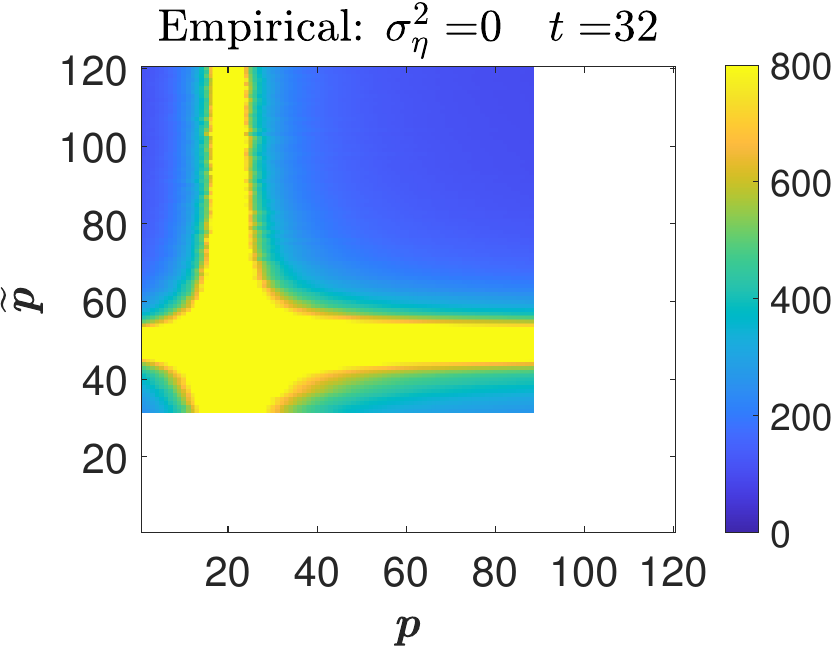} \label{fig:empirical_target_generalization_errors_eta_0_t32}}
		\subfloat[]{\includegraphics[width=0.24\textwidth]{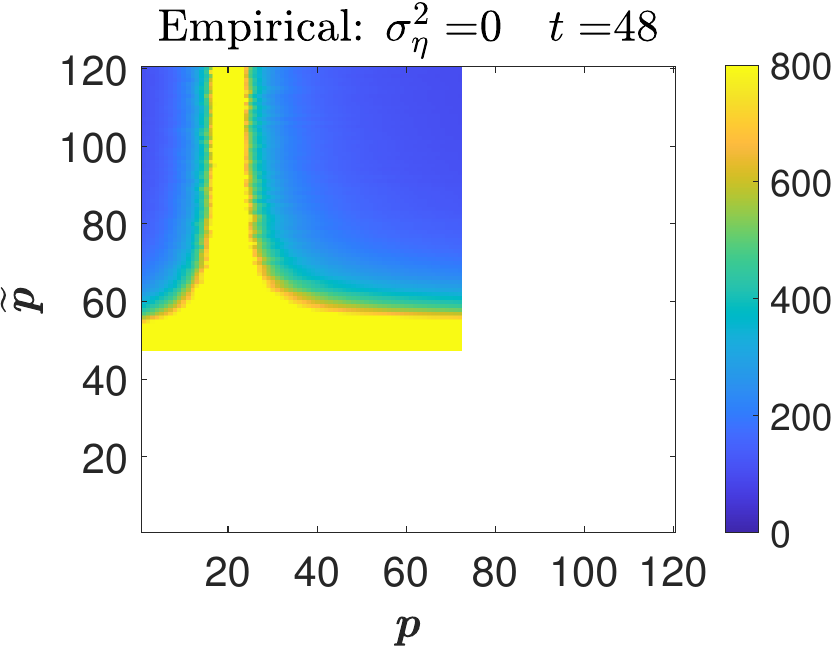} \label{fig:empirical_target_generalization_errors_eta_0_t48}}
		\\
		\subfloat[]{\includegraphics[width=0.24\textwidth]{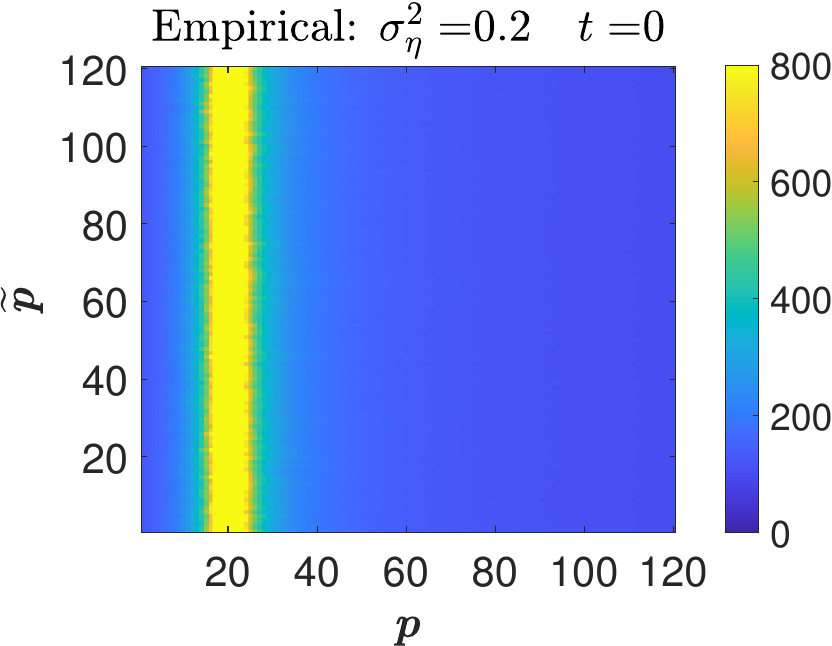}\label{fig:empirical_target_generalization_errors_eta_0.2_t0}}
		\subfloat[]{\includegraphics[width=0.24\textwidth]{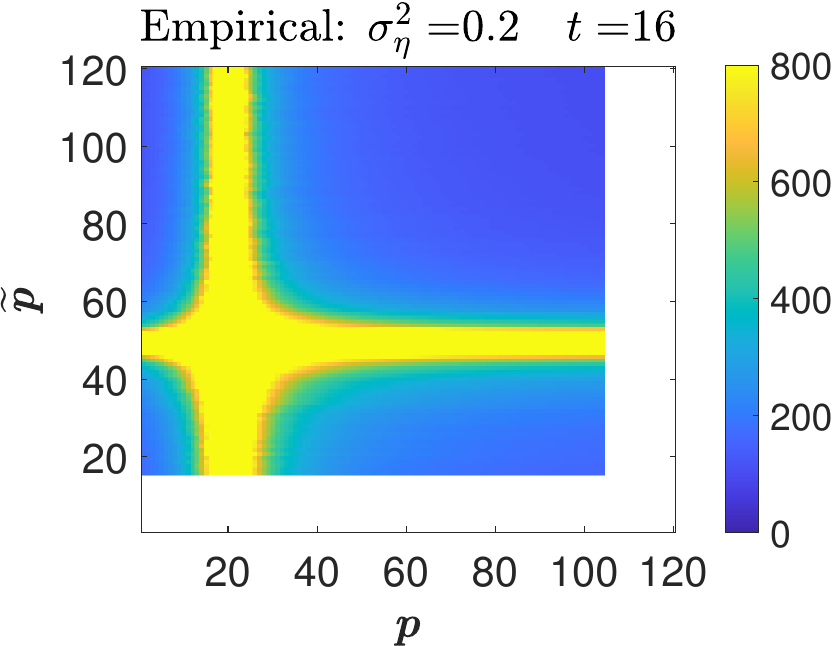} \label{fig:empirical_target_generalization_errors_eta_0.2_t16}}
		\subfloat[]{\includegraphics[width=0.24\textwidth]{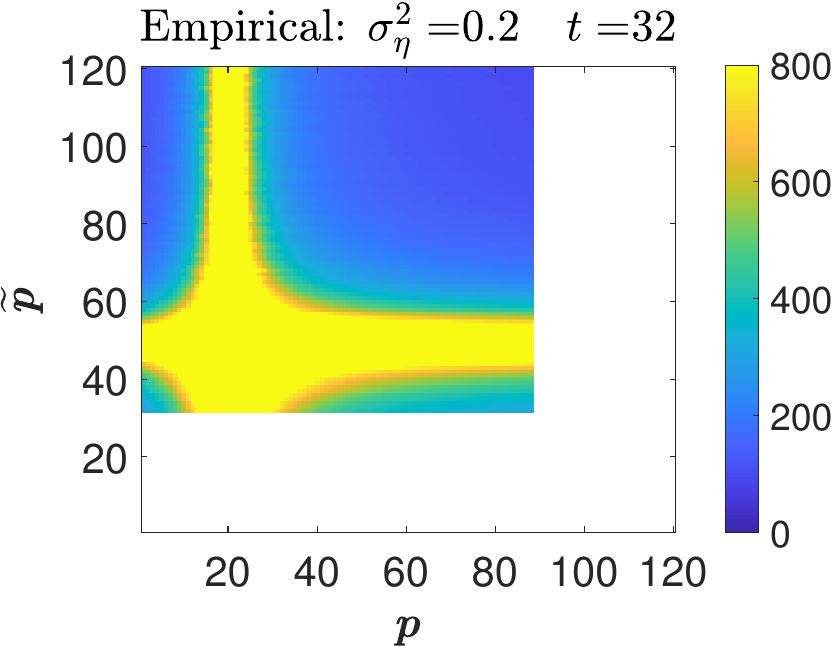} \label{fig:empirical_target_generalization_errors_eta_0.2_t32}}
		\subfloat[]{\includegraphics[width=0.24\textwidth]{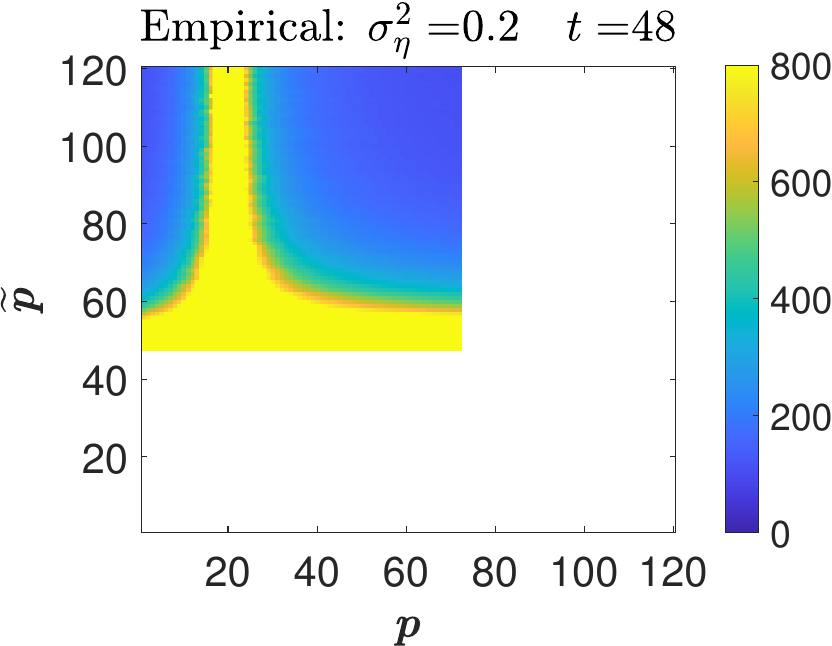} \label{fig:empirical_target_generalization_errors_eta_0.2_t48}}
		\\
		\subfloat[]{\includegraphics[width=0.24\textwidth]{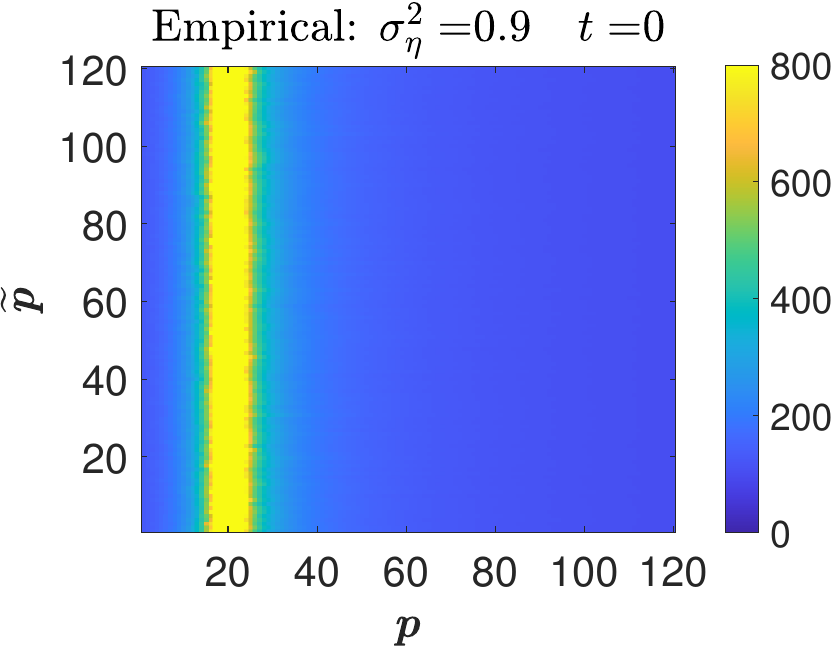}\label{fig:empirical_target_generalization_errors_eta_0.9_t0}}
		\subfloat[]{\includegraphics[width=0.24\textwidth]{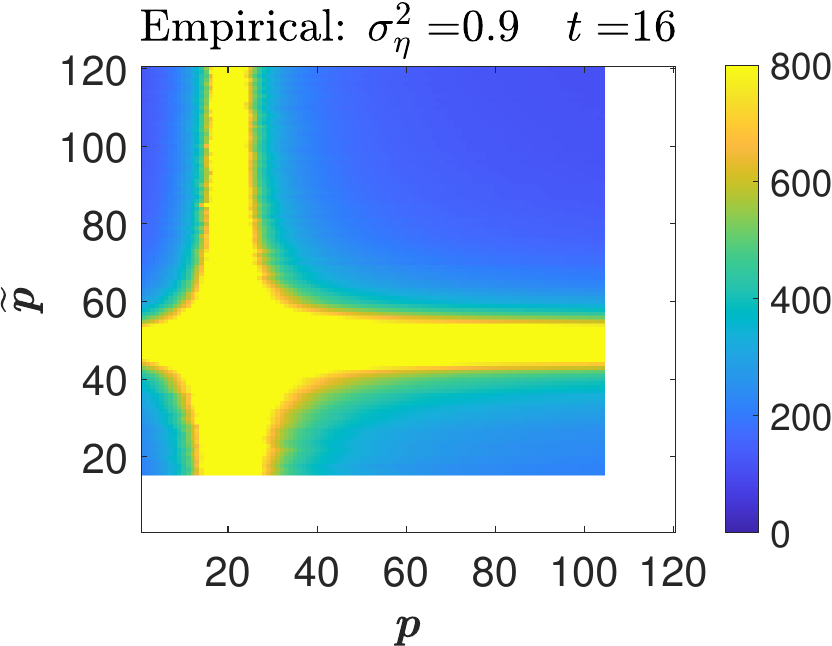} \label{fig:empirical_target_generalization_errors_eta_0.9_t16}}
		\subfloat[]{\includegraphics[width=0.24\textwidth]{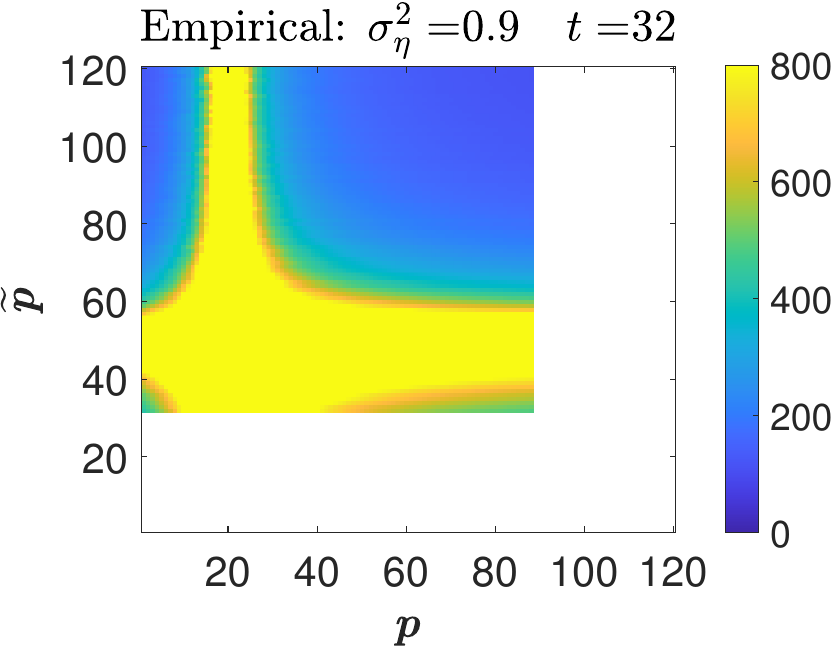} \label{fig:empirical_target_generalization_errors_eta_0.9_t32}}
		\subfloat[]{\includegraphics[width=0.24\textwidth]{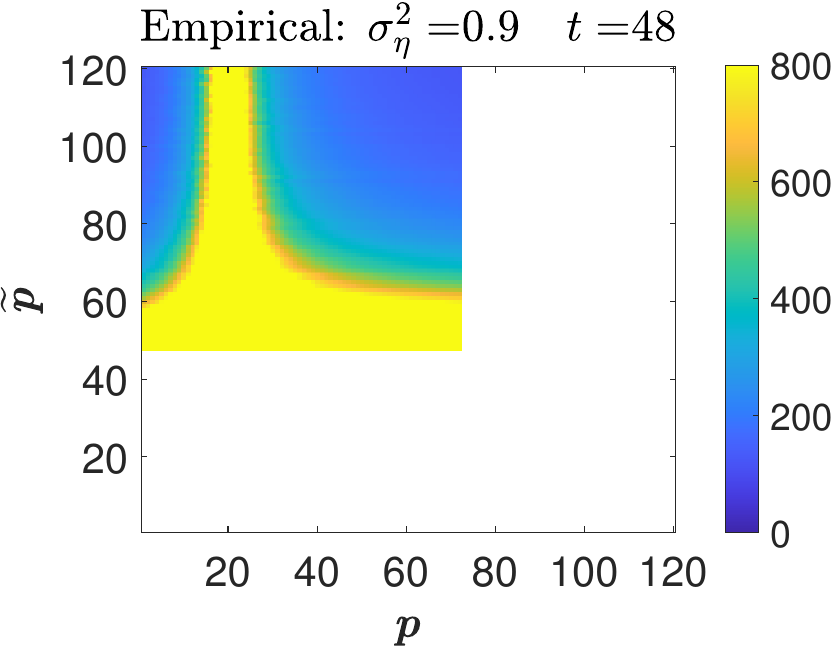} \label{fig:empirical_target_generalization_errors_eta_0.9_t48}}
		\\
		\subfloat[]{\includegraphics[width=0.24\textwidth]{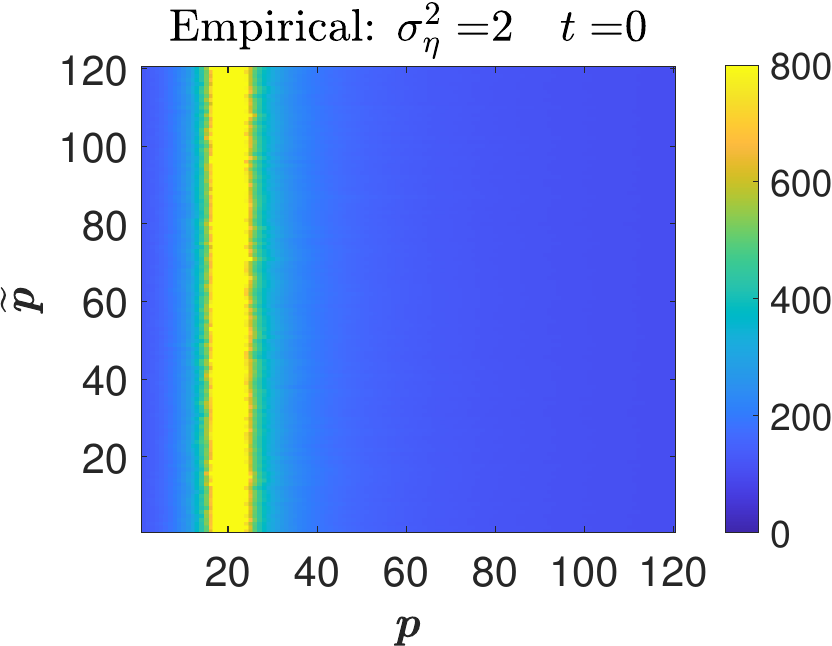}\label{fig:empirical_target_generalization_errors_eta_2_t0}}
		\subfloat[]{\includegraphics[width=0.24\textwidth]{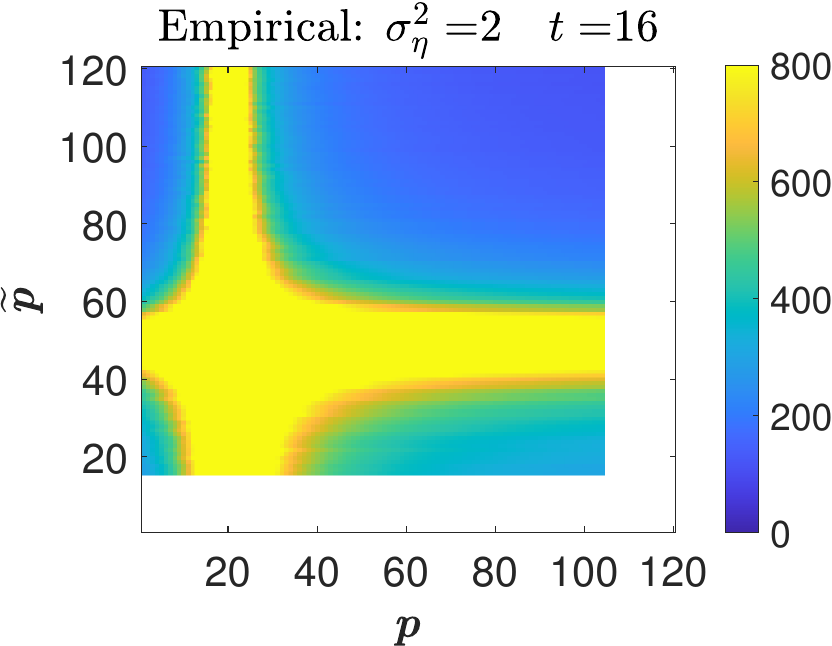} \label{fig:empirical_target_generalization_errors_eta_2_t16}}
		\subfloat[]{\includegraphics[width=0.24\textwidth]{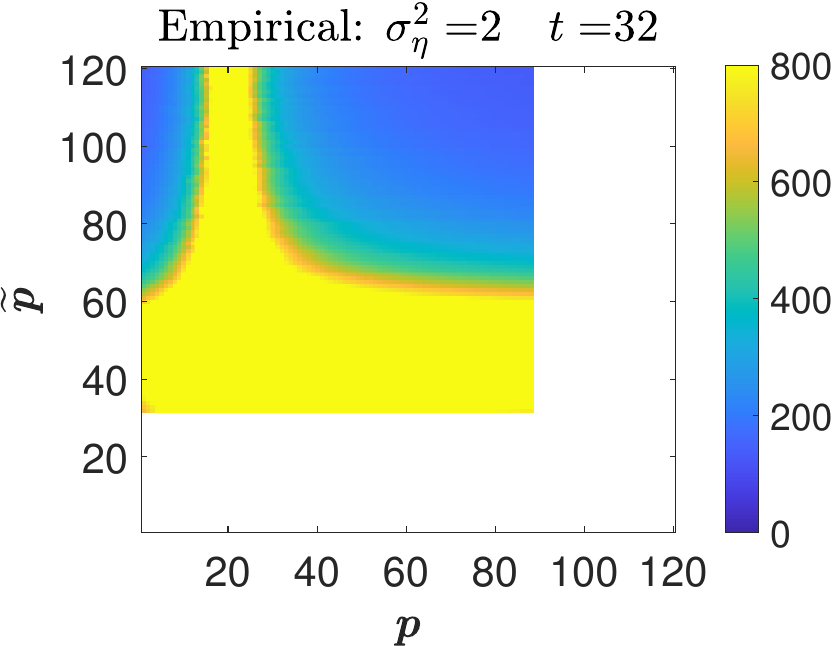} \label{fig:empirical_target_generalization_errors_eta_2_t32}}
		\subfloat[]{\includegraphics[width=0.24\textwidth]{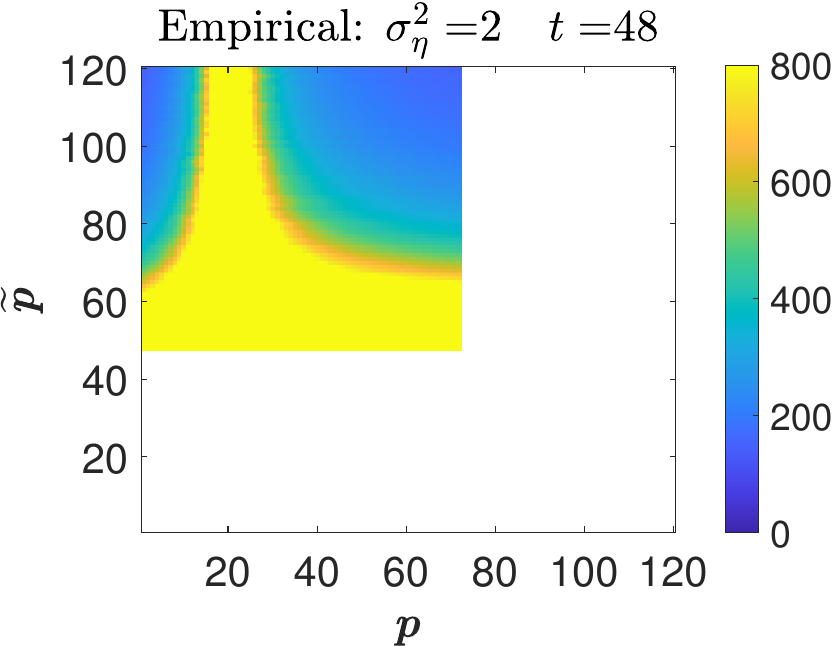} \label{fig:empirical_target_generalization_errors_eta_2_t48}}
		\caption{\textbf{Empirical} evaluation of the expected out-of-sample squared error of the target task, $\expectationwrt{ \mathcal{E}_{\rm out} }{\mathcal{L}}$, with respect to the number of free parameters $\widetilde{p}$ and $p$ (in the source and target tasks, respectively). The presented values obtained by averaging over 250 experiments. The figures here have settings as in the corresponding figures in the analytical evaluation in Fig.~\ref{appendix:fig:target_generalization_errors_p_vs_p_tilde_2D_planes}.}
		\label{appendix:fig:empirical_target_generalization_errors_p_vs_p_tilde_2D_planes}
	\end{center}
	\vspace*{-5mm}
\end{figure}

One can observe the excellent match between the empirical results in  Fig.~\ref{appendix:fig:empirical_target_generalization_errors_p_vs_p_tilde_2D_planes}
and the analytical results provided in Fig.~\ref{appendix:fig:target_generalization_errors_p_vs_p_tilde_2D_planes}. This further establishes the formulations given in Corollary \ref{corollary:out of sample error - target task}. 

\begin{figure}[t]
	\subfloat[{\small Dimensions as in Fig.~ \ref{fig:H_average5_target_generalization_errors_vs_p_eta_0_p_tilde_is_d}}]{\label{fig:target_generalization_errors_vs_p_eta_0_d_1x120_std_areas}\includegraphics[width=0.24\textwidth]{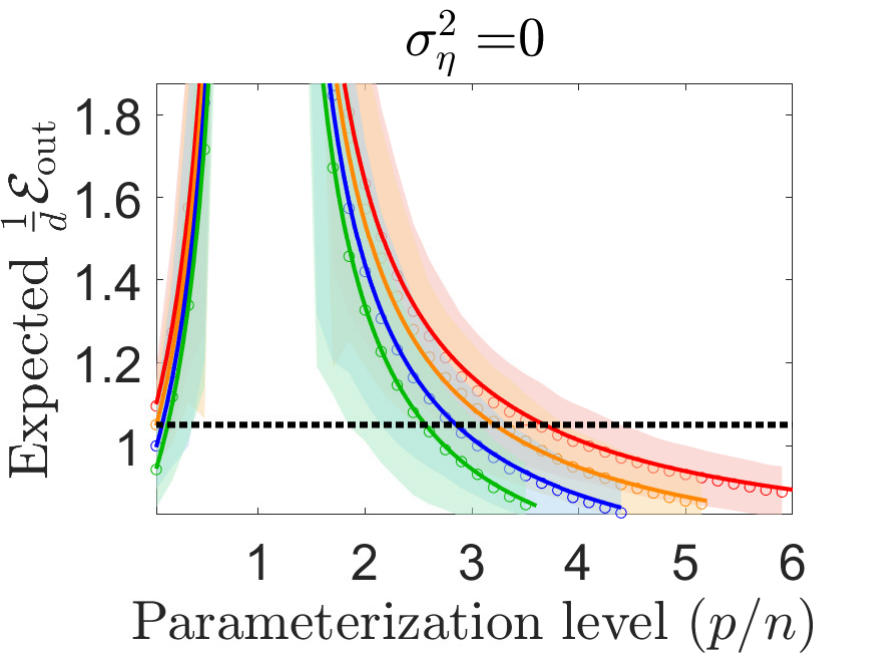}}
	~
	\subfloat[{\small Dimensions as in Fig.~ \ref{fig:H_average5_target_generalization_errors_vs_p_eta_0.2_p_tilde_is_d}}]{\label{fig:target_generalization_errors_vs_p_eta_0_2_d_1x120_std_areas}%
		\includegraphics[width=0.24\textwidth]{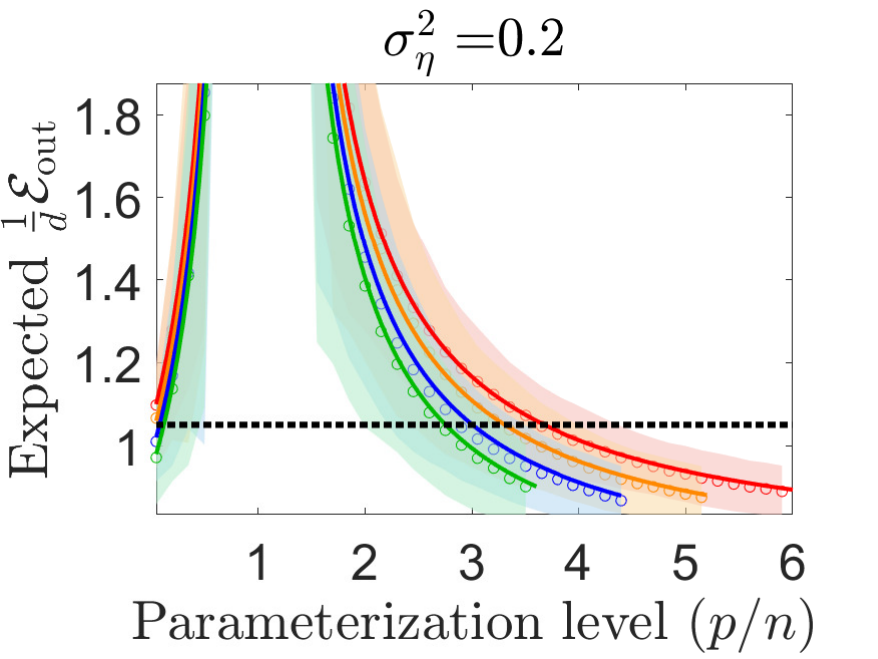}}
	~
	\subfloat[{\small Dimensions as in Fig.~ \ref{fig:H_average5_target_generalization_errors_vs_p_eta_0.9_p_tilde_is_d}}]{\label{fig:target_generalization_errors_vs_p_eta_0_9_d_1x120_std_areas}%
		\includegraphics[width=0.24\textwidth]{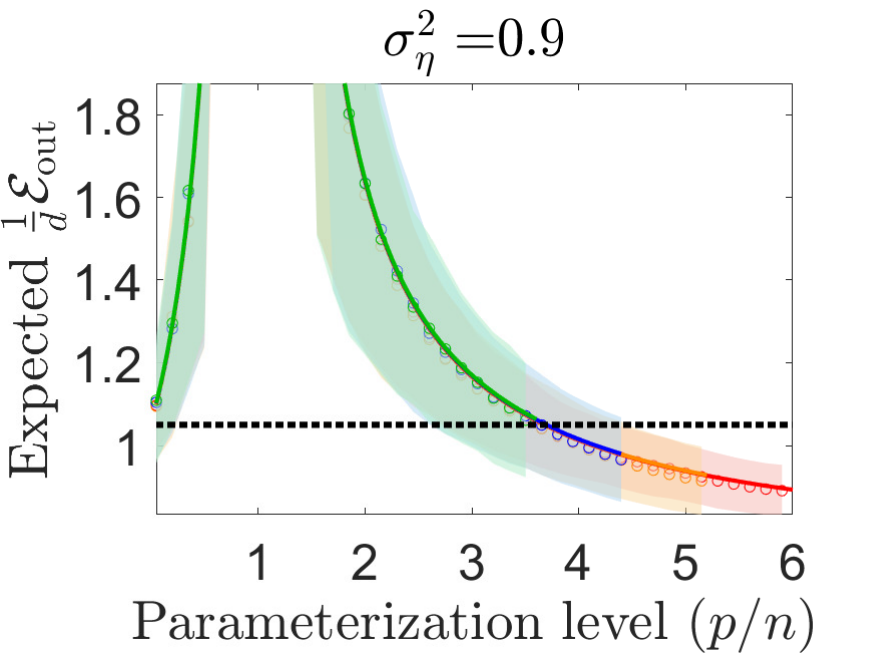}} 
	~
	\subfloat[{\small Dimensions as in Fig.~ \ref{fig:H_average5_target_generalization_errors_vs_p_eta_2_p_tilde_120}}]{\label{fig:target_generalization_errors_vs_p_eta_2_d_1x120_std_areas}%
		\includegraphics[width=0.24\textwidth]{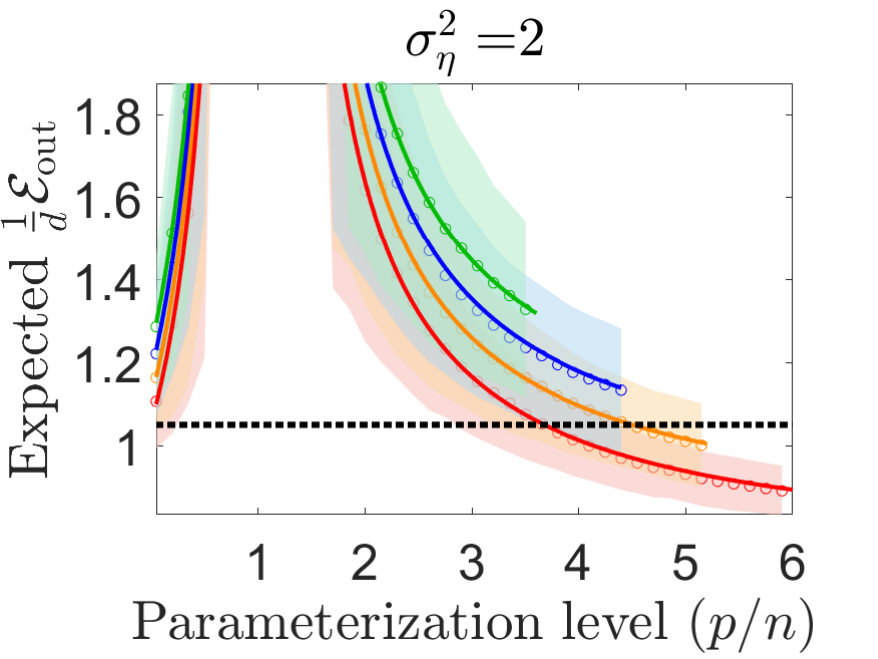}}
	\\~\\
	\subfloat[{\small Dimensions 3 times those in Fig.~\ref{fig:H_average5_target_generalization_errors_vs_p_eta_0_p_tilde_is_d}}]{\label{fig:target_generalization_errors_vs_p_eta_0_d_3x120_std_areas}\includegraphics[width=0.24\textwidth]{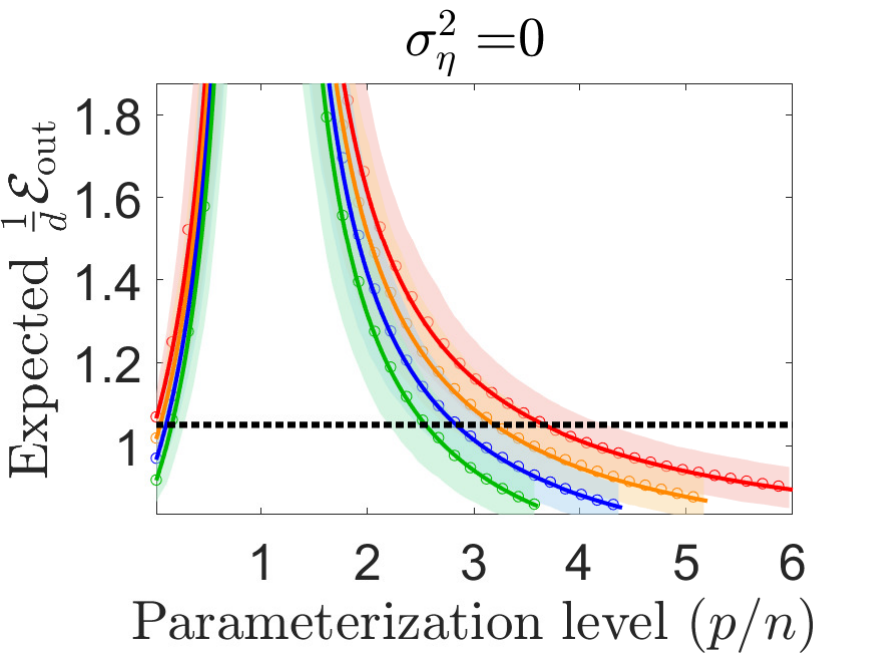}}
	~
	\subfloat[{\small Dimensions 3 times those in Fig.~\ref{fig:H_average5_target_generalization_errors_vs_p_eta_0.2_p_tilde_is_d}}]{\label{fig:target_generalization_errors_vs_p_eta_0_2_d_3x120_std_areas}%
		\includegraphics[width=0.24\textwidth]{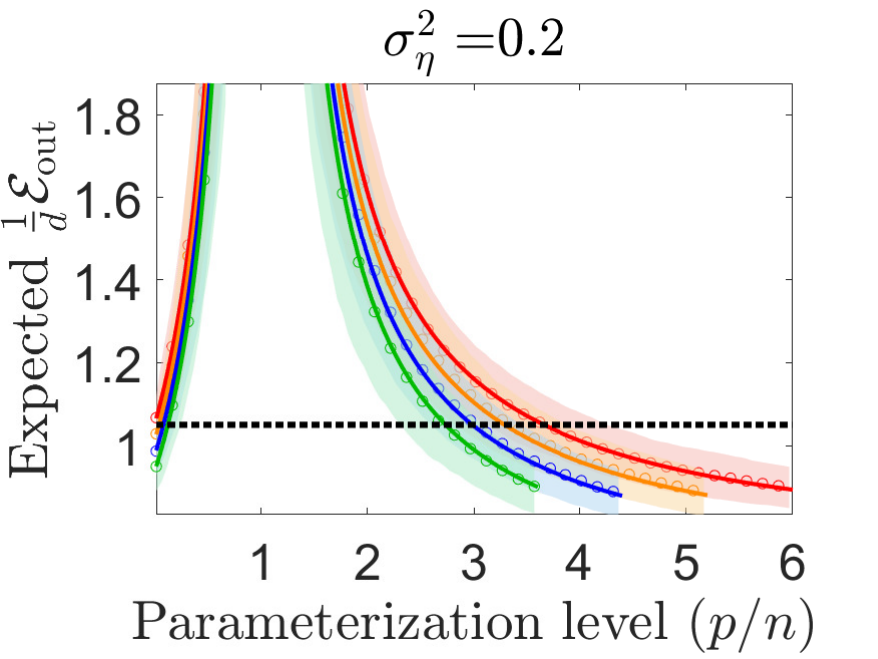}}
	~
	\subfloat[{\small Dimensions 3 times those in Fig.~\ref{fig:H_average5_target_generalization_errors_vs_p_eta_0.9_p_tilde_is_d}}]{\label{fig:target_generalization_errors_vs_p_eta_0_9_d_3x120_std_areas}%
		\includegraphics[width=0.24\textwidth]{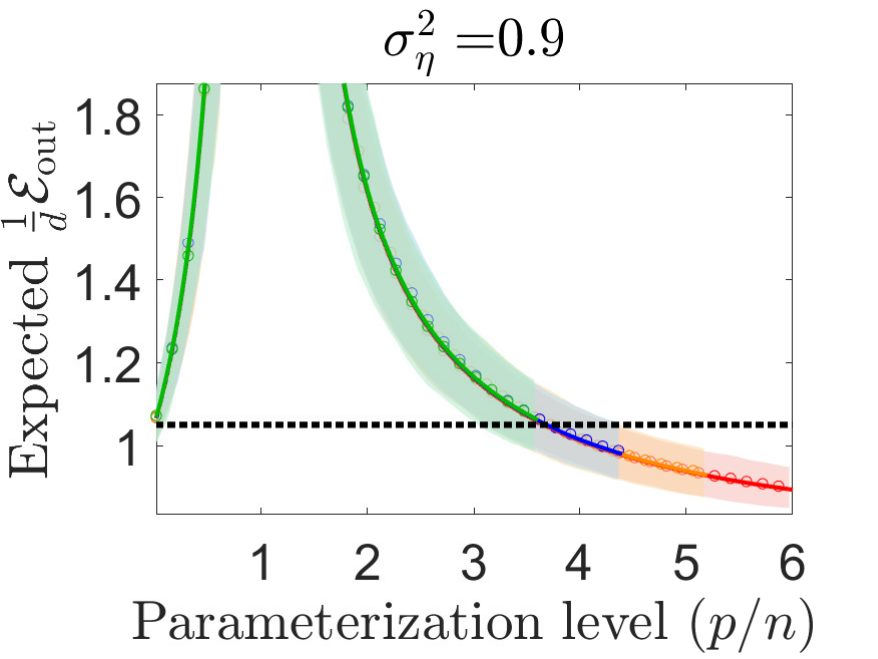}} 
	~
	\subfloat[{\small Dimensions 3 times those in Fig.~\ref{fig:H_average5_target_generalization_errors_vs_p_eta_2_p_tilde_120}}]{\label{fig:target_generalization_errors_vs_p_eta_2_d_3x120_std_areas}%
		\includegraphics[width=0.24\textwidth]{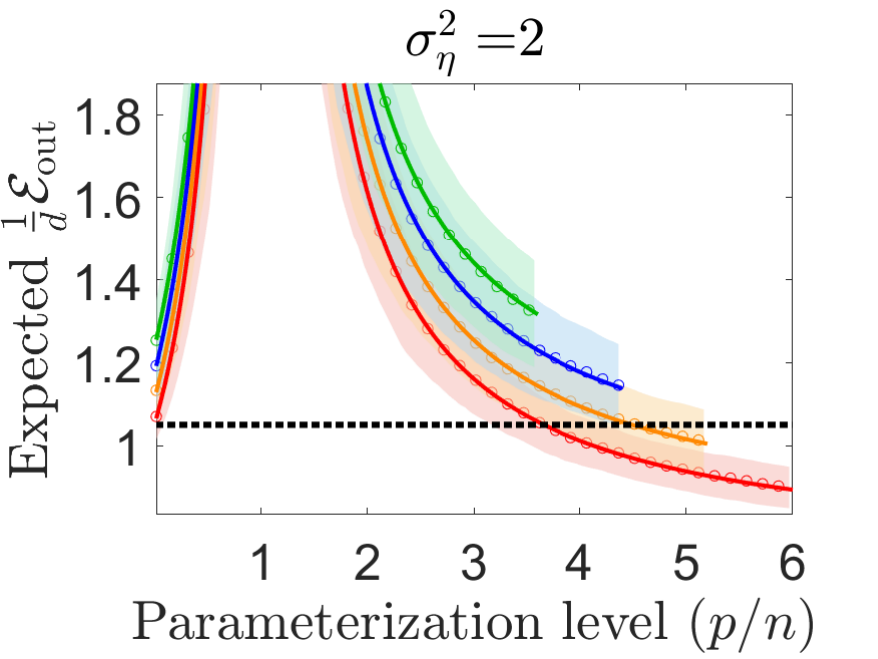}}
	\\~\\
	\subfloat[{\small Dimensions 5 times those in Fig.~\ref{fig:H_average5_target_generalization_errors_vs_p_eta_0_p_tilde_is_d}}]{\label{fig:target_generalization_errors_vs_p_eta_0_d_5x120_std_areas}\includegraphics[width=0.24\textwidth]{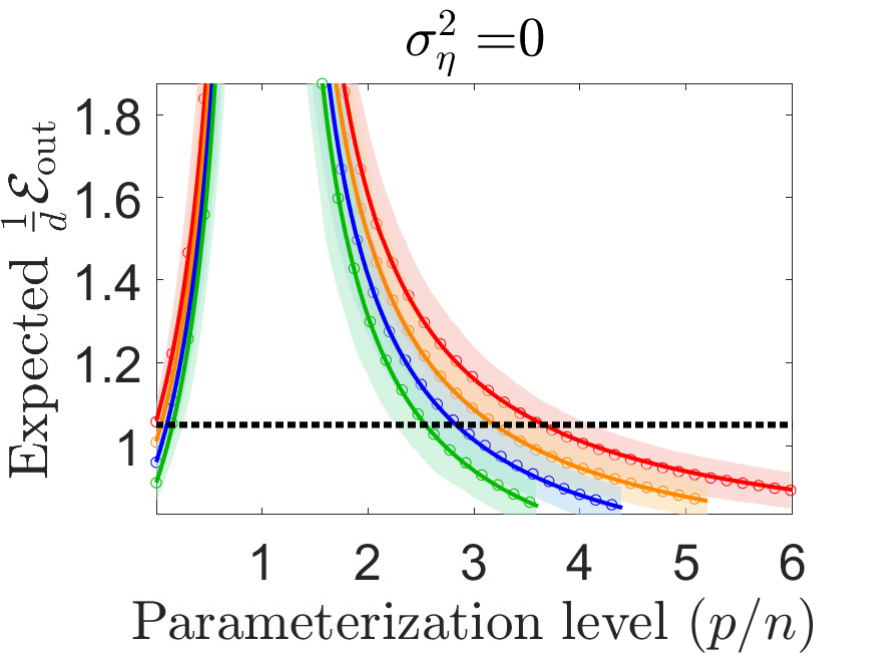}}
	~
	\subfloat[{\small Dimensions 5 times those in Fig.~\ref{fig:H_average5_target_generalization_errors_vs_p_eta_0.2_p_tilde_is_d}}]{\label{fig:target_generalization_errors_vs_p_eta_0_2_d_5x120_std_areas}%
		\includegraphics[width=0.24\textwidth]{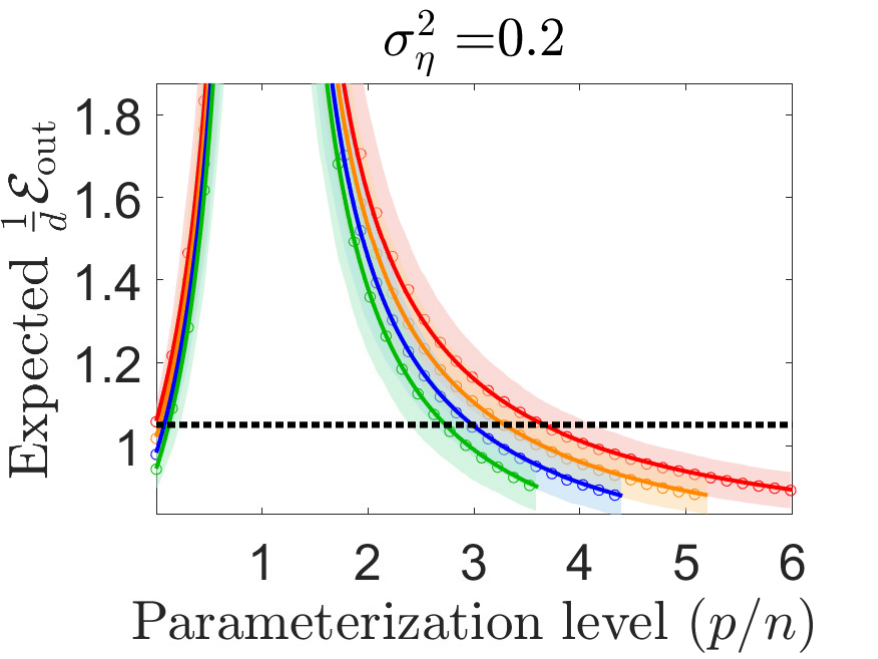}}
	~
	\subfloat[{\small Dimensions 5 times those in Fig.~\ref{fig:H_average5_target_generalization_errors_vs_p_eta_0.9_p_tilde_is_d}}]{\label{fig:target_generalization_errors_vs_p_eta_0_9_d_5x120_std_areas}%
		\includegraphics[width=0.24\textwidth]{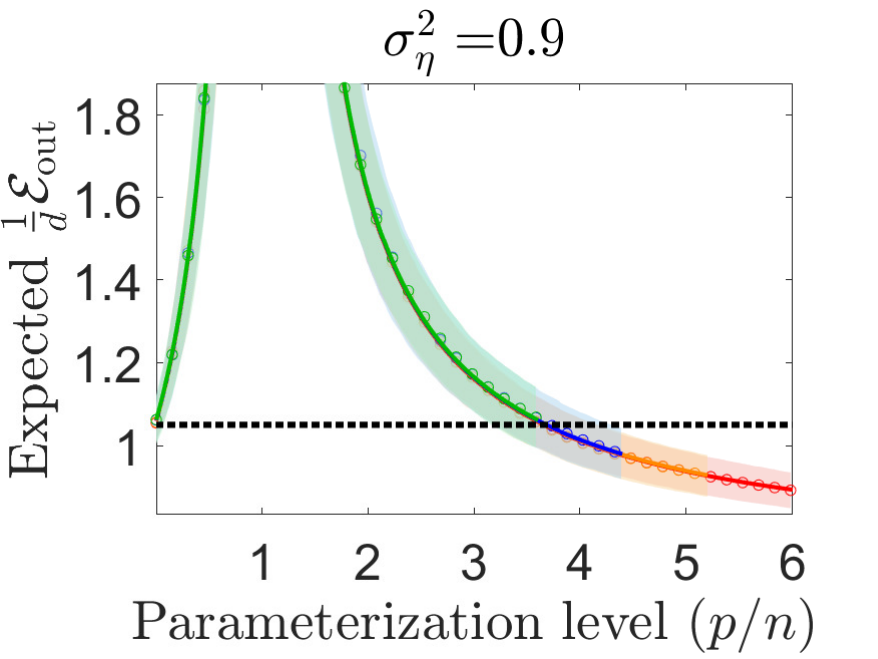}} 
	~
	\subfloat[{\small Dimensions 5 times those in Fig.~\ref{fig:H_average5_target_generalization_errors_vs_p_eta_2_p_tilde_120}}]{\label{fig:target_generalization_errors_vs_p_eta_2_d_5x120_std_areas}%
		\includegraphics[width=0.24\textwidth]{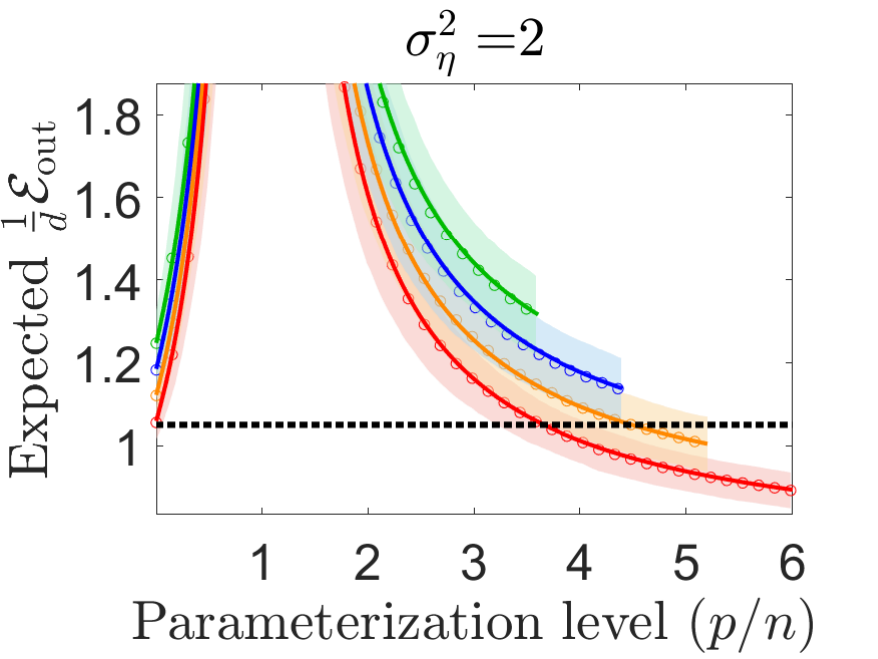}}
	\caption{Demonstration of error concentration. Each row of subfigures corresponds to the settings of Figures \ref{fig:H_average5_target_generalization_errors_vs_p_eta_0_p_tilde_is_d}-\ref{fig:H_average5_target_generalization_errors_vs_p_eta_2_p_tilde_120} but with a different proportional increase of dimensions and dimension-dependent quantities (see explanation in Section \ref{subsec:On-Average Analysis of Arbitrarily Selected Parameters} in the main text). The empirical standard deviations are denoted as shaded areas in colors corresponding to the on-average error curves (solid lines and markers denote the analytical and empirical evaluations of the expected error, respectively). Lines, markers and areas in red correspond to $t=0$ (no parameters are transferred); orange corresponds to transferring $t=16\times\frac{d}{120}$ parameters; blue corresponds to $t=32\times\frac{d}{120}$; green corresponds  to $t=48\times\frac{d}{120}$. Note that $\frac{d}{120}$ equals to 1, 3, 5, in the first, second, third rows of subfigures, respectively. The axes in this figure are normalized to be dimension-independent.}
	\label{appendix:fig:target_generalization_errors_vs_p__on_average_with_standard_deviations_full_comparison}
\end{figure}

\subsection{Additional Results for the Single-Layout Analysis in Section \ref{subsec:The Fragile Nature of Transfer Learning: Analysis of a Single Layout of Arbitrarily Selected Parameters}}
\label{appendix:sec:Transfer of Specific Sets of Parameters: Additional Analytical and Empirical Evaluations of Generalization Error Curves}

The following results are for two different forms of the true solution $\vecgreek{\beta}$: the first is a form with linearly increasing values (Fig.~\ref{appendix:fig:linear_beta_graph}), the second is a form with sparse values where only 25\% of coordinates have non-zero value (Fig.~\ref{appendix:fig:sparse_beta_graph}). Note that both forms satisfy $\Ltwonorm{\vecgreek{\beta}}=d$.

The three types of linear operator $\mtx{H}$ in the evaluations in Section \ref{subsec:The Fragile Nature of Transfer Learning: Analysis of a Single Layout of Arbitrarily Selected Parameters} are as follows. First, $\mtx{H}=\mtx{I}_{d}$ that is the identity operator. Second, is the circulant matrix $\mtx{H}$ that corresponds to a shift-invariant local averaging operator that uniformly considers 11-coordinates neighborhood around the computed coordinate (note that in other parts of this paper we consider also averaging operators with  neighborhood sizes other than 11). Third, is the circulant matrix $\mtx{H}$ that corresponds to discrete derivative operator based on the convolution kernel $[-0.5,0.5]$.

\begin{figure}[t]
	\begin{center}
		\subfloat[]{\includegraphics[width=0.24\textwidth]{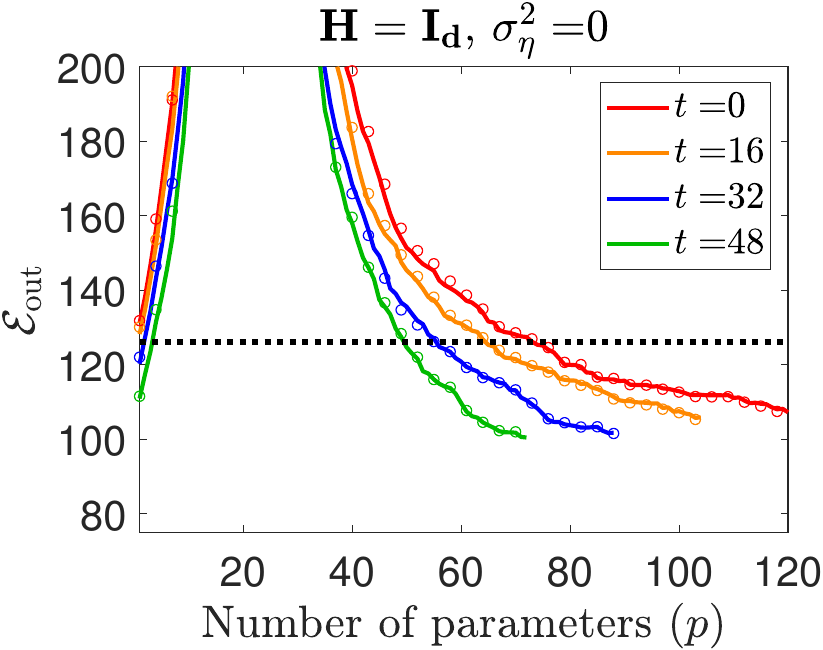}\label{appendix:fig:specific_linear_beta__H_is_I__target_generalization_errors_vs_p_eta_0_p_tilde_is_d}}
		\subfloat[]{\includegraphics[width=0.24\textwidth]{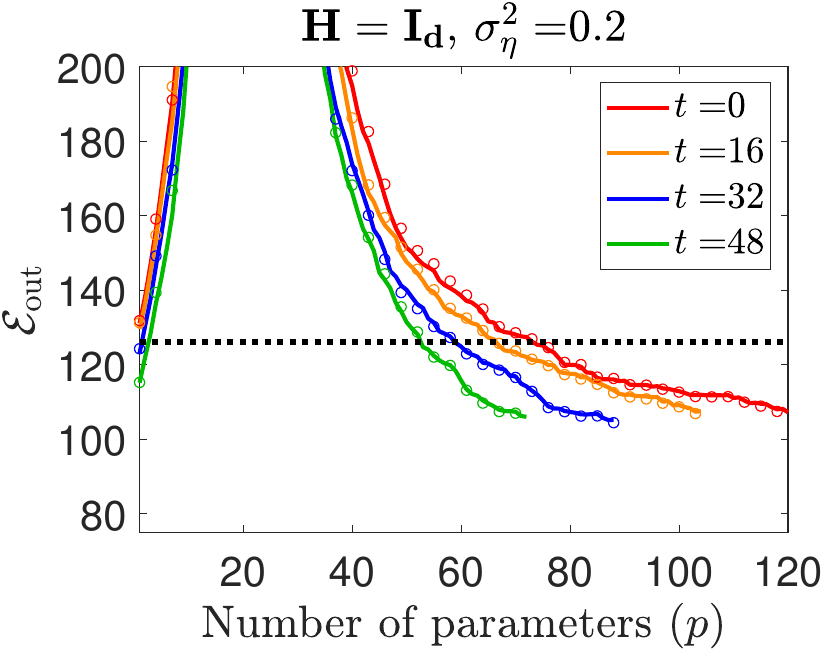} \label{appendix:fig:specific_linear_beta__H_is_I__target_generalization_errors_vs_p_eta_0.2_p_tilde_is_d}}
		\subfloat[]{\includegraphics[width=0.24\textwidth]{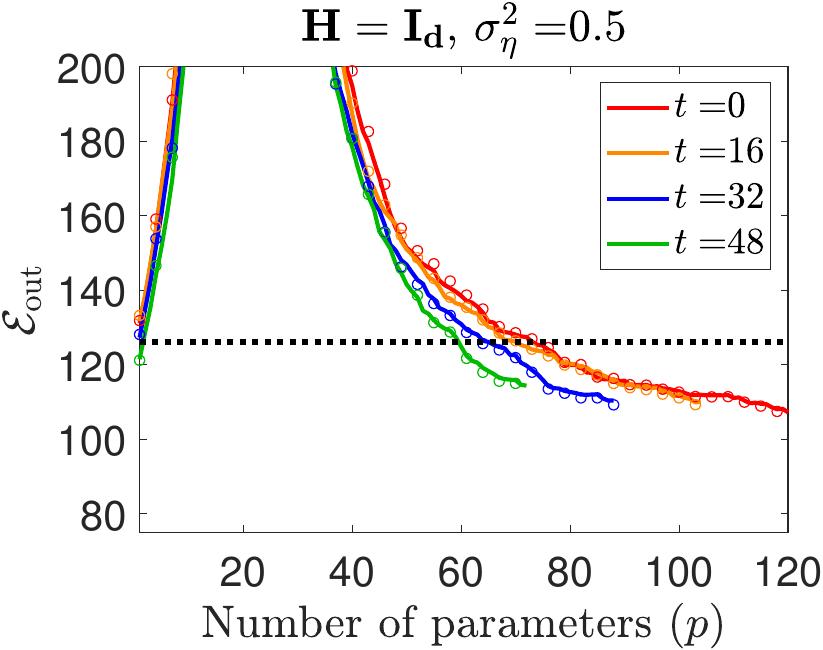} \label{appendix:fig:specific_linear_beta__H_is_I__target_generalization_errors_vs_p_eta_0_5_p_tilde_is_d}}
		\subfloat[]{\includegraphics[width=0.24\textwidth]{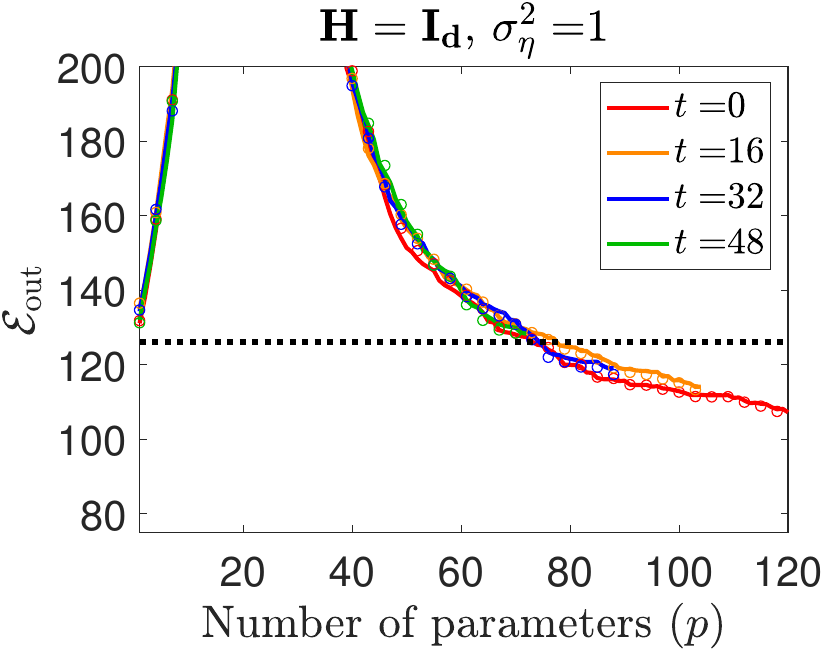} \label{appendix:fig:specific_linear_beta__H_is_I__target_generalization_errors_vs_p_eta_1_p_tilde_is_d}}
		\\
		\subfloat[]{\includegraphics[width=0.24\textwidth]{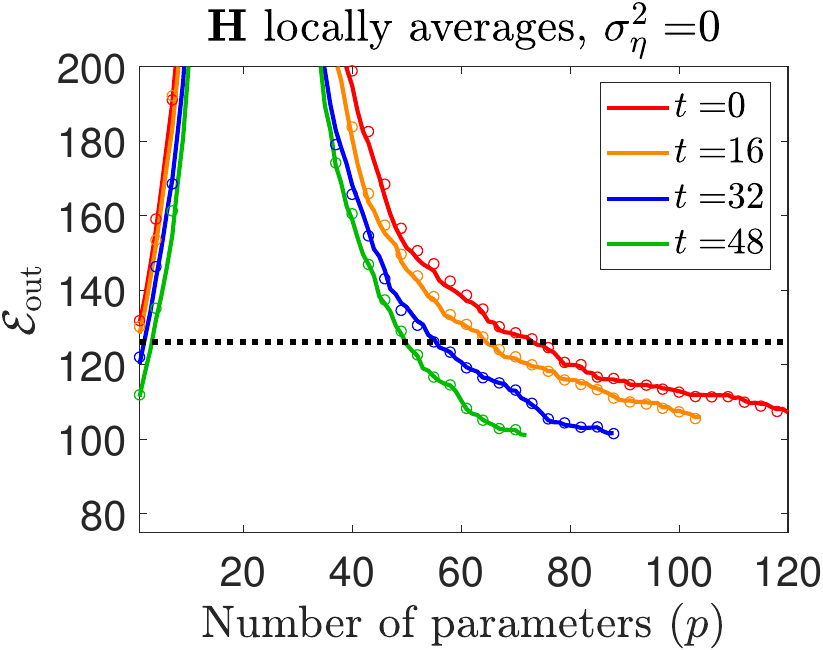}\label{appendix:fig:specific_linear_beta__H_is_local_average__target_generalization_errors_vs_p_eta_0_p_tilde_is_d}}
		\subfloat[]{\includegraphics[width=0.24\textwidth]{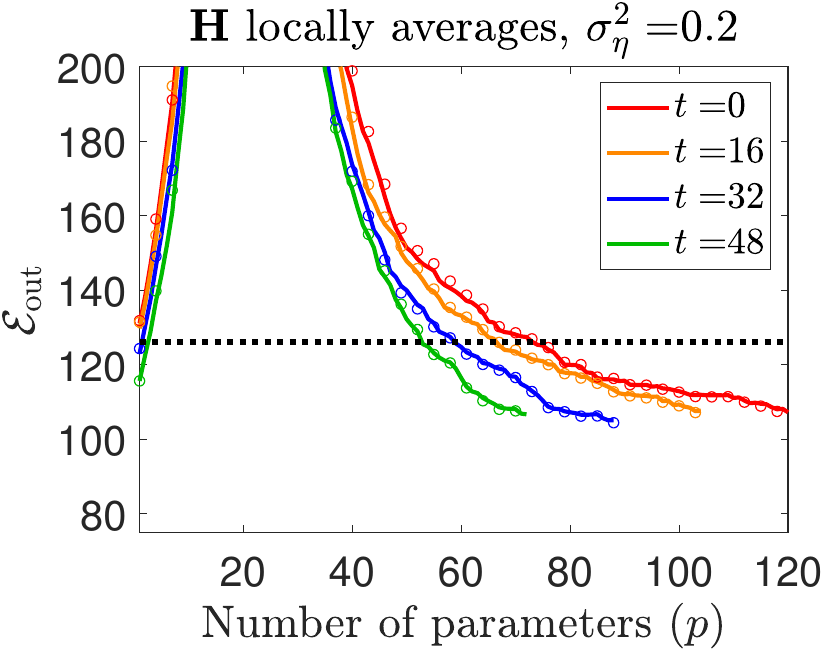} \label{appendix:fig:specific_linear_beta__H_is_local_average__target_generalization_errors_vs_p_eta_0.2_p_tilde_is_d}}
		\subfloat[]{\includegraphics[width=0.24\textwidth]{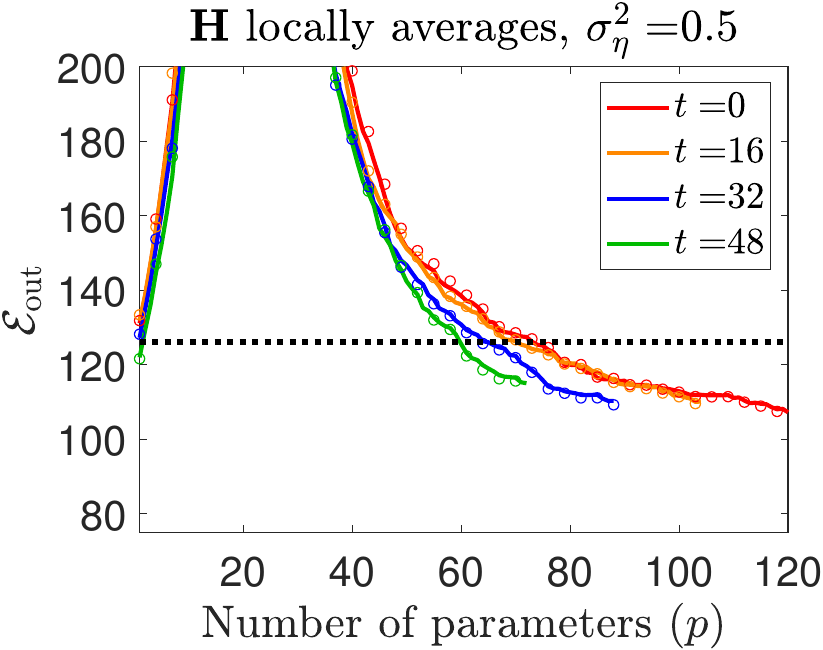} \label{appendix:fig:specific_linear_beta__H_is_local_average__target_generalization_errors_vs_p_eta_0.5_p_tilde_is_d}}
		\subfloat[]{\includegraphics[width=0.24\textwidth]{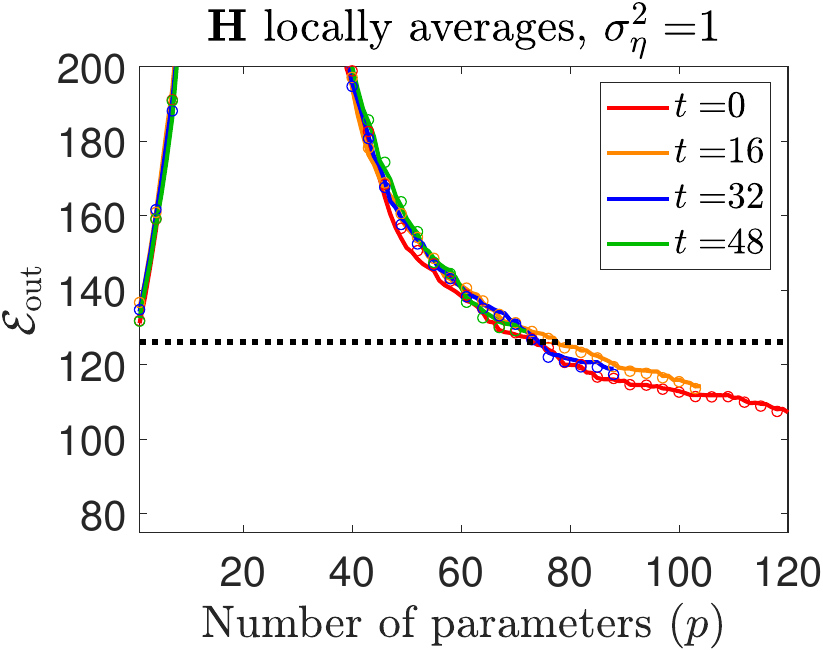} \label{appendix:fig:specific_linear_beta__H_is_local_average__target_generalization_errors_vs_p_eta_1_p_tilde_is_d}}
		\\
		\subfloat[]{\includegraphics[width=0.24\textwidth]{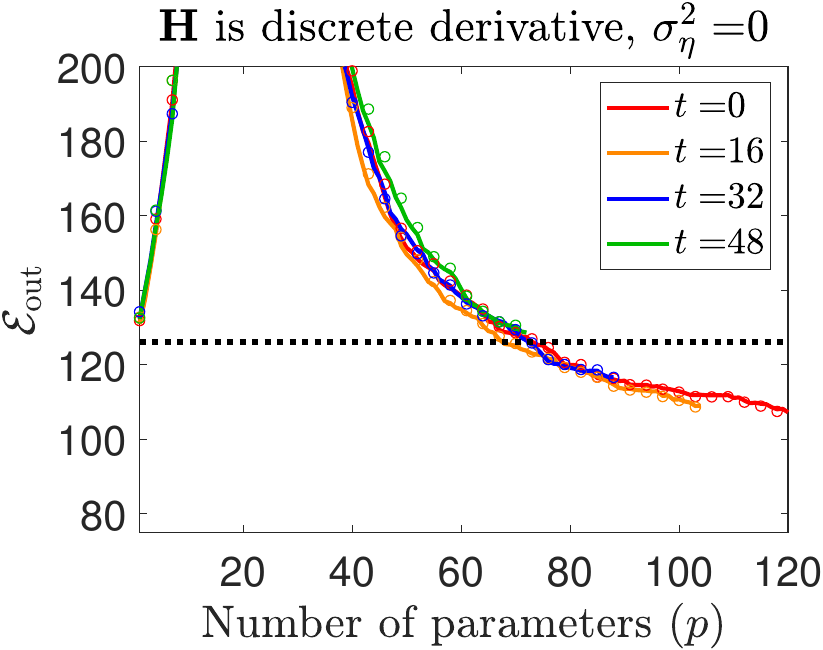}\label{appendix:fig:specific_linear_beta__H_is_derivative__target_generalization_errors_vs_p_eta_0_p_tilde_is_d}}
		\subfloat[]{\includegraphics[width=0.24\textwidth]{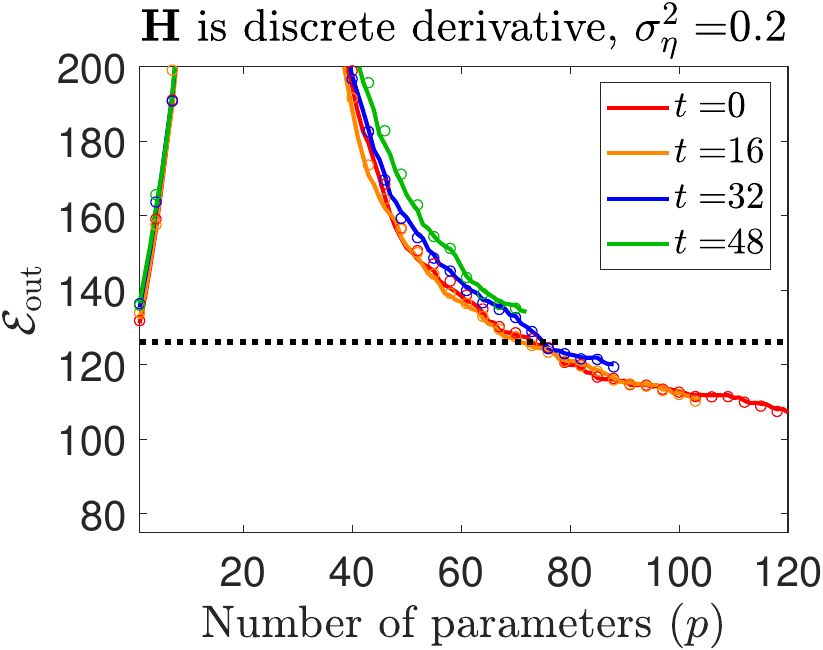} \label{appendix:fig:specific_linear_beta__H_is_derivative__target_generalization_errors_vs_p_eta_0.2_p_tilde_is_d}}
		\subfloat[]{\includegraphics[width=0.24\textwidth]{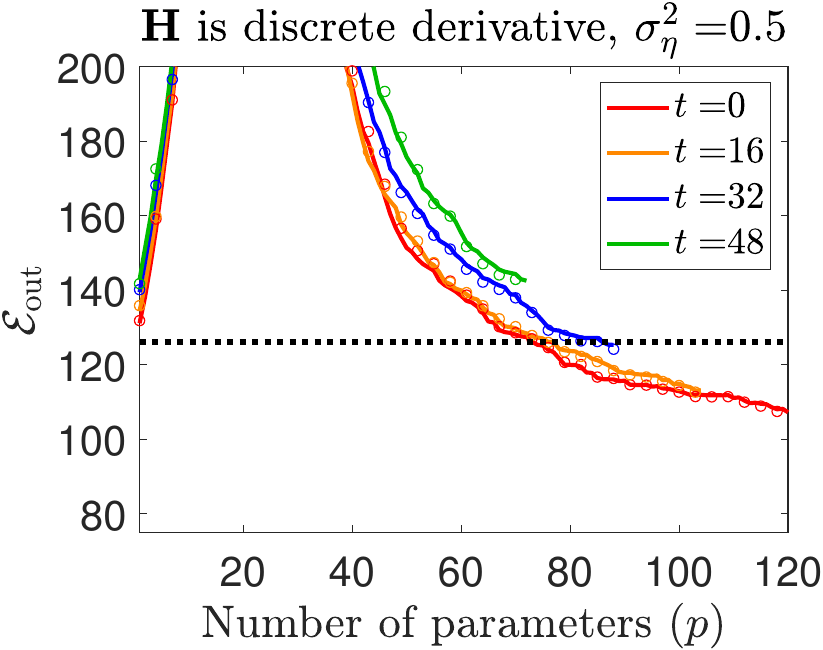} \label{appendix:fig:specific_linear_beta__H_is_derivative__target_generalization_errors_vs_p_eta_0.5_p_tilde_is_d}}
		\subfloat[]{\includegraphics[width=0.24\textwidth]{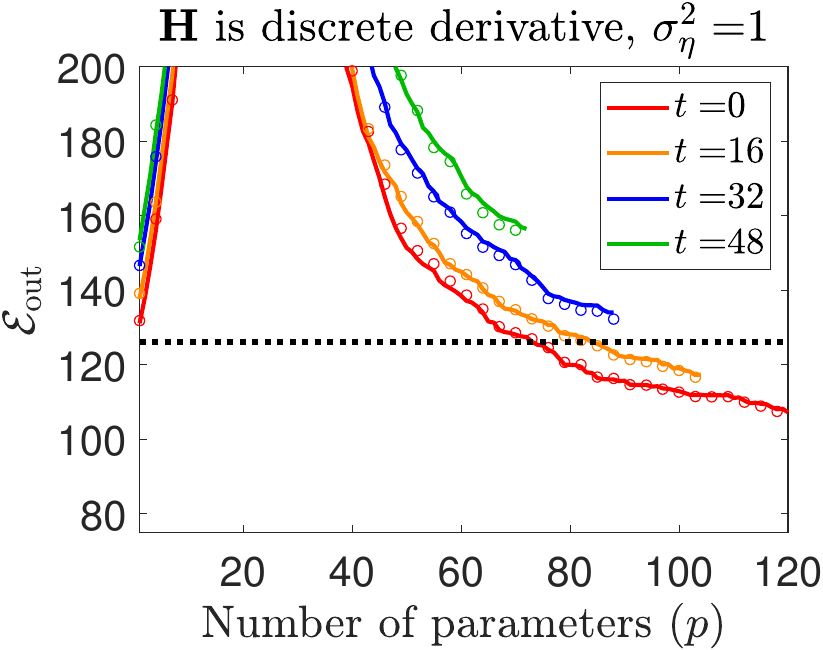} \label{appendix:fig:specific_linear_beta__H_is_derivative__target_generalization_errors_vs_p_eta_1_p_tilde_is_d}}
		\caption{Analytical (solid lines) and empirical (circle markers) values of $\mathcal{E}_{\rm out}^{(\mathcal{L})}$ for specific, non-random coordinate layouts. \textbf{The true solution $\vecgreek{\beta}$ has linearly-increasing values.}  All subfigures use the same sequential evolution of $\mathcal{L}$ with $p$. Each subfigure considers a different case of the relation (\ref{eq:theta-beta relation}) between the source and target tasks: each column of subfigures has a different $\sigma_{\eta}^2$ value, and each row of subfigures corresponds to a different linear operator $\mtx{H}$. The analytical values, computed using Theorem \ref{theorem:out of sample error - target task - specific layout}, are presented using solid-line curves, and the respective empirical results obtained from averaging over 250 experiments are denoted by circle markers.  Each curve color refers to a different number of transferred parameters.}
		\label{appendix:fig:target_generalization_errors_vs_p__specific_extended_for_linear_beta}
	\end{center}
	\vspace*{-5mm}
\end{figure}

\begin{figure}[t]
	\begin{center}
		\subfloat[]{\includegraphics[width=0.24\textwidth]{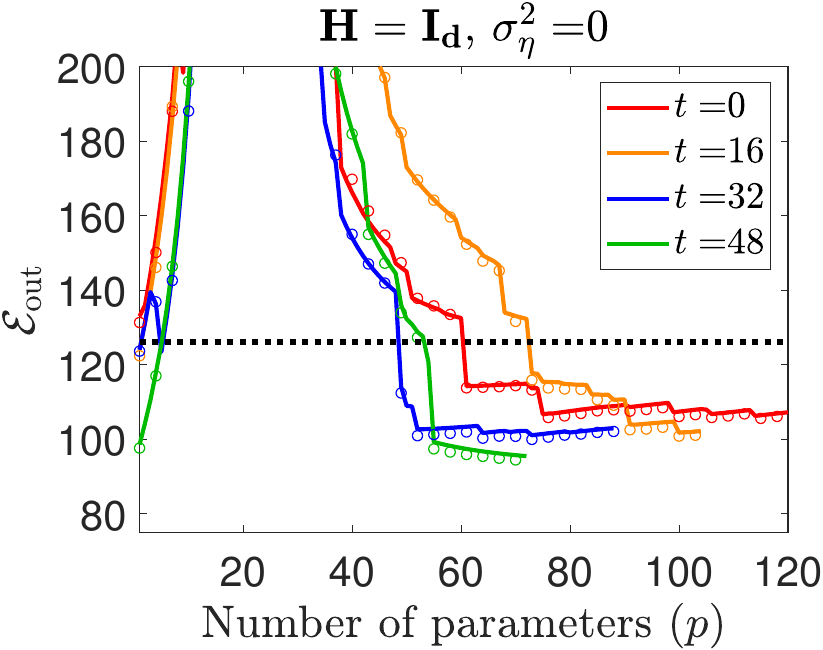}\label{appendix:fig:specific_sparse_beta__H_is_I__target_generalization_errors_vs_p_eta_0_p_tilde_is_d}}
		\subfloat[]{\includegraphics[width=0.24\textwidth]{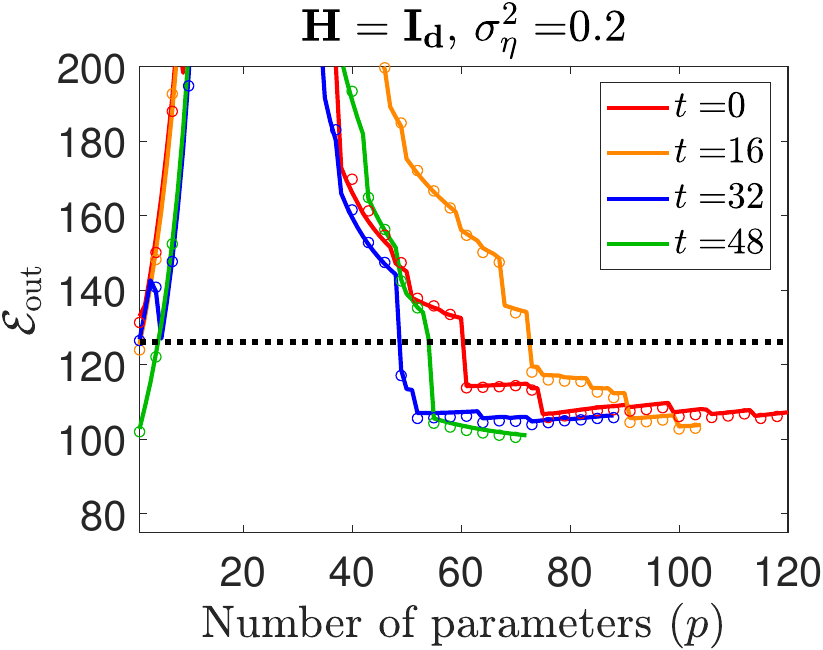} \label{appendix:fig:specific_sparse_beta__H_is_I__target_generalization_errors_vs_p_eta_0.2_p_tilde_is_d}}
		\subfloat[]{\includegraphics[width=0.24\textwidth]{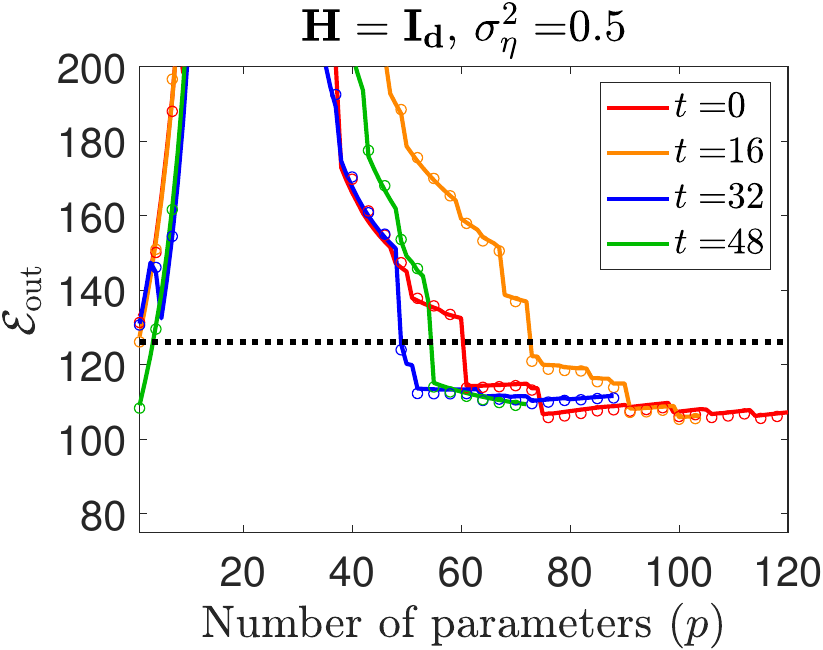} \label{appendix:fig:specific_sparse_beta__H_is_I__target_generalization_errors_vs_p_eta_0.5_p_tilde_is_d}}
		\subfloat[]{\includegraphics[width=0.24\textwidth]{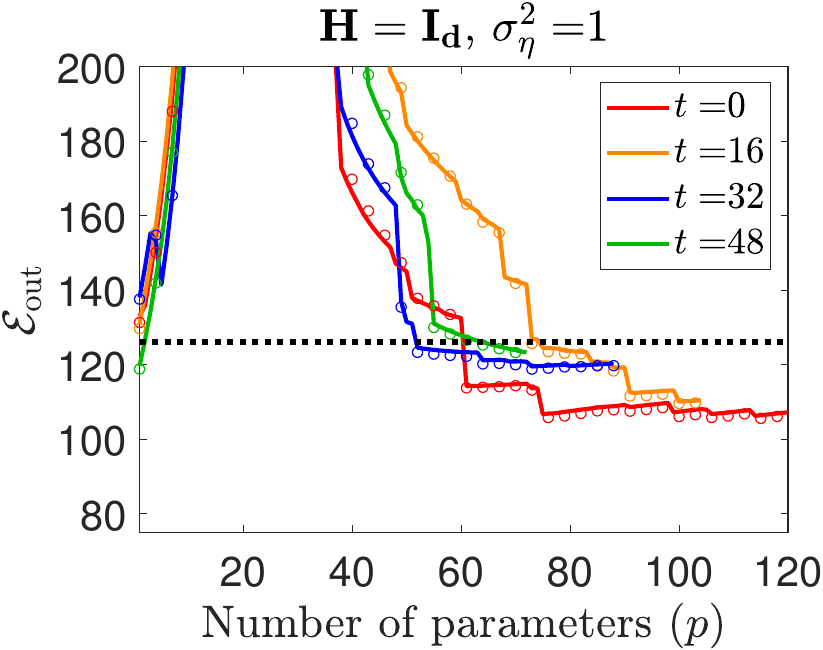} \label{appendix:fig:specific_sparse_beta__H_is_I__target_generalization_errors_vs_p_eta_1_p_tilde_is_d}}
		\\
		\subfloat[]{\includegraphics[width=0.24\textwidth]{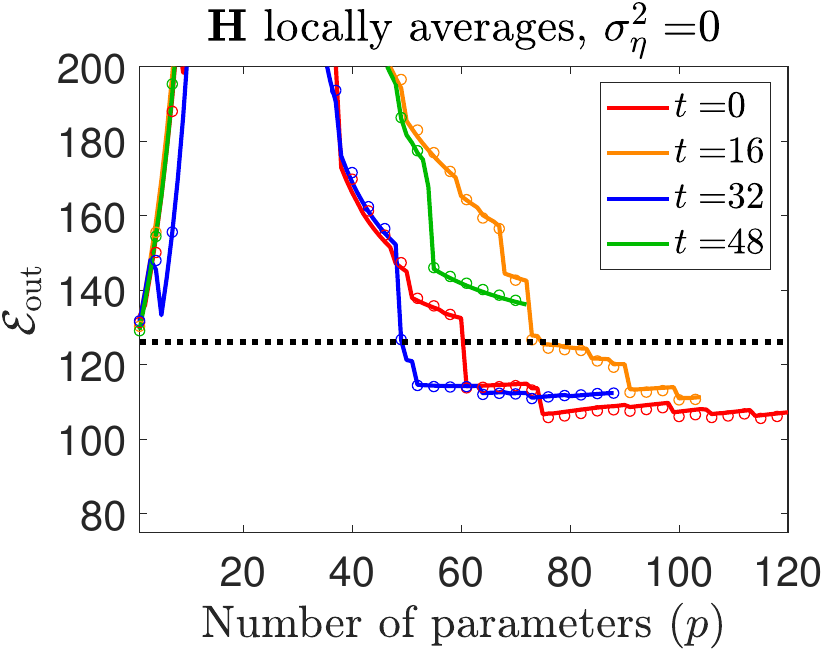}\label{appendix:fig:specific_sparse_beta__H_is_local_average__target_generalization_errors_vs_p_eta_0.5_p_tilde_is_d}}
		\subfloat[]{\includegraphics[width=0.24\textwidth]{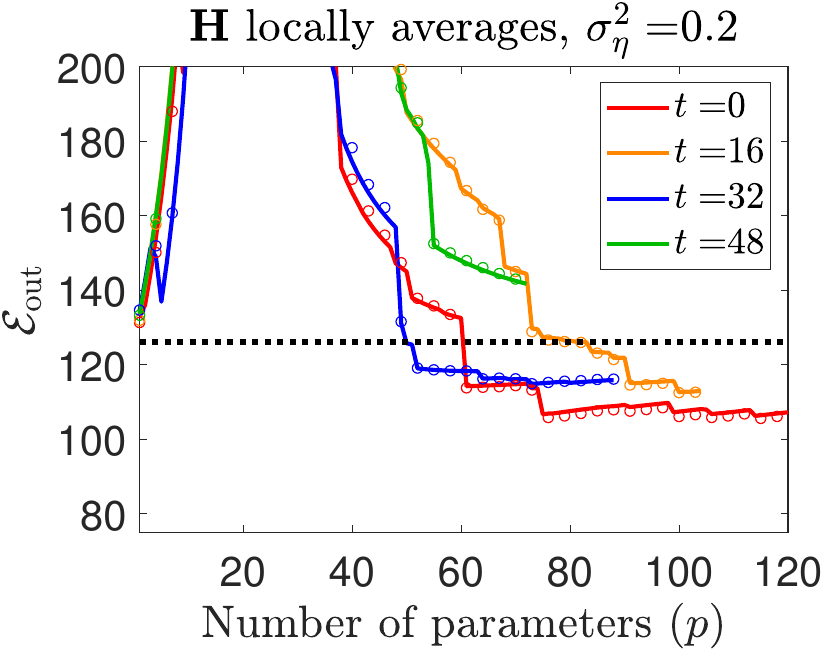} \label{appendix:fig:specific_sparse_beta__H_is_local_average__target_generalization_errors_vs_p_eta_0.2_p_tilde_is_d}}
		\subfloat[]{\includegraphics[width=0.24\textwidth]{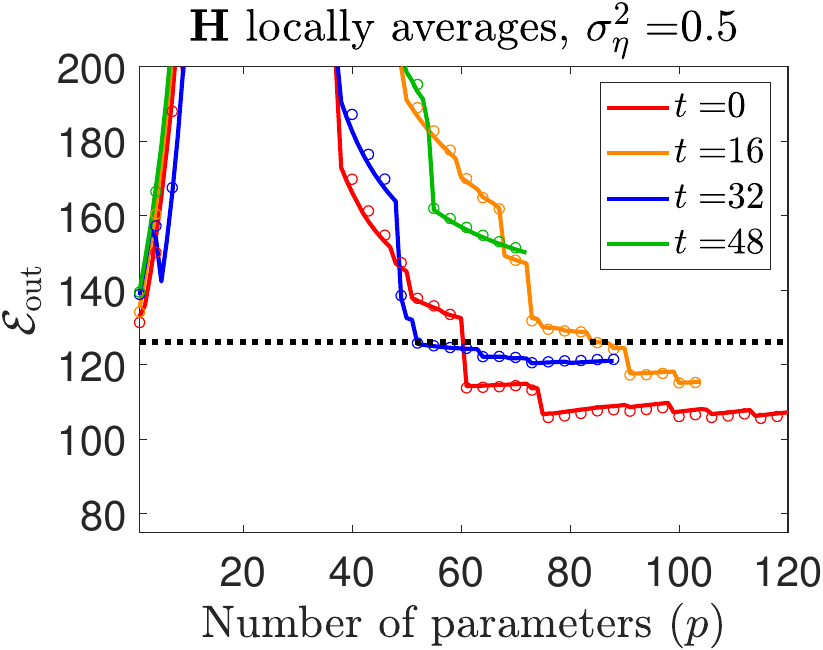} \label{appendix:fig:specific_sparse_beta__H_is_local_average__target_generalization_errors_vs_p_eta_0_p_tilde_is_d}}
		\subfloat[]{\includegraphics[width=0.24\textwidth]{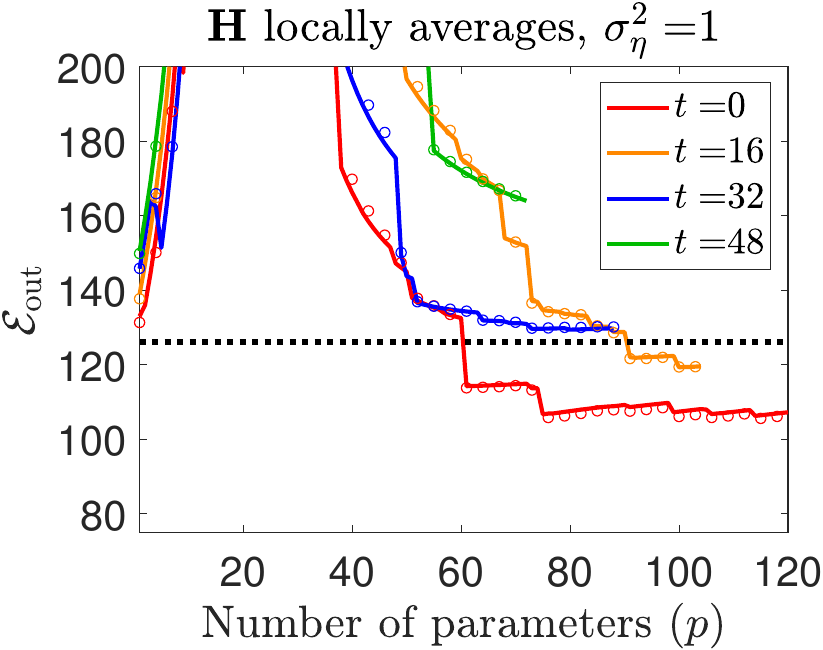} \label{appendix:fig:specific_sparse_beta__H_is_local_average__target_generalization_errors_vs_p_eta_1_p_tilde_is_d}}
		\\
		\subfloat[]{\includegraphics[width=0.24\textwidth]{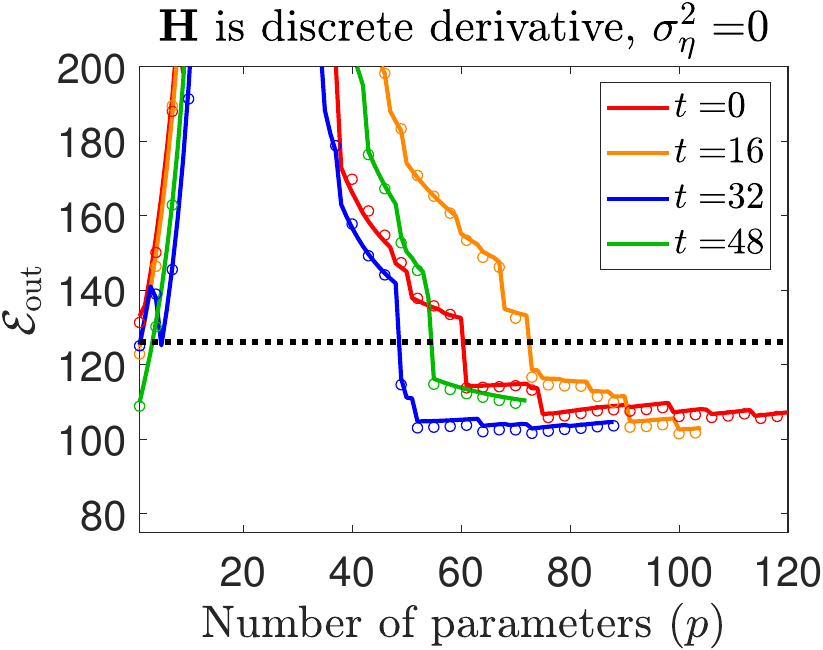}\label{appendix:fig:specific_sparse_beta__H_is_derivative__target_generalization_errors_vs_p_eta_0_p_tilde_is_d}}
		\subfloat[]{\includegraphics[width=0.24\textwidth]{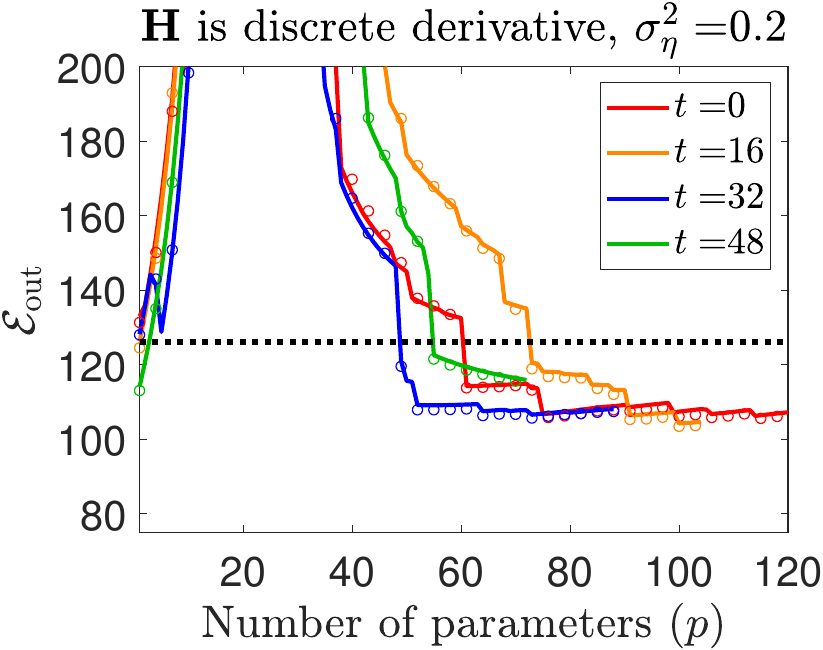} \label{appendix:fig:specific_sparse_beta__H_is_derivative__target_generalization_errors_vs_p_eta_0.2_p_tilde_is_d}}
		\subfloat[]{\includegraphics[width=0.24\textwidth]{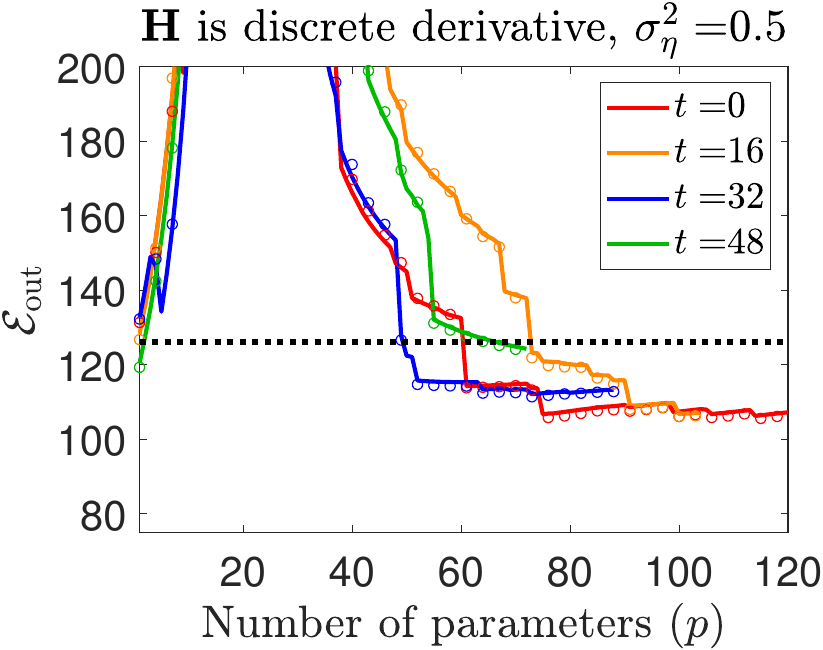} \label{appendix:fig:specific_sparse_beta__H_is_derivative__target_generalization_errors_vs_p_eta_0.5_p_tilde_is_d}}
		\subfloat[]{\includegraphics[width=0.24\textwidth]{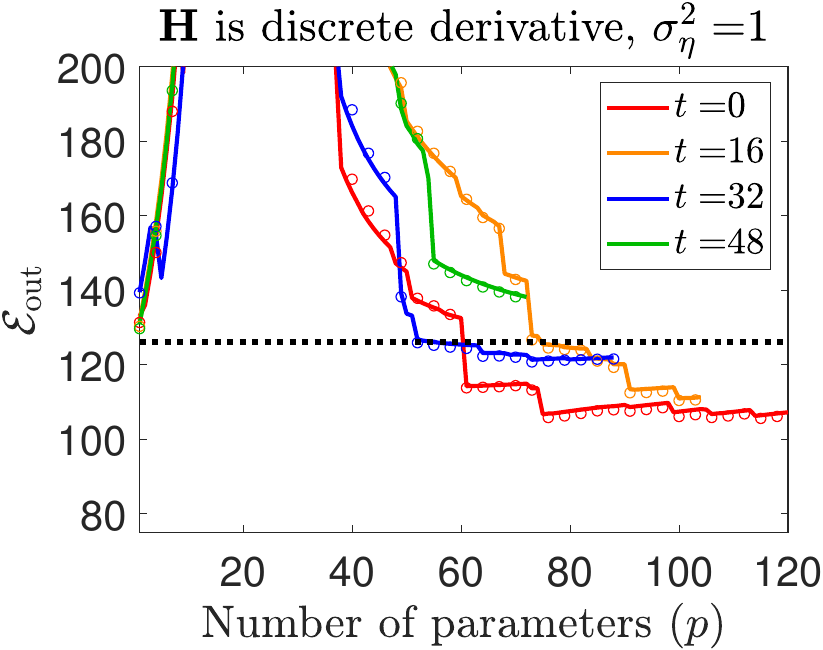} \label{appendix:fig:specific_sparse_beta__H_is_derivative__target_generalization_errors_vs_p_eta_1_p_tilde_is_d}}
		\caption{Analytical (solid lines) and empirical (circle markers) values of $\mathcal{E}_{\rm out}^{(\mathcal{L})}$ for specific, non-random coordinate layouts. \textbf{The true solution $\vecgreek{\beta}$ has a sparse form of values.}  All subfigures use the same sequential evolution of $\mathcal{L}$ with $p$. Each subfigure considers a different case of the relation (\ref{eq:theta-beta relation}) between the source and target tasks: each column of subfigures has a different $\sigma_{\eta}^2$ value, and each row of subfigures corresponds to a different linear operator $\mtx{H}$. The analytical values, computed using Theorem \ref{theorem:out of sample error - target task - specific layout}, are presented using solid-line curves, and the respective empirical results obtained from averaging over 250 experiments are denoted by circle markers.  Each curve color refers to a different number of transferred parameters.}
		\label{appendix:fig:target_generalization_errors_vs_p__specific_extended_for_sparse_beta}
	\end{center}
	\vspace*{-5mm}
\end{figure}

Figures \ref{appendix:fig:target_generalization_errors_vs_p__specific_extended_for_linear_beta}-\ref{appendix:fig:target_generalization_errors_vs_p__specific_extended_for_sparse_beta} present the analytical and empirical values of the generalization error of the target task with respect to specific coordinate layouts $\mathcal{L}$ that evolve with respect to the value of $p$ (this evolution of $\mathcal{L}$ is the same in each of the subfigures and it is not particularly designed to any of the combinations of the true $\vecgreek{\beta}$, $\mtx{H}$, and $\sigma_{\eta}^2$). 
It is clear from Figures \ref{appendix:fig:target_generalization_errors_vs_p__specific_extended_for_linear_beta}-\ref{appendix:fig:target_generalization_errors_vs_p__specific_extended_for_sparse_beta} that the increase in $\sigma_{\eta}^2$, which by its definition corresponds to less related source and target tasks, reduces the benefits or even increases the harm due to transfer of parameters (one can observe that in Figs.~\ref{appendix:fig:target_generalization_errors_vs_p__specific_extended_for_linear_beta}-\ref{appendix:fig:target_generalization_errors_vs_p__specific_extended_for_sparse_beta} by comparing the error curves among subfigures in the same row). 

The effect of $\mtx{H}$ with respect to the true $\vecgreek{\beta}$ is also evident. First, the identity operator $\mtx{H}=\mtx{I}_{d}$ does not reduce the relation between the source and target tasks and therefore does not degrade the parameter transfer performance by itself (i.e., for $\mtx{H}=\mtx{I}_{d}$, only the additive noise level $\sigma_{\eta}^2$ can reduce the relation between the tasks).
Second, when $\mtx{H}$ is a local averaging operator it does not reduce the benefits from transfer learning (e.g., compare second to first row of subfigures in Figs.~\ref{appendix:fig:target_generalization_errors_vs_p__specific_extended_for_linear_beta}-\ref{appendix:fig:target_generalization_errors_vs_p__specific_extended_for_sparse_beta}) in the case of linearly-increasing $\vecgreek{\beta}$ shape (because local averaging does not affect a linear function, except to the few first and last coordinates where the periodic averaging is applied), in contrast, the local averaging operator significantly degrades the parameter transfer performance in the case of the sparse $\vecgreek{\beta}$ form. 
Lastly, when $\mtx{H}$ is a discrete derivative operator it renders transfer learning harmful in the case of linearly-increasing $\vecgreek{\beta}$ shape (e.g., compare third to first row of subfigures in Fig.~\ref{appendix:fig:target_generalization_errors_vs_p__specific_extended_for_linear_beta}). In the case of the sparse $\vecgreek{\beta}$ form the discrete derivative reduces the potential benefits of the parameter transfer but does not eliminate them completely in case these benefits exist for $\mtx{H}=\mtx{I}_{d}$ (e.g., compare third to first row of subfigures in Fig.~\ref{appendix:fig:target_generalization_errors_vs_p__specific_extended_for_sparse_beta}). 


\section{Additional Empirical Results on Parameter Transfer Usefulness}
\label{appendix:sec:Proofs and Additional Details on Parameter Transfer Usefulness in the Setting of Uniformly-Distributed Coordinate Layouts}

\subsection{Details on the Empirical Evaluation of \texorpdfstring{$\Delta\mathcal{E}_{\rm transfer}$}{~}}
\label{appendix:subsec:Details on empirical evaluation of transfer learning error difference term}

The analytical formula for $\Delta\mathcal{E}_{\rm transfer}\triangleq\expectationwrt{\Delta\mathcal{E}_{\rm transfer}^{(\mathcal{T},\mathcal{S})}}{\mathcal{L}}$, which is based on (\ref{eq:error difference term - transfer versus zero - auxiliary term}) and Corollary \ref{corollary:out of sample error - target task}, measures the (normalized) expected difference in the generalization error (of the target task) due to transferring parameters instead of setting them to zero. 
Accordingly, the empirical evaluation of $\Delta\mathcal{E}_{\rm transfer}$ for a given $\widetilde{p}$ can be computed by 
\begin{align}
\label{appendix:eq:empirical evaluation of expected error difference due to parameter transfer}
&\widehat{\Delta}\mathcal{E}_{\rm transfer} = \frac{1}{d-3}
\sum_{p=1,\dots,n-2,n+2,\dots,d} {\frac{  {\widehat{\mathbb{E}}_{\mathcal{L}}\Big\{{\mathcal{E}^{(\widetilde{p},p,{t=m})}_{\rm out}}\Big\}} - {\widehat{\mathbb{E}}_{\mathcal{L}}\Big\{{\mathcal{E}^{({\widetilde{p},p,t=0})}_{\rm out}}\Big\}}  }{m\cdot\alpha(p)}}
\end{align}
where 
\begin{equation}
\label{appendix:eq:empirical evaluation of expected error difference due to parameter transfer - normalization term}
\alpha(p) \triangleq \begin{cases}
1 + \frac{p}{n-p-1}               & \text{for } p \le n-2, \\
1 + \frac{n}{p-n-1}              & \text{for } p \ge n+2  \\
\end{cases}
\end{equation}
is a normalization factor required for independence from $p$. The value measured in (\ref{appendix:eq:empirical evaluation of expected error difference due to parameter transfer}) is also normalized by the number of transferred parameters. 
Here ${\widehat{\mathbb{E}}_{\mathcal{L}}\Big\{{\mathcal{E}^{({\widetilde{p},p,{t=m}})}_{\rm out}}\Big\}}$ is the out-of-sample error of the target task that is \textit{empirically} computed for $m$ transferred parameters, $p$ free parameters in the target task, and $\widetilde{p}$ free parameters in the source task. Correspondingly,  ${\widehat{\mathbb{E}}_{\mathcal{L}}\Big\{{\mathcal{E}^{({\widetilde{p},p,t=0})}_{\rm out}}\Big\}}$ is the empirically computed error induced by avoiding parameter transfer.  
Therefore, the formula in (\ref{appendix:eq:empirical evaluation of expected error difference due to parameter transfer}) empirically measures the average error difference for a single transferred parameter by averaging over the various settings induced by different values of $p$ while $\widetilde{p}$ is kept fixed. To obtain a good numerical accuracy with averaging over a moderate number of experiments we use the value $m=5$. 

Each empirical evaluation of ${\widehat{\mathbb{E}}_{\mathcal{L}}\Big\{{\mathcal{E}^{({\widetilde{p},p,t})}_{\rm out}}\Big\}}$, for a specific set of values $\widetilde{p},p,t$ corresponds to averaging over 500 experiments 
where each experiment was conducted for new realizations of the data matrices, noise components, and the sequential order of adding coordinates to subsets. 
Each single evaluation of the expectation of the squared error for an out-of-sample data pair ${\left( { \vec{x}^{(\rm test)}, y^{(\rm test)} } \right)}$ was empirically computed by averaging over 1000 out-of-sample realizations of data pairs.

\begin{figure}
	\begin{center}
		{\includegraphics[width=0.24\textwidth]{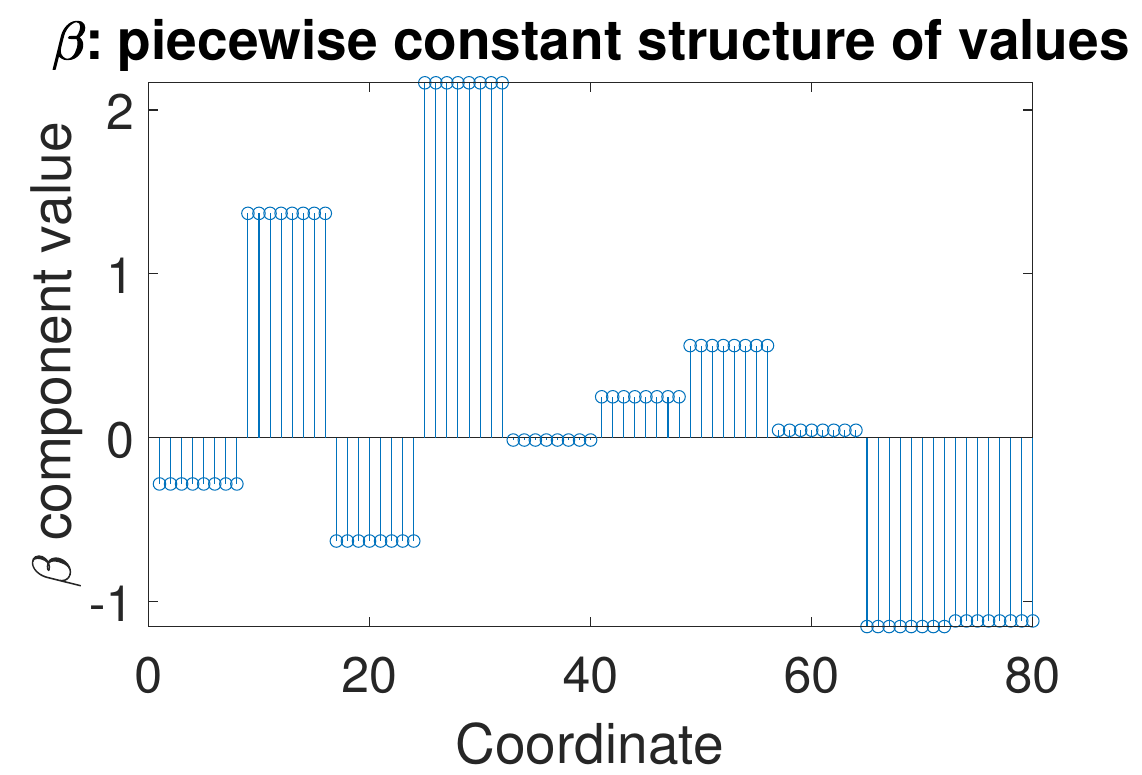}\label{appendix:fig:piecewise_constant_beta}}
		\caption{The piecewise-constant structure of $\vecgreek{\beta}$ that was used in part of the experiments. }
		\label{appendix:fig:piecewise_constant_beta_structure}
	\end{center}
	\vspace*{-5mm}
\end{figure}

\subsection{Results for $\widetilde{n}<d$}

\begin{figure}[t]
	\begin{center}
		\subfloat[{\small$\mtx{H}$:\hspace{.4em}local\hspace{.4em}averaging neighborhood size  3}]{\includegraphics[width=0.24\textwidth]{figures/transfer_learning_usefulness_plane__H_average3_analytic-eps-converted-to.pdf}\label{appendix:fig:transfer_learning_usefulness_plane__H_average3_analytic}}
		\subfloat[{\small$\mtx{H}$:\hspace{.4em}local\hspace{.4em}averaging neighborhood size  15}]{\includegraphics[width=0.24\textwidth]{figures/transfer_learning_usefulness_plane__H_average15_analytic-eps-converted-to.pdf} \label{appendix:fig:transfer_learning_usefulness_plane__H_average15_analytic}}
		\subfloat[{\small$\mtx{H}$:\hspace{.4em}local\hspace{.4em}averaging neighborhood size  59}]{\includegraphics[width=0.24\textwidth]{figures/transfer_learning_usefulness_plane__H_average59_analytic-eps-converted-to.pdf} \label{appendix:fig:transfer_learning_usefulness_plane__H_average59_analytic}}
		\subfloat[{\small$\mtx{H}$:\hspace{.4em}discrete\hspace{.4em}derivative}]{\includegraphics[width=0.24\textwidth]{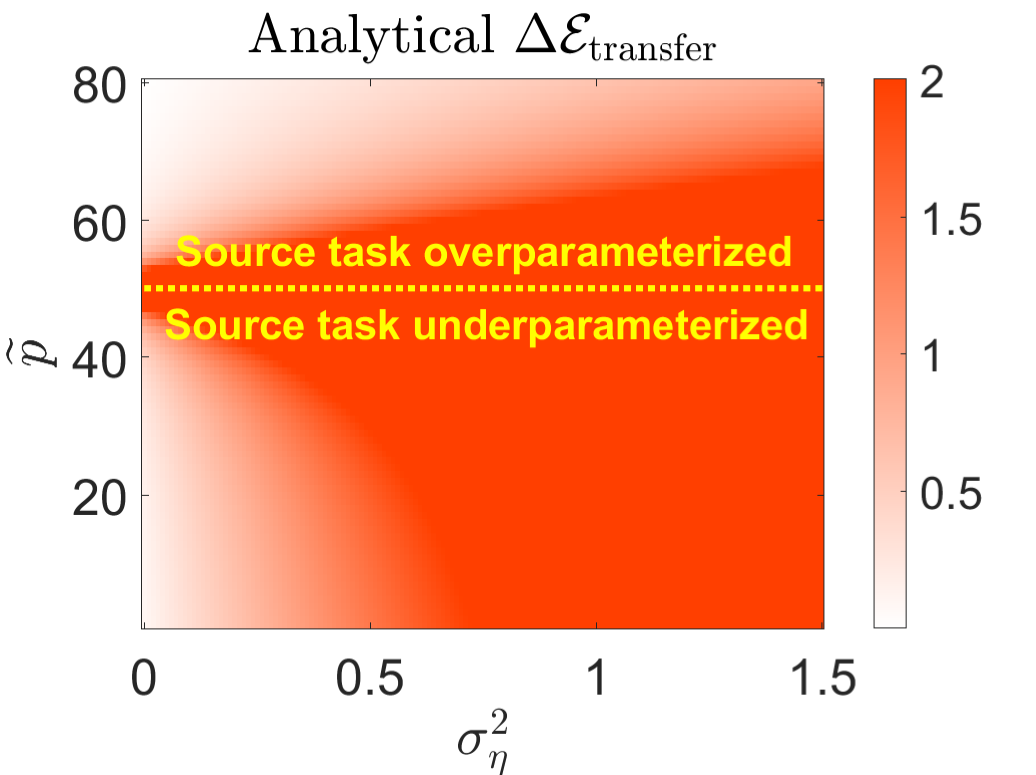} \label{appendix:fig:transfer_learning_usefulness_plane__H_derivative_analytic}}
		\\
		\subfloat[{\small$\mtx{H}$:\hspace{.4em}local\hspace{.4em}averaging neighborhood size  3}]{\includegraphics[width=0.24\textwidth]{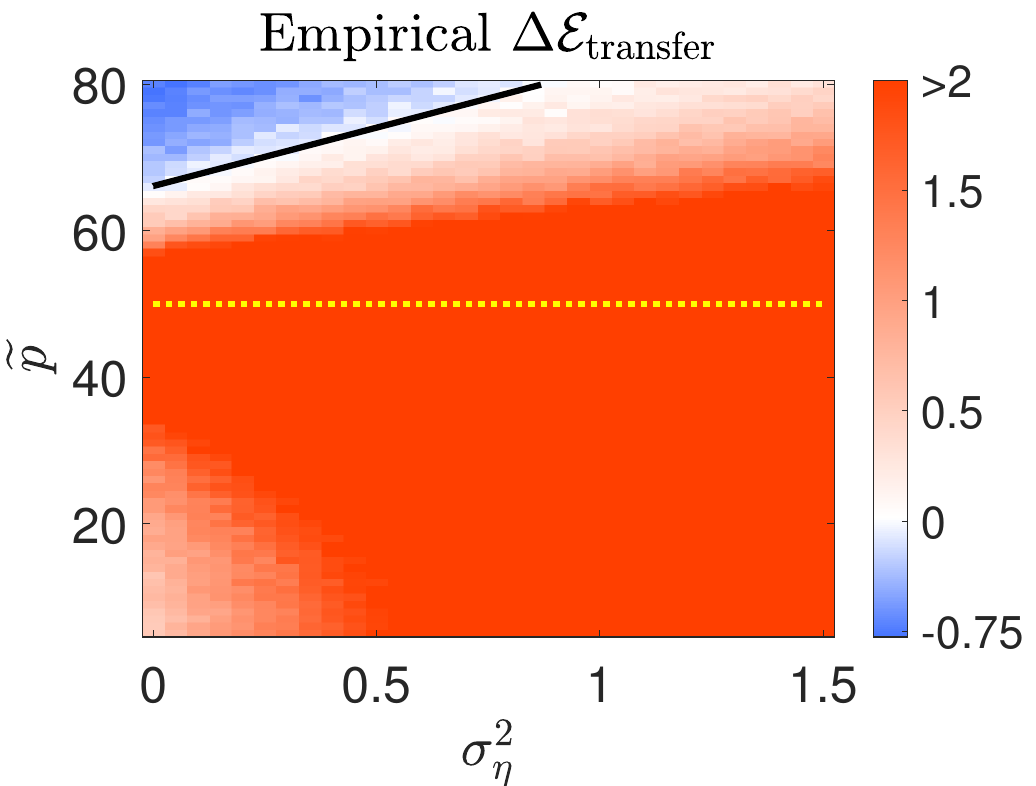}\label{appendix:fig:transfer_learning_usefulness_plane__H_average3_empirical}}
		\subfloat[{\small$\mtx{H}$:\hspace{.4em}local\hspace{.4em}averaging neighborhood size  15}]{\includegraphics[width=0.24\textwidth]{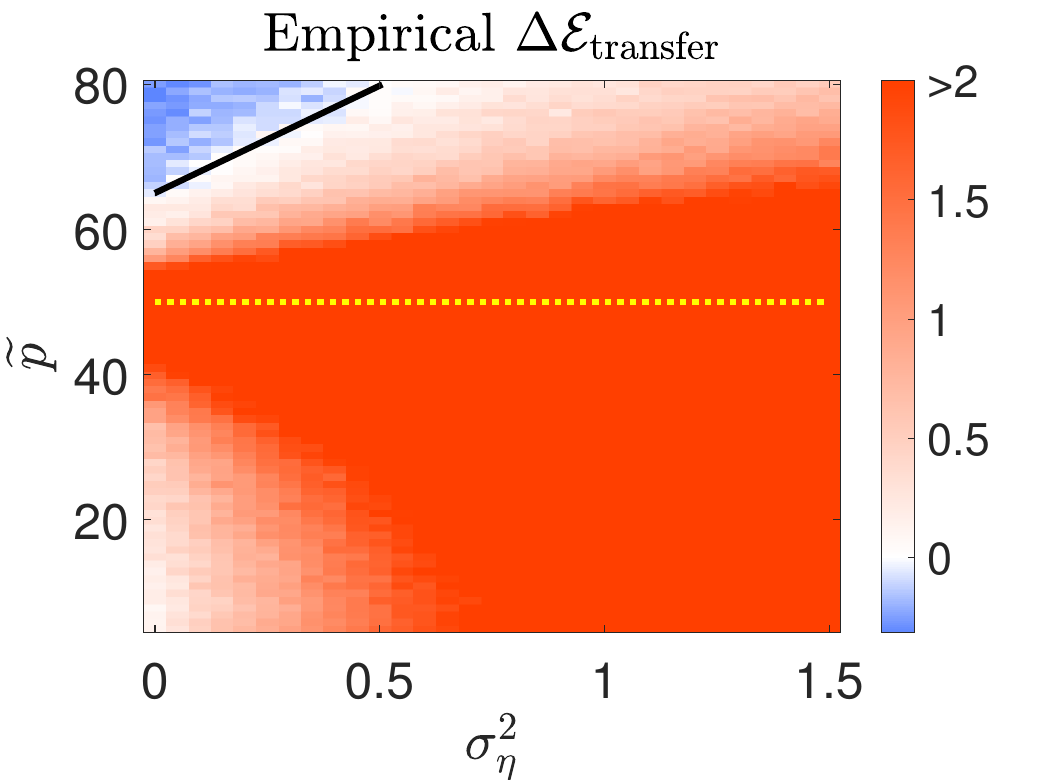} \label{appendix:fig:transfer_learning_usefulness_plane__H_average15_empirical}}
		\subfloat[{\small$\mtx{H}$:\hspace{.4em}local\hspace{.4em}averaging neighborhood size  59}]{\includegraphics[width=0.24\textwidth]{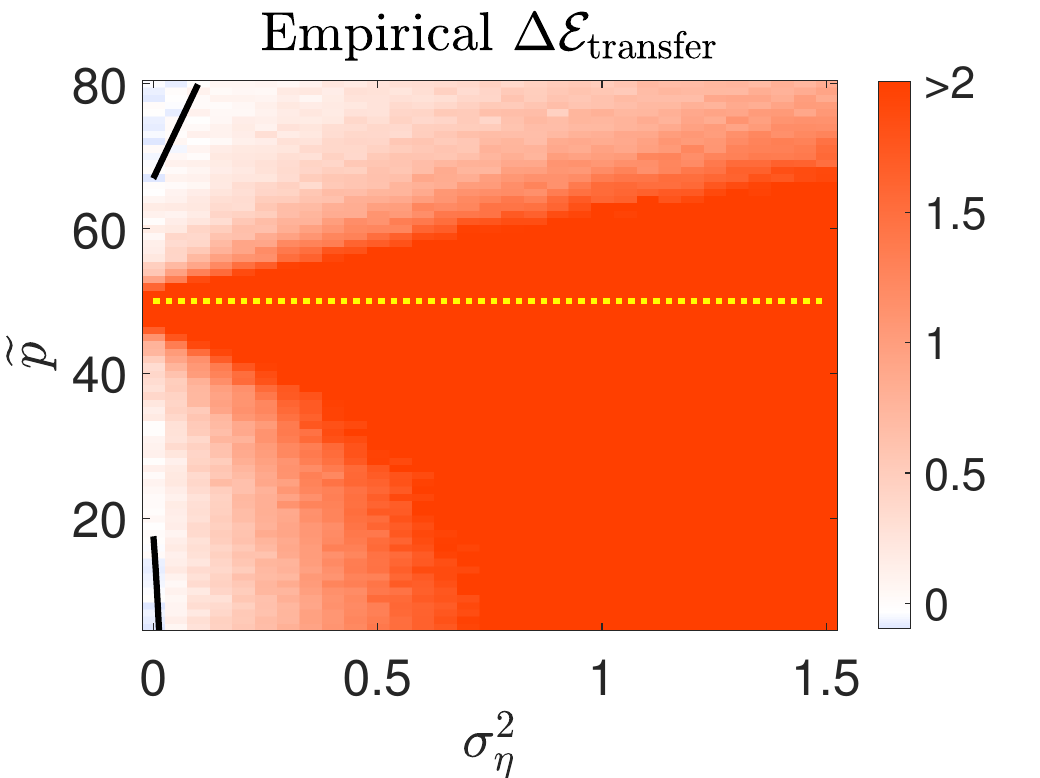} \label{appendix:fig:transfer_learning_usefulness_plane__H_average59_empirical}}
		\subfloat[{\small$\mtx{H}$:\hspace{.4em}discrete\hspace{.4em}derivative}]{\includegraphics[width=0.24\textwidth]{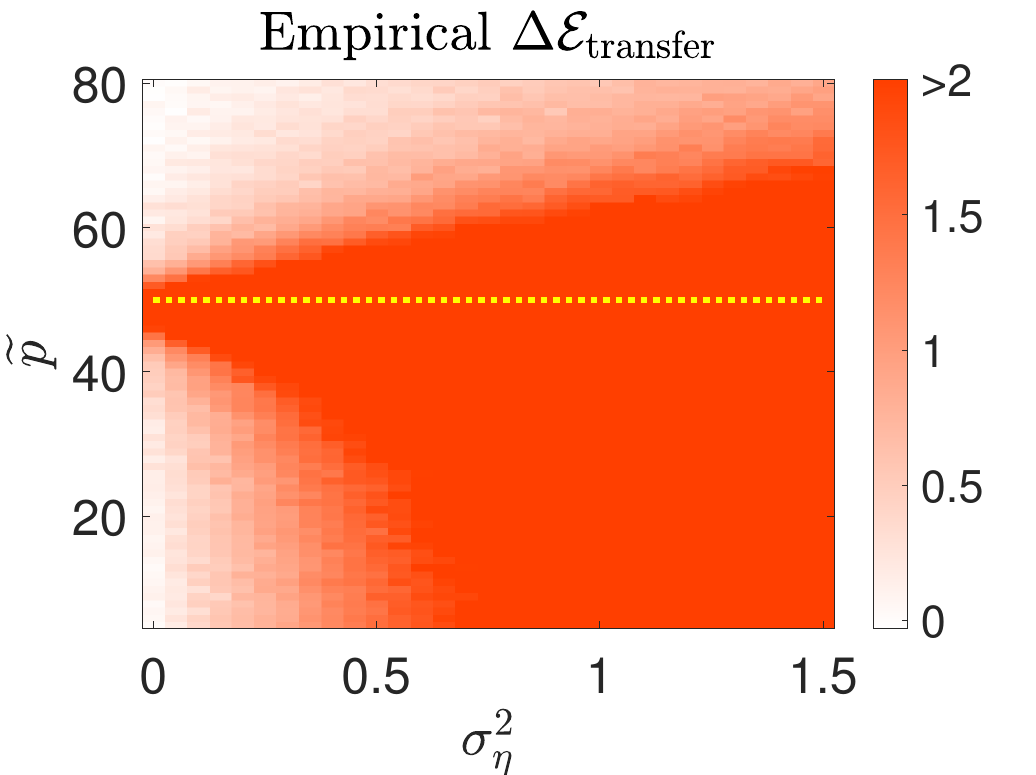} \label{appendix:fig:transfer_learning_usefulness_plane__H_derivative_empirical}}
		\caption{The analytical and empirical values of $\Delta\mathcal{E}_{\rm transfer}$ defined in Corollary \ref{corollary:out of sample error - target task} (here, normalized by $t$, namely, the expected error difference due to transfer of a parameter from the source to target task) as a function of $\widetilde{p}$ and $\sigma_{\eta}^2$. The positive and negative values of $\frac{1}{t}\Delta\mathcal{E}_{\rm transfer}$ appear in color scales of red and blue, respectively. 
			The regions of negative values (appear in shades of blue) correspond to beneficial transfer of parameters. 
			The positive values were truncated in the value of 2 for the clarity of visualization.  Each subfigure corresponds to a different task relation model induced by the definitions of $\mtx{H}$ as: \textit{(a)-(c),(e)-(g)} local averaging operators with different neighborhood sizes, \textit{(d),(h)} discrete derivative.
			For all the subfigures, $d=80$, $\widetilde{n}=50$, $\| \vecgreek{\beta} \|_2^2 = d$, $\sigma_{\xi}^2 = 0.025\cdot d$. Here, all the subfigures correspond to a $\vecgreek{\beta}$ vector with a piecewise-constant form (see Fig.~\ref{appendix:fig:piecewise_constant_beta_structure}).}
		\label{appendix:fig:transfer_learning_usefulness_plane - H is averaging}
	\end{center}
	\vspace*{-5mm}
\end{figure}

\begin{figure}[t]
	\begin{center}
		\subfloat[]{\includegraphics[width=0.3\textwidth]{figures/transfer_learning_usefulness_plane__H_is_0_5_I_analytic-eps-converted-to.pdf}\label{appendix:fig:transfer_learning_usefulness_plane__H_is_0_5_I_analytic}}
		\subfloat[]{\includegraphics[width=0.3\textwidth]{figures/transfer_learning_usefulness_plane__H_is_I_analytic-eps-converted-to.pdf} \label{appendix:fig:transfer_learning_usefulness_plane__H_is_I_analytic}}
		\subfloat[]{\includegraphics[width=0.3\textwidth]{figures/transfer_learning_usefulness_plane__H_is_1_5_I_analytic-eps-converted-to.pdf} \label{appendix:fig:transfer_learning_usefulness_plane__H_is_1_5_I_analytic}}
		\\
		\subfloat[]{\includegraphics[width=0.3\textwidth]{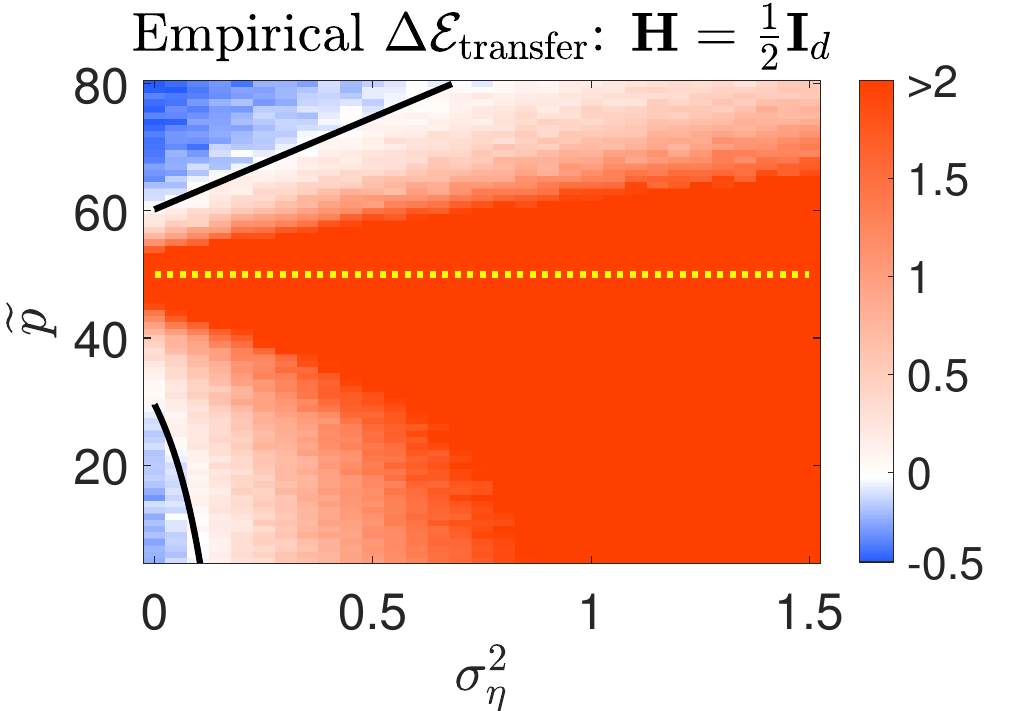}\label{appendix:fig:transfer_learning_usefulness_plane__H_is_0_5_I_empirical}}
		\subfloat[]{\includegraphics[width=0.3\textwidth]{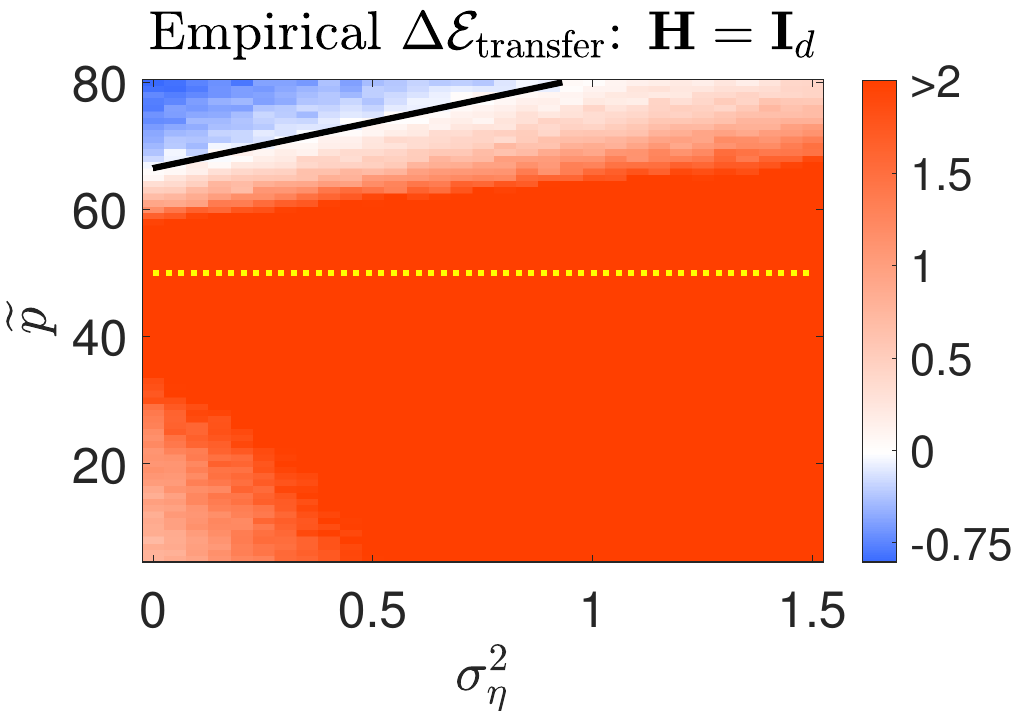} \label{appendix:fig:transfer_learning_usefulness_plane__H_is_I_empirical}}
		\subfloat[]{\includegraphics[width=0.3\textwidth]{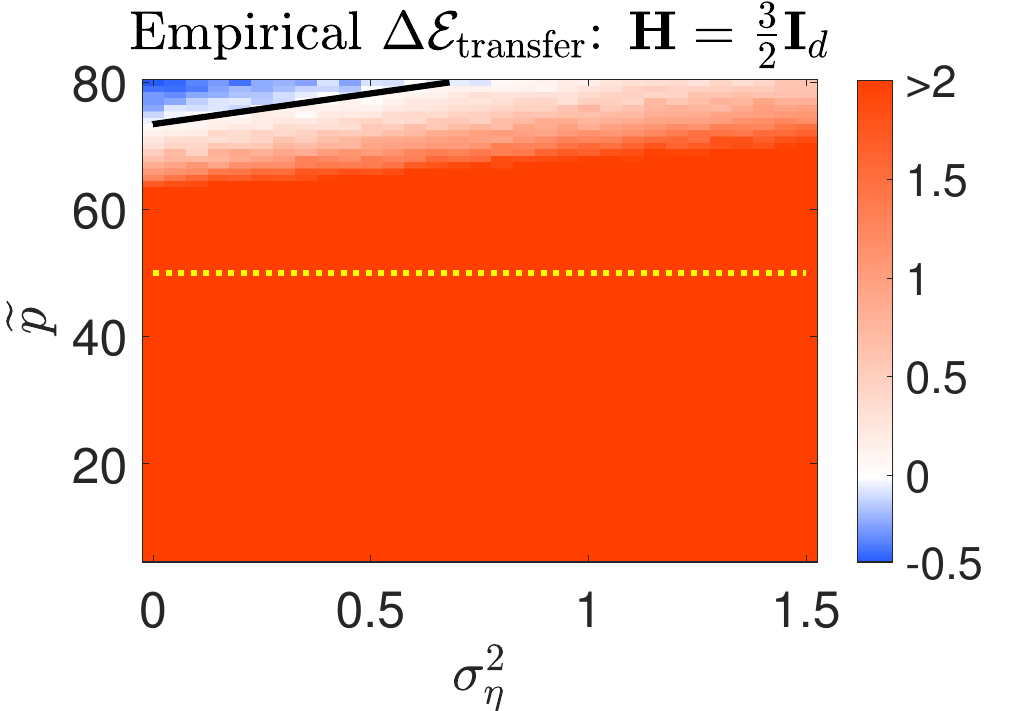} \label{appendix:fig:transfer_learning_usefulness_plane__H_is_1_5_I_empirical}}
		\caption{The analytical and empirical values of $\Delta\mathcal{E}_{\rm transfer}$ defined in Corollary \ref{corollary:out of sample error - target task} (here, normalized by $t$, namely, the expected error difference due to transfer of a parameter from the source to target task) as a function of $\widetilde{p}$ and $\sigma_{\eta}^2$. The positive and negative values of $\frac{1}{t}\Delta\mathcal{E}_{\rm transfer}$ appear in color scales of red and blue, respectively. 
			The regions of negative values (appear in shades of blue) correspond to beneficial transfer of parameters. 
			The positive values were truncated in the value of 2 for the clarity of visualization.  Each subfigure corresponds to a different task relation model induced by the definitions of $\mtx{H}$ as $\mtx{H}=\frac{1}{2}\mtx{I}_{d}$, $\mtx{H}=\mtx{I}_{d}$, and  $\mtx{H}=\frac{3}{2}\mtx{I}_{d}$. 
			For all the subfigures, $d=80$, $\widetilde{n}=50$, $\| \vecgreek{\beta} \|_2^2 = d$, $\sigma_{\xi}^2 = 0.025\cdot d$.
			Here, all the subfigures correspond to a $\vecgreek{\beta}$ vector with a linear form (see Fig.~\ref{appendix:fig:linear_beta_graph}).}
		\label{fig:transfer_learning_usefulness_plane - H is I}
	\end{center}
	\vspace*{-5mm}
\end{figure}

In Figures \ref{appendix:fig:transfer_learning_usefulness_plane - H is averaging}-\ref{fig:transfer_learning_usefulness_plane - H is I} we present analytical and empirical values of $\frac{1}{t}\Delta\mathcal{E}_{\rm transfer}$ induced by settings where $\widetilde{n}<d$ (specifically, $\widetilde{n}=50$ and $d=80$), which naturally enable the corresponding 
overparameterized (i.e., $\widetilde{n}<\widetilde{p}<d$) and underparameterized (i.e., $\widetilde{p}<\widetilde{n}<d$) settings of the source task. 
In Fig.~\ref{appendix:fig:transfer_learning_usefulness_plane - H is averaging} we provide the analytical and empirical results for cases where $\mtx{H}$ is local averaging and discrete derivative operators. In the main text only the analytical results were provided and here we show them again near their empirical counterparts that excellently match them (up to the resolution of the empirical settings).

In Figure \ref{fig:transfer_learning_usefulness_plane - H is I} we provide additional results for cases where the operator $\mtx{H}$ is a scaled identity matrix.

\subsection{Results for $\widetilde{n}>d$}

\begin{figure}[t]
	\begin{center}
		\subfloat[]{\includegraphics[width=0.32\textwidth]{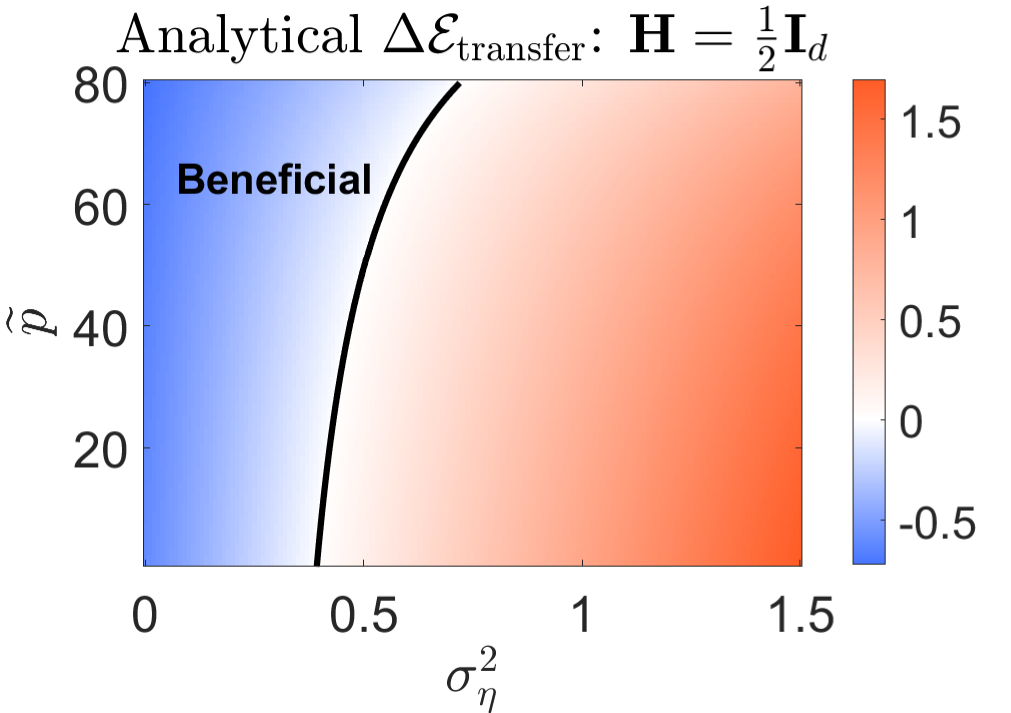}\label{appendix:fig:transfer_learning_usefulness_plane__H_is_0_5_I_analytic__n_tilde_greater_than_d}}
		\subfloat[]{\includegraphics[width=0.32\textwidth]{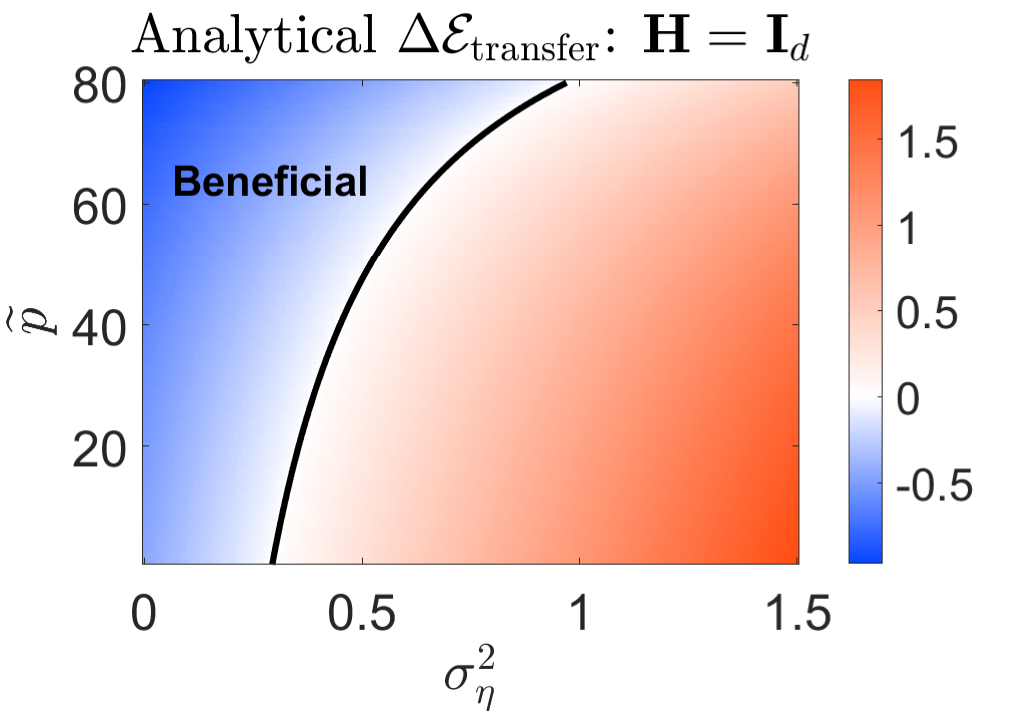} \label{appendix:fig:transfer_learning_usefulness_plane__H_is_I_analytic__n_tilde_greater_than_d}}
		\subfloat[]{\includegraphics[width=0.32\textwidth]{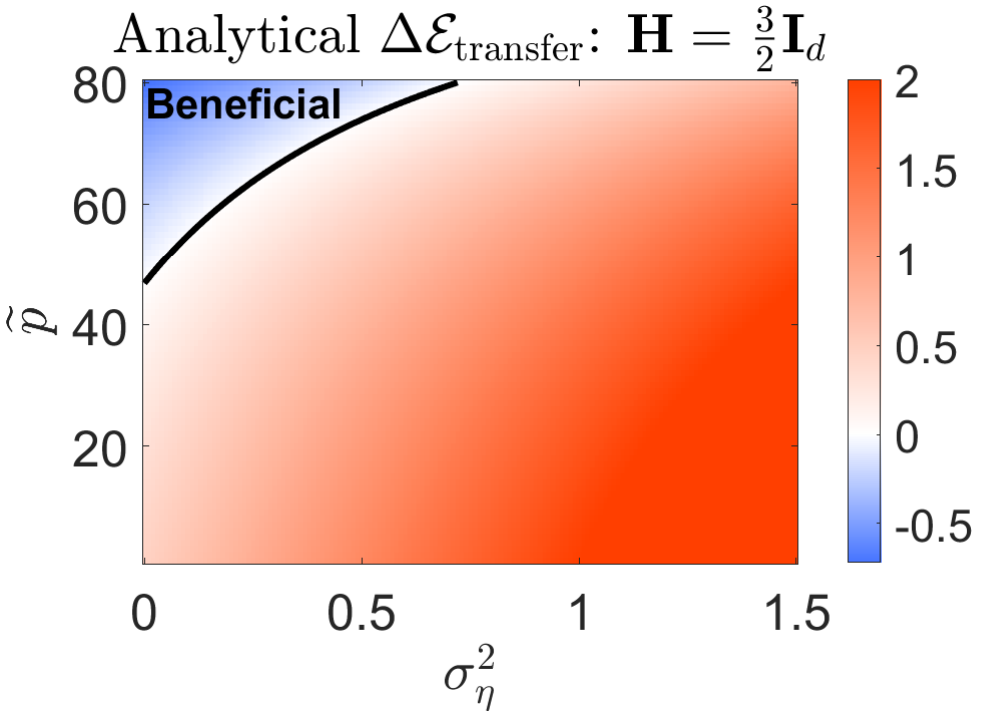} \label{appendix:fig:transfer_learning_usefulness_plane__H_is_1_5_I_analytic__n_tilde_greater_than_d}}
		\\
		\subfloat[]{\includegraphics[width=0.32\textwidth]{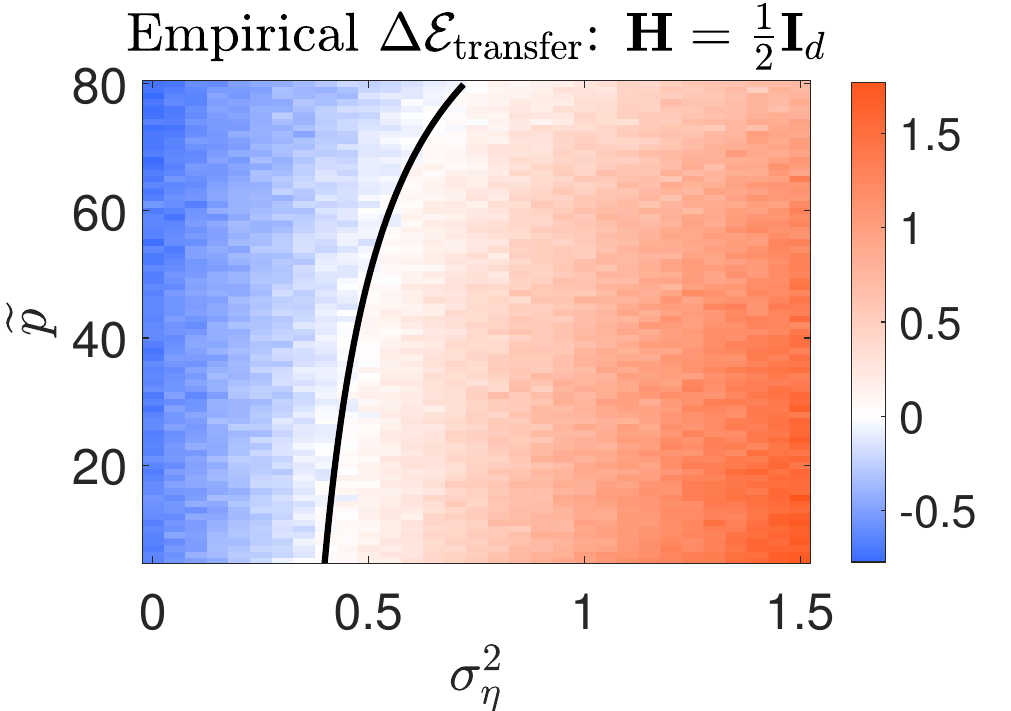}\label{appendix:fig:transfer_learning_usefulness_plane__H_is_0_5_I_empirical__n_tilde_greater_than_d}}
		\subfloat[]{\includegraphics[width=0.32\textwidth]{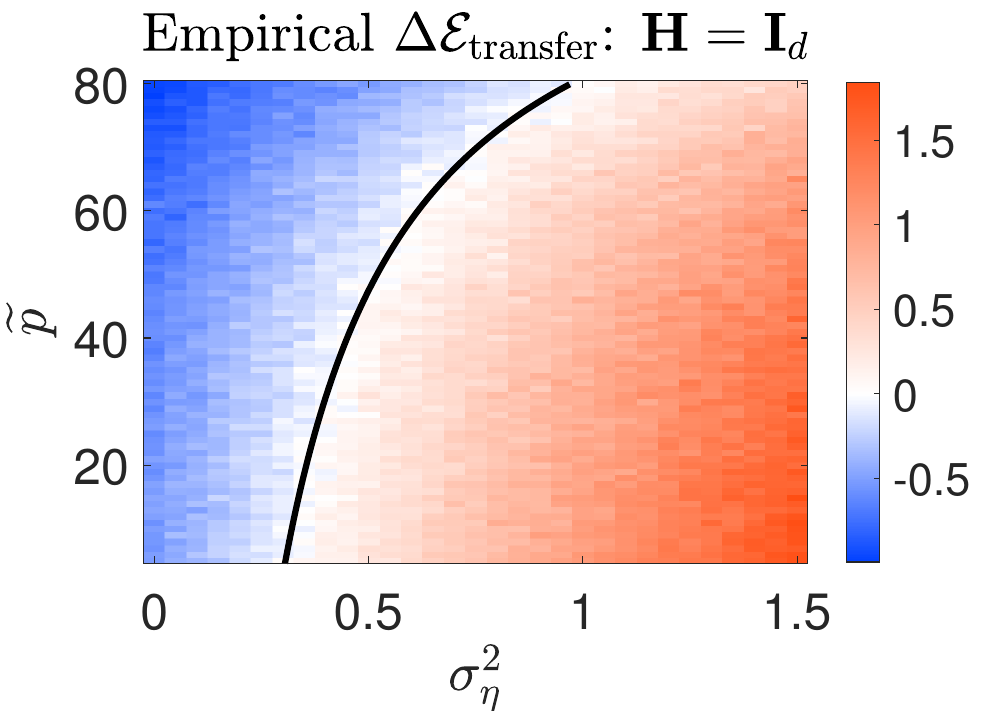} \label{appendix:fig:transfer_learning_usefulness_plane__H_is_I_empirical__n_tilde_greater_than_d}}
		\subfloat[]{\includegraphics[width=0.32\textwidth]{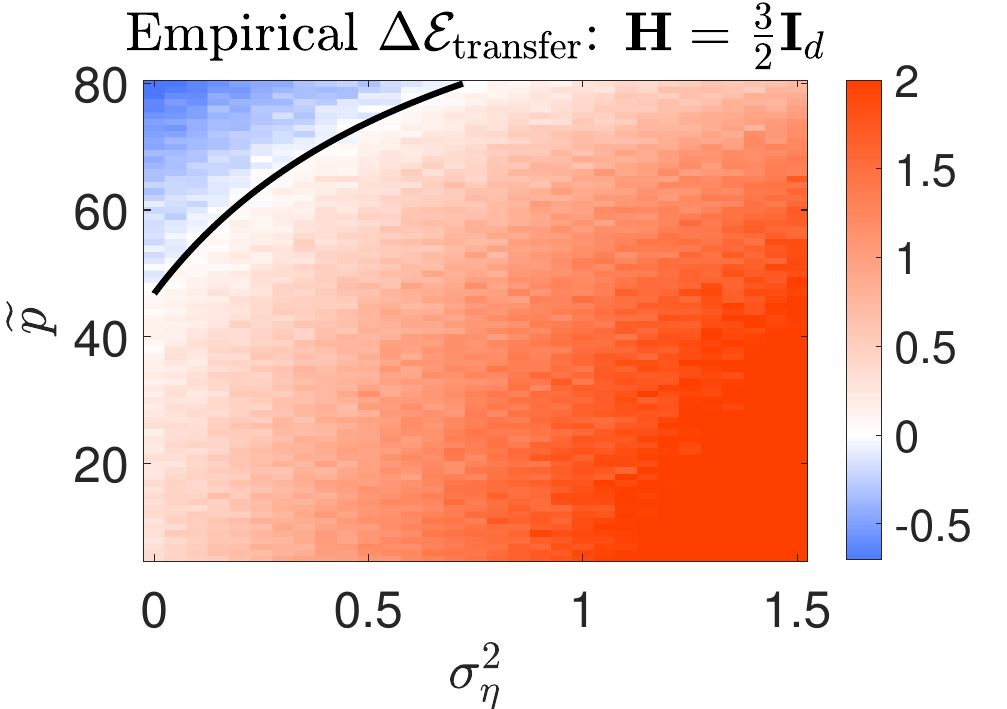} \label{appendix:fig:transfer_learning_usefulness_plane__H_is_1_5_I_empirical__n_tilde_greater_than_d}}
		\caption{The analytical (top row of subfigures) and empirical (bottom row of subfigures) values of $\Delta\mathcal{E}_{\rm transfer}$ defined in Corollary \ref{corollary:out of sample error - target task} (here, normalized by $t$, namely, the expected error difference due to transfer of a parameter from the source to target task) as a function of $\widetilde{p}$ and $\sigma_{\eta}^2$. The positive and negative values of $\frac{1}{t}\Delta\mathcal{E}_{\rm transfer}$ appear in color scales of red and blue, respectively. 
			The regions of negative values (appear in shades of blue) correspond to beneficial transfer of parameters. 
			The positive values were truncated in the value of 2 for the clarity of visualization.  Each column of subfigures correspond to a different task relation model induced by the definitions of $\mtx{H}$ as $\mtx{H}=\frac{1}{2}\mtx{I}_{d}$, $\mtx{H}=\mtx{I}_{d}$, and  $\mtx{H}=\frac{3}{2}\mtx{I}_{d}$. 
			For all the subfigures, $d=80$, $\widetilde{n}=150$, $\| \vecgreek{\beta} \|_2^2 = d$, $\sigma_{\xi}^2 = 0.025\cdot d$. Here, all the subfigures correspond to a $\vecgreek{\beta}$ vector with a linear form (see Fig.~\ref{appendix:fig:linear_beta_graph}). Note that the results in this figure are for $\widetilde{n}>d$. }
		\label{appendix:fig:transfer_learning_usefulness_plane__n_tilde_greater_than_d}
	\end{center}
	\vspace*{-5mm}
\end{figure}

Here we provide in Fig.~\ref{appendix:fig:transfer_learning_usefulness_plane__n_tilde_greater_than_d} the analytical and empirical evaluations of $\Delta\mathcal{E}_{\rm transfer}$ that correspond to settings where $\widetilde{n}>d$, we specifically consider $\widetilde{n}=150$ and $d=80$. Note that $\widetilde{n}>d$ implies that, by the definition of $\widetilde{p}$, the corresponding settings (of the source task) are underparameterized with $\widetilde{p}\le d < \widetilde{n}$. Like in Fig.~\ref{fig:transfer_learning_usefulness_plane - H is I}, the results in Fig.~\ref{appendix:fig:transfer_learning_usefulness_plane__n_tilde_greater_than_d} show the excellent match between the analytical and empirical results.

\subsection{Empirical Results for Benefits in Transferred versus Free Parameters}
\label{appendix:subsec:Empirical Results for Benefits in Transferred versus Free Parameters}

\begin{figure}[t]
	\begin{center}		
		\subfloat[{\small$\mtx{H}$: local averaging \hspace{4em}neighborhood size  3}]{\includegraphics[height=0.3\textheight]{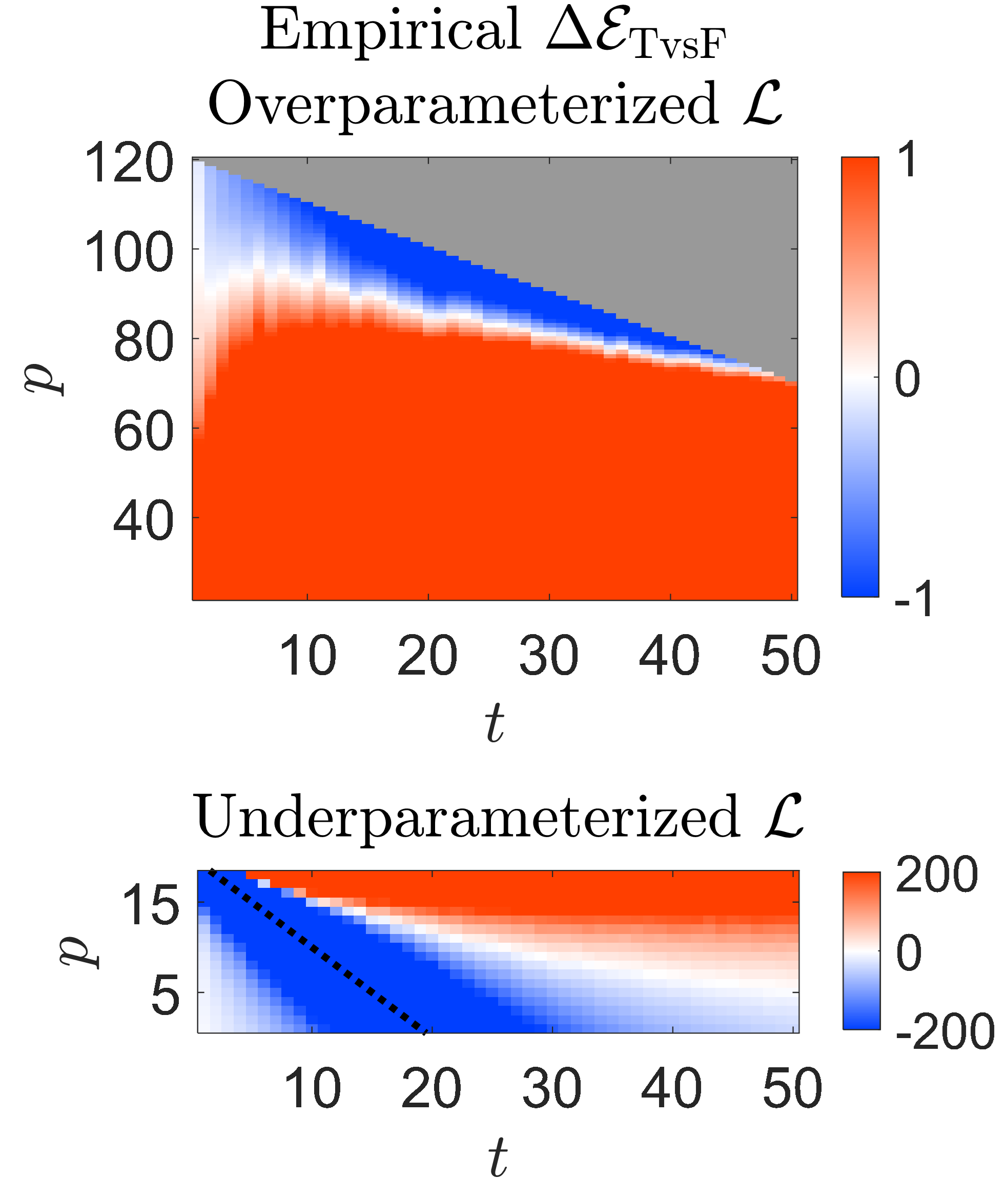}\label{fig:usefulness_transfer_vs_free_betaPWC_averaging3_empirical}}
		~~
		\subfloat[{\small$\mtx{H}$: local averaging \hspace{2em}neighborhood size  15}]{\includegraphics[height=0.3\textheight]{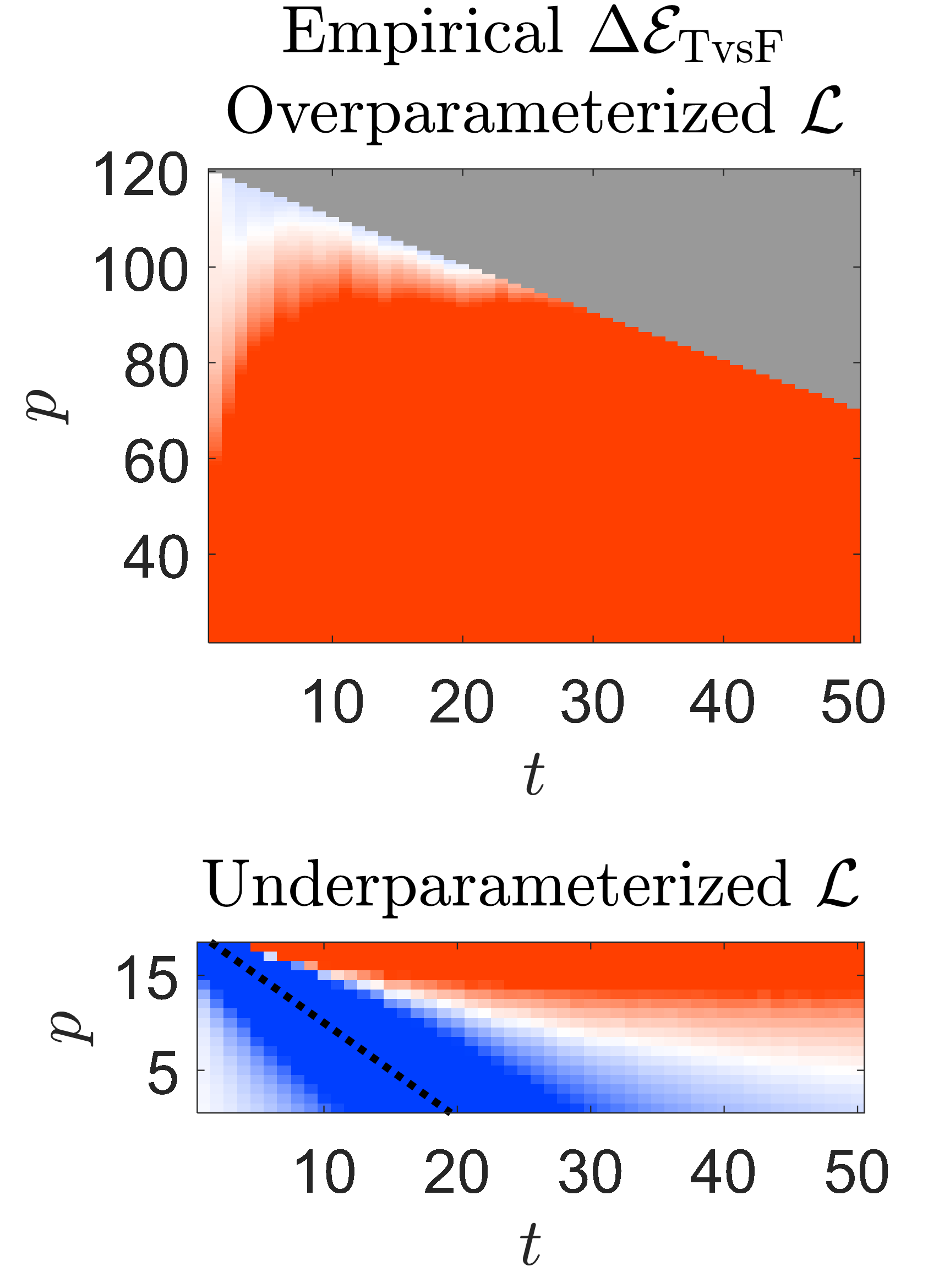}\label{fig:usefulness_transfer_vs_free_betaPWC_averaging15_empirical}}
		~~
		\subfloat[{\small$\mtx{H}=5\mtx{I}_d$}]{\includegraphics[height=0.3\textheight]{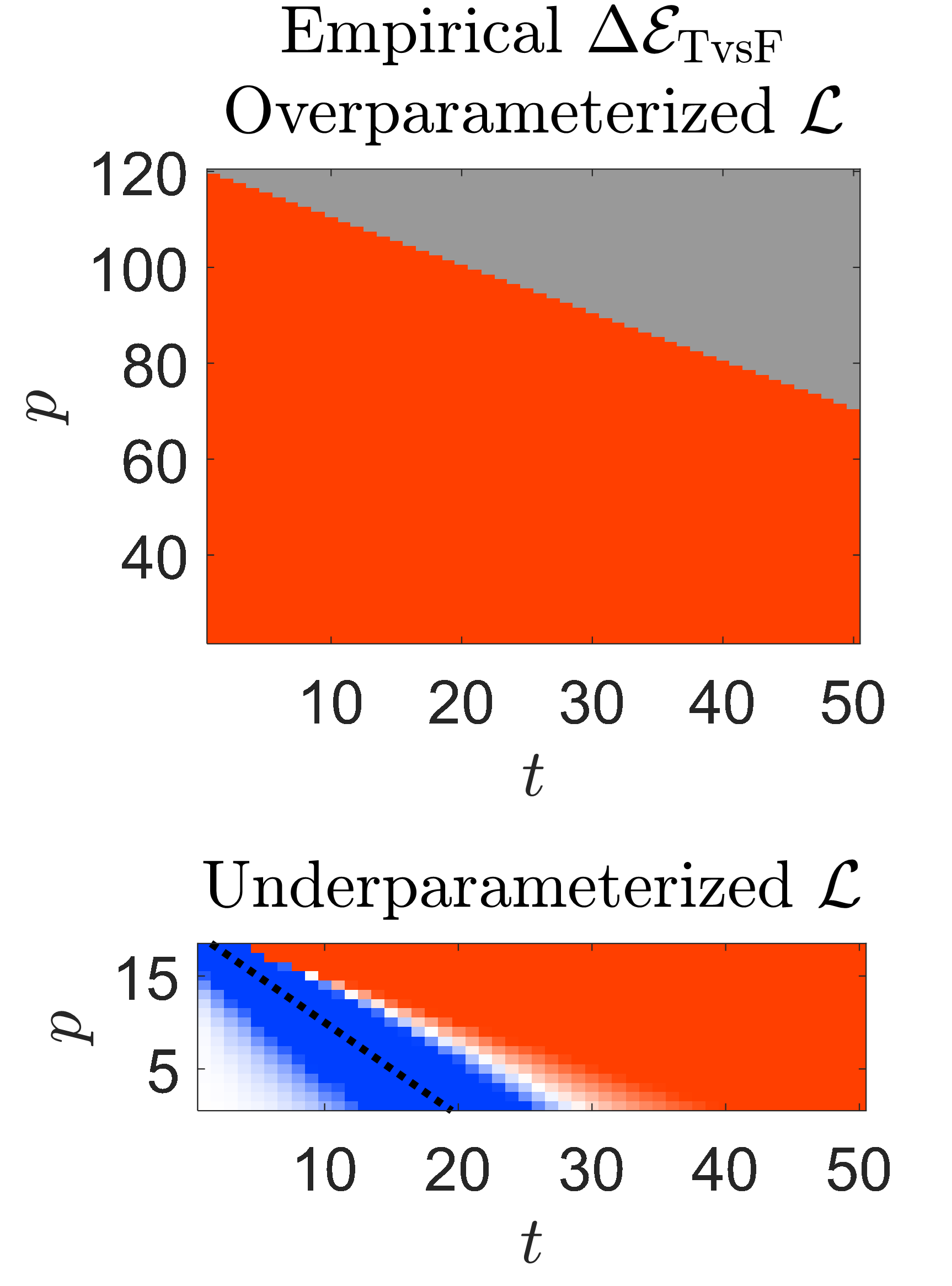}\label{fig:usefulness_transfer_vs_free_betaPWC_H_is_5I_empirical}}
		\caption{The empirical values of $\Delta\mathcal{E}_{\rm TvsF}$ (namely, the average error difference due to transfer of an arbitrarily-selected set of $t$ parameters versus setting them as free parameters) as a function of $t$ and $p$. The settings and visualization of results are as in the corresponding analytical results in Fig.~\ref{fig:transfer_learning_usefulness_plane - transfer versus free}. }
		\label{fig:transfer_learning_usefulness_plane - transfer versus free - empirical}
	\end{center}
	\vspace*{-5mm}
\end{figure}

We provide in Figure \ref{fig:transfer_learning_usefulness_plane - transfer versus free - empirical} the empirical evaluation of $\Delta\mathcal{E}_{\rm TvsF}$ that corresponds to the analytical evaluation in Figure \ref{fig:transfer_learning_usefulness_plane - transfer versus free} in the main text. 
The empirical evaluation was conducted by averaging over 750 experiments where, in each of them, $\Delta\mathcal{E}_{\rm TvsF}$ was evaluated based on its definition in (\ref{eq:error difference term - transfer versus free}) and using a different realization of $\mathcal{L}$ (from the uniform distribution that we use) and its corresponding $\mathcal{L}''$.

\section{The Optimal Componentwise Task Relation: Additional Analytical and Empirical Details}
\label{sec:The Optimal Componentwise Task Relation: Additional Analytical and Empirical Details}

\subsection{Proof of Theorem \ref{theorem:optimal eigenvalues of H}}
\label{subsec:Proof of Theorem optimal eigenvalues of H}

The proof outline for Theorem \ref{theorem:optimal eigenvalues of H}, which characterizes the optimal $\mtx{H}$ in a componentwise task relation, is as follows.
Recall that in this theorem we consider $\mtx{H}={\rm diag}\{\lambda_{\mtx{H}}^{(1)},\dots,\lambda_{\mtx{H}}^{(d)}\}$. 
The proof starts by setting the expressions from (\ref{eq:componentwise transfer bias})-(\ref{eq: zeta_S in terms of H eigenvalues}) in the formulations of the transfer bias and variance from Corollary \ref{corollary:out of sample error - target task - specific layout - detailed - specific layout}, then we can also reformulate the expression of $\mathcal{E}_{\rm transfer}^{(\mathcal{T},\mathcal{S})}$ from Theorem \ref{theorem:out of sample error - target task - specific layout}.

In the case of an underparameterized source task where $1\le \widetilde{p}\le \widetilde{n}-2$, we get 
\begin{equation}
\label{eq:optimal H proof - error difference term - underparameterized}
\mathcal{E}_{\rm transfer}^{(\mathcal{T},\mathcal{S})} = t\sigma_{\eta}^2 + \sum_{j\in\mathcal{T}} {\left({\lambda_{\mtx{H}}^{(j)} - 1}\right)^2 \left({\beta^{(j)} }\right)^2} + \frac{t}{\widetilde{n}-\widetilde{p}-1}\left({\left(d-\widetilde{p}\right)\sigma_{\eta}^2 + \sum_{j\in\mathcal{S}^c} {\left({\lambda_{\mtx{H}}^{(j)} \beta^{(j)} }\right)^2} + \sigma_{\xi}^2}\right)
\end{equation}
Recall that $\mathcal{T}\subseteq\mathcal{S}$ and hence $\mathcal{T}\cap\mathcal{S}^c = \emptyset$. Accordingly, the two sums in (\ref{eq:optimal H proof - error difference term - underparameterized}) include distinct sets of coordinates, which simplify the derivative of $\mathcal{E}_{\rm transfer}^{(\mathcal{T},\mathcal{S})}$ with respect to $\lambda_{\mtx{H}}^{(k)}$ for a particular $k$. Hence, in this case of an underparameterized source task and $\beta^{(k)}\neq 0$, the condition $\frac{\partial\mathcal{E}_{\rm transfer}^{(\mathcal{T},\mathcal{S})}}{\partial\lambda_{\mtx{H}}^{(k)}}=0$ is satisfied by $\lambda_{\mtx{H}}^{(k)}=1$ for $k\in\mathcal{T}$ and by $\lambda_{\mtx{H}}^{(k)}=0$ for $k\in\mathcal{S}^c$, which minimize $\mathcal{E}_{\rm transfer}^{(\mathcal{T},\mathcal{S})}$ due to convexity. Note that the eigenvalues $\{\lambda_{\mtx{H}}^{(k)}\}_{k\in\mathcal{S}\setminus\mathcal{T}}$ do not appear in (\ref{eq:optimal H proof - error difference term - underparameterized}) and hence they can have any value without affecting the minimization of $\mathcal{E}_{\rm transfer}^{(\mathcal{T},\mathcal{S})}$ for $\widetilde{p}\le\widetilde{n}-2$.

In the case of an overparameterized source task where $\widetilde{p}\ge \widetilde{n}+2$, we get 
\begin{align}
\label{eq:optimal H proof - error difference term - overparameterized}
\mathcal{E}_{\rm transfer}^{(\mathcal{T},\mathcal{S})} &= \frac{\widetilde{n}}{\widetilde{p}}\left( t \sigma_{\eta}^2 + \frac{\widetilde{p}-\widetilde{n}}{\widetilde{p}^2 - 1} t \sum_{j\in\mathcal{S}\setminus\mathcal{T}} {\left({\lambda_{\mtx{H}}^{(j)} \beta^{(j)} }\right)^2} 
\right.
\\\nonumber
& + \sum_{j\in\mathcal{T}} {\left({ \frac{\left( \widetilde{p}-\widetilde{n} \right)t + \widetilde{n}\widetilde{p}-1}{\widetilde{p}^2 - 1}  \lambda_{\mtx{H}}^{(j)} - 2}\right)\lambda_{\mtx{H}}^{(j)}\left({\beta^{(j)} }\right)^2} 
\\\nonumber
& \left. +\frac{t}{\widetilde{p}-\widetilde{n}-1}\left({\left(d-\widetilde{p}\right)\sigma_{\eta}^2 + \sum_{j\in\mathcal{S}^c} {\left({\lambda_{\mtx{H}}^{(j)} \beta^{(j)} }\right)^2} + \sigma_{\xi}^2}\right) \right)
\end{align}
where the three sums refer to disjoint sets of coordinates. Accordingly, in this case of an overparameterized source task and $\beta^{(k)}\neq 0$, the condition $\frac{\partial\mathcal{E}_{\rm transfer}^{(\mathcal{T},\mathcal{S})}}{\partial\lambda_{\mtx{H}}^{(k)}}=0$ is satisfied by $\lambda_{\mtx{H}}^{(k)}=\frac{\widetilde{p}^2 - 1}{\widetilde{n}\widetilde{p}-1+ t \left(\widetilde{p}-\widetilde{n}\right)}$ for $k\in\mathcal{T}$ and by $\lambda_{\mtx{H}}^{(k)}=0$ for $k\in\mathcal{T}^c$, which minimize $\mathcal{E}_{\rm transfer}^{(\mathcal{T},\mathcal{S})}$ due to convexity. 

In both cases (\ref{eq:optimal H proof - error difference term - underparameterized}) and (\ref{eq:optimal H proof - error difference term - overparameterized}), if $k\in\{1,\dots,d\}$ corresponds to $\beta^{(k)} = 0$ then $\lambda_{\mtx{H}}^{(k)}$ may have any value without affecting the minimization of $\mathcal{E}_{\rm transfer}^{(\mathcal{T},\mathcal{S})}$. 
By this we complete the proof outline for Theorem \ref{theorem:optimal eigenvalues of H}.

\subsection{Empirical Results for The Optimal $\mtx{H}$ in a Componentwise Task Relation}
\label{appendix:subsec:Empirical Results for The Optimal H in a Componentwise Task Relation}

\begin{figure}[t]
	\begin{center}		
		\subfloat[{\small$\sigma_{\eta}^2 = 0.2$; $\vecgreek{\beta}$ has the form in~Fig.~\ref{fig:beta_all_ones_usefulness}}]{\includegraphics[width=0.31\textwidth]{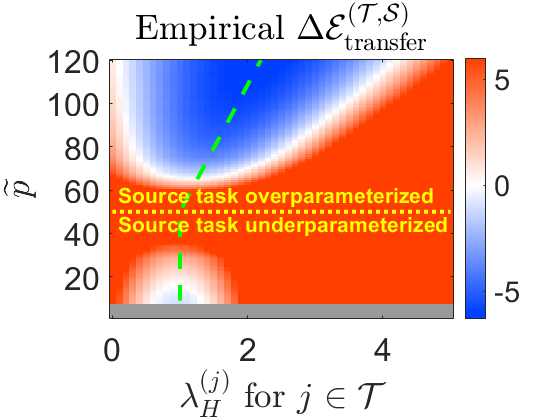}\label{fig:eigenvals_usefulness_beta_all__ones_t8__eta0_2_empirical}}
		\subfloat[{\small$\sigma_{\eta}^2 = 1$; $\vecgreek{\beta}$ has the form in~Fig.~\ref{fig:beta_all_ones_usefulness} }]{\includegraphics[width=0.28\textwidth]{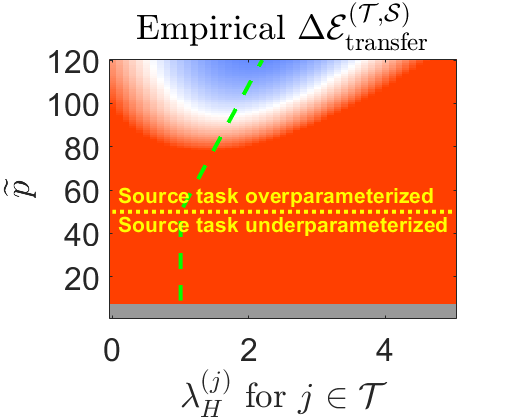}\label{fig:eigenvals_usefulness_beta_all__ones_t8__eta1_empirical}}			
		\subfloat[{\small$\sigma_{\eta}^2 = 0.2$;  $\vecgreek{\beta}$ has the form in~Fig.~\ref{fig:beta_high_in_non_transferred__usefulness} }]{\includegraphics[width=0.28\textwidth]{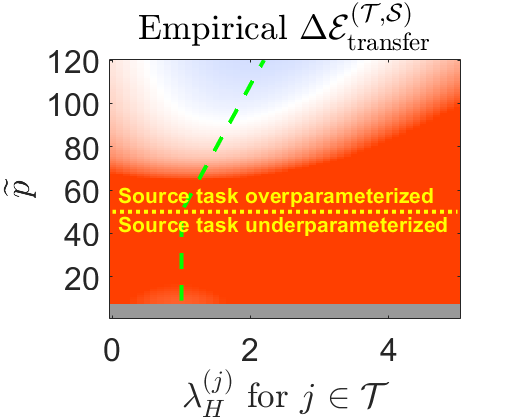}\label{fig:eigenvals_usefulness_beta_high_in_non_transferred_t8__eta0_2_empirical}}
		\caption{The empirical evaluation of $\Delta\mathcal{E}_{\rm transfer}^{(\mathcal{T},\mathcal{S})}$ as a function of $\widetilde{p}$ and a value that determines the eigenvalues of $\mtx{H}$ in $\mathcal{T}$. 
			The settings and visualization of results are as in the corresponding analytical results in Fig.~\ref{fig:transfer_learning_usefulness_plane - in terms of H eigenvalues}.
		}
		\label{appendix:fig:transfer_learning_usefulness_plane - in terms of H eigenvalues - empirical}
	\end{center}
	\vspace*{-5mm}
\end{figure}

In Figure \ref{appendix:fig:transfer_learning_usefulness_plane - in terms of H eigenvalues - empirical} we provide the empirical evaluations that correspond to the analytical results in Figure~\ref{fig:transfer_learning_usefulness_plane - in terms of H eigenvalues}. The empirical values were computed by averaging over 500 experiments. 

\section{Additional Details and Results for Section 6}
\label{appendix:sec:Additional Details and Results for Section 6}

\subsection{Ridge Regression: Error Expression and Optimal Tuning for the Source Task}
\label{appendix:subsec:Ridge Regression Error Expression and Optimal Tuning for the Source Task}

The out-of-sample error of the ridge solution (\ref{eq:constrained linear regression - source - ridge regression - free parameters})-(\ref{eq:constrained linear regression - source - ridge regression - free parameters - closed form}) to the source task is developed as follows. 
First, due to the layout of free parameters in $\widehat{\vecgreek{\theta}}$ we get that 
\begin{align}
\widetilde{\mathcal{E}}_{\rm out} & =   \sigma_{\xi}^2 + \expectation{ \left \Vert { \widehat{\vecgreek{\theta}} - \vecgreek{\theta} } \right \Vert _2^2 }  
\nonumber \\ 
& =   \sigma_{\xi}^2 +  \Ltwonorm{\vecgreek{\theta}_{\mathcal{S}^c}} 
+ 
\expectation{ \Ltwonorm{ \widehat{\vecgreek{\theta}}_{\mathcal{S}}  - \vecgreek{\theta}_{\mathcal{S}} } }.
\label{appendix:eq:out of sample error - source task - ridge - development - starting decomposition}
\end{align}
Now we set the closed-form ridge solution from (\ref{eq:constrained linear regression - source - ridge regression - free parameters - closed form}) to develop the third term in (\ref{appendix:eq:out of sample error - source task - ridge - development - starting decomposition}):
\begin{align}
&\expectation{ \Ltwonorm{ \widehat{\vecgreek{\theta}}_{\mathcal{S}}  - \vecgreek{\theta}_{\mathcal{S}} } } =
\expectation{ \Ltwonorm{\left( \mtx{Z}_{\mathcal{S}}^T \mtx{Z}_{\mathcal{S}} + \widetilde{\alpha} \mtx{I}_{\widetilde{p}} \right)^{-1} \mtx{Z}_{\mathcal{S}}^T \vec{v}  - \vecgreek{\theta}_{\mathcal{S}}} } 
\nonumber \\ 
& =  \expectation{ \Ltwonorm{\left( \mtx{Z}_{\mathcal{S}}^T \mtx{Z}_{\mathcal{S}} + \widetilde{\alpha} \mtx{I}_{\widetilde{p}} \right)^{-1} \mtx{Z}_{\mathcal{S}}^T \left({ \mtx{Z}_{\mathcal{S}^c} \vecgreek{\theta}_{\mathcal{S}^c} + \vecgreek{\xi} }\right) } } + \expectation{ \Ltwonorm{ \left( \left( \mtx{Z}_{\mathcal{S}}^T \mtx{Z}_{\mathcal{S}} + \widetilde{\alpha} \mtx{I}_{\widetilde{p}} \right)^{-1} \mtx{Z}_{\mathcal{S}}^T \mtx{Z}_{\mathcal{S}} - \mtx{I}_{\widetilde{p}} \right)\vecgreek{\theta}_{\mathcal{S}}} }.
\nonumber \\
\label{appendix:eq:out of sample error - source task - ridge - development - decomposition with ridge}
\end{align}
Next, note that 
\begin{align}
&\expectation{ \Ltwonorm{\left( \mtx{Z}_{\mathcal{S}}^T \mtx{Z}_{\mathcal{S}} + \widetilde{\alpha} \mtx{I}_{\widetilde{p}} \right)^{-1} \mtx{Z}_{\mathcal{S}}^T \left({ \mtx{Z}_{\mathcal{S}^c} \vecgreek{\theta}_{\mathcal{S}^c} + \vecgreek{\xi} }\right) } } = \left( \sigma_{\xi}^2 + \Ltwonorm{\vecgreek{\theta}_{\mathcal{S}^c}} \right) \expectation{ \mtxtrace{ \left( \mtx{Z}_{\mathcal{S}}^T \mtx{Z}_{\mathcal{S}} + \widetilde{\alpha} \mtx{I}_{\widetilde{p}} \right)^{-2} \mtx{Z}_{\mathcal{S}}^T \mtx{Z}_{\mathcal{S}} } } 
\label{appendix:eq:out of sample error - source task - ridge - development - decomposition with ridge - first term}
\end{align}
Also, we use the eigendecomposition $\mtx{Z}_{\mathcal{S}}^T \mtx{Z}_{\mathcal{S}}=\mtx{\Phi}_{\mtx{Z}_{\mathcal{S}}} \mtx{\Lambda}_{\mtx{Z}_{\mathcal{S}}} \mtx{\Phi}_{\mtx{Z}_{\mathcal{S}}}^{T}$ where $\mtx{\Phi}_{\mtx{Z}_{\mathcal{S}}}$ is a $\widetilde{p}\times\widetilde{p}$ orthonormal matrix and $\mtx{\Lambda}_{\mtx{Z}_{\mathcal{S}}}\triangleq {\rm{diag}}\{\lambda_{\mtx{Z}_{\mathcal{S}},1},\dots,\lambda_{\mtx{Z}_{\mathcal{S}},\widetilde{p}} \}$ is the $\widetilde{p}\times\widetilde{p}$ diagonal matrix formed by the eigenvalues of $\mtx{Z}_{\mathcal{S}}^T \mtx{Z}_{\mathcal{S}}$. 
By using this eigendecomposition and (\ref{appendix:eq:out of sample error - source task - ridge - development - decomposition with ridge - first term}), we develop (\ref{appendix:eq:out of sample error - source task - ridge - development - decomposition with ridge}) into the form of 
\begin{align}
&\expectation{ \Ltwonorm{ \widehat{\vecgreek{\theta}}_{\mathcal{S}}  - \vecgreek{\theta}_{\mathcal{S}} } } =
\nonumber \\ 
& =  \left( \sigma_{\xi}^2 + \Ltwonorm{\vecgreek{\theta}_{\mathcal{S}^c}} \right) \expectation{ \mtxtrace{ \left( \mtx{\Lambda}_{\mtx{Z}_{\mathcal{S}}} + \widetilde{\alpha} \mtx{I}_{\widetilde{p}} \right)^{-2} \mtx{\Lambda}_{\mtx{Z}_{\mathcal{S}}} } }  
+ \frac{\Ltwonorm{\vecgreek{\theta}_{\mathcal{S}}}}{\widetilde{p}}\widetilde{\alpha}^2 \expectation{ \mtxtrace{ \left( \mtx{\Lambda}_{\mtx{Z}_{\mathcal{S}}} + \widetilde{\alpha} \mtx{I}_{\widetilde{p}} \right)^{-2} } }.
\nonumber \\
& = \expectation{\sum_{k=1}^{\widetilde{p}}{\frac{ \left( \sigma_{\xi}^2 + \Ltwonorm{\vecgreek{\theta}_{\mathcal{S}^c}} \right) \lambda_{\mtx{Z}_{\mathcal{S}},k} + \frac{\Ltwonorm{\vecgreek{\theta}_{\mathcal{S}}}}{\widetilde{p}}\widetilde{\alpha}^2  }{\left( \lambda_{\mtx{Z}_{\mathcal{S}},k} + \widetilde{\alpha} \right)^2}} }.
\nonumber \\
\label{appendix:eq:out of sample error - source task - ridge - development - decomposition with ridge - eigenvalues form}
\end{align}
The optimal tuning is achieved by the $\widetilde{\alpha}$ value that satisfies $\frac{\partial \widetilde{\mathcal{E}}_{\rm out}}{\partial\widetilde{\alpha}}=0$, which by using (\ref{appendix:eq:out of sample error - source task - ridge - development - starting decomposition}) and (\ref{appendix:eq:out of sample error - source task - ridge - development - decomposition with ridge - eigenvalues form}) is $\widetilde{\alpha}=\frac{\widetilde{p}\left(\sigma_{\xi}^2 + \Ltwonorm{\vecgreek{\theta}_{\mathcal{S}^c}}\right)}{\Ltwonorm{\vecgreek{\theta}_{\mathcal{S}}}}$. In the main text we explain how we approximate the optimal $\widetilde{\alpha}$ in our experiments.

\subsection{Ridge Regression: Error Expression and Optimal Tuning for the Target Task}
\label{appendix:subsec:Ridge Regression Error Expression and Optimal Tuning for the Target Task}

Let us start by explaining in more detail the ridge solution in (\ref{eq:constrained linear regression - target - ridge regression - free parameters - closed form}). For this, note that the optimization constraints in (\ref{eq:constrained linear regression - target task - ridge regression}) imply $\Ltwonorm{\vec{y} - \mtx{X}\vec{b}}=\Ltwonorm{\vec{y} - \mtx{X}_{\mathcal{F}}\vec{b}_{\mathcal{F}} - \mtx{X}_{\mathcal{T}}\widehat{\vecgreek{\theta}}_{\mathcal{T}} }$. Hence, the solution of (\ref{eq:constrained linear regression - target task - ridge regression}) is equivalent to $\widehat{\vecgreek{\beta}}$ where $\widehat{\vecgreek{\beta}}_{\mathcal{Z}}=0$, $\widehat{\vecgreek{\beta}}_{\mathcal{T}}=\widehat{\vecgreek{\theta}}_{\mathcal{T}}$, and 
\begin{align} 
\label{eq:constrained linear regression - target - ridge regression - free parameters - standard ridge optimization form}
\widehat{\vecgreek{\beta}}_{\mathcal{F}} =  \argmin_{\vec{f}\in\mathbb{R}^{p}}  \Ltwonorm{ \left(\vec{y}-\mtx{X}_{\mathcal{T}}\widehat{\vecgreek{\theta}}_{\mathcal{T}} \right) - \mtx{X}_{\mathcal{F}} \vec{f} } + \alpha \Ltwonorm{\vec{f}}. 
\end{align}
The optimization in (\ref{eq:constrained linear regression - target - ridge regression - free parameters - standard ridge optimization form}) has a standard ridge regression form and, accordingly, its closed-form solution is provided in (\ref{eq:constrained linear regression - target - ridge regression - free parameters - closed form}). 

Let us develop the out-of-sample error expression of the target task. 
Based on the layout of free, transferred, and zeroed parameters in $\widehat{\vecgreek{\beta}}$ we have  
\begin{align}
\mathcal{E}_{\rm out} & =   \sigma_{\epsilon}^2 + \expectation{ \left \Vert { \widehat{\vecgreek{\beta}} - \vecgreek{\beta} } \right \Vert _2^2 }  
\nonumber \\ 
& =   \sigma_{\epsilon}^2 +  \Ltwonorm{\vecgreek{\beta}_{\mathcal{Z}}} 
+ 
\expectation{ \Ltwonorm{ \widehat{\vecgreek{\theta}}_{\mathcal{T}}  - \vecgreek{\beta}_{\mathcal{T}} } }
+ 
\expectation{ \Ltwonorm{ \widehat{\vecgreek{\beta}}_{\mathcal{F}}  - \vecgreek{\beta}_{\mathcal{F}} } }.
\label{appendix:eq:out of sample error - target task - ridge - development - starting decomposition}
\end{align}
Then, similar to the proof given for the source task in Section \ref{appendix:subsec:Ridge Regression Error Expression and Optimal Tuning for the Source Task}, we get that 
\begin{align}
&\expectation{ \Ltwonorm{ \widehat{\vecgreek{\beta}}_{\mathcal{F}}  - \vecgreek{\beta}_{\mathcal{F}} } } =
\nonumber \\ 
& =  \left( \sigma_{\epsilon}^2 + \Ltwonorm{\vecgreek{\beta}_{\mathcal{Z}}} + \expectation{\Ltwonorm{\vecgreek{\beta}_{\mathcal{T}}-\widehat{\vecgreek{\theta}}_{\mathcal{T}} }} \right) \expectation{ \mtxtrace{ \left( \mtx{\Lambda}_{\mtx{X}_{\mathcal{F}}} + \alpha \mtx{I}_{p} \right)^{-2} \mtx{\Lambda}_{\mtx{X}_{\mathcal{F}}} } }  
\nonumber \\ 
&\quad + \frac{\Ltwonorm{\vecgreek{\beta}_{\mathcal{F}}}}{p}\alpha^2 \expectation{ \mtxtrace{ \left( \mtx{\Lambda}_{\mtx{X}_{\mathcal{F}}} + \alpha \mtx{I}_{p} \right)^{-2} } }.
\nonumber \\
& = \expectation{\sum_{k=1}^{p}{\frac{ \left( \sigma_{\epsilon}^2 + \Ltwonorm{\vecgreek{\beta}_{\mathcal{Z}}} + \expectation{\Ltwonorm{\vecgreek{\beta}_{\mathcal{T}}-\widehat{\vecgreek{\theta}}_{\mathcal{T}} }}\right) \lambda_{\mtx{X}_{\mathcal{F}},k} + \frac{\Ltwonorm{\vecgreek{\beta}_{\mathcal{F}}}}{p}\alpha^2 }{\left( \lambda_{\mtx{X}_{\mathcal{F}},k} + \alpha \right)^2}} }, 
\nonumber \\
\label{appendix:eq:out of sample error - target task - ridge - development - decomposition with ridge - eigenvalues form}
\end{align}
where we use the eigendecomposition $\mtx{X}_{\mathcal{F}}^T \mtx{X}_{\mathcal{F}}=\mtx{\Phi}_{\mtx{X}_{\mathcal{F}}} \mtx{\Lambda}_{\mtx{X}_{\mathcal{F}}} \mtx{\Phi}_{\mtx{X}_{\mathcal{F}}}^{T}$ where $\mtx{\Phi}_{\mtx{X}_{\mathcal{F}}}$ is a $p\times p$ orthonormal matrix and $\mtx{\Lambda}_{\mtx{X}_{\mathcal{F}}}\triangleq {\rm{diag}}\{\lambda_{\mtx{X}_{\mathcal{F}},1},\dots,\lambda_{\mtx{X}_{\mathcal{F}},p} \}$ is the $p\times p$ diagonal matrix of the eigenvalues of $\mtx{X}_{\mathcal{F}}^T \mtx{X}_{\mathcal{F}}$. 

Here, the optimal tuning is obtained by the $\alpha$ value that provides $\frac{\partial \mathcal{E}_{\rm out}}{\partial\alpha}=0$. Then, (\ref{appendix:eq:out of sample error - target task - ridge - development - starting decomposition}) and (\ref{appendix:eq:out of sample error - target task - ridge - development - decomposition with ridge - eigenvalues form}) imply that the optimal tuning is given by
\begin{equation}
\label{appendix:eq:optimal alpha ridge target task}
\alpha=\frac{p\left(\sigma_{\epsilon}^2 + \Ltwonorm{\vecgreek{\beta}_{\mathcal{Z}}} + \expectation{\Ltwonorm{\vecgreek{\beta}_{\mathcal{T}}-\widehat{\vecgreek{\theta}}_{\mathcal{T}} }} \right)}{\Ltwonorm{\vecgreek{\beta}_{\mathcal{F}}}}.
\end{equation} 
In our experiments we assume that only $\Ltwonorm{\vecgreek{\beta}}$, $\Ltwonorm{\vecgreek{\beta}-\vecgreek{\theta}}$, $\sigma_{\epsilon}$, $p$, $t$, and $d$ are known; thus, we use the approximations $\Ltwonorm{\vecgreek{\beta}_{\mathcal{F}}}\approx\frac{p}{d}\Ltwonorm{\vecgreek{\beta}}$, $\Ltwonorm{\vecgreek{\beta}_{\mathcal{Z}}}\approx\left( 1 - \frac{p+t}{d}\right)\Ltwonorm{\vecgreek{\beta}}$, and 
\begin{equation}
\nonumber
\expectation{\Ltwonorm{\vecgreek{\beta}_{\mathcal{T}}-\widehat{\vecgreek{\theta}}_{\mathcal{T}} }} \approx \frac{t}{d} \expectation{\Ltwonorm{\vecgreek{\beta}-\vecgreek{\theta}}} = \frac{t}{d} \expectation{\Ltwonorm{\vecgreek{\beta}-\left(\mtx{H}\vecgreek{\beta}+\vecgreek{\eta}\right)}} = \frac{t}{d} \Ltwonorm{\left(\mtx{I}_{d}-\mtx{H}\right)\vecgreek{\beta}} + t\sigma_{\eta}^2, 
\end{equation}
to approximate (\ref{appendix:eq:optimal alpha ridge target task}) as 
$\alpha = \frac{d\sigma_{\epsilon}^2}{\Ltwonorm{\vecgreek{\beta}}} + d - p - t + t\frac{\Ltwonorm{\left(\mtx{I}_d - \mtx{H}\right)\vecgreek{\beta}} + d\sigma_{\eta}^2}{\Ltwonorm{\vecgreek{\beta}}}$ in our experiments.

\bibliographystyle{siamplain}
\bibliography{transfer_learning_references}

\end{document}